\documentclass[10pt,twocolumn,letterpaper]{article}

\usepackage[pagenumbers]{cvpr} % To force page numbers, e.g. for an arXiv version

\definecolor{cvprblue}{rgb}{0.21,0.49,0.74}
\usepackage[pagebackref,breaklinks,colorlinks,allcolors=cvprblue]{hyperref}

\usepackage{hyperref}
\usepackage{url}
\usepackage{algorithm}
\usepackage[margin=1in]{geometry}
\usepackage{amsmath, amssymb, amsthm, mathtools}
\usepackage{bm}
\usepackage{microtype}
\usepackage{booktabs}
\usepackage{graphicx}
\usepackage{xcolor}
\usepackage{enumitem}
\usepackage{multicol}
\usepackage{thmtools}
\usepackage{algorithm}
\usepackage[noend]{algpseudocode}
\usepackage{hyperref}
\usepackage{tabularx}
\usepackage{subcaption}
\usepackage{float}

\usepackage{makecell}
\usepackage[utf8]{inputenc} % allow utf-8 input
\usepackage[T1]{fontenc}    % use 8-bit T1 fonts
\usepackage{hyperref}       % hyperlinks
\usepackage{url}            % simple URL typesetting
\usepackage{booktabs}       % professional-quality tables
\usepackage{amsfonts}       % blackboard math symbols
\usepackage{nicefrac}       % compact symbols for 1/2, etc.
\usepackage{microtype}      % microtypography
\usepackage{xcolor}         % colors
\usepackage{graphicx}
\usepackage{array}
\usepackage{adjustbox}
\usepackage{float} \usepackage{wrapfig} 
\usepackage{newunicodechar}
\newunicodechar{↙}{\swarrow}
\usepackage[most]{tcolorbox}

\usepackage{tcolorbox}
\newtcolorbox{respfigbox}[1][]{
  enhanced,
  colback=white, colframe=black!40, boxrule=0.4pt, arc=2mm,
  left=2mm, right=2mm, top=2mm, bottom=2mm, boxsep=1mm,
  breakable=false,              % <-- important
  #1
}

\usepackage[most]{tcolorbox}
\newtcolorbox{panelbox}[1][]{
  enhanced, breakable=false,
  colframe=black!35, boxrule=0.3pt, arc=2pt,
  left=1mm, right=1mm, top=1mm, bottom=1mm, boxsep=0.5mm,
  #1
}

\def\paperID{20387} % *** Enter the Paper ID here
\def\confName{CVPR}
\def\confYear{2026}

\title{\textsc{Sphinx}: A Synthetic Environment for Visual Perception and Reasoning}

\author{Md Tanvirul Alam\\
Rochester Institute of Technology\\
Rochester, NY, USA\\
{\tt\small ma8235@rit.edu}
\and
Saksham Aggarwal\\
Rochester Institute of Technology\\
Rochester, NY, USA\\
{\tt\small sxavse@rit.edu}
\and
Justin Yang Chae\\
University of Washington\\
Seattle, WA, USA\\
{\tt\small jchae3@uw.edu}
\and
Nidhi Rastogi\\
Rochester Institute of Technology\\
Rochester, NY, USA\\
{\tt\small nxrvse@rit.edu}
}

\begin{document}
\maketitle

\begin{abstract}
We present \textsc{Sphinx}, a synthetic environment for visual perception and reasoning that targets core cognitive primitives. \textsc{Sphinx} procedurally generates puzzles using motifs, tiles, charts, icons, and geometric primitives, each paired with verifiable ground-truth solutions, enabling both precise evaluation and large-scale dataset construction. The benchmark covers 25 task types spanning symmetry detection, geometric transformations, spatial reasoning, chart interpretation, and sequence prediction. Evaluating recent large vision–language models (LVLMs) shows that even state-of-the-art GPT-5 attains only 51.1\% accuracy, well below human performance. Finally, we demonstrate that reinforcement learning with verifiable rewards (RLVR) substantially improves model accuracy on these tasks and yields gains on external visual reasoning benchmarks, highlighting its promise for advancing multimodal reasoning. Project page, code, and dataset available at \url{https://maveryn.github.io/sphinx/}.
\end{abstract}

\section{Introduction}

Large language models (LLMs) have recently demonstrated striking advances in reasoning, achieving gold medal level performance at the International Mathematical Olympiad~\citep{castelvecchi2025ai} and strong results across mathematics, logical reasoning, and coding~\citep{guo2025deepseek,jaech2024openai,wu2024reasoning,comanici2025gemini,yang2025qwen3}. Because reasoning is a core component of human intelligence, it has become a central benchmark for progress toward Artificial General Intelligence (AGI)~\citep{goertzel2014artificial}. Techniques such as Chain-of-Thought prompting~\citep{wei2022chain}, test-time compute scaling~\citep{jaech2024openai}, and post-training strategies such as rule-based reinforcement learning in DeepSeek-R1 have further improved model performance, helping mitigate reward hacking~\citep{guo2025deepseek} and allowing more robust generalization across domains~\citep{xie2025logic,albalak2025big,he2025deepmath}.

\begin{figure}[t]
    \centering
    \includegraphics[width=0.42\textwidth]{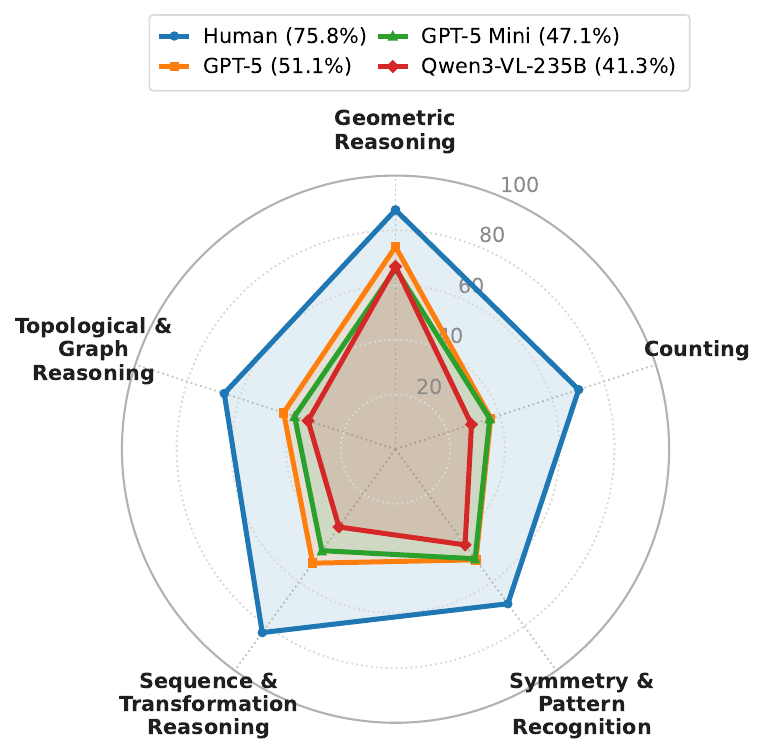}
    \caption{Radar plot shows accuracies (\%) achieved by LVLMs and by humans on the broad categories of \textsc{Sphinx}.}
    \label{fig:radar-llm}
\end{figure}

In contrast to the rapid progress of LLMs, large vision-language models (LVLMs) remain far less capable of visual reasoning~\cite{yuksekgonul2022and,schulze2025visual,malkinski2024reasoning,smart101}. Unlike text-based systems that can leverage structured prompts and post-training strategies, LVLMs must jointly parse visual inputs and integrate them with language, a substantially more complex challenge~\citep{gandhi2025cognitive,guo2025deepseek,xie2025logic,wang2025jigsaw,cao2024visual}. Current models often fail to construct coherent reasoning chains and stumble on tasks trivial to humans~\citep{yang2025r1}. Although reinforcement learning has been applied to strengthen LVLMs~\citep{liu2025visual,peng2025lmm}, progress is constrained by benchmarks that emphasize perception over reasoning, such as referring to expression comprehension or math-with-diagram datasets, where models frequently reduce visual inputs to text and rely on language reasoning~\citep{xu2025visulogic,zhang2024mathverse}.

More recently, several works have begun to investigate abstract visual reasoning (AVR) in LVLMs~\citep{xu2025visulogic,cao2024visual,malkinski2024reasoning,jiang2024marvel,arc_analysis,chollet2025arc}, yet these efforts still fall short of systematically evaluating core perceptual primitives such as symmetry detection, mental rotation, and structured pattern matching. Cognitive science has long established that these abilities underpin fluid intelligence and matrix reasoning~\citep{fisher1981role,carpenter1990one,pizlo2021concept,shepard1986mental}, implying that practical machine-learning evaluation must directly target such primitives through controlled tasks that disentangle perception from abstraction. To address this gap, we introduce \textsc{Sphinx}, a synthetic environment that programmatically generates visual perception and reasoning tasks centered on symmetry, transformation, and related spatial operations. Each instance includes an unambiguous ground-truth solution, enabling a precise evaluation and systematic investigation of failure modes. The framework also supports the generation of scalable datasets for reinforcement learning, paralleling synthetic reasoning environments shown to benefit text-based LLMs~\citep{reasoninggym,chen2025enigmata}.

We make the following key contributions:

\begin{enumerate}[leftmargin=*, itemsep=2pt]
    \item We introduce \textsc{Sphinx}, a synthetic environment for generating visual perception and reasoning datasets, comprising 25 tasks in five broad categories (see Figure~\ref{fig:radar-llm}). To the best of our knowledge, this represents the largest-scale synthetic environment designed for such tasks.  
    \item We construct a benchmark dataset with 2,500 questions using \textsc{Sphinx} and evaluate a range of proprietary and open-source LVLMs. We provide a comparative analysis between human performance and LVLM performance across task categories.  
    \item We apply reinforcement learning with verifiable rewards (RLVR) on a separate training set derived from \textsc{Sphinx}, achieving consistent gains in in-distribution tasks, improved generalization to held-out task families, and measurable performance improvements on multiple external benchmarks. 
\end{enumerate}

\section{\textsc{Sphinx} Design}

\textsc{\textsc{Sphinx}} is a modular framework for programmatically generating visual reasoning data with verifiable ground truth. Its central idea is to decouple the appearance from the rule structure through three composable modules: \textit{motifs}, \textit{tilings}, and \textit{tasks}, allowing each dimension to be flexibly combined or independently varied.

\subsection{Design Principles}
\begin{enumerate}[leftmargin=*, itemsep=1pt]
    \item \textbf{Factorized control of variation.} Appearance (\emph{motifs}), spatial layout (\emph{tilings}), and reasoning rules (\emph{tasks}) are separated, enabling systematic exploration across perceptual diversity, geometric structures, and rule families.
    \item \textbf{Verifiable supervision.} Each instance is paired with a deterministic checker that certifies rule satisfaction and guarantees a single correct answer; this eliminates ambiguity and supports exact evaluation.
    \item \textbf{Distribution and difficulty control.} Weighted samplers govern the mix of tasks and motifs, while difficulty knobs (e.g., steps, tile size) provide fine-grained control over problem complexity.
    \item \textbf{Standardized artifacts.} Every sample exports a composite image, natural-language prompt, ground-truth answer, distractors (if any), and rich metadata in analysis-ready formats.
\end{enumerate}

\subsection{Building Blocks}
\noindent\textbf{Motifs (rendered primitives).} A motif is a parameterized renderer $m(\theta)$ that produces an RGBA tile from attributes such as kind, size, count, angle, and stroke. Families include dots, rings, polygons, star polygons, crescents, glyphs, and other iconographic primitives. Motifs expose attribute ranges and a clamp to guarantee validity; tasks can bias selection through per‑task motif weights and request asymmetric variants to avoid trivial self-mappings in symmetry/transform problems. Example motifs are shown in Figure~\ref{fig:motifs-row} and the rest in Appendix~\ref{app:summary}.

\noindent\textbf{Geometric primitives.} Beyond motifs, \textsc{Sphinx} renders canonical geometry shapes that include circles, n-gons, angles, polylines constrained to grid edges, grids, and Venn-style region unions. These support tasks hinge on spatial relations and combinatorial structure (e.g., symmetry classification, shortest paths, connected components, region area/perimeter).

\begin{figure*}[t]
  \centering
  \setlength{\tabcolsep}{2pt}
  \begin{tabular}{@{}cccccc@{}}
    \includegraphics[width=0.2\textwidth]{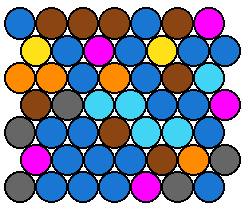} &
        \includegraphics[width=0.2\textwidth]{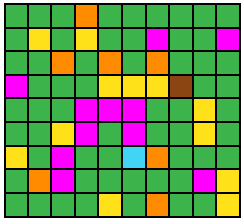} &

        \includegraphics[width=0.2\textwidth]{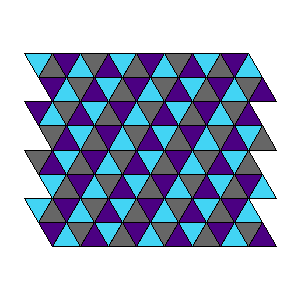}
&
    \includegraphics[width=0.2\textwidth]{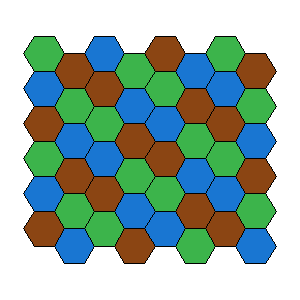} &
    \includegraphics[width=0.2\textwidth]{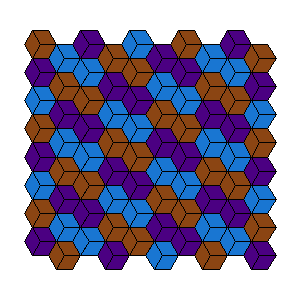} 
  \end{tabular}
  \caption{Example Tilings (from left): circles, square, triangular, hexagonal, rhombille.}
  \label{fig:tilings-row}
\end{figure*}

\noindent\textbf{Tilings (geometric canvases).} Tilings define cell layouts and adjacency (square, triangular, hexagonal, rhombille, and circle variants). Tiling specs control grid size, margins, adjacency notion, and coloring regime. Uniform schemes and palette‑driven non-uniform schemes produce structured variation. Example tilings are shown in Figure~\ref{fig:tilings-row}.

\begin{figure*}[t]
  \centering
  \setlength{\tabcolsep}{2pt} % tighter gaps between images
  \begin{tabular}{@{}cccccc@{}}
    \includegraphics[width=0.16\textwidth]{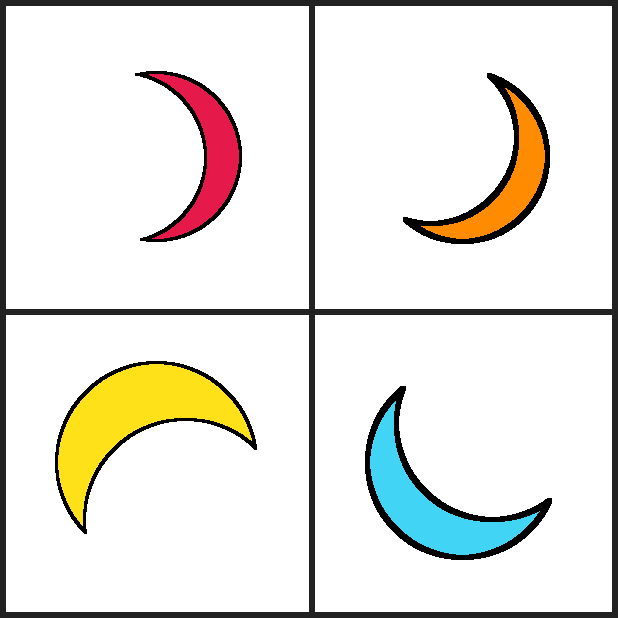} &
    \includegraphics[width=0.16\textwidth]{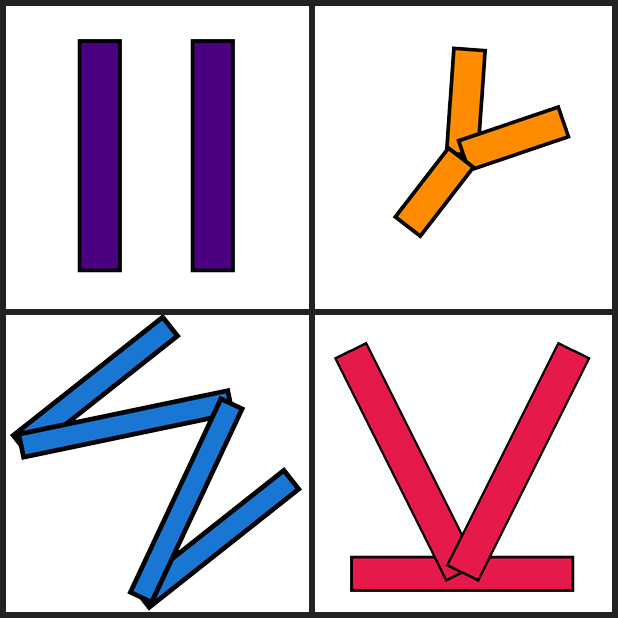} &
    \includegraphics[width=0.16\textwidth]{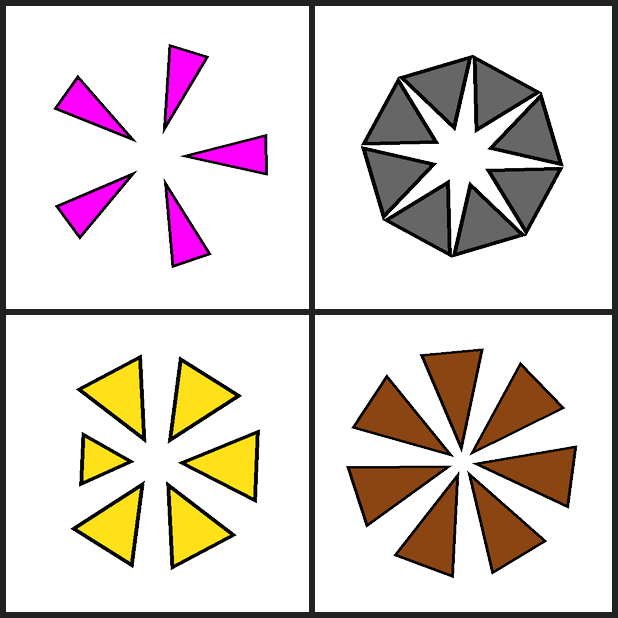} &
    \includegraphics[width=0.16\textwidth]{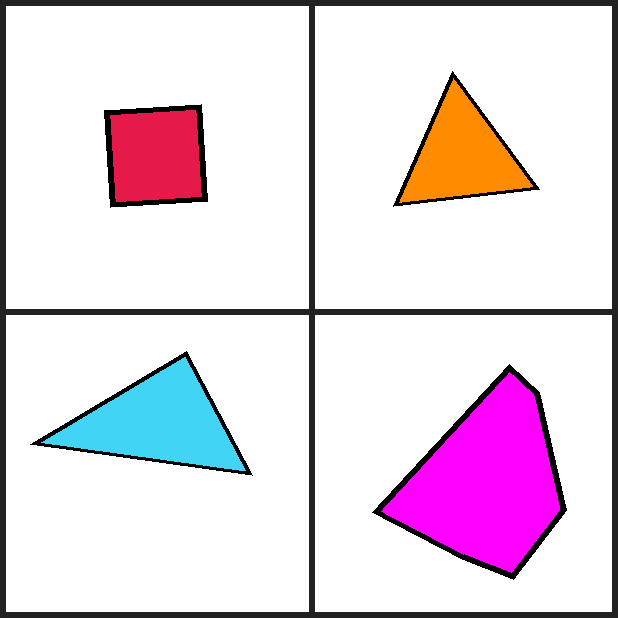} &
    \includegraphics[width=0.16\textwidth]{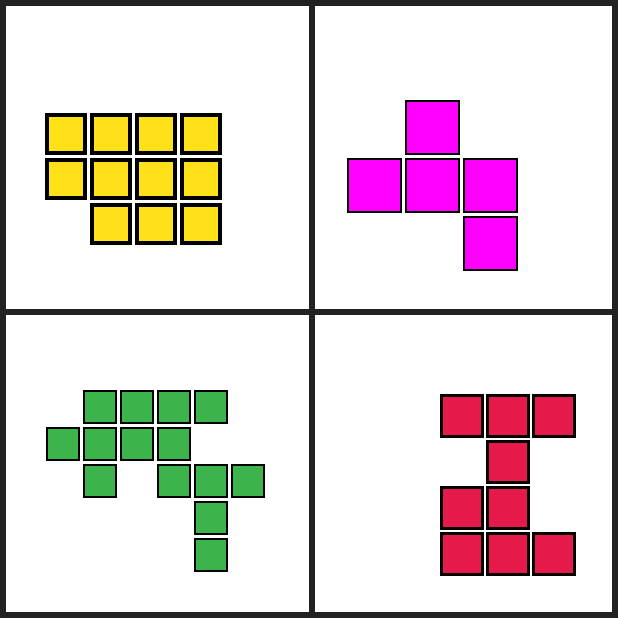} &
    \includegraphics[width=0.16\textwidth]{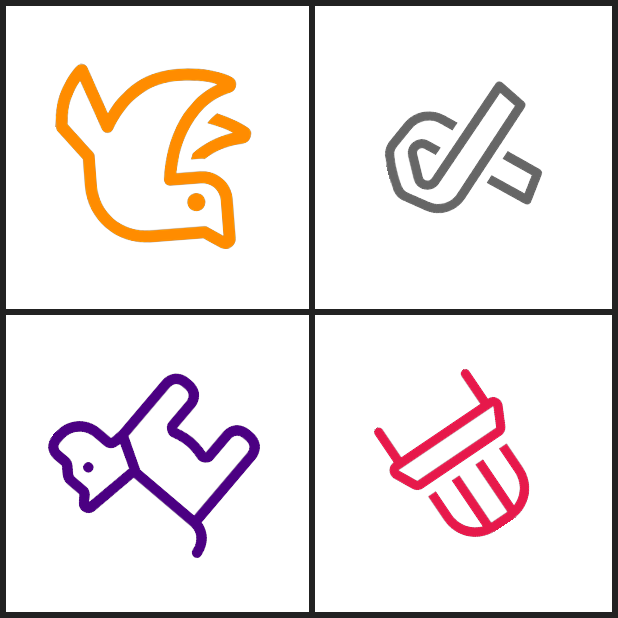} 
  \end{tabular}
  \caption{Example Motifs (from left): Crescent, Glyph, Pinwheel, Polygon, Polyomino and Icons}
  \label{fig:motifs-row}
\end{figure*}

\newcolumntype{P}[1]{>{\centering\arraybackslash}p{#1}}

% \subsection{Verification and Option Quality}
% Let $O=\{o_0,\dots,o_K\}$ denote the option set and $o_y$ the correct option. \textsc{\textsc{Sphinx}} enforces (i) rule validity via task‑specific checkers (e.g., permitted isometries, graph distances, region measures) and (ii) pairwise option uniqueness via pixel and perceptual‑hash gates:
% \[
% \mathrm{accept}(O)
% = \mathbf{1}\!\left\{
%  V_r=1 \;\land\;
%  \min_{i\ne j} d_{\mathrm{pix}}(o_i,o_j) \ge \delta_{\mathrm{pix}} \;\land\;
%  \min_{i\ne j} d_{\mathrm{hash}}(o_i,o_j) \ge \delta_{\mathrm{hash}}
% \right\}.
% \]
% Here $V_r\!\in\!\{0,1\}$ is the task‑specific verifier; $d_{\mathrm{pix}}$ is a normalized fraction of differing pixels; and $d_{\mathrm{hash}}$ is a channel‑aware perceptual‑hash Hamming distance. Global thresholds $(\delta_{\mathrm{pix}}, \delta_{\mathrm{hash}})$ are shared across tasks for consistent quality. Instances failing any gate (including ambiguous symmetries or visually indistinguishable progressions) are resampled within a fixed retry budget, yielding unambiguous supervision at scale.

% \taskcard{image}{task name}{question}{answer}
\newcommand{\taskcard}[4]{%
  \begin{minipage}[t]{\linewidth}
    \centering
    \includegraphics[width=\linewidth]{#1}\par\vspace{1pt}
    \begin{flushleft}
      \scriptsize Task: #2\par
      \scriptsize Q: #3\par
      \scriptsize A: #4
    \end{flushleft}
  \end{minipage}%
}

\begin{figure*}[t]
  \centering
  \captionsetup{skip=2pt}
  \begin{adjustbox}{max width=0.86\textwidth}
  \setlength{\tabcolsep}{3pt}
  \begin{adjustbox}{max width=\textwidth}
  \begin{tabular}{@{}P{0.32\textwidth} P{0.32\textwidth} P{0.32\textwidth}@{}}

    \taskcard{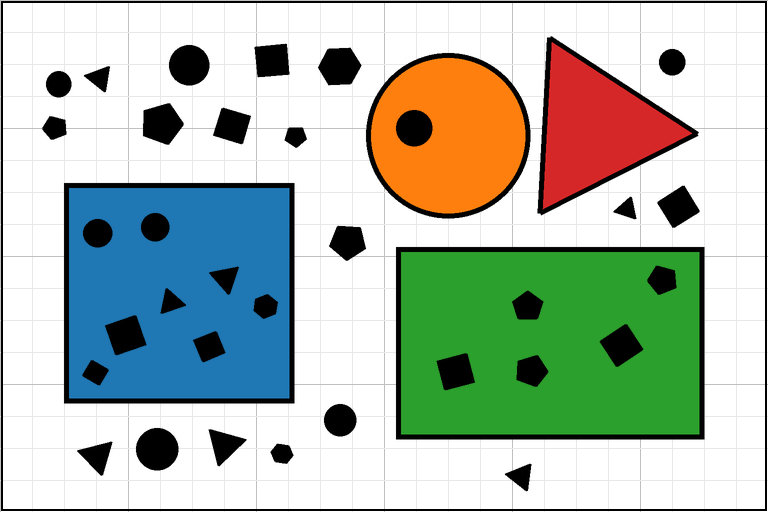}
      {Positional Count}
      {What is the total number of triangles strictly below the red triangle (below its bottommost point)?}
      {5} &
      
    \taskcard{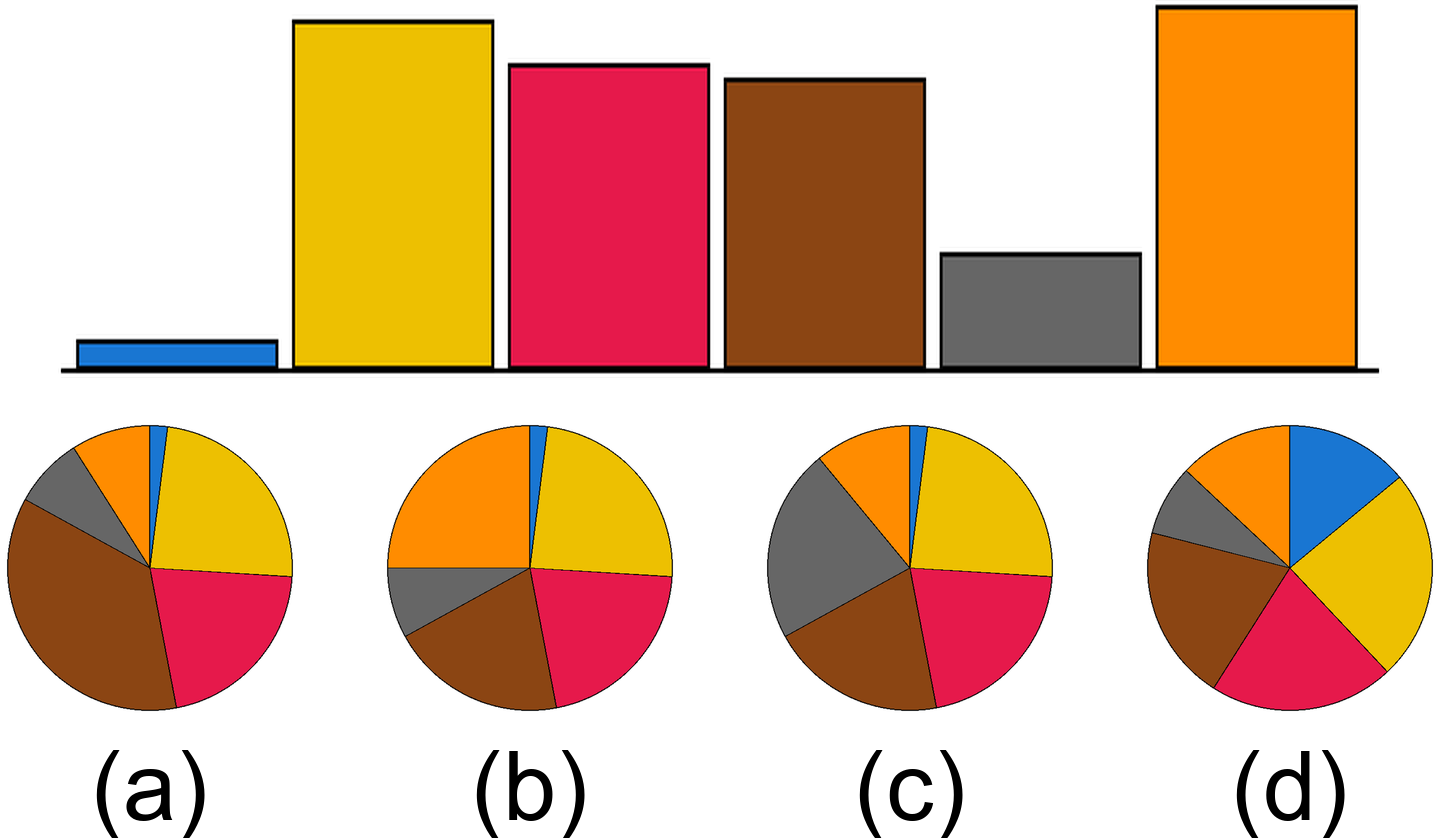}
      {Chart Comparison}
      {The top diagram is a bar chart. The bottom row contains four pie charts. Which pie chart matches the same proportional breakdown by color as the bar chart?}
      {(b)} &
      
    \taskcard{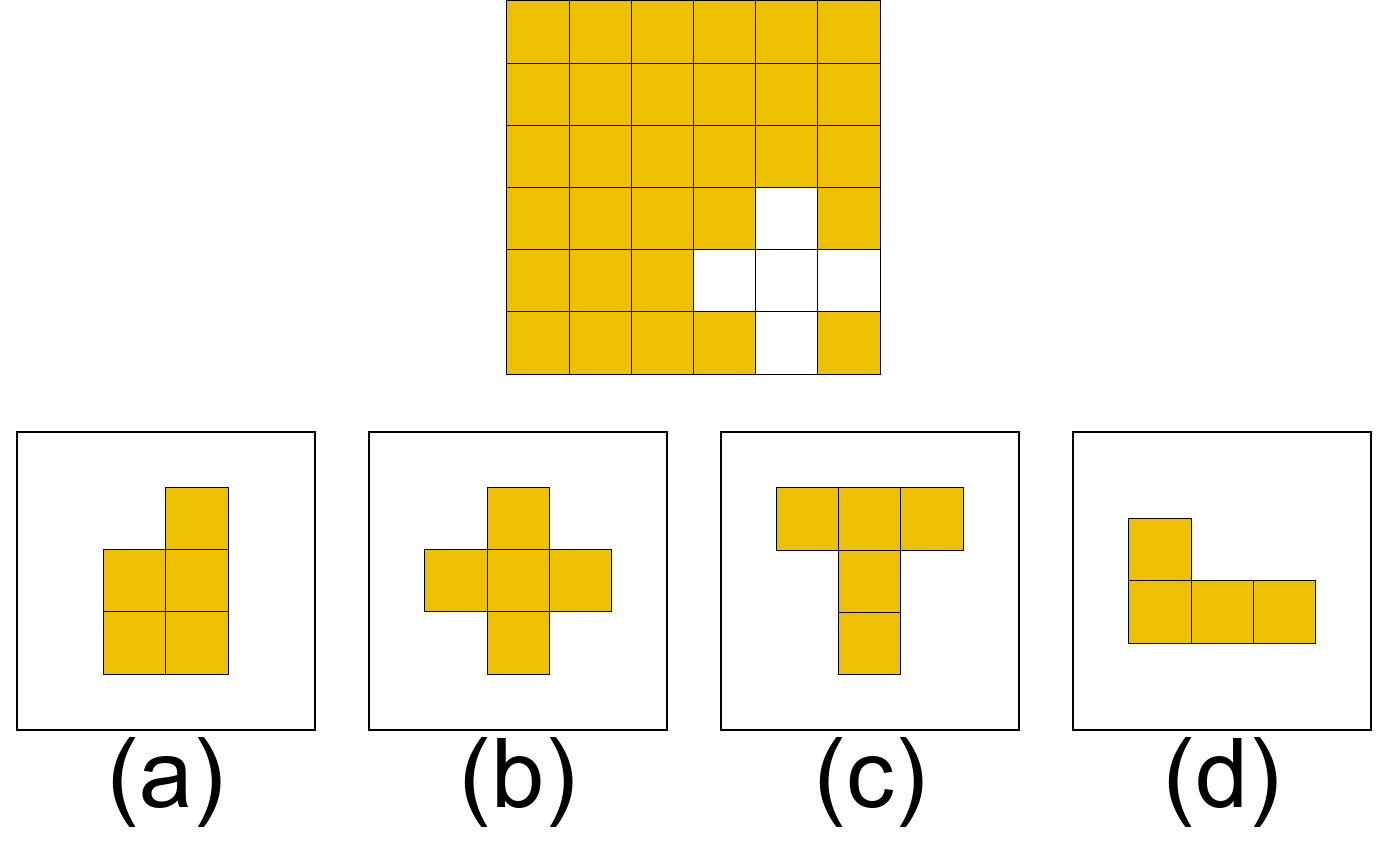}
      {Missing Tiles}
      {Look at the top image: one region is uncolored. Which option (a)–(d) provides the exact missing colors?}
      {(d)} \\[6pt]

    \taskcard{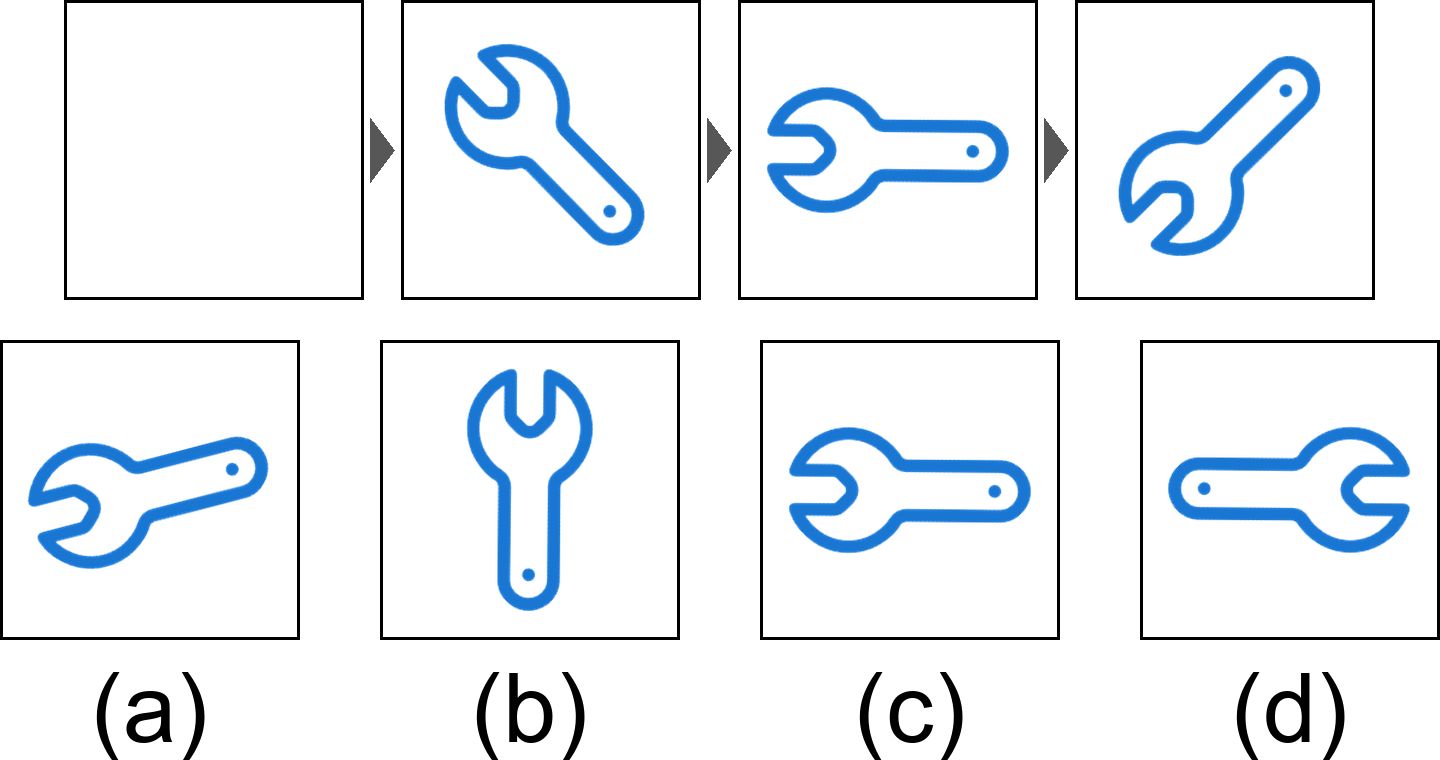}
      {Sequence Rotation}
      {In the top row, the motif rotates by a constant angle each step. Which option (a)–(d) below fills the blank?}
      {(b)} &

    \taskcard{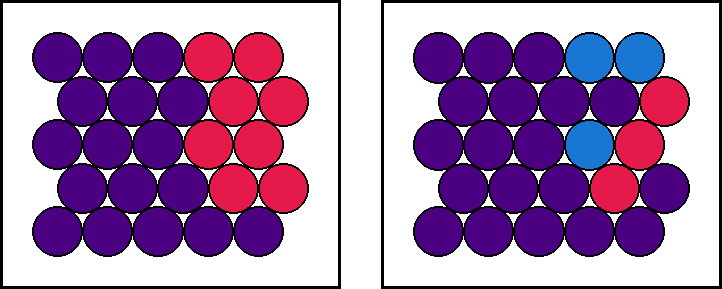}
      {Tiles Recoloring}
      {Two tiles are shown (left/right). A cell counts as different if its color differs (including filled vs. blank). How many cells differ?}
      {5} &
      
    \taskcard{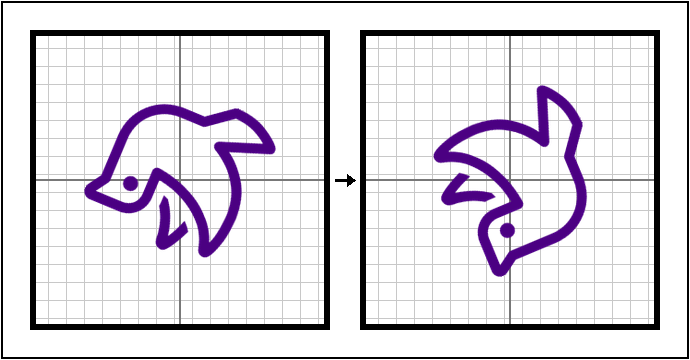}
      {Transform Pair Infer}
      {Identify the transformation applied between the left and right images. (a) pure translation (shift), (b) rotate 270° counterclockwise, (c) reflect across the anti-diagonal $(↙)$, (d) main-diagonal reflection, (e) quarter-turn CCW (90°), (f) horizontal line symmetry}
      {(c)} \\[6pt]

    % \taskcard{figures/task_illustrations/wallpaper_groups_040.png}
    %   {Wallpaper Groups}
    %   {Three options belong to the same wallpaper symmetry. Which tile in the 2×2 grid (a–d) belongs to a different group?}
    %   {(b)} &
      
    % \taskcard{figures/task_illustrations/tiles_shortest_path_010.png}
    %   {Tiles Shortest Path}
    %   {Using edge adjacency, dark gray tiles are impassable and light gray tiles are passable. Start = blue (\#1976d2), End = red (\#e6194b). What is the length of the shortest path in steps?}
    %   {10} &

    % \taskcard{figures/task_illustrations/venn_diagram_010.png}
    %   {Venn Diagram}
    %   {Determine the sum of values that exist solely within the purple rectangle(s), not in any overlapping area.}
    %   {5} \\[6pt]

  \end{tabular}
  \end{adjustbox}
  \end{adjustbox}
  \caption{\textsc{Sphinx} task illustrations}
  \label{fig:tasks-3x4}
\vspace{-0.4em}
\end{figure*}

% in preamble:
% \usepackage{enumitem}
% (we set a slightly inward left margin so numbers don't hang into the section margin)
% \setlist[enumerate,1]{leftmargin=1.6em,label=\textbf{\arabic*.},labelsep=0.5em,itemsep=2pt,topsep=2pt,align=left}

\subsection{Tasks}
A task maps one or more motif instances and/or tiled regions to a well-defined question, optionally with multiple-choice options. Each instance outputs a composite image, a natural-language prompt, and precisely one unique correct answer, along with distractors, when applicable. A key design principle in our task formulation is that questions should be visually answerable directly from the image by a human, without requiring detailed, paper-and-pencil style reasoning.  

We categorize the tasks into five broad families. Figure~\ref{fig:tasks-3x4} illustrates representative examples, with additional cases provided in Appendix~\ref{app:tasks}.

\paragraph{Geometric Reasoning.} These tasks probe spatial relations and geometric properties, such as area, perimeter, and relative size. They align with the relational and formal geometric reasoning studied in previous work~\citep{lu2021inter,zhang2024mathverse,mathvision,yuksekgonul2022and}. The tasks include:

\begin{enumerate}
  \item \textbf{Positional Count}: Count how many small shapes satisfy a specific spatial relation (inside, outside, above, below) relative to larger reference shapes.
  \item \textbf{Shape Sorting}: Sort a set of geometric shapes (polygons, ellipses, angles, lines) by area, perimeter, or angle measure.
  \item \textbf{Stack Count}: Count objects that lie strictly inside a specified sheet in a stack of overlapping shapes, where only the top shapes are fully visible.
  \item \textbf{Pie Chart}: Sort the slices of a pie chart by their visual size.
  \item \textbf{Chart Comparison}: Match a pie chart with a bar chart by visually comparing the relative proportions of their segments.
\end{enumerate}

\paragraph{Counting.} The tasks in this group focus on counting discrete elements or measuring linear features in visual scenes, similar to the counting and comparison tasks emphasized by early diagnostic benchmarks such as CLEVR~\citep{johnson2017clevr}. They include:
\begin{enumerate}[resume]
  \item \textbf{Venn Diagram}: Compute sums in different regions of a Venn diagram rendered with overlapping shapes.
  \item \textbf{Shape Counting}: Count the number of sub-shapes (e.g., rectangles, squares, triangles, parallelograms) contained within a composite figure.
  \item \textbf{Tiles Line Length}: Measure the length of a highlighted polyline in a tiling by counting edge steps.
  \item \textbf{Tiles Line Intersections}: Count the intersection points between colored polylines constrained to tile edges.
  \item \textbf{Tiles Recoloring}: Count the number of cells that differ between two colored boards, typically reflecting the size of a modified region.
\end{enumerate}

\paragraph{Symmetry \& Pattern Recognition.} These tasks require detecting symmetry, periodicity, or odd-one-out patterns. Similar phenomena are explored in visual oddity and abstract reasoning benchmarks, where participants must identify the element that violates a geometric rule or pattern~\cite{zerroug2022benchmark,wozniak2023visual}. The \textsc{Sphinx} tasks are:
\begin{enumerate}[resume]
  \item \textbf{Mirror Identification}: Classify an image according to the type of mirror symmetry present.
  \item \textbf{Symmetry Fill}: Complete a $2\times2$ grid by selecting the tile that satisfies a specified mirror symmetry.
  \item \textbf{Frieze Groups}: In a set of four \textit{frieze patterns}, identify the one that belongs to a different symmetry group.
  \item \textbf{Wallpaper Groups}: Identify the odd patch among four \textit{wallpaper patterns}.
\end{enumerate}

\paragraph{Sequence \& Transformation Reasoning.} This category encompasses tasks involving temporal sequences, rotation progressions, or transformation inference. These tasks correspond to temporal reasoning and mental-rotation challenges~\citep{wexler1998motor,xu2025visulogic}. The tasks include:
\begin{enumerate}[resume]
  \item \textbf{Transform Result Identify}: Choose the correct result when a specific transformation is applied to an image.
  \item \textbf{Transform Pair Infer}: Given two tiles, determine the transformation that maps the source to the target.
  \item \textbf{Transform Similarity Identify}: Identify which option is similar or dissimilar to a base shape under Euclidean similarity transformations (uniform scaling, rotation, reflection).
  \item \textbf{Sequence Rotation}: Predict the missing panel in a sequence of rotated motifs.
  \item \textbf{Sequence Arithmetic}: Predict the missing panel in a numeric progression of shapes.
  \item \textbf{Sequence Multi-Column Arithmetic}: Predict the next panel when each column in a grid independently undergoes its own arithmetic progression.
\end{enumerate}

\paragraph{Topological \& Graph Reasoning.} These tasks involve reasoning about connectivity, paths, and assembly on tilings or grids~\citep{wei2024rendering,qi2021pqa,jiang2024marvel}. The tasks are:
\begin{enumerate}[resume]
  \item \textbf{Tiles Geometry}: Compute areas, perimeters, number of holes, or union perimeters of colored regions on a tiling.
  \item \textbf{Tiles Connected Component}: Determine the size or number of connected components of a specified color under different adjacency notions.
  \item \textbf{Tiles Shortest Path}: Find the minimal number of steps between two tiles or determine that no path exists.
  \item \textbf{Missing Tiles}: Restore missing tiles by selecting shapes or color assignments that fit the blanked region.
  \item \textbf{Tiles Composition}: Decompose a connected region into smaller pieces or compose multiple pieces into a single connected shape.
\end{enumerate}

\begin{table*}[t]
\centering
\caption{Performance comparison of human, closed-source, and open-source LVLMs across multiple reasoning categories.}
\label{tab:llm-performance}
\resizebox{0.8\textwidth}{!}{%
\begin{tabular}{lcccccc}
\toprule
\textbf{Models} &
\multicolumn{1}{c}{\textbf{Geometric}} &
\multicolumn{1}{c}{\textbf{Counting}} &
\multicolumn{1}{c}{\textbf{Symmetry \&}} &
\multicolumn{1}{c}{\textbf{Sequence \&}} &
\multicolumn{1}{c}{\textbf{Topological \&}} &
\multicolumn{1}{c}{\textbf{Average}} \\
 & 
 & 
 & \textbf{Pattern Recognition} 
 & \textbf{Transformation} 
 & \textbf{Graph Reasoning}
 & \textbf{(25 Tasks)} \\
\midrule
Human           & 87.4 & 70.4 & 69.9 & 82.9 & 65.8 & 75.8 \\
\midrule
GPT-5           & \textbf{74.0} & 36.6 & \textbf{50.0} & \textbf{51.5} & \textbf{43.0} & \textbf{51.1} \\
GPT-5 Mini      & 66.0 & 36.2 & 49.5 & 45.8 & 38.8 & 47.1 \\
GPT-5 Nano      & 44.4 & 24.8 & 40.8 & 31.8 & 25.8 & 33.2 \\
Gemini3-Flash   & 30.0 & 30.6 & 26.0 & 19.0 & 13.8 & 23.6 \\
Claude4.5-Opus  & 57.8 & 35.4 & 44.3 & 32.2 & 32.4 & 39.9 \\
Claude4.5-Sonnet & 68.0 & \textbf{40.2} & 34.3 & 37.3 & 42.6 & 44.6 \\
\midrule
InternVL3-8B    & 33.4 & 14.2 & 30.0 & 23.8 & 20.0 & 24.0 \\
InternVL3-38B   & 45.4 & 18.8 & 33.8 & 26.3 & 26.0 & 29.8 \\
Llama-3.2-11B   & 22.4 &  9.2 & 21.2 & 23.7 & 17.2 & 18.8 \\
Qwen2.5-VL-3B   & 31.0 &  8.2 & 22.8 & 22.2 & 17.0 & 20.2 \\
Qwen2.5-VL-7B   & 38.2 & 14.6 & 28.7 & 23.5 & 19.6 & 24.7 \\
Qwen2.5-VL-32B  & 52.4 & 21.8 & 33.5 & 26.7 & 28.2 & 32.2 \\
Qwen3-VL-4B     & 42.4 & 18.6 & 36.2 & 29.8 & 21.2 & 29.4 \\
Qwen3-VL-8B     & 44.8 & 20.0 & 34.2 & 33.0 & 23.8 & 31.1 \\
Qwen3-VL-30B    & 47.4 & 26.4 & 34.0 & 28.8 & 23.6 & 31.8 \\
Qwen3-VL-235B   & 66.8 & 29.2 & 43.6 & 35.2 & 33.6 & 41.3 \\
Molmo2-4B       & 30.6 & 13.4 & 24.0 & 21.2 & 22.4 & 22.2 \\
Molmo2-8B       & 17.8 & 14.0 & 23.3 & 22.8 & 20.4 & 19.6 \\
Molmo2-O-7B     & 15.8 & 12.8 & 28.5 & 21.3 & 16.4 & 18.7 \\

\bottomrule
\end{tabular}%
}

\end{table*}

\section{Benchmark}

% \begin{figure}[t]
%   \centering
%   \includegraphics[width=0.4\textwidth]{figures/qualitative_analysis/gpt5_vs_human_scatterplot.png}
%   \caption{Comparison of human and GPT-5 accuracy.}
%   \label{fig:humanvsgpt}
% \end{figure}

We curated the \textsc{Sphinx} benchmark to include 2,500 questions, with 100 instances per task. We evaluated six proprietary LVLMs: three GPT-5 variants (regular, mini, and nano) using their default reasoning settings~\cite{openai2025introducinggpt5}, Gemini3-Flash, Claude4.5-Opus, and Claude4.5-Sonnet. In addition, we evaluated 13 open-source vision-language models, including the Qwen2.5-VL family~\cite{bai2025qwen2}, Llama 3.2~\cite{meta2024llama}, InternVL3~\cite{zhu2025internvl3}, Molmo2~\cite{clark2026molmo2}, and the Qwen3-VL family~\cite{qwen3_vl_2025}, covering parameter scales from 3B to 235B. The evaluation prompt is provided in the Appendix~\ref{app:evalprompt}. We use the \texttt{mathruler} library~\cite{mathruler} for answer extraction and verification, with GPT-5-mini serving as a fallback when extraction is ambiguous. For open-source models, we use the VLMEvalKit framework for inference~\cite{duan2024vlmevalkit}.

Results are summarized in Table~\ref{tab:llm-performance}. Overall, GPT-5 achieves the best performance with an average accuracy of 51.1\% on all tasks, although it still falls short of human accuracy by 24.7\%. GPT-5 also achieves the strongest results on geometric reasoning, symmetry and pattern recognition, sequence and transformation, and topological and graph reasoning, while Claude4.5-Sonnet attains the best counting score at 40.2\%. GPT-5 Mini remains close to the full model, with a 4.0\% drop in overall accuracy. Among open-source models, Qwen3-VL-235B achieves the highest accuracy at 41.3\%, outperforming Claude4.5-Opus and substantially surpassing the smaller open-weight baselines.

% Performance varies substantially between task categories. The most significant gap between models and human evaluators occurs in \emph{Sequence} and \emph{Transformation} tasks, where GPT-5 lags human accuracy by 39.2\%. In contrast, the gap is less pronounced on \emph{Geometric Reasoning} and \emph{Tiles-based} tasks that emphasize topological or graph-structured reasoning. Figure~\ref{fig:humanvsgpt} shows the performance in all 25 tasks, comparing GPT-5 with human precision. Although there is an overall positive correlation, several tasks exhibit substantial disparities, which we analyze in detail below.

\section{Analysis }

\subsection{Human Evaluation}
\label{sec:human-eval}

\begin{figure}[t]
  \centering
  \captionsetup{skip=2pt}
  \includegraphics[width=0.74\linewidth]{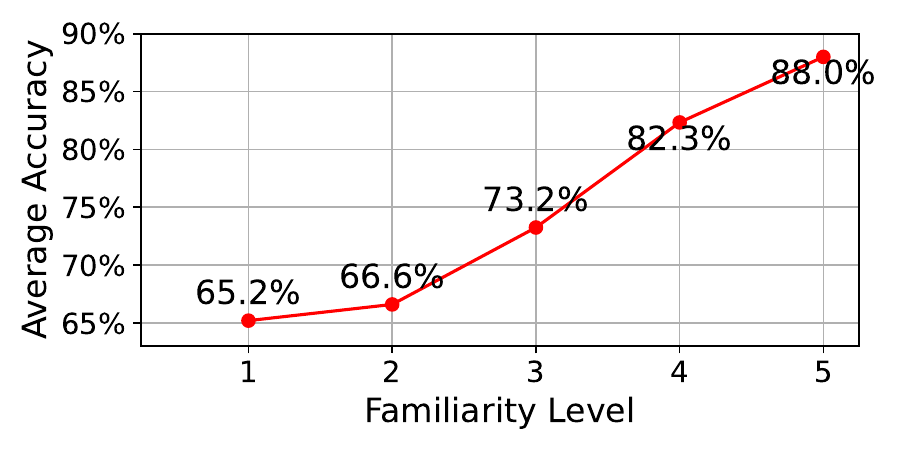}
\caption{Familiarity vs Accuracy - Human Evaluators}
  \label{fig:familiarity-vs-accuracy}
\vspace{-0.5em}
\end{figure}

In our evaluation, participants received no task-specific training or tutorial on visual reasoning, and many reported limited prior exposure to symmetry, pattern induction, or spatial composition tasks. In a post-study questionnaire, participants rated their familiarity on a scale of 1 to 5 (where 5 indicates the highest level of familiarity); as shown in Fig.~\ref{fig:familiarity-vs-accuracy}, familiarity strongly correlates with accuracy, and inexperienced participants performed markedly worse.

\begin{figure}[t]
  \centering
  \captionsetup{skip=2pt}
  \includegraphics[width=0.74\linewidth]{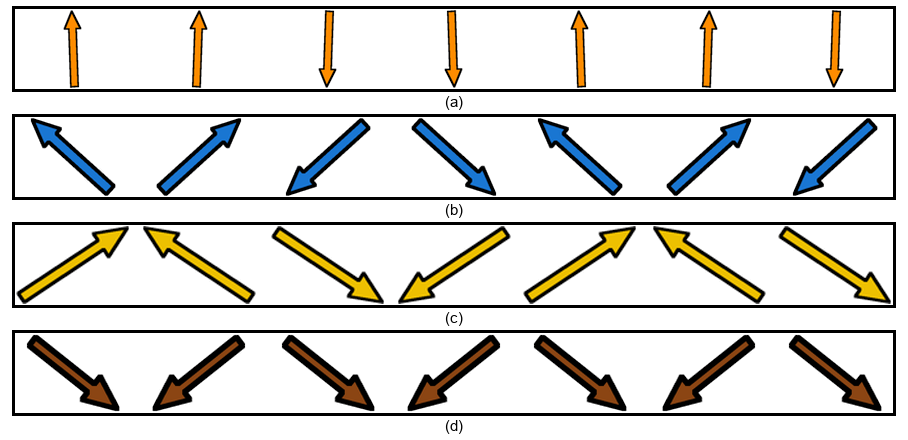}
\caption{Example Frieze group task. Odd one out is (d).}
  \label{fig:example-frieze}
\vspace{-0.5em}
\end{figure}

The lowest human accuracy was observed on the \textit{Frieze Group Odd-One-Out} task (48.4\%), where subtle differences between frieze symmetry classes often led participants to rely on superficial visual cues. For instance, in Fig.~\ref{fig:example-frieze}, option (\textit{a}) appears visually distinct, yet the true odd-one-out is (\textit{d}), whose symmetry class (p1m1) differs from the others (p2mg), illustrating errors driven by first-glance impressions rather than symmetry reasoning.

\begin{figure}[t]
  \centering
  \captionsetup{skip=2pt}
  \includegraphics[width=0.60\linewidth]{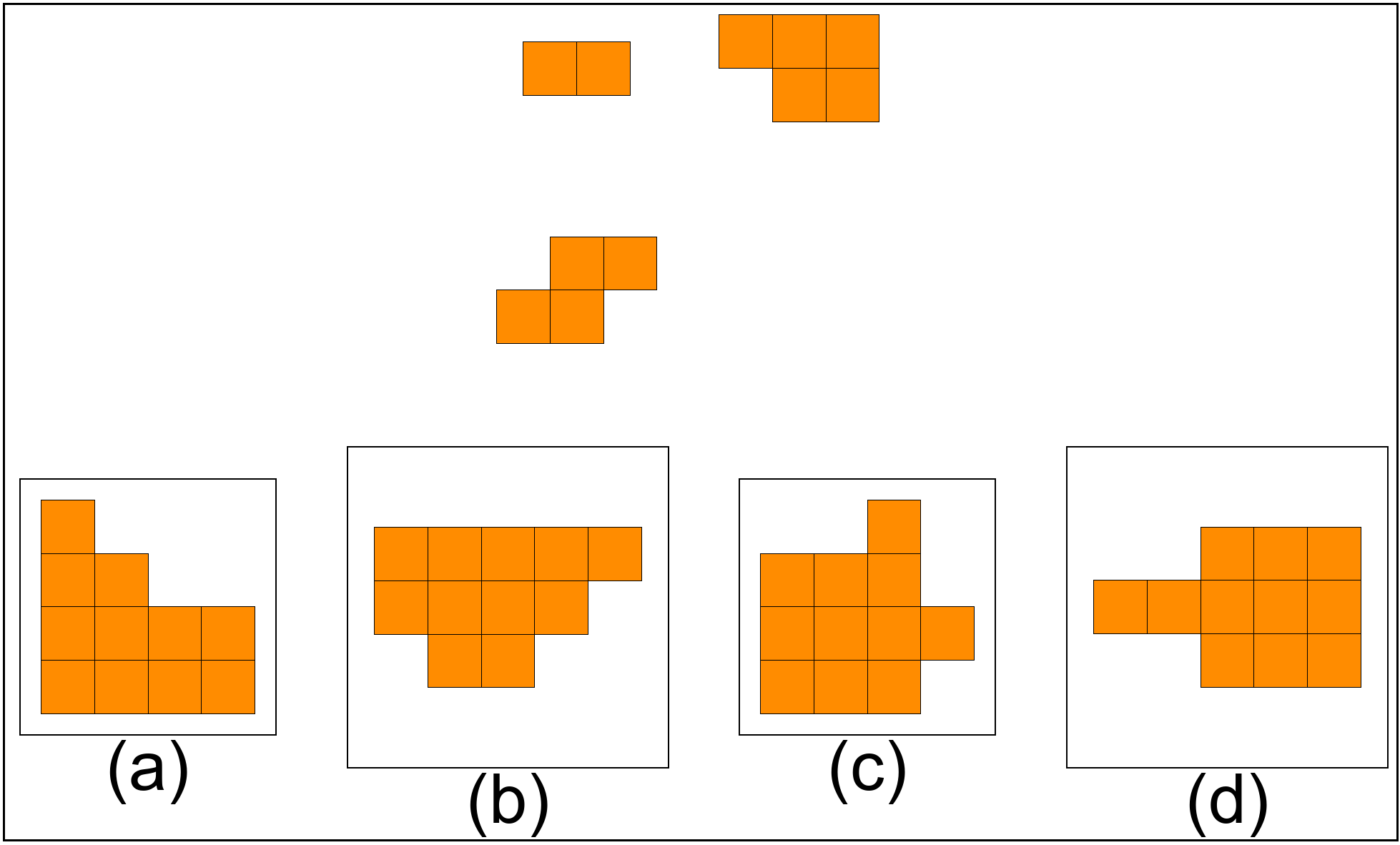}
\caption{Example Tiles Composition task. Correct answer:(b)}
  \label{fig:example-tiles-composition}
\vspace{-0.5em}
\end{figure}

The next most difficult task was \textit{Tiles Composition}, which involves mentally rotating and assembling multiple tile pieces (see Fig.~\ref{fig:example-tiles-composition}). Since trivial distractors are not used (e.g. mismatched tile counts), participants must consider only valid geometric combinations. This task had low accuracy and the longest average response time (Appendix~\ref{app:humaneval}), indicating a genuine cognitive load rather than inattention.

Other tasks with lower performance include those requiring mental arithmetic (\textit{Venn Diagram}, \textit{Shape Counting}) and grid-based enumeration (\textit{Tiles Geometry}, \textit{Tiles Connected Component}, \textit{Shortest Path}). Since participants completed the study without pencil-and-paper aids, their scores on these counting-heavy tasks likely underestimate true ability.

Overall, the human study shows that (i) prior familiarity with visual reasoning puzzles strongly influences performance, and (ii) tasks involving mental rotation, multi-step spatial reasoning, or fine-grained symmetry classification are challenging even for humans, underscoring the diagnostic value of the \textsc{Sphinx} benchmark.

\subsection{GPT-5 vs. Humans}

% \begin{figure}
%     \centering % Center the row of figures
    
%     \begin{minipage}{0.49\linewidth}
%         \centering
%         \includegraphics[width=\linewidth]{figures/qualitative_analysis/human_vs_gpt5_wins.pdf}
%         \captionof{figure}{Tasks where humans exceed GPT-5.}
%         \label{fig:human_win}
%     \end{minipage}
%     \hfill % Adds flexible space between the two figures
%     \begin{minipage}{0.49\linewidth}
%         \centering
%         \includegraphics[width=\linewidth]{figures/qualitative_analysis/gpt5_vs_human_wins.pdf}
%         \captionof{figure}{Tasks where GPT-5 exceeds or is close to human performance.}
%         \label{fig:gpt_win}
%     \end{minipage}

% \end{figure}

\begin{figure}[t]
  \centering
  \captionsetup{skip=2pt}
  \includegraphics[width=0.82\linewidth]{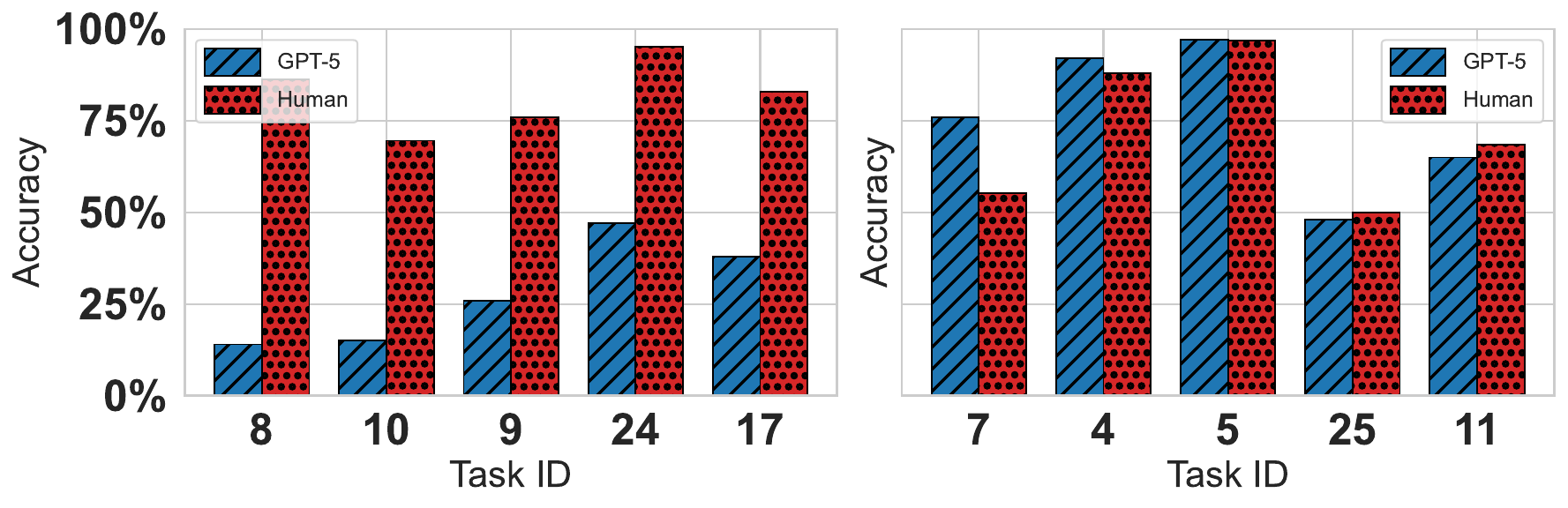}
\caption{Per–task accuracy comparison between GPT-5 and humans. 
\textbf{Left:} five tasks where humans most strongly outperform GPT-5. 
\textbf{Right:} five tasks where GPT-5 is closest to or exceeds human performance. 
\textbf{Task IDs:} 
8 (Tiles Line Length),  
10 (Tiles Recoloring),    
9 (Tiles Line Intersections),
24 (Missing Tiles),
17 (Transform Similarity Identify),  
7 (Shape Counting),  
4 (Pie Chart),  
5 (Chart Comparison),  
25 (Tiles Composition),  
11 (Mirror Identification).}

  \label{human-gpt-barplot}
\vspace{-0.5em}
\end{figure}

In Figure~\ref{human-gpt-barplot}, we show the five tasks where humans most strongly outperform GPT-5 (left) and the five tasks where GPT-5 is closest to or exceeds human performance (right). GPT-5’s largest deficits occur on tile-based tasks, including \emph{Tiles Line Length}, \emph{Tiles Recoloring}, \emph{Tiles Line Intersections}, and \emph{Missing Tiles}, all of which humans find relatively intuitive. Its remaining major failure arises in \emph{Transform Similarity Identify}. In contrast, GPT-5 outperforms humans on the \emph{Shape Counting} task and performs comparably on the two chart-based tasks. Unlike other tile-based tasks, it also performs comparably to humans on \emph{Tiles Composition}, where human performance is relatively poor. We provide a more in-depth analysis of GPT-5’s performance in Appendix~\ref{app:gpt5-analysis}.

\subsection{GPT-5 vs. GPT-5 Mini}

\begin{figure}[t]
  \centering
  \captionsetup{skip=2pt}
  \includegraphics[width=0.82\linewidth, height=0.82\textheight, keepaspectratio]{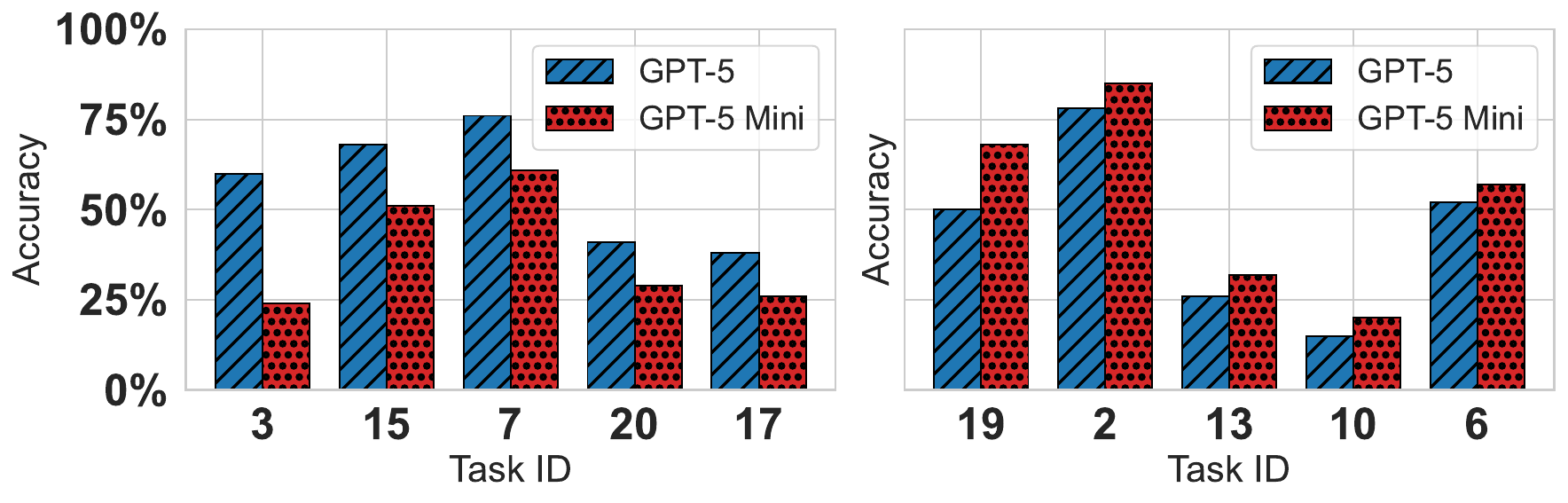}

 \caption{Per–task accuracy comparison between GPT-5 and GPT-5 Mini. 
\textbf{Left:} five tasks where GPT-5 outperforms GPT-5 Mini the most. 
\textbf{Right:} five tasks where GPT-5 Mini leads. 
\textbf{Task IDs:} 
3 (Stack Count),  
15 (Transform Result Identify),  
7 (Shape Counting),  
20 (Sequence Multi-Column Arithmetic),  
17 (Transform Similarity Identify),  
19 (Sequence Arithmetic),  
2 (Shape Sorting),  
13 (Frieze Groups)
,  
10 (Tiles Recoloring),  
6 (Venn Diagram).}

  \label{gpt-mini-barplot}
\vspace{-0.5em}
\end{figure}

% \begin{figure}
%     \centering % Center the whole row of figures
    
%     \begin{minipage}{0.49\linewidth}
%         \centering
%         \includegraphics[width=\linewidth]{figures/qualitative_analysis/gpt5_vs_mini_wins.pdf}
%         \captionof{figure}{Tasks where GPT-5 exceeds GPT-5 mini}
%         \label{fig:gpt5_w2}
%     \end{minipage}
%     \hfill % Adds flexible space between the two minipages
%     \begin{minipage}{0.49\linewidth}
%         \centering
%         \includegraphics[width=\linewidth]{figures/qualitative_analysis/mini_vs_gpt5_wins.pdf}
%         \captionof{figure}{Tasks where GPT-5 mini exceeds GPT-5.}
%         \label{fig:mini_wins}
%     \end{minipage}

% \end{figure}

Figure~\ref{gpt-mini-barplot} compares the five tasks where GPT-5 most strongly outperforms GPT-5 Mini (left) with the five tasks where GPT-5 Mini has an advantage (right). GPT-5 performs best on tasks with explicit procedural structure, such as counting or applying a specified transformation, where it can leverage step-by-step reasoning to follow instructions faithfully. It also shows a more apparent ability to identify the global structure of the image than GPT-5 Mini, even though both models exhibit weaknesses in low-level visual perception. A more detailed comparison between the two models is provided in Appendix~\ref{app:gpt5vsmini}.

\subsection{Task Complexity vs. Model Accuracy}

\begin{figure}[t]
  \centering
  \captionsetup{skip=2pt}
  \includegraphics[width=0.78\linewidth]{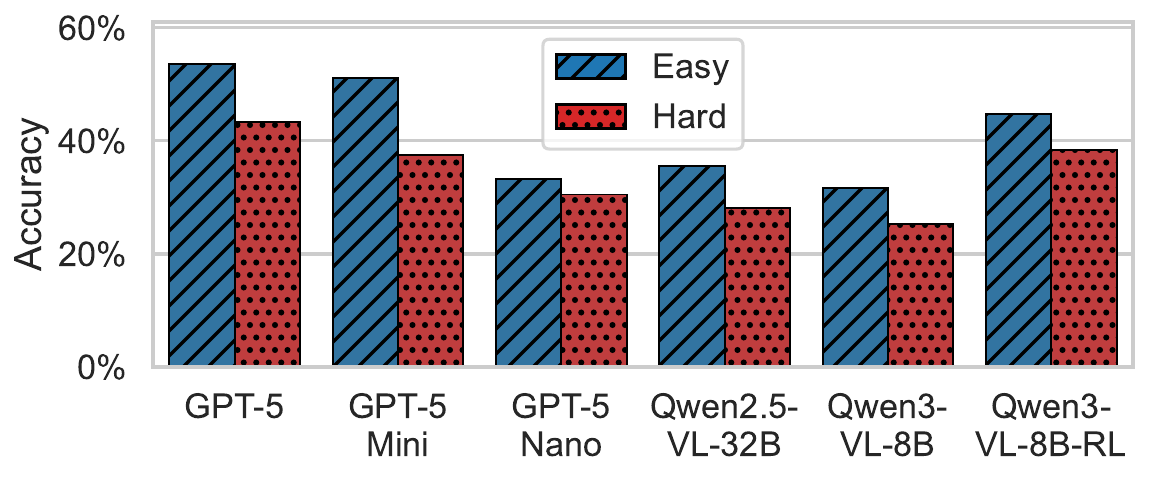}
  \caption{Accuracy by complexity for different LVLMs}
  \label{fig:complexity-acc}
\vspace{-0.5em}
\end{figure}

Many \textsc{Sphinx} tasks provide explicit knobs to adjust difficulty. In particular, motif–counting and tile–based tasks admit natural controls (e.g., number of motifs, grid density, tile size), from which we can construct a single monotone complexity axis. In contrast, for tasks such as symmetry detection, the apparent difficulty depends strongly on motif geometry and arrangement, making a scalar complexity measure ambiguous. We exclude those tasks and retain 17 tasks for which complexity is well defined.

For retained tasks, we introduce a normalized complexity parameter \(c \in [0,1]\) that maps the underlying task settings to a common scale. Implementation details are provided in the Appendix~\ref{app:tasks}. We bin the instances with \(c<0.75\) as \emph{Easy} and with \(c\ge 0.75\) as \emph{Hard}. Figure~\ref{fig:complexity-acc} reports model accuracy at these two levels. Accuracy decreases consistently as complexity increases in all evaluated models, including the RLVR–trained model. For these tasks, \textsc{Sphinx} can scale complexity even further, for example, by increasing the maximum tile size in tile-based tasks. This extensibility enables the benchmark to evolve along with future model capabilities.

\begin{table*}[t]
\centering
\captionsetup{skip=2pt}
\caption{Performance on \textsc{Sphinx} IID and OOD tasks. All values are accuracies (\%). RL indicates models trained with RLVR on the \textsc{Sphinx} train set; subscripts show absolute gains over the corresponding base model.}
\label{tab:sphinx_iid_ood}
\scriptsize
\setlength{\tabcolsep}{2.7pt}
\renewcommand{\arraystretch}{0.95}
\resizebox{0.89\textwidth}{!}{%
\begin{tabular}{l |c | ccccc c}
\toprule
\textbf{Model} &
\multicolumn{1}{c|}{\textbf{Sphinx IID}} &
\multicolumn{6}{c}{\textbf{Sphinx OOD}} \\
\cmidrule(lr){3-8}
 & \multicolumn{1}{c|}{} &
\multicolumn{1}{c}{\textbf{Positional}} &
\multicolumn{1}{c}{\textbf{Tiles}} &
\multicolumn{1}{c}{\textbf{Wallpaper}} &
\multicolumn{1}{c}{\textbf{Sequence}} &
\multicolumn{1}{c}{\textbf{Tiles}} &
\multicolumn{1}{c}{\textbf{Avg.}} \\
 & \multicolumn{1}{c|}{} &
\multicolumn{1}{c}{\textbf{Count}} &
\multicolumn{1}{c}{\textbf{Recoloring}} &
\multicolumn{1}{c}{\textbf{Groups}} &
\multicolumn{1}{c}{\textbf{Multi-Column Arith.}} &
\multicolumn{1}{c}{\textbf{Composition}} &
\\
\midrule
Qwen2.5-VL-7B &
25.5 & 14.0 &  8.0 & 28.0 & 27.0 & 31.0 & 21.6 \\
Qwen2.5-VL-7B-RL &
42.6\textsubscript{\scriptsize(+17.1)} &
30.0\textsubscript{\scriptsize(+16.0)} &
10.0\textsubscript{\scriptsize(+2.0)}  &
33.0\textsubscript{\scriptsize(+5.0)}  &
29.0\textsubscript{\scriptsize(+2.0)}  &
32.0\textsubscript{\scriptsize(+1.0)}  &
26.8\textsubscript{\scriptsize(+5.2)}  \\
\midrule
Qwen2.5-VL-3B &
20.6 & 15.0 &  9.0 & 23.0 & 25.0 & 21.0 & 18.6 \\
Qwen2.5-VL-3B-RL &
32.0\textsubscript{\scriptsize(+11.4)} &
29.0\textsubscript{\scriptsize(+14.0)} &
 3.0\textsubscript{\scriptsize(-6.0)}  &
27.0\textsubscript{\scriptsize(+4.0)} &
34.0\textsubscript{\scriptsize(+9.0)} &
20.0\textsubscript{\scriptsize(-1.0)}  &
22.6\textsubscript{\scriptsize(+4.0)}  \\
\midrule
Qwen3-VL-8B &
32.3 & 23.0 & 15.0 & 39.0 & 26.0 & 29.0 & 26.4 \\
Qwen3-VL-8B-RL &
44.6\textsubscript{\scriptsize(+12.3)} &
33.0\textsubscript{\scriptsize(+10.0)} &
12.0\textsubscript{\scriptsize(-3.0)}  &
45.0\textsubscript{\scriptsize(+6.0)}  &
25.0\textsubscript{\scriptsize(-1.0)}  &
37.0\textsubscript{\scriptsize(+8.0)}  &
30.4\textsubscript{\scriptsize(+4.0)}  \\
\midrule
Qwen3-VL-4B &
30.6 & 31.0 & 10.0 & 38.0 & 23.0 & 22.0 & 24.8 \\
Qwen3-VL-4B-RL &
44.0\textsubscript{\scriptsize(+13.4)} &
37.0\textsubscript{\scriptsize(+6.0)} &
10.0\textsubscript{\scriptsize(+0.0)}  &
50.0\textsubscript{\scriptsize(+12.0)} &
32.0\textsubscript{\scriptsize(+9.0)} &
39.0\textsubscript{\scriptsize(+17.0)} &
33.6\textsubscript{\scriptsize(+8.8)} \\
\bottomrule
\end{tabular}%
}
\normalsize
\end{table*}

\begin{table*}[t]
\centering
\captionsetup{skip=2pt}
\caption{Results on multimodal math and reasoning benchmarks. RL indicates model trained with RLVR on \textsc{Sphinx} train set. Subscripts show absolute gains over the corresponding base model.}
\label{tab:multimodal_math_benchmarks}
\scriptsize
\setlength{\tabcolsep}{2.7pt}
\renewcommand{\arraystretch}{0.95}
\resizebox{0.89\textwidth}{!}{%
\begin{tabular}{lccccccccc}
\toprule
Model & MathVista & MathVision & MathVerse & LogicVista & BLINK & MMT & MMVP & VStarBench & Avg. \\
\midrule
Qwen3-VL-8B      & 76.0 & 45.8 & 57.6 & 53.7 & 68.2 & 67.6 & 80.3 & 81.7 & 66.4 \\
Qwen3-VL-8B-RL   & 77.0\textsubscript{\scriptsize(+1.0)} & 50.2\textsubscript{\scriptsize(+4.4)} & 61.7\textsubscript{\scriptsize(+4.1)} & 60.6\textsubscript{\scriptsize(+6.9)} & 68.9\textsubscript{\scriptsize(+0.7)} & 68.2\textsubscript{\scriptsize(+0.6)} & 80.3\textsubscript{\scriptsize(+0.0)} & 83.3\textsubscript{\scriptsize(+1.6)} & 68.8\textsubscript{\scriptsize(+2.4)} \\
\midrule
Qwen3-VL-4B      & 73.0 & 42.7 & 57.1 & 51.0 & 65.0 & 65.0 & 80.0 & 79.1 & 64.1 \\
Qwen3-VL-4B-RL   & 74.4\textsubscript{\scriptsize(+1.4)} & 47.1\textsubscript{\scriptsize(+4.4)} & 62.1\textsubscript{\scriptsize(+4.9)} & 59.3\textsubscript{\scriptsize(+8.3)} & 66.8\textsubscript{\scriptsize(+1.8)} & 66.9\textsubscript{\scriptsize(+1.9)} & 80.3\textsubscript{\scriptsize(+0.3)} & 78.5\textsubscript{\scriptsize(-0.6)} & 66.9\textsubscript{\scriptsize(+2.8)} \\
\midrule
Qwen2.5-VL-7B    & 68.8 & 25.1 & 41.0 & 45.2 & 55.3 & 62.1 & 77.0 & 76.4 & 56.4 \\
Qwen2.5-VL-7B-RL & 69.7\textsubscript{\scriptsize(+0.9)} & 26.8\textsubscript{\scriptsize(+1.7)} & 41.2\textsubscript{\scriptsize(+0.3)} & 45.4\textsubscript{\scriptsize(+0.2)} & 57.6\textsubscript{\scriptsize(+2.3)} & 62.8\textsubscript{\scriptsize(+0.7)} & 78.3\textsubscript{\scriptsize(+1.3)} & 78.5\textsubscript{\scriptsize(+2.1)} & 57.5\textsubscript{\scriptsize(+1.2)} \\
\midrule
Qwen2.5-VL-3B    & 62.2 & 22.2 & 29.8 & 39.4 & 49.0 & 60.8 & 71.3 & 75.4 & 51.3 \\
Qwen2.5-VL-3B-RL & 62.6\textsubscript{\scriptsize(+0.4)} & 21.9\textsubscript{\scriptsize(-0.3)} & 34.1\textsubscript{\scriptsize(+4.3)} & 40.0\textsubscript{\scriptsize(+0.6)} & 48.7\textsubscript{\scriptsize(-0.3)} & 61.6\textsubscript{\scriptsize(+0.8)} & 71.3\textsubscript{\scriptsize(+0.0)} & 74.9\textsubscript{\scriptsize(-0.5)} & 51.9\textsubscript{\scriptsize(+0.6)} \\
\bottomrule
\end{tabular}%
}
\normalsize
\end{table*}

\section{Reinforcement Learning}

% We perform reinforcement learning with verifiable rewards on synthetic datasets generated by \textsc{Sphinx}.

\paragraph{Data Split.}  
We designate 20 tasks as in-distribution and withhold 5 for testing to assess generalization to unseen tasks. The withheld tasks are \emph{Positional Count}, \emph{Tiles Recoloring}, \emph{Wallpaper Groups}, \emph{Sequence Multi-Column Arithmetic}, and \emph{Tiles Composition}. We generate \(100{,}000\) synthetic samples using a fixed random seed. From these, we select \(1{,}600\) samples per in-distribution task (a total of \(32{,}000\) training samples) chosen to maximize the minimum semantic similarity to evaluation samples from the same task. Semantic similarity is computed using the \texttt{sentence-transformers} library~\citep{reimers2019sentence}, employing the CLIP ViT-B/32 embedding model.

\paragraph{Model Training.}
We train our models using GRPO (Group Relative Policy Optimization), which avoids a separate critic network by ranking multiple sampled outputs per prompt and using their relative ordering as the learning signal~\citep{shao2024deepseekmath}. 
We fine-tune four base models—Qwen2.5-VL-7B, Qwen2.5-VL-3B~\citep{bai2025qwen2}, and Qwen3-VL-4B, Qwen3-VL-8B~\citep{qwen3_vl_2025} using the EasyR1 framework~\citep{zheng2025easyr1}. 
Training is carried out for 500 iterations with the following hyperparameters: 
\(\mathrm{kl\_coef} = 1.0 \times 10^{-2}\), maximum response length \(= 2048\), optimizer \texttt{adamw} (learning rate \(1.0 \times 10^{-6}\), weight decay \(1.0 \times 10^{-2}\)), rollout count \(n = 5\), sampling temperature \(= 1.0\) and batch size \(= 128\).

We adopt the default prompting and reward configuration provided by EasyR1. 
The reward function consists of binary correctness (1 for a correct answer, 0 otherwise) and an auxiliary formatting reward. 
The final reward used for optimization is
\[
\text{reward} 
= \lambda \cdot \text{format\_reward} 
+ (1 - \lambda) \cdot \text{correctness\_reward},
\]
with \(\lambda = 0.1\). 
Correctness is verified against ground truth using the \texttt{mathruler} library~\citep{mathruler}.

% \begin{table}[h]
% \centering
% \caption{Performance of Qwen2.5 models with and without RLVR across benchmarks. Values are accuracies (\%).}
% \label{tab:rlvr-results}
% \begin{tabular}{lccccc}
% \toprule
% \textbf{Model} & \textbf{Sphinx IID} & \textbf{Sphinx OOD} & \textbf{MathVision} & \textbf{MM-IQ} & \textbf{Geo3k} \\
% \midrule
% Qwen2.5-7B        & 25.15 & 19.8 & 16.8 & 24.6 & 37.9 \\
% \quad +RLVR       & 42.55 & 26.8 & 23.9 & 25.9 & 37.1 \\
% Qwen2.5-3B        & 17.55 & 14.6 & 21.8 & 22.8 & 24.5 \\
% \quad +RLVR       & 31.65 & 22.2 & 21.8 & 24.7 & 29.0 \\
% \bottomrule
% \end{tabular}
% \end{table}

% Requires in preamble:
% \usepackage{booktabs}
% \usepackage{graphicx}   % for \resizebox
% \usepackage{makecell}   % for multi-line headers

\paragraph{Performance on the \textsc{Sphinx} test set.}
Table~\ref{tab:sphinx_iid_ood} summarizes the performance of the RLVR-trained models on both the IID and OOD subsets of \textsc{Sphinx}. 
We observe consistent and substantial gains in the IID split in all four models, indicating that RLVR effectively improves performance in the 20 types of tasks included during training. 
These gains also transfer to the held-out OOD tasks, though with greater variation across task families. 

Among the OOD tasks, \emph{Tiles Recoloring} remains the most challenging: two models show degraded performance relative to their base versions. 
This suggests that cell-wise comparison between two independently rendered boards is particularly difficult for current LVLMs. 
In contrast, the largest improvements occur on the \emph{Positional Count} task. 
We hypothesize that this transferability arises from its structural similarity to the \emph{Stack Count} task included in the training set, enabling RLVR to generalize the underlying spatial-relation reasoning.

\paragraph{Performance on External Benchmarks.}
We further evaluate the four RLVR-trained models on eight external visual reasoning benchmarks: 
MathVista~\cite{lu2023mathvista}, 
MathVision~\cite{mathvision}, 
MathVerse~\cite{zhang2024mathverse}, 
LogicVista~\cite{xiao2024logicvista}, 
BLINK~\cite{fu2024blink}, 
MMT-Bench~\cite{ying2024mmt}, 
MMVP~\cite{tong2024eyes} 
and VStarBench~\cite{cheng2025v}. 
For all open-source models, we use VLMEvalKit~\cite{duan2024vlmevalkit} to ensure consistent and standardized evaluation. 
The results  are summarized in Table~\ref{tab:multimodal_math_benchmarks}.

Across all four base models, RLVR consistently improves the average accuracy. 
Of the 32 model--dataset combinations, we observe performance gains in 26 cases, declines in 4, and no change in the remaining 2. 
The largest average gains are observed for the Qwen3 variants: Qwen3-VL-4B improves by +2.8\%, and Qwen3-VL-8B improves by +2.4\%. 
These results suggest that RLVR is especially effective for the Qwen3 family under our training setup. Due to resource constraints, we were unable to train larger models such as Qwen3-VL-30B or Qwen3-VL-235B. 
Studying whether similarly strong gains extend to larger architectures remains an open future direction.

\begin{figure}[t]
    \centering
    \captionsetup{skip=2pt}
    \begin{subfigure}{0.40\linewidth}
        \centering
        \includegraphics[width=\linewidth]{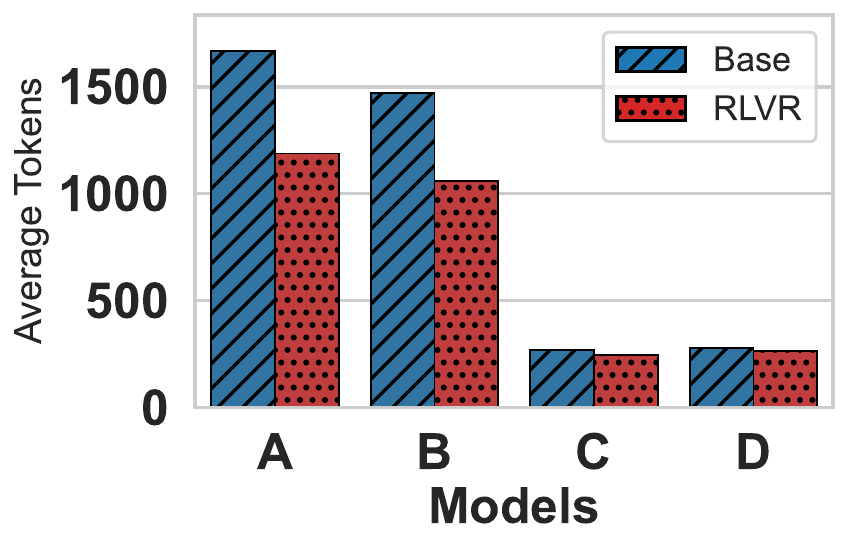}
        \caption{\textsc{Sphinx}}
    \end{subfigure}
    \hspace{0.008\linewidth}
    \begin{subfigure}{0.40\linewidth}
        \centering
        \includegraphics[width=\linewidth]{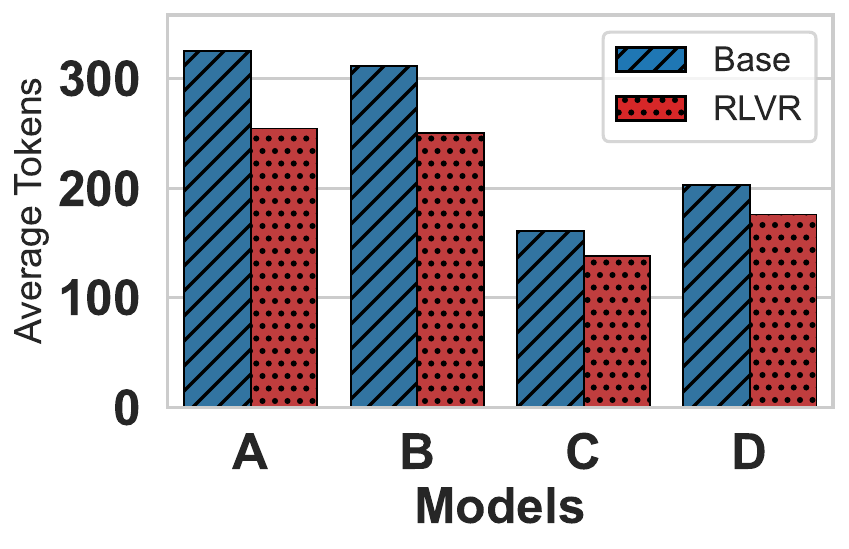}
        \caption{External Benchmarks}
    \end{subfigure}

    \caption{
    Average prediction token lengths between Base and RLVR models. 
    \textbf{Models} A (Qwen3-VL-4B), B (Qwen3-VL-8B), C (Qwen2.5-VL-3B), D (Qwen2.5-VL-7B).
    }
    \label{fig:response}
\vspace{-0.5em}
\end{figure}

\paragraph{Response Length.}

Figure~\ref{fig:response} shows the average response length of the RLVR-trained models compared to their base counterparts on both \textsc{Sphinx} and external benchmarks. 
Interestingly, models trained with RLVR consistently produce shorter responses in both settings, a trend that may be partly driven by the maximum 2048-token response length enforced during training. 
A detailed analysis of the responses of the RLVR model is provided in the Appendix~\ref{app:rlvranalysis}.

\section{Related works}

% \cite{xu2025visulogic} propsoed VisuLogic: a benchmark of 1,000 human-verified problems across six categories (e.g., quantitative shifts, spatial relations, attribute comparisons) with most MLLMs achieving near chance accuracy

% Research on visual reasoning benchmarks spans synthetic environments, cognitive-inspired tests, and reinforcement learning with verifiable rewards. We situate our work with respect to these efforts.

Research on visual reasoning originates in psychology and cognitive science, where human abilities are evaluated using tests such as Raven’s Progressive Matrices (RPM)~\citep{carpenter1990one} and the Wechsler Intelligence Scale for Children (WISC)~\citep{wechsler1949wechsler}. These assessments probe core perceptual and reasoning primitives such as symmetry detection, pattern completion, spatial transformation that underpin abstraction and fluid intelligence and remain challenging for artificial systems.

\noindent\textbf{Datasets and fixed benchmarks.}
Cognitive-test benchmarks such as ARC~\citep{arc_analysis}, Bongard Problems~\citep{malkinski2024reasoning}, and BONGARD-LOGO~\citep{bongardlogo} probe concept learning and analogy-making, while IQ-inspired datasets including MM-IQ~\citep{mmiq}, MARVEL~\citep{jiang2024marvel}, SMART-101~\citep{smart101}, and MaRs-VQA~\citep{cao2024visual} assess abstraction and generalization. MATH-Vision~\citep{mathvision} focuses on multimodal mathematical reasoning. Surveys of RPM-style tasks~\citep{rpm_survey} highlight persistent human–model performance gaps, particularly in zero-shot generalization. Although informative, these datasets are fixed in size and limited in visual and structural diversity.

\noindent\textbf{Synthetic and procedural benchmarks.}
Procedural datasets address these limitations by enabling controlled variation: CVR~\citep{zerroug2022benchmark}, A-I-RAVEN, I-RAVEN-Mesh~\citep{airaven}, and NTSEBench~\citep{ntsebench} expand RPM-style designs; IconQA~\citep{iconqa}, VisuLogic~\citep{xu2025visulogic}, and Visual Riddles~\citep{visualriddles} generate diagrammatic and abstraction-focused puzzles. Broader synthetic environments such as Reasoning Gym~\citep{reasoninggym}, Enigmata~\citep{chen2025enigmata}, and UniBench~\citep{unibench} provide scalable generator–verifier frameworks. \textsc{Sphinx} extends this line of work by offering a diverse suite of procedurally generated tasks, each paired with deterministic verifiers for reliable and reproducible evaluation.

\noindent\textbf{Reinforcement learning for visual reasoning.}
RL with verifiable rewards (RLVR) has shown promise for improving visual reasoning, with consistent gains reported by Reason-RFT~\citep{reasonrft}, Visual-RFT~\citep{liu2025visual}, and Jigsaw-R1~\citep{wang2025jigsaw}. 
Other work explores grounded reasoning~\citep{vigorl}, data mixing~\citep{modomodo}, or RL-driven self-reflection and curricula~\citep{vlrethinker,vlcogito}. 
Generator–verifier frameworks such as Reasoning Gym~\citep{reasoninggym} and Enigmata~\citep{chen2025enigmata} further demonstrate the value of scalable and verifiable reward signals. 
\textsc{Sphinx} extends this line of work by providing a synthetic environment with deterministic verifiers for each task, making it well-suited for RLVR.

% \noindent\textbf{Reinforcement learning for visual reasoning.}
% Recent work has explored reinforcement learning with verifiable rewards (RLVR) to improve model reasoning. Reinforcement fine-tuning improves generalization in visual reasoning, as shown by Reason-RFT~\citep{reasonrft}, Visual-RFT~\citep{visualrft}, and Jigsaw-R1~\citep{jigsawr1}. ViGoRL \citep{vigorl} spatially grounds the reasoning steps for interpretability, while MoDoMoDo \citep{modomodo} investigates data mixture strategies. VL-Rethinker~\citep{vlrethinker} and VL-Cogito~\citep{vlcogito} incorporate RL-driven self-reflection and curriculum-based learning. Generator–verifier frameworks such as Reasoning Gym~\citep{reasoninggym} and Enigmata~\citep{enigmata} highlight the value of scalable reward signals. \textsc{Sphinx} complements these approaches by providing a synthetic gym where every task has a verifiable ground-truth solution, making it naturally suited for RLVR experiments.

% \paragraph{Summary.}
% Prior work has revealed important insights into visual reasoning but remains limited by fixed datasets, narrow procedural domains, or ad hoc feedback mechanisms. \textsc{Sphinx} integrates insights from cognitive testing, synthetic generation, and RLVR into a unified environment for systematic evaluation and training of multimodal reasoning models.

\section{Limitations \& Future Work}
While \textsc{Sphinx} offers a large-scale synthetic environment for visual perception and reasoning, the current study focuses on a subset of task families. Consequently, the observed performance gains may not fully translate to broader multimodal benchmarks. Future work can expand the diversity of task types and incorporate more visually complex or noisy inputs to better approximate real-world conditions.

Another limitation arises in our RLVR setup: even after training, in-distribution accuracy remains below 50\%, indicating that many \textsc{Sphinx} tasks are still difficult for current LVLMs. 
Low accuracy can produce scarce rewards during RL training, where some prompts do not yield correct traces, making target solution effectively unreachable under RLVR~\citep{yue2025does}. 
Addressing this challenge may require combining RLVR with supervised fine-tuning on reasoning traces generated by stronger teacher models~\citep{yan2025learning,ma2025learning}, thus improving the quality and density of learning signals.

In addition, curriculum-based strategies that explicitly incorporate task difficulty could further enhance generalization~\citep{reasoninggym}. 
Another promising direction is to reduce the guessability of multiple-choice formats during RL, ensuring that improvements reflect genuine reasoning rather than shortcut exploitation~\citep{guo2025deepseek}.

\section{Conclusion}
We introduced \emph{\textsc{Sphinx}}, a synthetic environment for visual perception and reasoning. 
Across its 25 tasks, state-of-the-art LVLMs still struggle, while reinforcement learning with verifiable rewards yields meaningful improvements and enhances generalization to external visual reasoning benchmarks. 
Future work will expand \emph{\textsc{Sphinx}} with additional task families and more advanced RL paradigms. 
We have released the framework as open source to support broader adoption and community-driven extensions.

% \clearpage

{
    \small
    \bibliographystyle{ieeenat_fullname}
    \bibliography{reference}

\begin{thebibliography}{74}
\providecommand{\natexlab}[1]{#1}
\providecommand{\url}[1]{\texttt{#1}}
\expandafter\ifx\csname urlstyle\endcsname\relax
  \providecommand{\doi}[1]{doi: #1}\else
  \providecommand{\doi}{doi: \begingroup \urlstyle{rm}\Url}\fi

\bibitem[Al-Tahan et~al.(2024)Al-Tahan, Garrido, Balestriero, Bouchacourt,
  Hazirbas, Ibrahim, et~al.]{unibench}
Haider Al-Tahan, Quentin Garrido, Randall Balestriero, Diane Bouchacourt, Caner
  Hazirbas, Mark Ibrahim, et~al.
\newblock Unibench: Visual reasoning requires rethinking vision-language beyond
  scaling.
\newblock \emph{NeurIPS Datasets and Benchmarks}, 2024.

\bibitem[Albalak et~al.(2025)Albalak, Phung, Lile, Rafailov, Gandhi,
  Castricato, Singh, Blagden, Xiang, Mahan, et~al.]{albalak2025big}
Alon Albalak, Duy Phung, Nathan Lile, Rafael Rafailov, Kanishk Gandhi, Louis
  Castricato, Anikait Singh, Chase Blagden, Violet Xiang, Dakota Mahan, et~al.
\newblock Big-math: A large-scale, high-quality math dataset for reinforcement
  learning in language models.
\newblock \emph{arXiv preprint arXiv:2502.17387}, 2025.

\bibitem[Bai et~al.(2025)Bai, Chen, Liu, Wang, Ge, Song, Dang, Wang, Wang,
  Tang, et~al.]{bai2025qwen2}
Shuai Bai, Keqin Chen, Xuejing Liu, Jialin Wang, Wenbin Ge, Sibo Song, Kai
  Dang, Peng Wang, Shijie Wang, Jun Tang, et~al.
\newblock Qwen2. 5-vl technical report.
\newblock \emph{arXiv preprint arXiv:2502.13923}, 2025.

\bibitem[Bitton-Guetta et~al.(2024)Bitton-Guetta, Slobodkin, Maimon, Habba,
  Rassin, Bitton, Szpektor, Globerson, Elovici, et~al.]{visualriddles}
Nitzan Bitton-Guetta, Aviv Slobodkin, Aviya Maimon, Eliya Habba, Royi Rassin,
  Yonatan Bitton, Idan Szpektor, Amir Globerson, Yuval Elovici, et~al.
\newblock Visual riddles: A commonsense and world knowledge challenge for large
  vision and language models.
\newblock \emph{NeurIPS Dataset / arXiv}, 2024.

\bibitem[Cai et~al.(2025)Cai, Yang, and Hu]{mmiq}
Huanqia Cai, Yijun Yang, and Winston Hu.
\newblock Mm-iq: Benchmarking human-like abstraction and reasoning in
  multimodal models.
\newblock 2025.

\bibitem[Cao et~al.(2024)Cao, Lai, Ye, Ma, Heintz, Chen, Cao, and
  Rehg]{cao2024visual}
Xu Cao, Bolin Lai, Wenqian Ye, Yunsheng Ma, Joerg Heintz, Jintai Chen, Jianguo
  Cao, and James~M Rehg.
\newblock What is the visual cognition gap between humans and multimodal llms?
\newblock \emph{arXiv preprint arXiv:2406.10424}, 2024.

\bibitem[Carpenter et~al.(1990)Carpenter, Just, and Shell]{carpenter1990one}
Patricia~A Carpenter, Marcel~A Just, and Peter Shell.
\newblock What one intelligence test measures: a theoretical account of the
  processing in the raven progressive matrices test.
\newblock \emph{Psychological review}, 97\penalty0 (3):\penalty0 404, 1990.

\bibitem[Castelvecchi(2025)]{castelvecchi2025ai}
Davide Castelvecchi.
\newblock Ai models solve maths problems at level of top students.
\newblock \emph{Nature}, 644:\penalty0 7, 2025.

\bibitem[Chen et~al.(2025)Chen, He, Yuan, Chen, Cai, Dai, Yu, Yu, Li, Chen,
  et~al.]{chen2025enigmata}
Jiangjie Chen, Qianyu He, Siyu Yuan, Aili Chen, Zhicheng Cai, Weinan Dai,
  Hongli Yu, Qiying Yu, Xuefeng Li, Jiaze Chen, et~al.
\newblock Enigmata: Scaling logical reasoning in large language models with
  synthetic verifiable puzzles.
\newblock \emph{arXiv preprint arXiv:2505.19914}, 2025.

\bibitem[Cheng et~al.(2025)Cheng, Hu, Liu, Si, Li, and Gong]{cheng2025v}
Zixu Cheng, Jian Hu, Ziquan Liu, Chenyang Si, Wei Li, and Shaogang Gong.
\newblock V-star: Benchmarking video-llms on video spatio-temporal reasoning.
\newblock \emph{arXiv preprint arXiv:2503.11495}, 2025.

\bibitem[Cherian et~al.(2023)Cherian, Peng, Lohit, Smith, and
  Tenenbaum]{smart101}
Anoop Cherian, Kuan-Chuan Peng, Suhas Lohit, Kevin~A Smith, and Joshua~B
  Tenenbaum.
\newblock Are deep neural networks smarter than second graders?
\newblock In \emph{Proceedings of CVPR (Open Access)}, 2023.

\bibitem[Chollet et~al.(2025)Chollet, Knoop, Kamradt, Landers, and
  Pinkard]{chollet2025arc}
Francois Chollet, Mike Knoop, Gregory Kamradt, Bryan Landers, and Henry
  Pinkard.
\newblock Arc-agi-2: A new challenge for frontier ai reasoning systems.
\newblock \emph{arXiv preprint arXiv:2505.11831}, 2025.

\bibitem[Clark et~al.(2026)Clark, Zhang, Ma, Park, Salehi, Tripathi, Lee, Ren,
  Kim, Yang, et~al.]{clark2026molmo2}
Christopher Clark, Jieyu Zhang, Zixian Ma, Jae~Sung Park, Mohammadreza Salehi,
  Rohun Tripathi, Sangho Lee, Zhongzheng Ren, Chris~Dongjoo Kim, Yinuo Yang,
  et~al.
\newblock Molmo2: Open weights and data for vision-language models with video
  understanding and grounding.
\newblock \emph{arXiv preprint arXiv:2601.10611}, 2026.

\bibitem[Comanici et~al.(2025)Comanici, Bieber, Schaekermann, Pasupat,
  Sachdeva, Dhillon, Blistein, Ram, Zhang, Rosen, et~al.]{comanici2025gemini}
Gheorghe Comanici, Eric Bieber, Mike Schaekermann, Ice Pasupat, Noveen
  Sachdeva, Inderjit Dhillon, Marcel Blistein, Ori Ram, Dan Zhang, Evan Rosen,
  et~al.
\newblock Gemini 2.5: Pushing the frontier with advanced reasoning,
  multimodality, long context, and next generation agentic capabilities.
\newblock \emph{arXiv preprint arXiv:2507.06261}, 2025.

\bibitem[Duan et~al.(2024)Duan, Yang, Qiao, Fang, Chen, Liu, Dong, Zang, Zhang,
  Wang, et~al.]{duan2024vlmevalkit}
Haodong Duan, Junming Yang, Yuxuan Qiao, Xinyu Fang, Lin Chen, Yuan Liu, Xiaoyi
  Dong, Yuhang Zang, Pan Zhang, Jiaqi Wang, et~al.
\newblock Vlmevalkit: An open-source toolkit for evaluating large
  multi-modality models.
\newblock In \emph{Proceedings of the 32nd ACM international conference on
  multimedia}, pages 11198--11201, 2024.

\bibitem[Fisher et~al.(1981)Fisher, Ferdinandsen, and
  Bornstein]{fisher1981role}
Celia~B Fisher, Kay Ferdinandsen, and Marc~H Bornstein.
\newblock The role of symmetry in infant form discrimination.
\newblock \emph{Child development}, pages 457--462, 1981.

\bibitem[Fu et~al.(2024)Fu, Hu, Li, Feng, Wang, Lin, Roth, Smith, Ma, and
  Krishna]{fu2024blink}
Xingyu Fu, Yushi Hu, Bangzheng Li, Yu Feng, Haoyu Wang, Xudong Lin, Dan Roth,
  Noah~A Smith, Wei-Chiu Ma, and Ranjay Krishna.
\newblock Blink: Multimodal large language models can see but not perceive.
\newblock In \emph{European Conference on Computer Vision}, pages 148--166.
  Springer, 2024.

\bibitem[Gandhi et~al.(2025)Gandhi, Chakravarthy, Singh, Lile, and
  Goodman]{gandhi2025cognitive}
Kanishk Gandhi, Ayush Chakravarthy, Anikait Singh, Nathan Lile, and Noah~D
  Goodman.
\newblock Cognitive behaviors that enable self-improving reasoners, or, four
  habits of highly effective stars.
\newblock \emph{arXiv preprint arXiv:2503.01307}, 2025.

\bibitem[Goertzel(2014)]{goertzel2014artificial}
Ben Goertzel.
\newblock Artificial general intelligence: Concept, state of the art, and
  future prospects.
\newblock \emph{Journal of Artificial General Intelligence}, 5\penalty0
  (1):\penalty0 1, 2014.

\bibitem[Guo et~al.(2025)Guo, Yang, Zhang, Song, Zhang, Xu, Zhu, Ma, Wang, Bi,
  et~al.]{guo2025deepseek}
Daya Guo, Dejian Yang, Haowei Zhang, Junxiao Song, Ruoyu Zhang, Runxin Xu,
  Qihao Zhu, Shirong Ma, Peiyi Wang, Xiao Bi, et~al.
\newblock Deepseek-r1: Incentivizing reasoning capability in llms via
  reinforcement learning.
\newblock \emph{arXiv preprint arXiv:2501.12948}, 2025.

\bibitem[He et~al.(2025)He, Liang, Xu, Liu, Chen, Wang, Song, Yu, Liang, Wang,
  et~al.]{he2025deepmath}
Zhiwei He, Tian Liang, Jiahao Xu, Qiuzhi Liu, Xingyu Chen, Yue Wang, Linfeng
  Song, Dian Yu, Zhenwen Liang, Wenxuan Wang, et~al.
\newblock Deepmath-103k: A large-scale, challenging, decontaminated, and
  verifiable mathematical dataset for advancing reasoning.
\newblock \emph{arXiv preprint arXiv:2504.11456}, 2025.

\bibitem[hiyouga(2025)]{mathruler}
hiyouga.
\newblock Mathruler.
\newblock \url{https://github.com/hiyouga/MathRuler}, 2025.

\bibitem[Jaech et~al.(2024)Jaech, Kalai, Lerer, Richardson, El-Kishky, Low,
  Helyar, Madry, Beutel, Carney, et~al.]{jaech2024openai}
Aaron Jaech, Adam Kalai, Adam Lerer, Adam Richardson, Ahmed El-Kishky, Aiden
  Low, Alec Helyar, Aleksander Madry, Alex Beutel, Alex Carney, et~al.
\newblock Openai o1 system card.
\newblock \emph{arXiv preprint arXiv:2412.16720}, 2024.

\bibitem[Jiang et~al.(2024)Jiang, Sun, Sourati, Ahrabian, Ma, Ilievski, Pujara,
  et~al.]{jiang2024marvel}
Yifan Jiang, Kexuan Sun, Zhivar Sourati, Kian Ahrabian, Kaixin Ma, Filip
  Ilievski, Jay Pujara, et~al.
\newblock Marvel: Multidimensional abstraction and reasoning through visual
  evaluation and learning.
\newblock \emph{Advances in Neural Information Processing Systems},
  37:\penalty0 46567--46592, 2024.

\bibitem[Johnson et~al.(2017)Johnson, Hariharan, Van Der~Maaten, Fei-Fei,
  Lawrence~Zitnick, and Girshick]{johnson2017clevr}
Justin Johnson, Bharath Hariharan, Laurens Van Der~Maaten, Li Fei-Fei, C
  Lawrence~Zitnick, and Ross Girshick.
\newblock Clevr: A diagnostic dataset for compositional language and elementary
  visual reasoning.
\newblock In \emph{Proceedings of the IEEE conference on computer vision and
  pattern recognition}, pages 2901--2910, 2017.

\bibitem[Lee et~al.(2024)Lee, Sim, Shin, Seo, Park, Lee, Hwang, Kim, and
  Kim]{arc_analysis}
Seungpil Lee, Woochang Sim, Donghyeon Shin, Wongyu Seo, Jiwon Park, Seokki Lee,
  Sanha Hwang, Sejin Kim, and Sundong Kim.
\newblock Reasoning abilities of large language models: In-depth analysis on
  the abstraction and reasoning corpus.
\newblock 2024.

\bibitem[Liang et~al.(2025)Liang, Qiu, Ding, Liu, Tompkin, Xu, Xia, Tu, Shi,
  and Zhu]{modomodo}
Yiqing Liang, Jielin Qiu, Wenhao Ding, Zuxin Liu, James Tompkin, Mengdi Xu,
  Mengzhou Xia, Zhengzhong Tu, Laixi Shi, and Jiacheng Zhu.
\newblock Modomodo: Multi-domain data mixtures for multimodal llm reinforcement
  learning.
\newblock 2025.

\bibitem[Liu et~al.(2025)Liu, Sun, Zang, Dong, Cao, Duan, Lin, and
  Wang]{liu2025visual}
Ziyu Liu, Zeyi Sun, Yuhang Zang, Xiaoyi Dong, Yuhang Cao, Haodong Duan, Dahua
  Lin, and Jiaqi Wang.
\newblock Visual-rft: Visual reinforcement fine-tuning.
\newblock \emph{arXiv preprint arXiv:2503.01785}, 2025.

\bibitem[Lu et~al.(2021{\natexlab{a}})Lu, Gong, Jiang, Qiu, Huang, Liang, and
  Zhu]{lu2021inter}
Pan Lu, Ran Gong, Shibiao Jiang, Liang Qiu, Siyuan Huang, Xiaodan Liang, and
  Song-Chun Zhu.
\newblock Inter-gps: Interpretable geometry problem solving with formal
  language and symbolic reasoning.
\newblock \emph{arXiv preprint arXiv:2105.04165}, 2021{\natexlab{a}}.

\bibitem[Lu et~al.(2021{\natexlab{b}})Lu, Qiu, Chen, Xia, Zhao, Zhang, Yu,
  Liang, and Zhu]{iconqa}
Pan Lu, Liang Qiu, Jiaqi Chen, Tony Xia, Yizhou Zhao, Wei Zhang, Zhou Yu,
  Xiaodan Liang, and Song-Chun Zhu.
\newblock Icon{QA}: A new benchmark for abstract diagram understanding and
  visual language reasoning.
\newblock In \emph{Thirty-fifth Conference on Neural Information Processing
  Systems Datasets and Benchmarks Track (Round 2)}, 2021{\natexlab{b}}.

\bibitem[Lu et~al.(2023)Lu, Bansal, Xia, Liu, Li, Hajishirzi, Cheng, Chang,
  Galley, and Gao]{lu2023mathvista}
Pan Lu, Hritik Bansal, Tony Xia, Jiacheng Liu, Chunyuan Li, Hannaneh
  Hajishirzi, Hao Cheng, Kai-Wei Chang, Michel Galley, and Jianfeng Gao.
\newblock Mathvista: Evaluating mathematical reasoning of foundation models in
  visual contexts.
\newblock \emph{arXiv preprint arXiv:2310.02255}, 2023.

\bibitem[Ma et~al.(2025)Ma, Liang, Qiang, Tang, Ma, Wong, Niu, Shen, He, Li,
  et~al.]{ma2025learning}
Lu Ma, Hao Liang, Meiyi Qiang, Lexiang Tang, Xiaochen Ma, Zhen~Hao Wong, Junbo
  Niu, Chengyu Shen, Runming He, Yanhao Li, et~al.
\newblock Learning what reinforcement learning can't: Interleaved online
  fine-tuning for hardest questions.
\newblock \emph{arXiv preprint arXiv:2506.07527}, 2025.

\bibitem[Ma{\l}ki{\'n}ski et~al.(2024)Ma{\l}ki{\'n}ski, Pawlonka, and
  Ma{\'n}dziuk]{malkinski2024reasoning}
Miko{\l}aj Ma{\l}ki{\'n}ski, Szymon Pawlonka, and Jacek Ma{\'n}dziuk.
\newblock Reasoning limitations of multimodal large language models. a case
  study of bongard problems.
\newblock \emph{arXiv preprint arXiv:2411.01173}, 2024.

\bibitem[Małkiński and Mańdziuk(2025{\natexlab{a}})]{airaven}
Mikołaj Małkiński and Jacek Mańdziuk.
\newblock A-i-raven and i-raven-mesh: Two new benchmarks for abstract visual
  reasoning.
\newblock 2025{\natexlab{a}}.

\bibitem[Małkiński and Mańdziuk(2025{\natexlab{b}})]{rpm_survey}
Mikołaj Małkiński and Jacek Mańdziuk.
\newblock Deep learning methods for abstract visual reasoning: A survey on
  raven’s progressive matrices.
\newblock \emph{ACM Computing Surveys}, 57\penalty0 (7):\penalty0 1–36,
  2025{\natexlab{b}}.

\bibitem[Meta(2024)]{meta2024llama}
AI Meta.
\newblock Llama 3.2: Revolutionizing edge ai and vision with open, customizable
  models.
\newblock \emph{Meta AI Blog. Retrieved December}, 20:\penalty0 2024, 2024.

\bibitem[Nie et~al.(2020)Nie, Yu, Mao, Patel, Zhu, Anandkumar,
  et~al.]{bongardlogo}
Weili Nie, Zhiding Yu, Lei Mao, Ankit~B Patel, Yuke Zhu, Animashree Anandkumar,
  et~al.
\newblock Bongard-logo: A new benchmark for human-level concept learning and
  reasoning.
\newblock \emph{NeurIPS 2020 (Dataset)}, 2020.

\bibitem[{OpenAI}(2025)]{openai2025introducinggpt5}
{OpenAI}.
\newblock Introducing gpt-5.
\newblock \url{https://openai.com/index/introducing-gpt-5/}, 2025.
\newblock Accessed August 2025.

\bibitem[Pandya et~al.(2025)Pandya, Gupta, Talwarr, Kataria, Roth, and
  Gupta]{ntsebench}
Pranshu Pandya, Vatsal Gupta, Agney~S Talwarr, Tushar Kataria, Dan Roth, and
  Vivek Gupta.
\newblock Ntsebench: Cognitive reasoning benchmark for vision language models.
\newblock 2025.

\bibitem[Peng et~al.(2025)Peng, Zhang, Zhang, You, Liu, Zhu, Yang, Xu, Geng,
  and Yang]{peng2025lmm}
Yingzhe Peng, Gongrui Zhang, Miaosen Zhang, Zhiyuan You, Jie Liu, Qipeng Zhu,
  Kai Yang, Xingzhong Xu, Xin Geng, and Xu Yang.
\newblock Lmm-r1: Empowering 3b lmms with strong reasoning abilities through
  two-stage rule-based rl.
\newblock \emph{arXiv preprint arXiv:2503.07536}, 2025.

\bibitem[Pizlo and De~Barros(2021)]{pizlo2021concept}
Zygmunt Pizlo and J~Acacio De~Barros.
\newblock The concept of symmetry and the theory of perception.
\newblock \emph{Frontiers in Computational Neuroscience}, 15:\penalty0 681162,
  2021.

\bibitem[Qi et~al.(2021)Qi, Zhang, Sain, and Song]{qi2021pqa}
Yonggang Qi, Kai Zhang, Aneeshan Sain, and Yi-Zhe Song.
\newblock Pqa: Perceptual question answering.
\newblock In \emph{Proceedings of the IEEE/CVF Conference on Computer Vision
  and Pattern Recognition}, pages 12056--12064, 2021.

\bibitem[Reimers and Gurevych(2019)]{reimers2019sentence}
Nils Reimers and Iryna Gurevych.
\newblock Sentence-bert: Sentence embeddings using siamese bert-networks.
\newblock \emph{arXiv preprint arXiv:1908.10084}, 2019.

\bibitem[Sarch et~al.(2025)Sarch, Saha, Khandelwal, Jain, Tarr, Kumar, and
  Fragkiadaki]{vigorl}
Gabriel Sarch, Snigdha Saha, Naitik Khandelwal, Ayush Jain, Michael~J. Tarr,
  Aviral Kumar, and Katerina Fragkiadaki.
\newblock Grounded reinforcement learning for visual reasoning.
\newblock 2025.

\bibitem[Schulze~Buschoff et~al.(2025)Schulze~Buschoff, Akata, Bethge, and
  Schulz]{schulze2025visual}
Luca~M Schulze~Buschoff, Elif Akata, Matthias Bethge, and Eric Schulz.
\newblock Visual cognition in multimodal large language models.
\newblock \emph{Nature Machine Intelligence}, 7\penalty0 (1):\penalty0 96--106,
  2025.

\bibitem[Shao et~al.(2024)Shao, Wang, Zhu, Xu, Song, Bi, Zhang, Zhang, Li, Wu,
  et~al.]{shao2024deepseekmath}
Zhihong Shao, Peiyi Wang, Qihao Zhu, Runxin Xu, Junxiao Song, Xiao Bi, Haowei
  Zhang, Mingchuan Zhang, YK Li, Yang Wu, et~al.
\newblock Deepseekmath: Pushing the limits of mathematical reasoning in open
  language models.
\newblock \emph{arXiv preprint arXiv:2402.03300}, 2024.

\bibitem[Shepard and Cooper(1986)]{shepard1986mental}
Roger~N Shepard and Lynn~A Cooper.
\newblock \emph{Mental images and their transformations.}
\newblock The MIT Press, 1986.

\bibitem[Stojanovski et~al.(2025)Stojanovski, Stanley, Sharratt, Jones,
  Adefioye, Kaddour, K{\"o}pf, et~al.]{reasoninggym}
Zafir Stojanovski, Oliver Stanley, Joe Sharratt, Richard Jones, Abdulhakeem
  Adefioye, Jean Kaddour, Andreas K{\"o}pf, et~al.
\newblock Reasoning gym: Reasoning environments for reinforcement learning with
  verifiable rewards.
\newblock \emph{arXiv preprint arXiv:2505.24760}, 2025.

\bibitem[Tan et~al.(2025)Tan, Ji, Hao, Chen, Wang, Wang, and Zhang]{reasonrft}
Huajie Tan, Yuheng Ji, Xiaoshuai Hao, Xiansheng Chen, Pengwei Wang, Zhongyuan
  Wang, and Shanghang Zhang.
\newblock Reason-rft: Reinforcement fine-tuning for visual reasoning of vision
  language models.
\newblock \emph{arXiv preprint arXiv:2503.20752}, 2025.

\bibitem[Team(2025)]{qwen3_vl_2025}
Qwen Team.
\newblock Qwen3-vl.
\newblock \url{https://github.com/QwenLM/Qwen3-VL}, 2025.
\newblock Accessed: 2025-11-14.

\bibitem[Tong et~al.(2024)Tong, Liu, Zhai, Ma, LeCun, and Xie]{tong2024eyes}
Shengbang Tong, Zhuang Liu, Yuexiang Zhai, Yi Ma, Yann LeCun, and Saining Xie.
\newblock Eyes wide shut? exploring the visual shortcomings of multimodal llms.
\newblock In \emph{Proceedings of the IEEE/CVF Conference on Computer Vision
  and Pattern Recognition}, pages 9568--9578, 2024.

\bibitem[Wang et~al.(2025{\natexlab{a}})Wang, Qu, Huang, Chu, Lin, and
  Chen]{vlrethinker}
Haozhe Wang, Chao Qu, Zuming Huang, Wei Chu, Fangzhen Lin, and Wenhu Chen.
\newblock Vl-rethinker: Incentivizing self-reflection of vision-language models
  with reinforcement learning.
\newblock \emph{arXiv preprint}, 2025{\natexlab{a}}.

\bibitem[Wang et~al.(2024)Wang, Pan, Shi, Lu, Ren, Zhou, Zhan, and
  Li]{mathvision}
Ke Wang, Junting Pan, Weikang Shi, Zimu Lu, Houxing Ren, Aojun Zhou, Mingjie
  Zhan, and Hongsheng Li.
\newblock Measuring multimodal mathematical reasoning with math-vision dataset.
\newblock \emph{Advances in Neural Information Processing Systems},
  37:\penalty0 95095--95169, 2024.

\bibitem[Wang et~al.(2025{\natexlab{b}})Wang, Zhu, Tang, Li, Xiong, Yu, and
  Blaschko]{wang2025jigsaw}
Zifu Wang, Junyi Zhu, Bo Tang, Zhiyu Li, Feiyu Xiong, Jiaqian Yu, and Matthew~B
  Blaschko.
\newblock Jigsaw-r1: A study of rule-based visual reinforcement learning with
  jigsaw puzzles.
\newblock \emph{arXiv preprint arXiv:2505.23590}, 2025{\natexlab{b}}.

\bibitem[Wechsler(1949)]{wechsler1949wechsler}
David Wechsler.
\newblock Wechsler intelligence scale for children.
\newblock 1949.

\bibitem[Wei et~al.(2022)Wei, Wang, Schuurmans, Bosma, Xia, Chi, Le, Zhou,
  et~al.]{wei2022chain}
Jason Wei, Xuezhi Wang, Dale Schuurmans, Maarten Bosma, Fei Xia, Ed Chi, Quoc~V
  Le, Denny Zhou, et~al.
\newblock Chain-of-thought prompting elicits reasoning in large language
  models.
\newblock \emph{Advances in neural information processing systems},
  35:\penalty0 24824--24837, 2022.

\bibitem[Wei et~al.(2024)Wei, Fu, Jiang, Kwok, and Zhang]{wei2024rendering}
Yanbin Wei, Shuai Fu, Weisen Jiang, James~T Kwok, and Yu Zhang.
\newblock Rendering graphs for graph reasoning in multimodal large language
  models.
\newblock \emph{arXiv preprint arXiv:2402.02130}, 1, 2024.

\bibitem[Wexler et~al.(1998)Wexler, Kosslyn, and Berthoz]{wexler1998motor}
Mark Wexler, Stephen~M Kosslyn, and Alain Berthoz.
\newblock Motor processes in mental rotation.
\newblock \emph{Cognition}, 68\penalty0 (1):\penalty0 77--94, 1998.

\bibitem[Wo{\'z}niak et~al.(2023)Wo{\'z}niak, J{\'o}nsson, Cherubini, Pantazi,
  and Eleftheriou]{wozniak2023visual}
Stanis{\l}aw Wo{\'z}niak, Hlynur J{\'o}nsson, Giovanni Cherubini, Angeliki
  Pantazi, and Evangelos Eleftheriou.
\newblock On the visual analytic intelligence of neural networks.
\newblock \emph{Nature Communications}, 14\penalty0 (1):\penalty0 5978, 2023.

\bibitem[Wu et~al.(2024)Wu, Qiu, Ross, Aky{\"u}rek, Chen, Wang, Kim, Andreas,
  and Kim]{wu2024reasoning}
Zhaofeng Wu, Linlu Qiu, Alexis Ross, Ekin Aky{\"u}rek, Boyuan Chen, Bailin
  Wang, Najoung Kim, Jacob Andreas, and Yoon Kim.
\newblock Reasoning or reciting? exploring the capabilities and limitations of
  language models through counterfactual tasks.
\newblock Association for Computational Linguistics, 2024.

\bibitem[Xiao et~al.(2024)Xiao, Sun, Liu, and Wang]{xiao2024logicvista}
Yijia Xiao, Edward Sun, Tianyu Liu, and Wei Wang.
\newblock Logicvista: Multimodal llm logical reasoning benchmark in visual
  contexts.
\newblock \emph{arXiv preprint arXiv:2407.04973}, 2024.

\bibitem[Xie et~al.(2025)Xie, Gao, Ren, Luo, Hong, Dai, Zhou, Qiu, Wu, and
  Luo]{xie2025logic}
Tian Xie, Zitian Gao, Qingnan Ren, Haoming Luo, Yuqian Hong, Bryan Dai, Joey
  Zhou, Kai Qiu, Zhirong Wu, and Chong Luo.
\newblock Logic-rl: Unleashing llm reasoning with rule-based reinforcement
  learning.
\newblock \emph{arXiv preprint arXiv:2502.14768}, 2025.

\bibitem[Xu et~al.(2025)Xu, Wang, Wang, Chen, Zhou, Yang, Lu, Li, Wang, Zhu,
  et~al.]{xu2025visulogic}
Weiye Xu, Jiahao Wang, Weiyun Wang, Zhe Chen, Wengang Zhou, Aijun Yang, Lewei
  Lu, Houqiang Li, Xiaohua Wang, Xizhou Zhu, et~al.
\newblock Visulogic: A benchmark for evaluating visual reasoning in multi-modal
  large language models.
\newblock \emph{arXiv preprint arXiv:2504.15279}, 2025.

\bibitem[Yan et~al.(2025)Yan, Li, Hu, Wang, Cui, Qu, Cheng, and
  Zhang]{yan2025learning}
Jianhao Yan, Yafu Li, Zican Hu, Zhi Wang, Ganqu Cui, Xiaoye Qu, Yu Cheng, and
  Yue Zhang.
\newblock Learning to reason under off-policy guidance.
\newblock \emph{arXiv preprint arXiv:2504.14945}, 2025.

\bibitem[Yang et~al.(2025{\natexlab{a}})Yang, Li, Yang, Zhang, Hui, Zheng, Yu,
  Gao, Huang, Lv, et~al.]{yang2025qwen3}
An Yang, Anfeng Li, Baosong Yang, Beichen Zhang, Binyuan Hui, Bo Zheng, Bowen
  Yu, Chang Gao, Chengen Huang, Chenxu Lv, et~al.
\newblock Qwen3 technical report.
\newblock \emph{arXiv preprint arXiv:2505.09388}, 2025{\natexlab{a}}.

\bibitem[Yang et~al.(2025{\natexlab{b}})Yang, He, Pan, Jiang, Deng, Yang, Lu,
  Yin, Rao, Zhu, et~al.]{yang2025r1}
Yi Yang, Xiaoxuan He, Hongkun Pan, Xiyan Jiang, Yan Deng, Xingtao Yang, Haoyu
  Lu, Dacheng Yin, Fengyun Rao, Minfeng Zhu, et~al.
\newblock R1-onevision: Advancing generalized multimodal reasoning through
  cross-modal formalization.
\newblock \emph{arXiv preprint arXiv:2503.10615}, 2025{\natexlab{b}}.

\bibitem[Ying et~al.(2024)Ying, Meng, Wang, Li, Lin, Yang, Zhang, Zhang, Lin,
  Liu, et~al.]{ying2024mmt}
Kaining Ying, Fanqing Meng, Jin Wang, Zhiqian Li, Han Lin, Yue Yang, Hao Zhang,
  Wenbo Zhang, Yuqi Lin, Shuo Liu, et~al.
\newblock Mmt-bench: A comprehensive multimodal benchmark for evaluating large
  vision-language models towards multitask agi.
\newblock \emph{arXiv preprint arXiv:2404.16006}, 2024.

\bibitem[Yuan et~al.(2025)Yuan, Xiao, Leng, Wang, Li, Xu, Wei, Zhang, Rong,
  et~al.]{vlcogito}
Ruifeng Yuan, Chenghao Xiao, Sicong Leng, Jianyu Wang, Long Li, Tingyang Xu,
  Zhongyu Wei, Hao Zhang, Yu Rong, et~al.
\newblock Vl-cogito: Progressive curriculum reinforcement learning for advanced
  multimodal reasoning.
\newblock \emph{arXiv preprint}, 2025.

\bibitem[Yue et~al.(2025)Yue, Chen, Lu, Zhao, Wang, Song, and
  Huang]{yue2025does}
Yang Yue, Zhiqi Chen, Rui Lu, Andrew Zhao, Zhaokai Wang, Shiji Song, and Gao
  Huang.
\newblock Does reinforcement learning really incentivize reasoning capacity in
  llms beyond the base model?
\newblock \emph{arXiv preprint arXiv:2504.13837}, 2025.

\bibitem[Yuksekgonul et~al.(2022)Yuksekgonul, Bianchi, Kalluri, Jurafsky, and
  Zou]{yuksekgonul2022and}
Mert Yuksekgonul, Federico Bianchi, Pratyusha Kalluri, Dan Jurafsky, and James
  Zou.
\newblock When and why vision-language models behave like bags-of-words, and
  what to do about it?
\newblock \emph{arXiv preprint arXiv:2210.01936}, 2022.

\bibitem[Zerroug et~al.(2022)Zerroug, Vaishnav, Colin, Musslick, and
  Serre]{zerroug2022benchmark}
Aimen Zerroug, Mohit Vaishnav, Julien Colin, Sebastian Musslick, and Thomas
  Serre.
\newblock A benchmark for compositional visual reasoning.
\newblock \emph{Advances in neural information processing systems},
  35:\penalty0 29776--29788, 2022.

\bibitem[Zhang et~al.(2024)Zhang, Jiang, Zhang, Lin, Guo, Qiu, Zhou, Lu, Chang,
  Qiao, et~al.]{zhang2024mathverse}
Renrui Zhang, Dongzhi Jiang, Yichi Zhang, Haokun Lin, Ziyu Guo, Pengshuo Qiu,
  Aojun Zhou, Pan Lu, Kai-Wei Chang, Yu Qiao, et~al.
\newblock Mathverse: Does your multi-modal llm truly see the diagrams in visual
  math problems?
\newblock In \emph{European Conference on Computer Vision}, pages 169--186.
  Springer, 2024.

\bibitem[Zheng et~al.(2025)Zheng, Lu, Wang, Feng, Kuang, and
  Xiong]{zheng2025easyr1}
Yaowei Zheng, Junting Lu, Shenzhi Wang, Zhangchi Feng, Dongdong Kuang, and
  Yuwen Xiong.
\newblock Easyr1: An efficient, scalable, multi-modality rl training framework,
  2025.

\bibitem[Zhu et~al.(2025)Zhu, Wang, Chen, Liu, Ye, Gu, Tian, Duan, Su, Shao,
  et~al.]{zhu2025internvl3}
Jinguo Zhu, Weiyun Wang, Zhe Chen, Zhaoyang Liu, Shenglong Ye, Lixin Gu, Hao
  Tian, Yuchen Duan, Weijie Su, Jie Shao, et~al.
\newblock Internvl3: Exploring advanced training and test-time recipes for
  open-source multimodal models.
\newblock \emph{arXiv preprint arXiv:2504.10479}, 2025.

\end{thebibliography}
}


\clearpage
\newpage

% \section*{Appendix}

\appendix
\section{Implementation Summary}
\label{app:summary}

\subsection{Overview}

\emph{\textsc{Sphinx}} is a framework for programmatically generating visual reasoning tasks by pairing a registry of parameterized motifs and tilings with a registry of task classes. Each task produces a complete instance consisting of (i) a rendered composite image, (ii) the exact specifications of all constituent motifs or tiles, and (iii) structured task metadata such as question, answer, reasoning type, and difficulty attributes. Task families are sampled according to configurable weights, allowing controlled variation over categories during dataset creation.

The generation engine follows a modular pipeline. A task class first samples its internal parameters (e.g., symmetry axis, sequence rule, region-of-interest) and draws motif or tiling instances from their respective registries. Motif objects expose sampling routines for geometry, appearance, and layout, while tilings produce full cell-level polygonal patches that downstream tasks query for adjacency or geometric measurements. All tasks support multiple output formats: some yield visual multiple-choice items with rendered distractor images, whereas others produce text-based options or integer-valued answers. Multiple-choice distractors are programmatically constructed to be unique and non-trivial, avoiding degenerate overlaps with the ground truth.

To further diversify the dataset, each task is associated with a bank of natural-language prompt templates. During generation, the engine selects a template, instantiates it with task-specific details, renders the composite scene at a fixed resolution, and logs all metadata: motif specifications, sampled parameters, answer, distractors, and provenance in a structured record. This design enables the construction of large-scale datasets with rich annotations, supports verifiable reward functions, and provides a unified interface for training and evaluating multimodal reasoning models.

\begin{figure*}[t]
  \centering
  \includegraphics[page=1,width=1\linewidth]{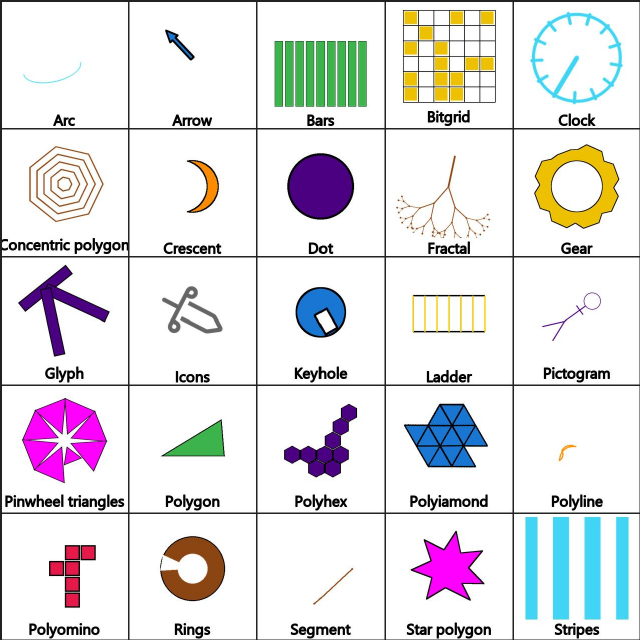}
  \caption{Randomly sampled example from each Motif family.}
  \label{fig:motif_gallery}
\end{figure*}

\subsection{Motif Library}
\label{sec:motifs}

\textsc{Sphinx} includes 25 procedurally generated visual motifs that serve as atomic building blocks to construct puzzles, scenes, and transformations. Each motif is parameterized by attributes of geometry, appearance, and layout, enabling broad variation and precise control. Figure~\ref{fig:motif_gallery} shows representative instances of the 25 motifs currently implemented in \textsc{Sphinx}. 
In the following, we list the full set.

\begin{enumerate}[leftmargin=*]

\item \textbf{Arc.}
Circular arc defined by center, radius, start angle, and sweep; optionally closed into a sector. Geometry varies in radius, sweep angle, thickness, and end-cap style. Appearance includes stroke color/width, optional fills, and dashed or solid rendering. Layout covers position, rotation, and multi-arc groupings.

\item \textbf{Arrow.}
Vector-like shape with a shaft and triangular or chevron head, optionally double-headed. Geometry varies via head angle, shaft width/length, curvature, and tail caps. Appearance includes filled or outlined styles, color palettes, and shading suppression. Layout controls orientation, alignment, and crowding.

\item \textbf{Bars.}
Parallel rectangular bars (horizontal or vertical) used for counts or measurements. Geometry includes bar count, width/height, spacing, and jitter. Appearance supports solid or gradient fills and optional outlines. Layout covers grouping, stacking, and background grid usage.

\item \textbf{Bitgrid.}
Binary on/off cell grid. Geometry includes grid size, cell aspect ratio, bit density, and mask structure. Appearance includes on/off colors, padding, rounding, and borders. Layout supports margins, rotation, and embedding within scenes.

\item \textbf{Clock.}
Analog clock with ticks, numerals, and hands. Geometry varies via tick count, numeral style, and hand lengths/angles. Appearance includes face and background styles. Layout supports centering and partial occlusion.

\item \textbf{Concentric Polygon.}
Multiple nested regular polygons sharing a center, optionally with rotation offsets. Geometry varies in side count, number of layers, spacing, and relative rotation. Appearance includes filled or outlined layers and alternating colors. Layout controls scale and juxtaposition.

\item \textbf{Crescent.}
Lune formed by subtracting one disk from another. Geometry parameters include radii ratio, center offset, orientation, and crescent thickness. Appearance includes fill color, outline, and boundary smoothing. Layout covers placement and combinations with other shapes.

\item \textbf{Dot.}
Single filled circle or square used as a point marker. Geometry includes radius and shape type. Appearance includes color, optional halo, and stroke. Layout spans isolated dots or constellations.

\item \textbf{Fractal.}
Deterministic fractal curves or sets (e.g., Koch, Sierpiński). Geometry varies via generator type, iteration depth, base size, and orientation. Appearance supports stroke or filled-region modes. Layout includes centering and multi-fractal combinations.

\item \textbf{Gear.}
Cogwheel with teeth and a central bore. Geometry varies in tooth count, rim/hub radii, and fillet style. Appearance includes solid fills, outlines, and simplified shading. Layout supports single gears, meshing pairs, and rotations.

\item \textbf{Glyph.}
Typographic outlines of letters, numerals, or symbols. Geometry varies via font family, weight, character set, and size. Appearance supports filled or outlined text and anti-aliasing. Layout covers single glyphs or short strings, with rotation options.

\item \textbf{Icons.}
Simple pictorial symbols from a curated set. Geometry includes icon category, stroke/fill structure, and detail level. Appearance includes monochrome or bicolor styles. Layout supports tiling and consistent scaling.

\item \textbf{Keyhole.}
Classic keyhole silhouette with circular/oval head and tapered slot. Geometry varies in head aspect ratio, slot width/length, fillets, and taper angle. Appearance includes filled or outlined styles. Layout covers orientation and grouping.

\item \textbf{Ladder.}
Two rails with evenly spaced rungs. Geometry parameters include rail spacing, rung count, rail/rung thickness, and tilt. Appearance supports stroke or filled-rectangle rendering. Layout includes vertical, horizontal, or tilted placement.

\item \textbf{Pictogram.}
Silhouette-style signs (e.g., person or restroom icons). Geometry varies by category, pose, simplification level, and aspect ratio. Appearance features filled silhouettes or minor stroke accents. Layout supports single or grid-aligned arrangements.

\item \textbf{Pinwheel Triangle.}
Radial arrangement of isosceles or right triangles forming a pinwheel. Geometry parameters include blade count, blade angle, inner radius, and chirality. Appearance includes alternating fills and optional center hubs. Layout uses centered radial symmetry.

\item \textbf{Polygon.}
Regular or irregular polygons, optionally star-convex. Geometry includes vertex count, side lengths, jitter, and rotation. Appearance supports fill, stroke, corner rounding, and hatch patterns. Layout may involve multiple polygons or packing constraints.

\item \textbf{Polyhex.}
Connected unions of unit hexagons. Geometry includes cell count, topology, and boundary complexity. Appearance supports cell outlines or merged silhouettes with fills. Layout aligns shapes to a hexagonal grid.

\item \textbf{Polyamond.}
Connected unions of unit equilateral triangles. Geometry includes cell count, triangulation orientation, and possible holes. Appearance includes outlined or merged fills. Layout aligns motifs to triangular lattice axes.

\item \textbf{Polyline.}
Open piecewise-linear path with ordered vertices. Geometry varies via vertex count, segment lengths, angle distributions, and self-avoidance. Appearance includes stroke width, joint style, and dash patterns. Layout supports start/end markers and crossings.

\item \textbf{Polyomino.}
Connected unions of unit squares. Geometry includes cell count, aspect ratio, perimeter length, and possible holes. Appearance includes borders or merged silhouettes with optional patterned fills. Layout aligns to a square grid.

\item \textbf{Rings.}
Circular annuli, single or multiple. Geometry includes outer radius, thickness, ring count, and spacing. Appearance supports filled or stroked annuli with alternating colors. Layout includes centering and multi-ring clusters.

\item \textbf{Segment.}
Straight line segment with optional endpoint markers. Geometry includes length, orientation, and end-cap style. Appearance includes stroke width, color, and dash patterns. Layout spans independent or bundled segments.

\item \textbf{Star Polygon.}
Regular star polygons $\{n,k\}$ defined by step-$k$ vertex connections. Geometry varies in $n$, $k$, radius, and rotation. Appearance includes outlined or filled modes and inner-polygon visibility. Layout supports single or layered stars.

\item \textbf{Stripes.}
Repeating parallel bands. Geometry varies in stripe width, spacing, orientation, and phase offset. Appearance includes alternating colors, gradients, and edge softness. Layout supports full-canvas or masked-region coverage.

\end{enumerate}

\subsection{Tilings}
\label{sec:tilings}

\textsc{Sphinx} provides a unified interface for generating polygonal tilings used in tile-based reasoning tasks. Each tiling generator returns a \texttt{TilingPatch} containing vertices, cells, and adjacency information, together with canonical lattice coordinates and optional Wythoffian color classes. The rendering maps each patch to a fixed-resolution canvas with consistent margins. All tilings—regular or irregular—support dual-graph construction for connectivity queries and geometry extraction via exact per-cell polygons. In addition, the tiling subsystem standardizes cell metadata, grid coordinates, and palette assignment, allowing tasks to operate uniformly across geometries. Irregular tilings use vertex pooling and quantization to maintain robust topology, while regular tilings expose symmetry-aware color classes that downstream tasks can query to form structured patterns or control distractor design.

\noindent\textbf{Implementation Overview.}
A \texttt{TilingPatch} consists of a pooled vertex array, a list of \texttt{Cell} objects (each with vertex indices, a \texttt{kind} tag, and a grid coordinate), and optional edge records. Generators accept a \texttt{TilingSpec} that specifies \texttt{width}, \texttt{height}, and \texttt{margin\_frac}. Regular tilings (square, triangular, hexagonal, rhombille) implement canonical Wythoffian colorings, while irregular tilings (e.g., circles) rely on uniform or heuristic palettes. The dual graph is built through \texttt{build\_dual\_graph}, where nodes are cells and edges represent shared boundaries (or, optionally, shared vertices when \texttt{connect\_on\_touch} is enabled).

\noindent\textbf{Implemented Tilings} We implement the following five tilings:

\begin{enumerate}[leftmargin=*]

\item \textbf{Circles (Circle Packing).}
Equal-radius disks are placed on a triangular lattice and approximated by $m$-gons whose vertices align with the six tangential directions. Grid coordinates index lattice centers. Because neighboring polygons often touch at single vertices, adjacency uses \texttt{connect\_on\_touch}=True. This tiling does not expose Wythoffian color classes. Variants include grid size, radius, polygon fidelity, and margins.

\item \textbf{Square.}
Axis-aligned unit squares arranged on an integer lattice form a standard $\,\texttt{width} \times \texttt{height}\,$ rectangular grid. Each cell is indexed by its lower-left $(i,j)$ coordinate. Adjacency is defined by shared edges, and a four-class Wythoffian coloring is given by $((i \& 1) \ll 1) | (j \& 1)$. Variants include grid dimensions, margins, and rendering styles.

\item \textbf{Triangular (Equilateral).}
Point-up and point-down equilateral triangles form a row/column lattice, with alternating rows shifted by half a cell horizontally. Cells are indexed by integer coordinates consistent with this construction. Adjacent cells share full edges, and a three-class Wythoffian coloring is provided by $(i + 2j) \bmod 3$. Variants include grid size, triangle side length normalization, and margin.

\item \textbf{Hexagonal (Flat-Top, odd-\boldmath$q$).}
Regular flat-top hexagons are arranged on an odd-$q$ axial grid. Each hexagon is generated from a fixed radius using six vertices spaced at $60^\circ$. Cells have up to six neighbors via edge adjacency. A three-class Wythoffian coloring is obtained from $(q - r) \bmod 3$. Variants include grid dimensions (axial columns/rows), hex radius, and margins.

\item \textbf{Rhombille.}
Each hexagon of a flat-top hex grid is subdivided into three congruent $60^\circ/120^\circ$ rhombi. Cells are indexed by axial $(q,r)$ plus a sector identifier $(0,1,2)$. Vertex pooling ensures a shared center for the three rhombi. Adjacency connects rhombi both within each subdivided hex and across boundaries of the underlying hex grid. A Wythoffian three-class coloring is available via axial parity. Variants include grid size, radius scaling, and margins.

\end{enumerate}

\paragraph{Notes for Downstream Tasks.}
Connectivity-based tasks operate on the dual graph; shortest-path tasks use BFS or 0–1 BFS since edges are unweighted. Geometric and counting tasks use \texttt{cell\_polygons()} for exact polygon shapes, with vertex pooling providing numerical stability in irregular tilings.

\section{Task Descriptions}
\label{app:tasks}

\subsection{Geometric Reasoning}

\begin{figure*}[t]
    \centering
    \begin{subfigure}{\textwidth}
        \centering
        \begin{tabular}{ccc}
            \includegraphics[width=0.3\textwidth]{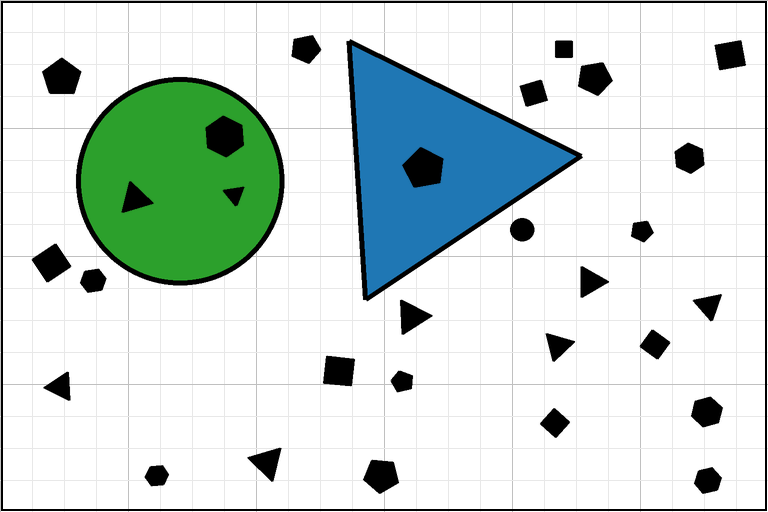} &
            \includegraphics[width=0.3\textwidth]{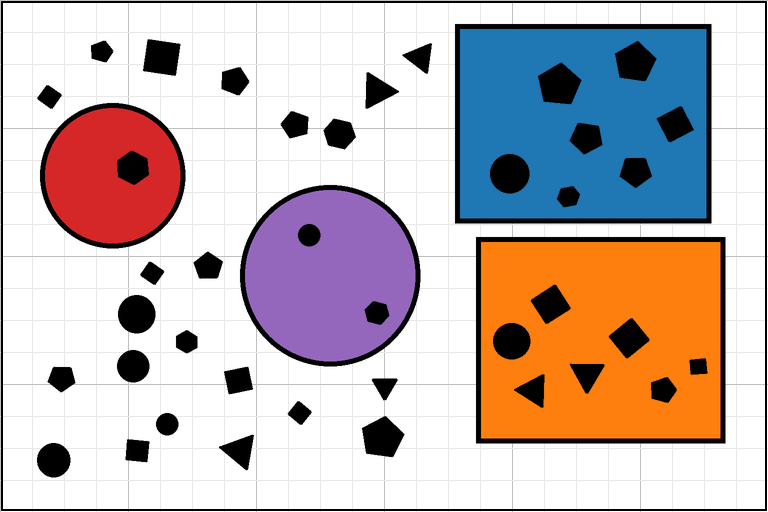} &
            \includegraphics[width=0.3\textwidth]{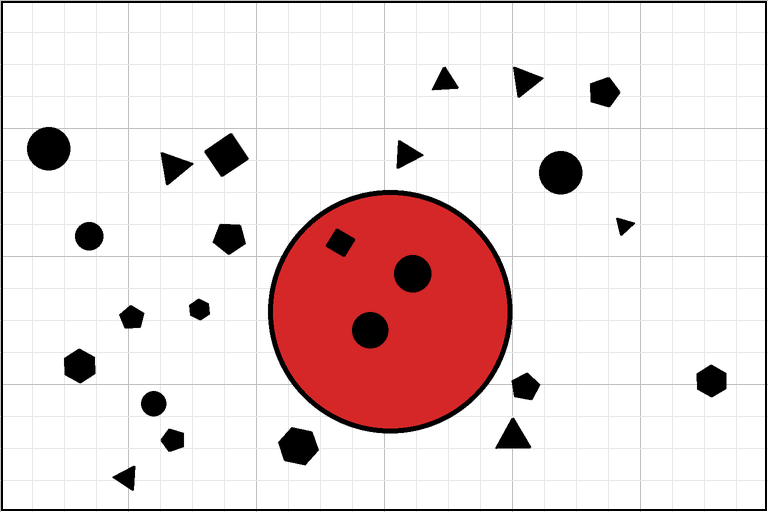} \\
        \end{tabular}
        \subcaption{\textbf{Positional Count} Count the small shapes that satisfy a specific spatial relation to a larger shape.}
        \label{fig:row_pos_count}
    \end{subfigure}
    
    \begin{subfigure}{\textwidth}
        \centering
        \begin{tabular}{ccc}
            \includegraphics[width=0.3\textwidth]{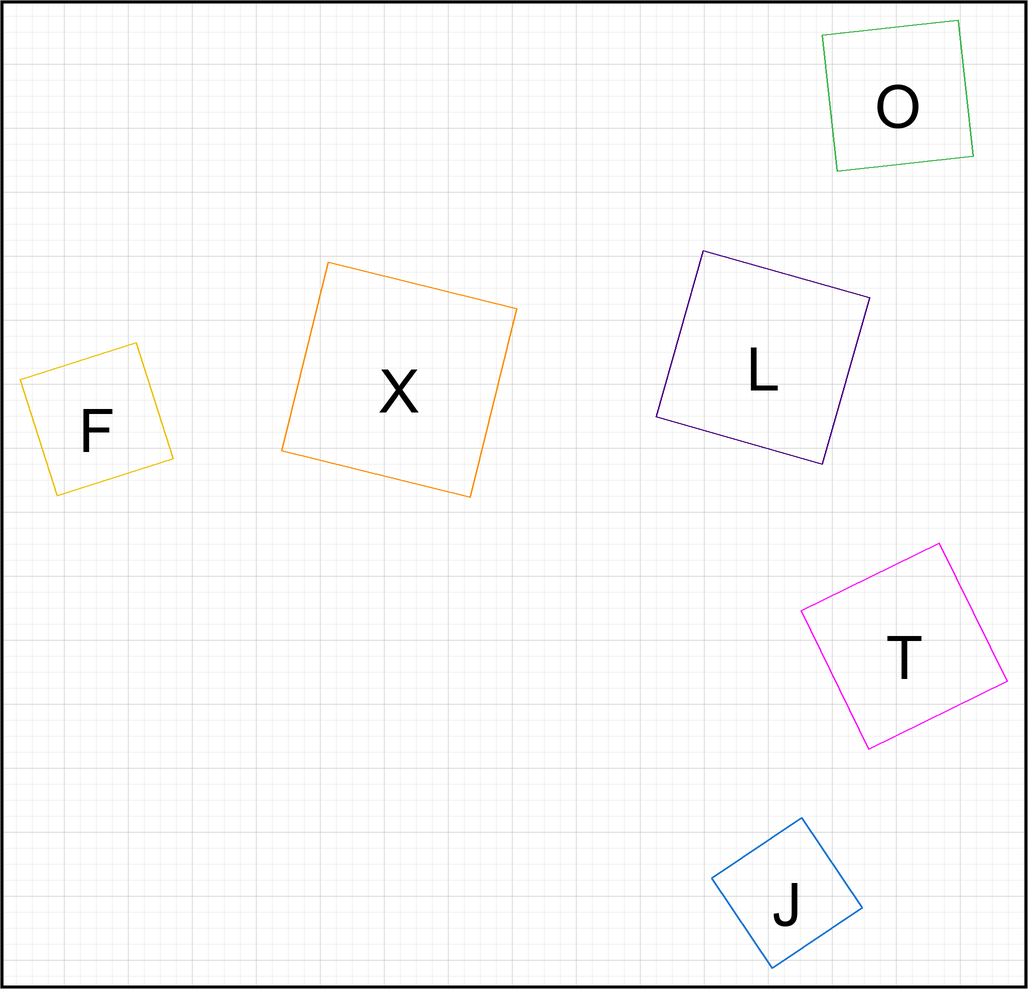} &
            \includegraphics[width=0.3\textwidth]{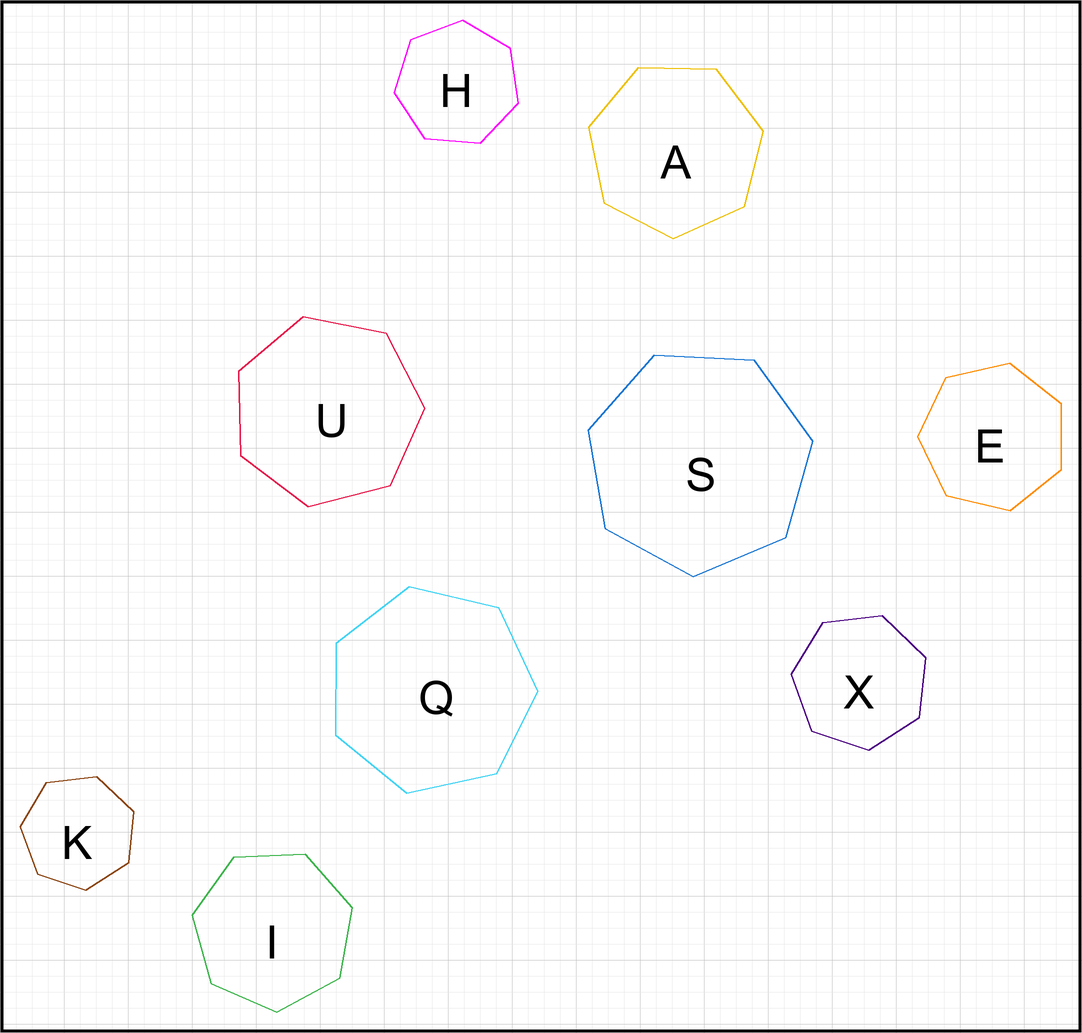} &
            \includegraphics[width=0.3\textwidth]{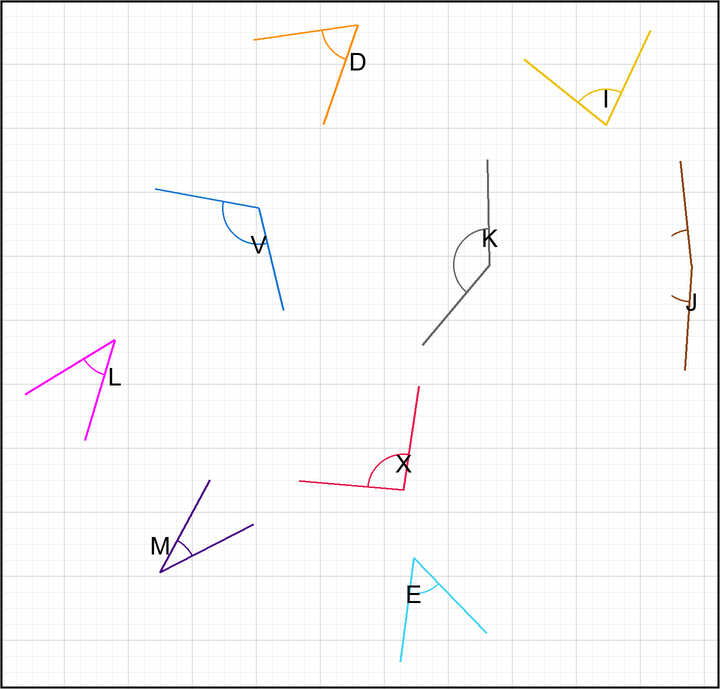} \\
        \end{tabular}
        \subcaption{\textbf{Shape Sorting} Sort the labeled shapes by a given metric, such as area or angle.}
        \label{fig:row_shape_sort}
    \end{subfigure}

    \begin{subfigure}{\textwidth}
        \centering
        \begin{tabular}{ccc}
            \includegraphics[width=0.3\textwidth]{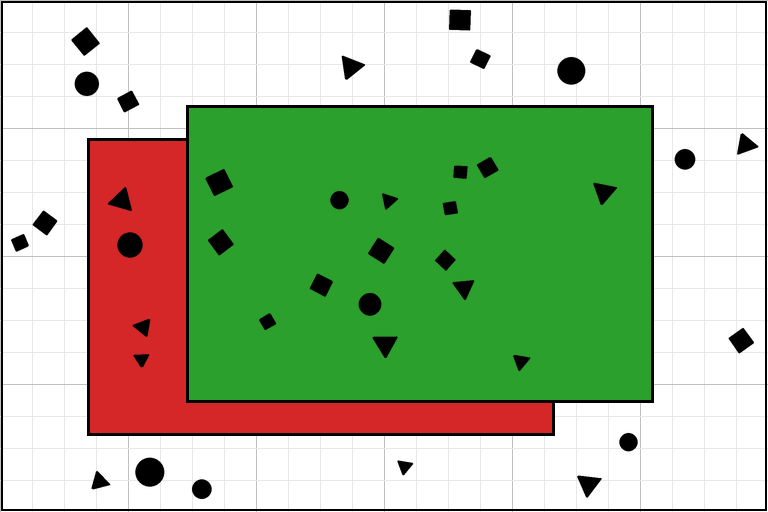} &
            \includegraphics[width=0.3\textwidth]{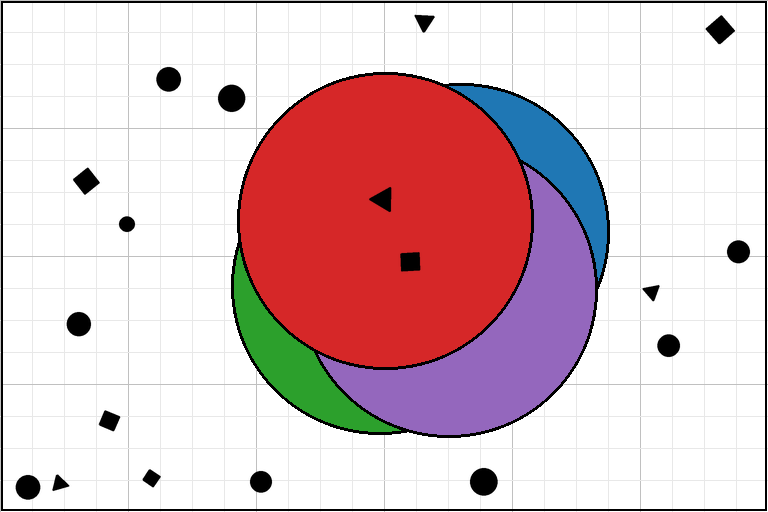} &
            \includegraphics[width=0.3\textwidth]{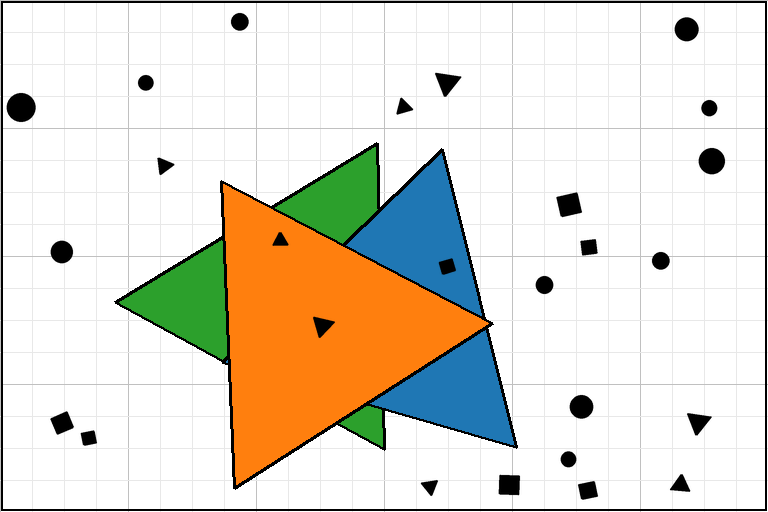} \\
        \end{tabular}
        \subcaption{\textbf{Stack Count} Count the number of a certain small shape that are fully inside one of the occluded, overlapping sheets.}
        \label{fig:row_stack_count}
    \end{subfigure}

    \begin{subfigure}{\textwidth}
        \centering
        \begin{tabular}{ccc}
            \includegraphics[width=0.3\textwidth]{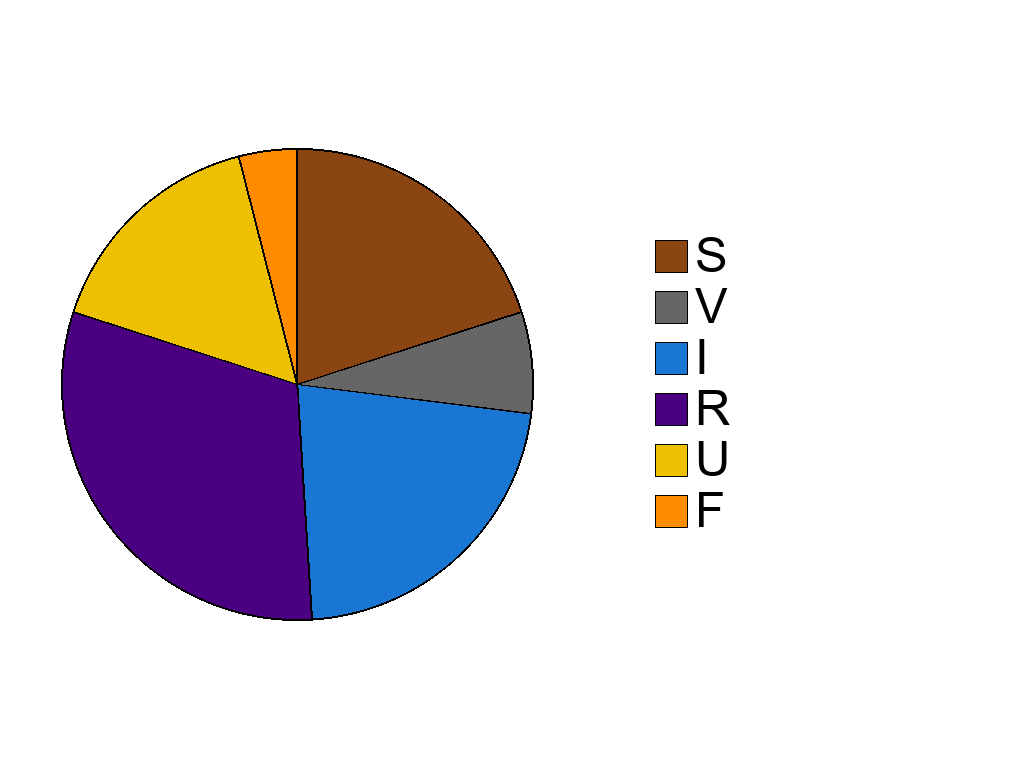} &
            \includegraphics[width=0.3\textwidth]{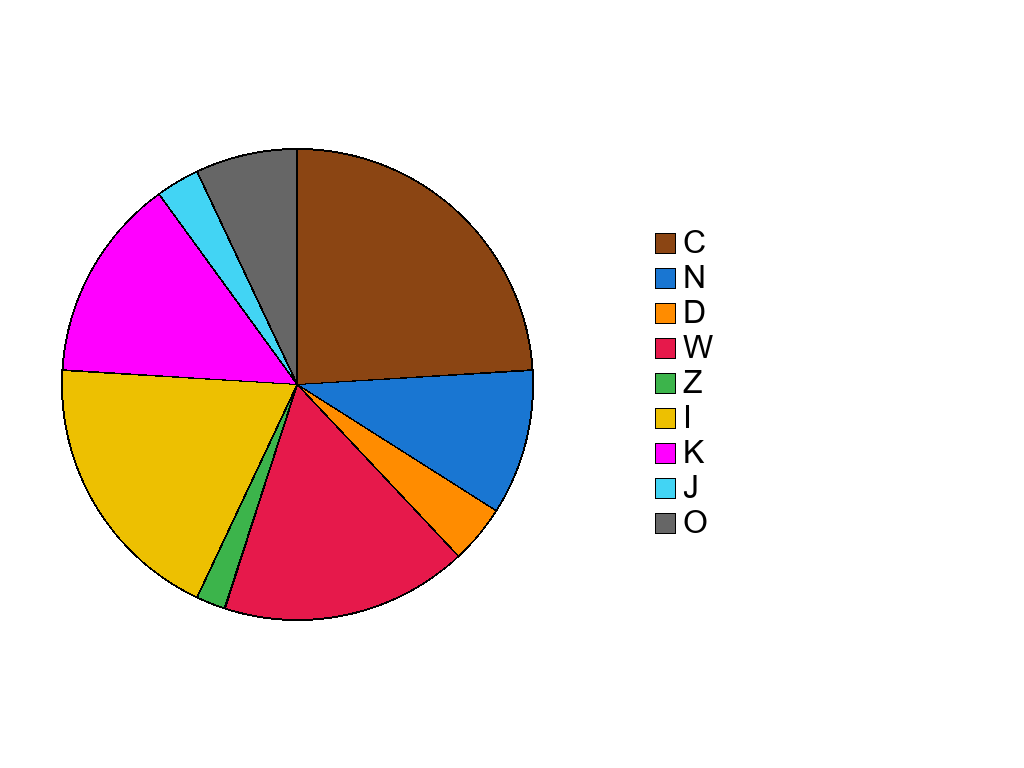} &
            \includegraphics[width=0.3\textwidth]{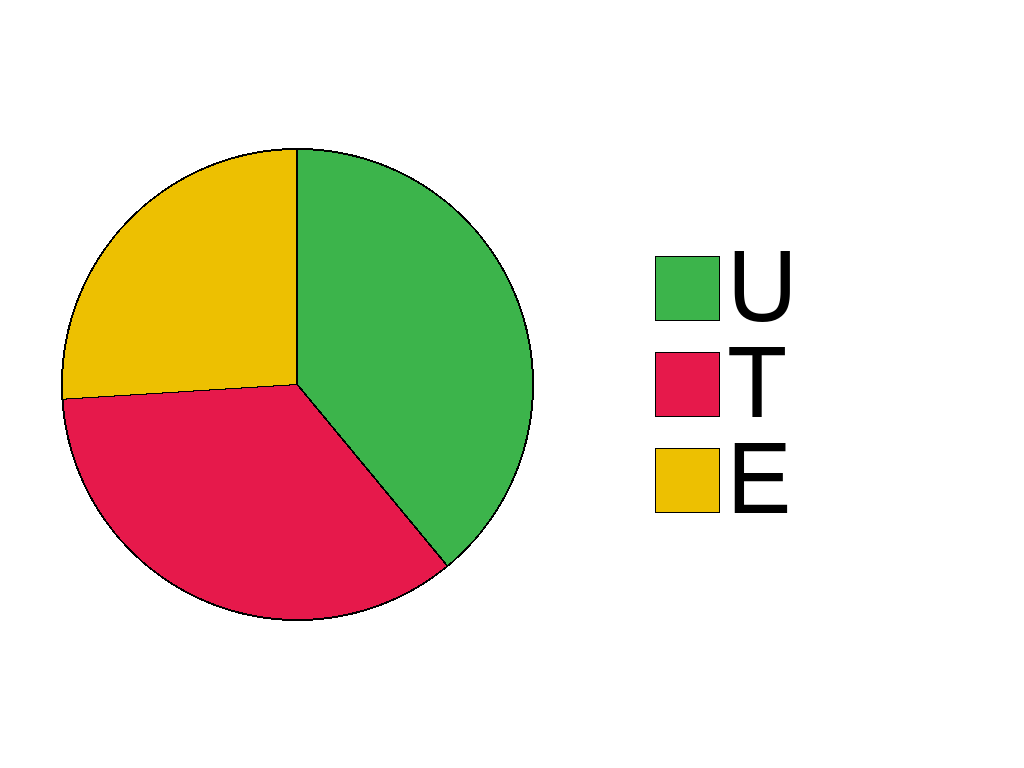} \\
        \end{tabular}
        \subcaption{\textbf{Pie Chart} Rank the slices of the pie chart by size.}
        \label{fig:row_pie_chart}
    \end{subfigure}

    \begin{subfigure}{\textwidth}
        \centering
        \begin{tabular}{ccc}
            \includegraphics[width=0.3\textwidth]{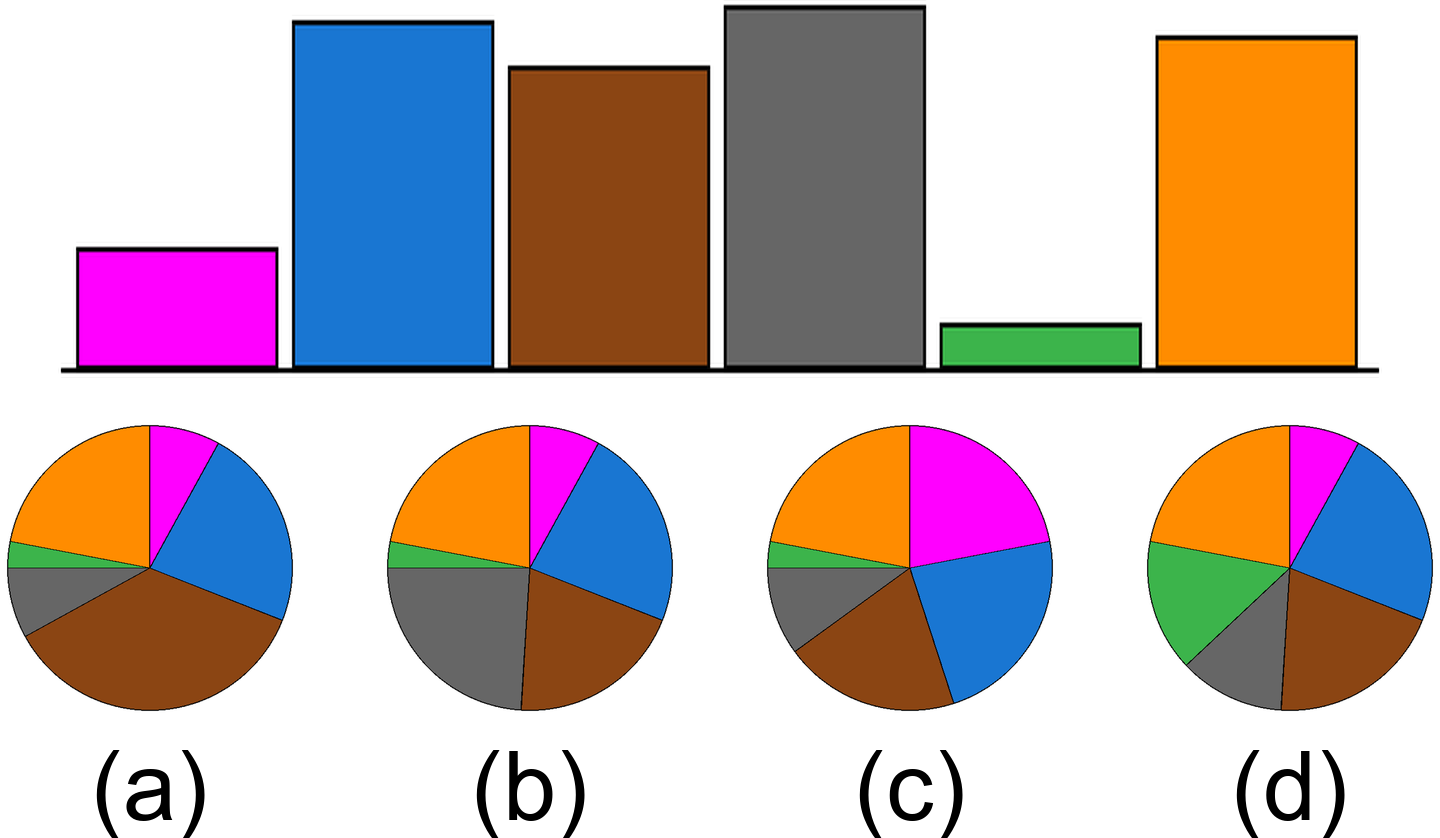} &
            \includegraphics[width=0.3\textwidth]{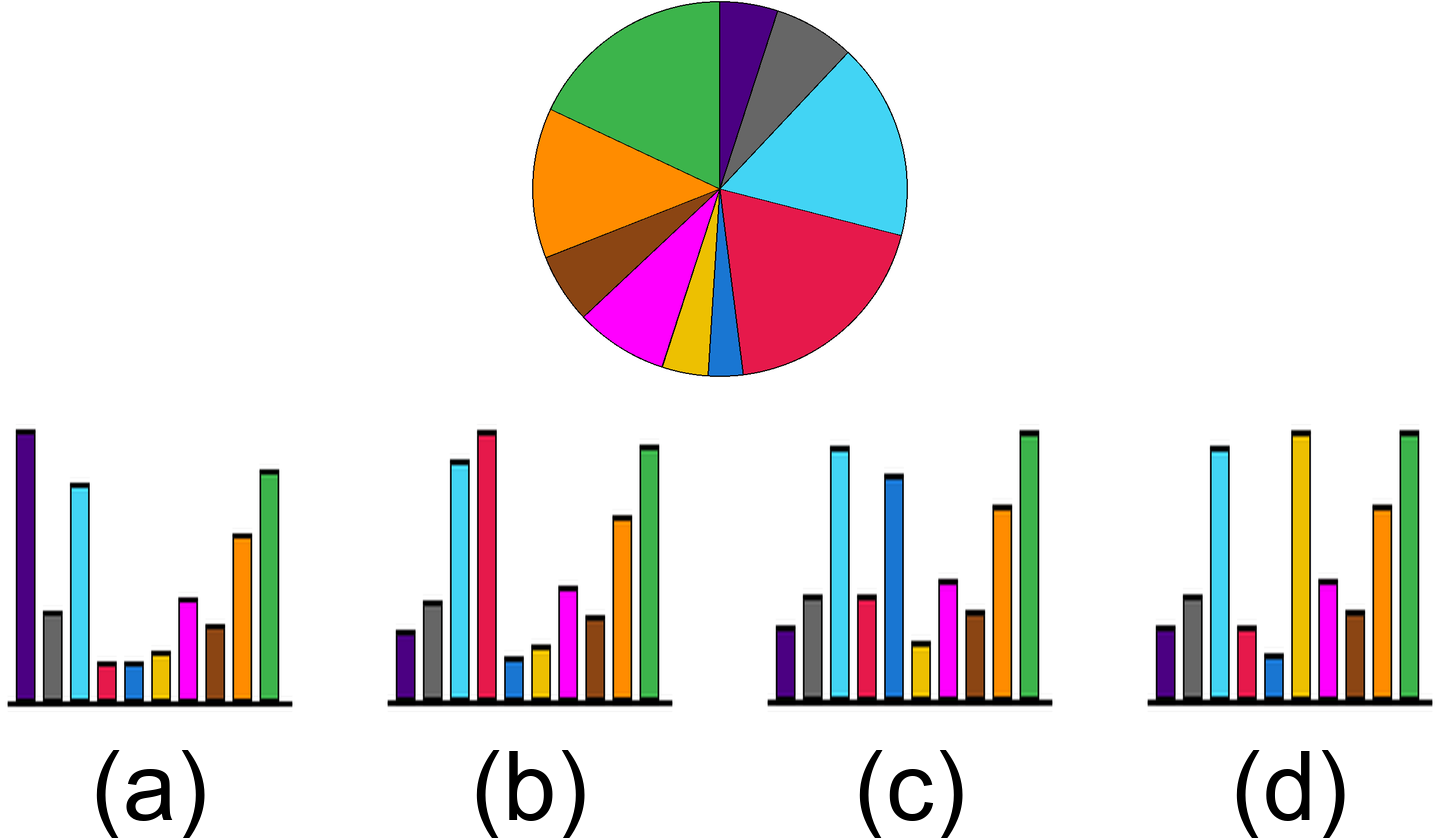} &
            \includegraphics[width=0.3\textwidth]{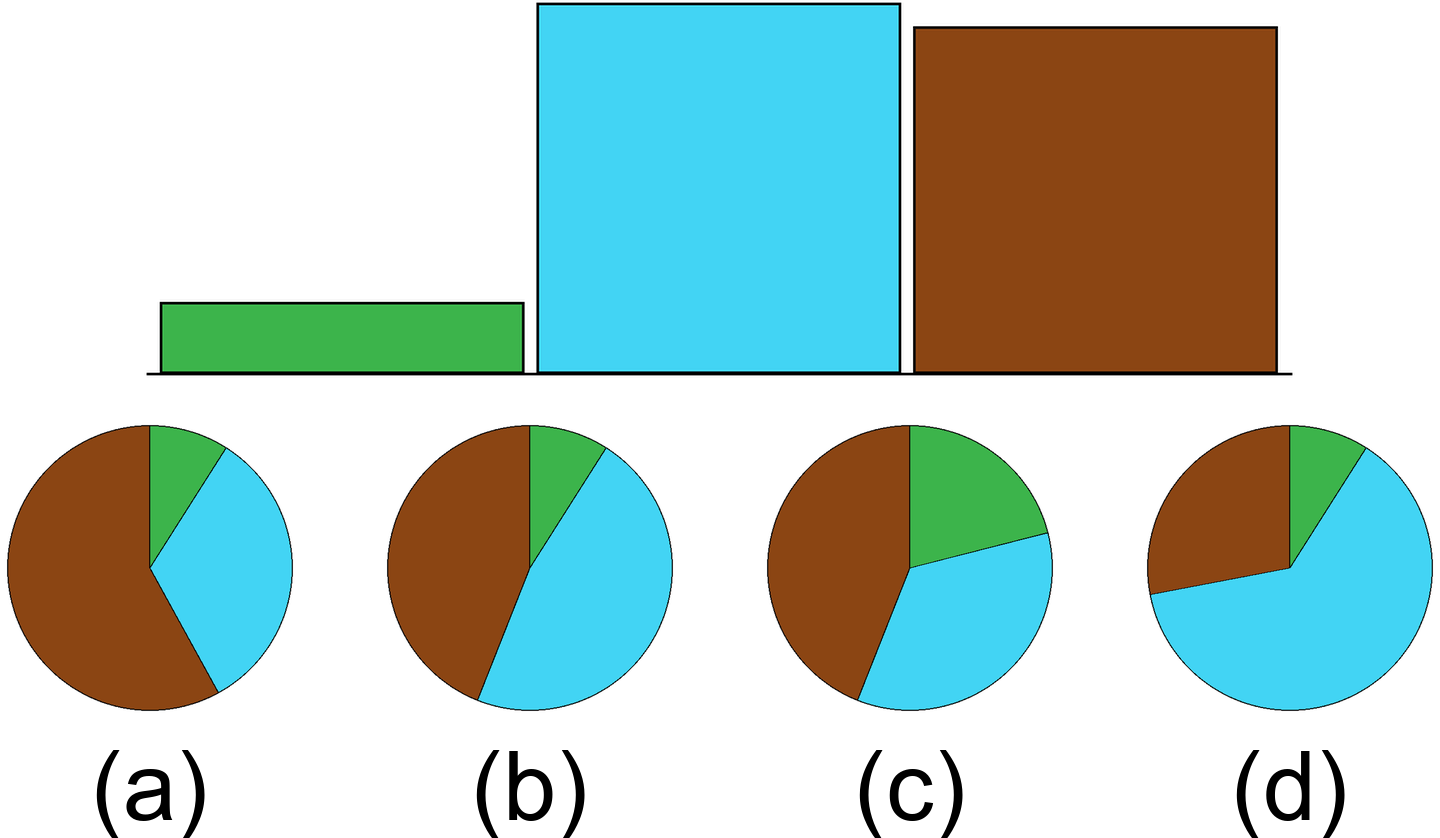} \\
        \end{tabular}
        \subcaption{\textbf{Chart Comparison} Find the bar/pie chart that correctly represents the proportions in the top chart.}
        \label{fig:row_chart_comp}
    \end{subfigure}
    
    \caption{Examples of Geometric Reasoning and Chart tasks.}
    \label{fig:grid_geometric}
\end{figure*}

Figure~\ref{fig:grid_geometric} shows examples of this type of task.

\subsubsection{Positional Count}

\noindent \textbf{Problem.} Positional counting relative to non-overlapping reference shapes (rectangles, circles, and triangles). The objective is to count small shapes that satisfy a strict spatial relation to a chosen reference.

\noindent \textbf{Construction.} Place 1-4 large reference shapes with enough separation. Sample small shapes (circle, triangle, square, pentagon, hexagon) with pairwise non-overlap and strict visual separation from all reference boundaries. Evaluate strict radius-aware predicates (inside, outside, above, below, left, right) to form the label.

\noindent \textbf{Variants.} Six relation categories crossed with multiple small-shape kinds; background and counts vary with seed.

\noindent  \textbf{Complexity.} We measure complexity with the count of larger reference shapes.

\noindent \textbf{Answer type.} Integer count.

\subsubsection{Shape Sorting}
\noindent  \textbf{Problem.} Ordinal sorting over labeled geometric primitives under a specified metric.

\noindent  \textbf{Construction.} Sample a family (polygon, ellipse, angle, line) and a metric (polygon/ellipse area or perimeter; angle measure; line length). Sample values with a minimum relative gap and render using a random-pack layout with uniform-height font for the label.

\noindent  \textbf{Variants.} Four families with metrics as above; the number of items $k$ is drawn from configurable bounds.

\noindent  \textbf{Complexity.} We measure complexity with the number of items sampled $k$.

\noindent  \textbf{Distractors.} Multiple-choice over comma-separated orderings; distractors are produced by swapping two positions in the true ordering (with random shuffles as a fallback to maintain four unique options).

\subsubsection{Stack Count}
\noindent  \textbf{Problem.} Given overlapping sheets of equal area, count small objects that lie strictly inside a designated sheet (excluding the topmost one).

\noindent \textbf{Construction.} Choose a stack type (rectangle, circle, equilateral triangle). Generate $k$ sheets with controlled pairwise overlap ratios and identical area; draw small objects (circle, triangle, square) on top of the stack. Pose an inside-of-border query about an occluded sheet.

\noindent  \textbf{Variants.} Three stack families $\times$ three kinds of small-objects. The prompts vary in target sheet (color) and object kind.

\noindent  \textbf{Complexity.} We measure complexity with the number of large stacked shapes.

\noindent  \textbf{Answer type.} Integer count.

\subsubsection{Pie Chart}
\noindent  \textbf{Problem.} Ordinal reasoning over a single pie chart. The model must rank categories by slice size (ascending or descending) without access to numeric labels.

\noindent  \textbf{Construction.} Sample $k$ categories with percentages that meet a strict relative gap; optionally, derive consistent integer counts for provenance. Render a legend-only chart.

\noindent  \textbf{Variants.} Two variants induced by the crossing of sort direction (ascending/descending, 50/50).

\noindent  \textbf{Complexity.} The number of $k$ categories is used to measure complexity.

\noindent  \textbf{Distractors.} Multiple-choice over comma-separated rankings; distractors come from swapping two positions in the true ordering (with random shuffles as a fallback to keep four unique options).

\subsubsection{Chart Comparison}
\noindent  \textbf{Problem.} Matching of proportions on two charts. A top chart (pie or bar) defines the color$\rightarrow$ percent mapping; the set of options comprises four options of the opposite chart type. Exactly one option preserves the mapping.

\noindent  \textbf{Construction.} Sample $k$ categories, distinct integer percentages for the categories that sum up to 100, and a distinct color palette. 

\noindent  \textbf{Variants.} Two display regimes with the top chart as a pie chart or a bar chart and the options as the opposite chart type.

\noindent  \textbf{Complexity.} We measure complexity with the number of $k$ categories in the charts.

\noindent  \textbf{Distractors.} Wrong options are produced by jittering and/or permuting the percentage vector. Candidates are admitted only if they pass absolute/relative difference thresholds and pairwise image-level distinctness checks.

\subsection{Counting}

\begin{figure*}[htp]
    \centering
    \begin{subfigure}{\textwidth}
        \centering
        \begin{tabular}{ccc}
            \includegraphics[width=0.3\textwidth]{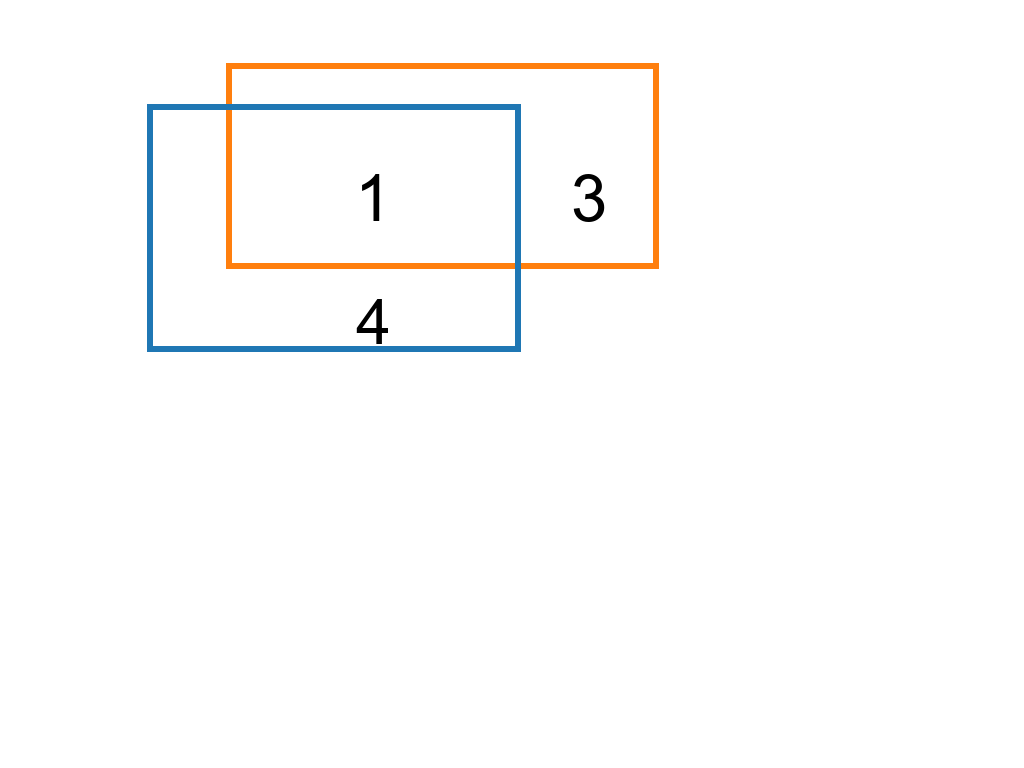} &
            \includegraphics[width=0.3\textwidth]{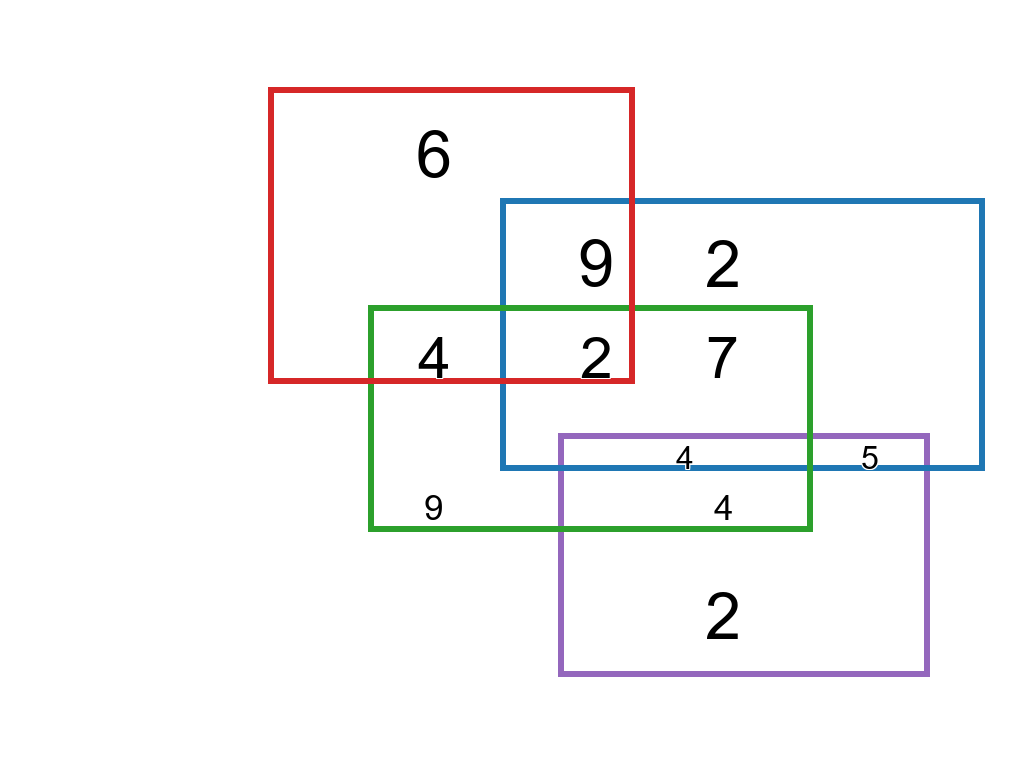} &
            \includegraphics[width=0.3\textwidth]{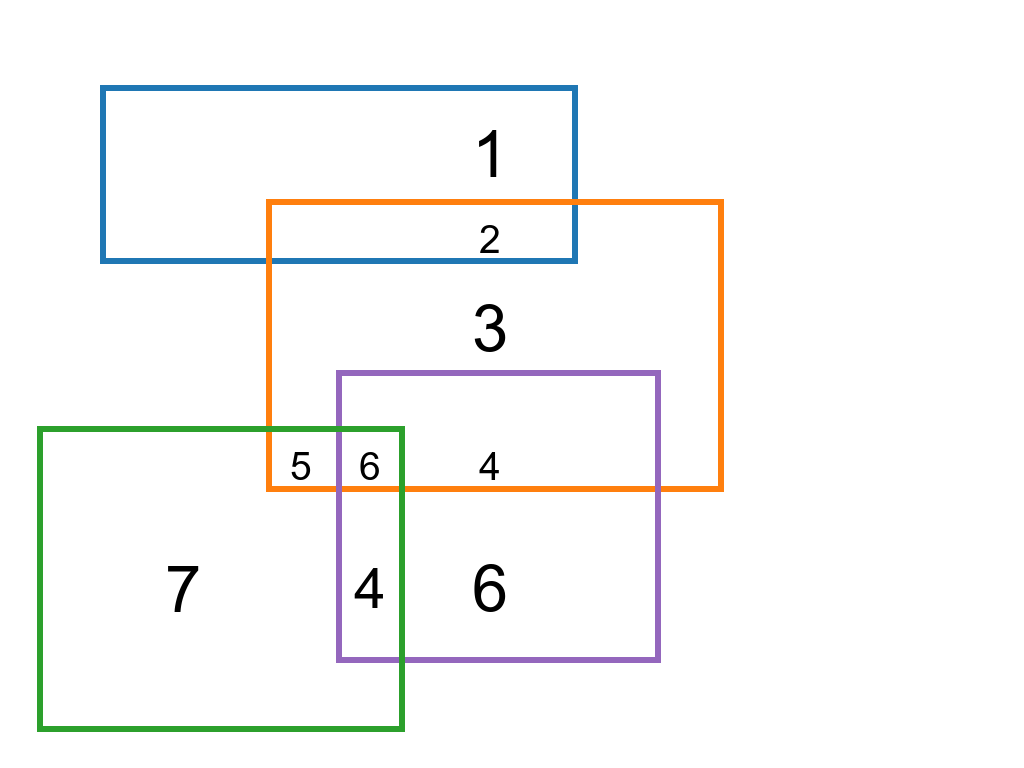} \\
        \end{tabular}
        \subcaption{\textbf{Venn Diagram} Calculate the sum of numbers in specified regions in the Venn diagram.}
        \label{fig:row_venn}
    \end{subfigure}

    \begin{subfigure}{\textwidth}
        \centering
        \begin{tabular}{ccc}
            \includegraphics[width=0.3\textwidth]{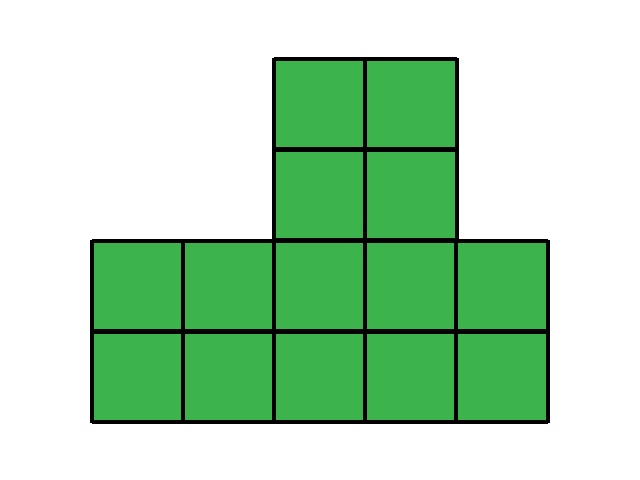} &
            \includegraphics[width=0.3\textwidth]{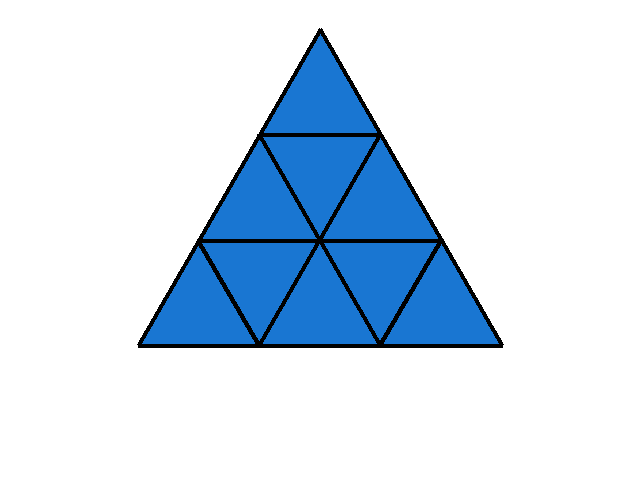} &
            \includegraphics[width=0.3\textwidth]{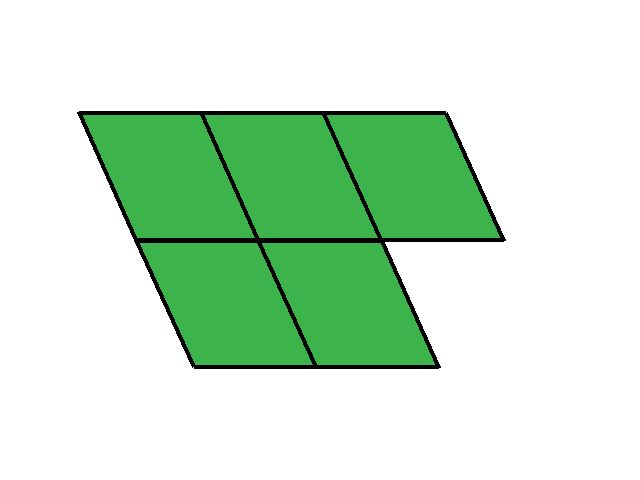} \\
        \end{tabular}
        \subcaption{\textbf{Shape Counting} Count the total number of a specific sub-shape within the larger figure.}
        \label{fig:row_shape_count}
    \end{subfigure}
    
    \begin{subfigure}{\textwidth}
        \centering
        \begin{tabular}{ccc}
            \includegraphics[width=0.25\textwidth]{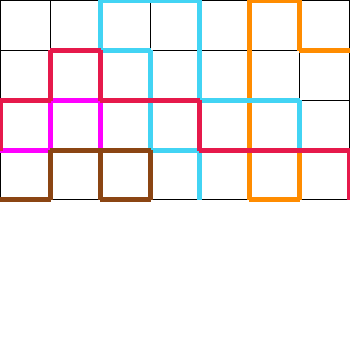} &
            \includegraphics[width=0.25\textwidth]{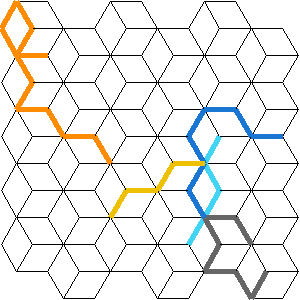} &
            \includegraphics[width=0.25\textwidth]{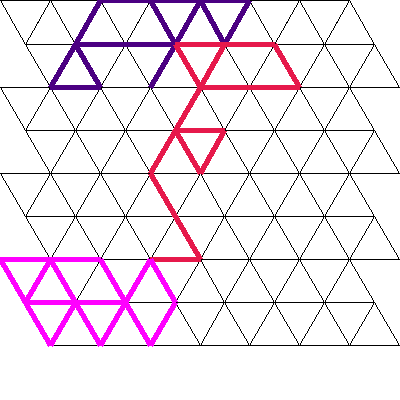} \\
        \end{tabular}
        \subcaption{\textbf{Tiles Line Length} Count the number of tile edges that make up the highlighted line.}
        \label{fig:row_line_length}
    \end{subfigure}

    \begin{subfigure}{\textwidth}
        \centering
        \begin{tabular}{ccc}
            \includegraphics[width=0.25\textwidth]{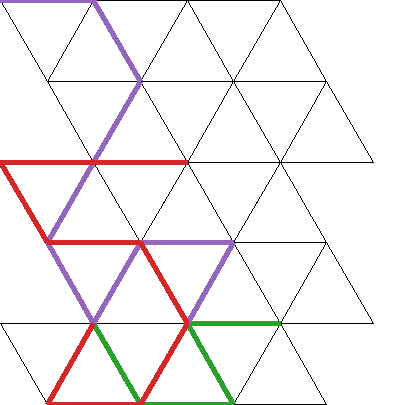} &
            \includegraphics[width=0.25\textwidth]{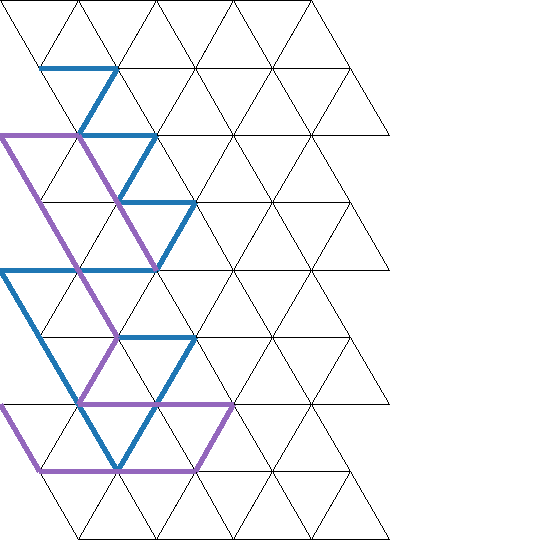} &
            \includegraphics[width=0.25\textwidth]{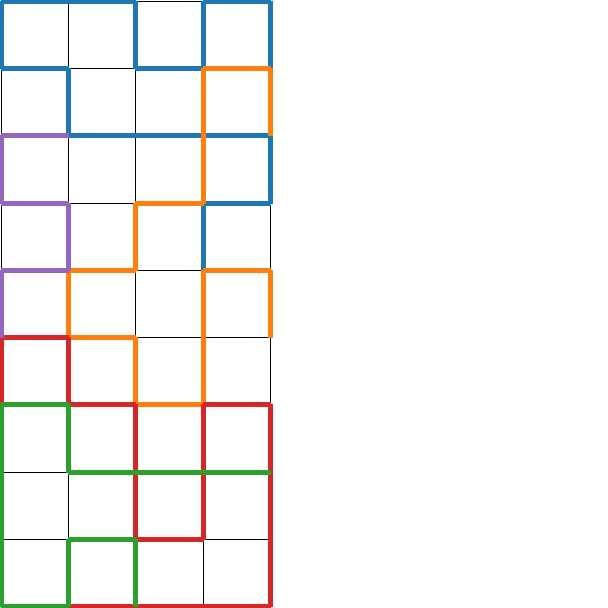} \\
        \end{tabular}
        \subcaption{\textbf{Tiles Line Intersections} Count the number of points where the two colored lines intersect.}
        \label{fig:row_line_intersect}
    \end{subfigure}

    \begin{subfigure}{\textwidth}
        \centering
        \begin{tabular}{ccc}
            \includegraphics[width=0.3\textwidth]{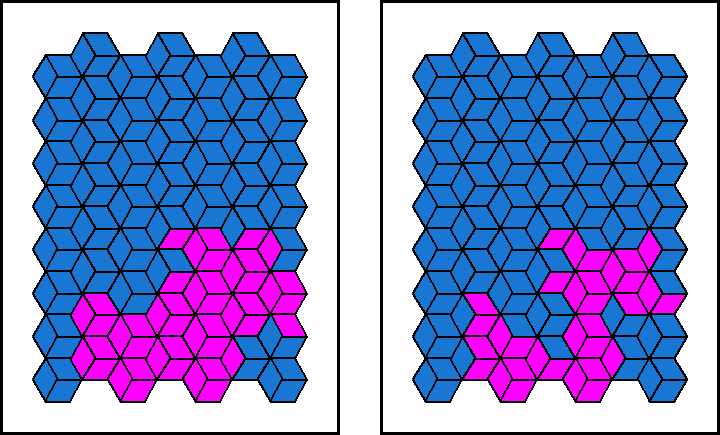} &
            \includegraphics[width=0.3\textwidth]{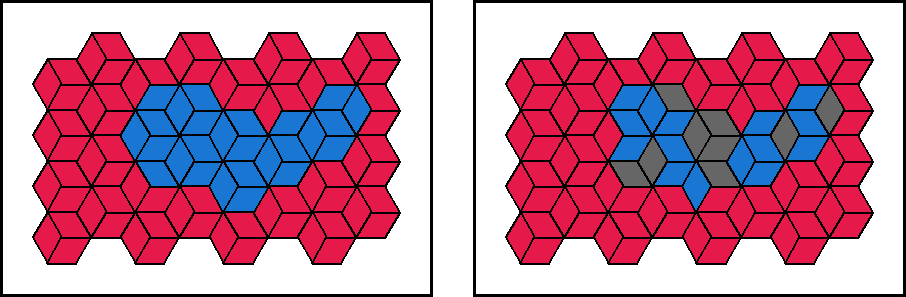} &
            \includegraphics[width=0.3\textwidth]{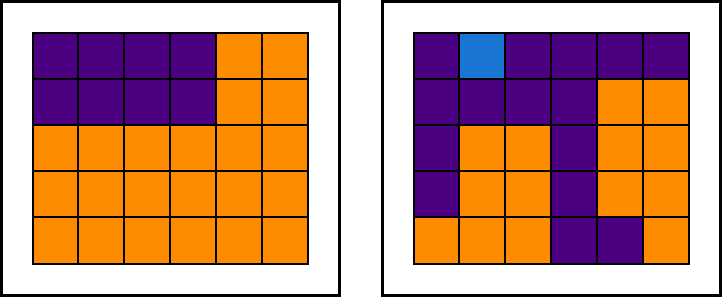} \\
        \end{tabular}
        \subcaption{\textbf{Tiles Recoloring} Count the number of cells that have different colors between the two boards.}
        \label{fig:row_recoloring}
    \end{subfigure}

    \caption{Examples of Counting tasks.}
    \label{fig:grid_counting}
\end{figure*}

Figure~\ref{fig:grid_counting} shows examples of this type of task.

\subsubsection{Venn Diagram}
\noindent \textbf{Problem.} Inclusion/exclusion over axis-aligned shapes with per-region numeric labels.

\noindent \textbf{Construction.} Sample 2-4 axes-aligned rectangles with a connected union. Induce a partition grid, place one integer in each non-empty atomic region (with skinny-region fallbacks), and pose include/exclude queries whose truth set uniquely determines the sum.

\noindent \textbf{Variants.} Four query modes sampled from (“only”, “intersection”, “butnot”, “union”); include/exclude masks are resampled until the target region is unique and non-empty.

\noindent \textbf{Complexity.} We measure complexity with the number of rectangles.

\noindent \textbf{Answer type.} Integer sum.

\subsubsection{Shape Counting}
\noindent  \textbf{Problem.} Counting of sub-shapes (rectangles, squares, triangles, parallelograms) within a single connected figure.

\noindent \textbf{Construction.} Draw one connected figure using one of several generators (axis-aligned polyomino, skewed poly-parallelogram, irregular/regular grids, staircase, triangular lattice, inscribed overlay). Render on a plain white background and compute the ground-truth count using exact combinatorial routines matched to the generator.

\noindent \textbf{Variants.} Eleven generator families (as above), each paired with appropriate query types. Instances are only emitted when the computed answer lies within configured bounds.

\noindent \textbf{Complexity.} The number of shapes in a figure (answer).

\noindent \textbf{Answer type.} Integer count.

\subsubsection{Tiles Line Length}
\noindent  \textbf{Problem.} Edge-step length estimation for a highlighted colored polyline.

\noindent \textbf{Construction.} On a chosen tiling, sample $k$ non-overlapping polylines, record their lengths, and ask for the length of one specified by color.

\noindent \textbf{Variants.} Four different types of tilings (square, triangular, hexagonal, rhombille).

\noindent \textbf{Complexity.} The number of cells in the tiling.

\noindent \textbf{Answer type.} Integer length.

\subsubsection{Tiles Line Intersections}
\textbf{Problem.} Intersection counting over colored polylines constrained to tile edges.

\noindent \textbf{Construction.} Build a vertex graph for the selected tiling; lay out $k$ vertex-simple polylines with distinct colors and no shared edges. 

\noindent \textbf{Variants.} Two different types of tiling (square, triangular).

\noindent \textbf{Complexity.} The number of intersections measures difficulty.

\noindent \textbf{Answer type.} Integer number of shared vertices (including endpoints).

\subsubsection{Tiles Recoloring}
\textbf{Problem.} Cell-wise recoloring/difference counting between two related boards.

\noindent \textbf{Construction.} Grow a connected region on the left board; derive the right board by adding/removing a connected set (same-color variant) or additionally recoloring overlap (color-change variant). 

\noindent \textbf{Variants.} Two variants - same color vs. color change - across several tiling families.

\noindent \textbf{Complexity.} The number of different cells measures the difficulty.

\noindent \textbf{Answer type.} Integer number of differing cells.

\subsection{Symmetry \& Pattern Recognition}

\begin{figure*}[t]
    \centering
    \begin{subfigure}{\textwidth}
        \centering
        \begin{tabular}{ccc}
            \includegraphics[width=0.3\textwidth]{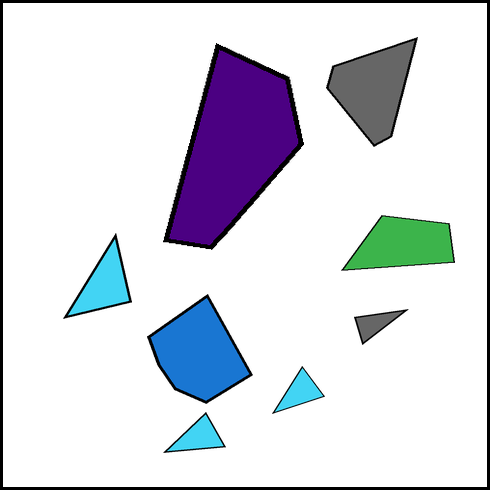} &
            \includegraphics[width=0.3\textwidth]{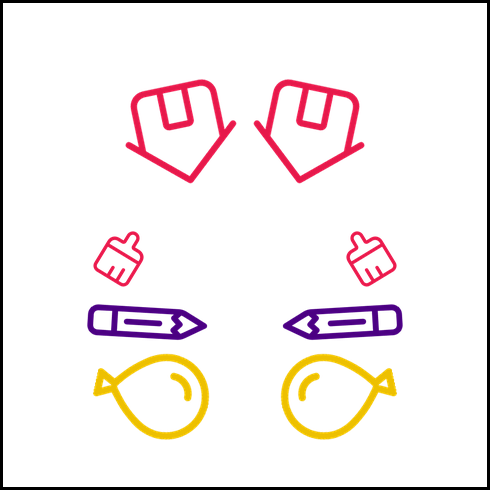} &
            \includegraphics[width=0.3\textwidth]{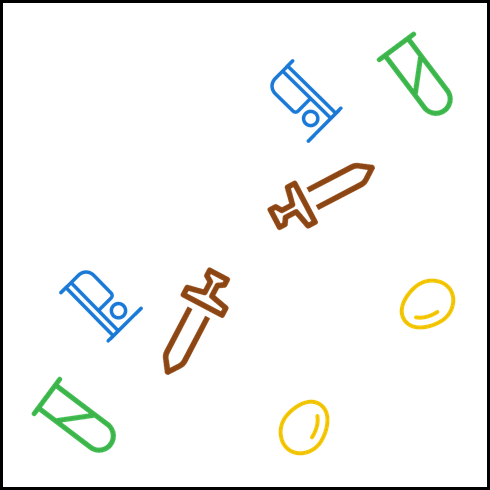} \\
        \end{tabular}
        \subcaption{\textbf{Mirror Identification} Identify the axis of mirror symmetry in the image if there is one.}
        \label{fig:row_mirror_id}
    \end{subfigure}

    \vspace{2em}
    \begin{subfigure}{\textwidth}
        \centering
        \begin{tabular}{ccc}
            \includegraphics[width=0.3\textwidth]{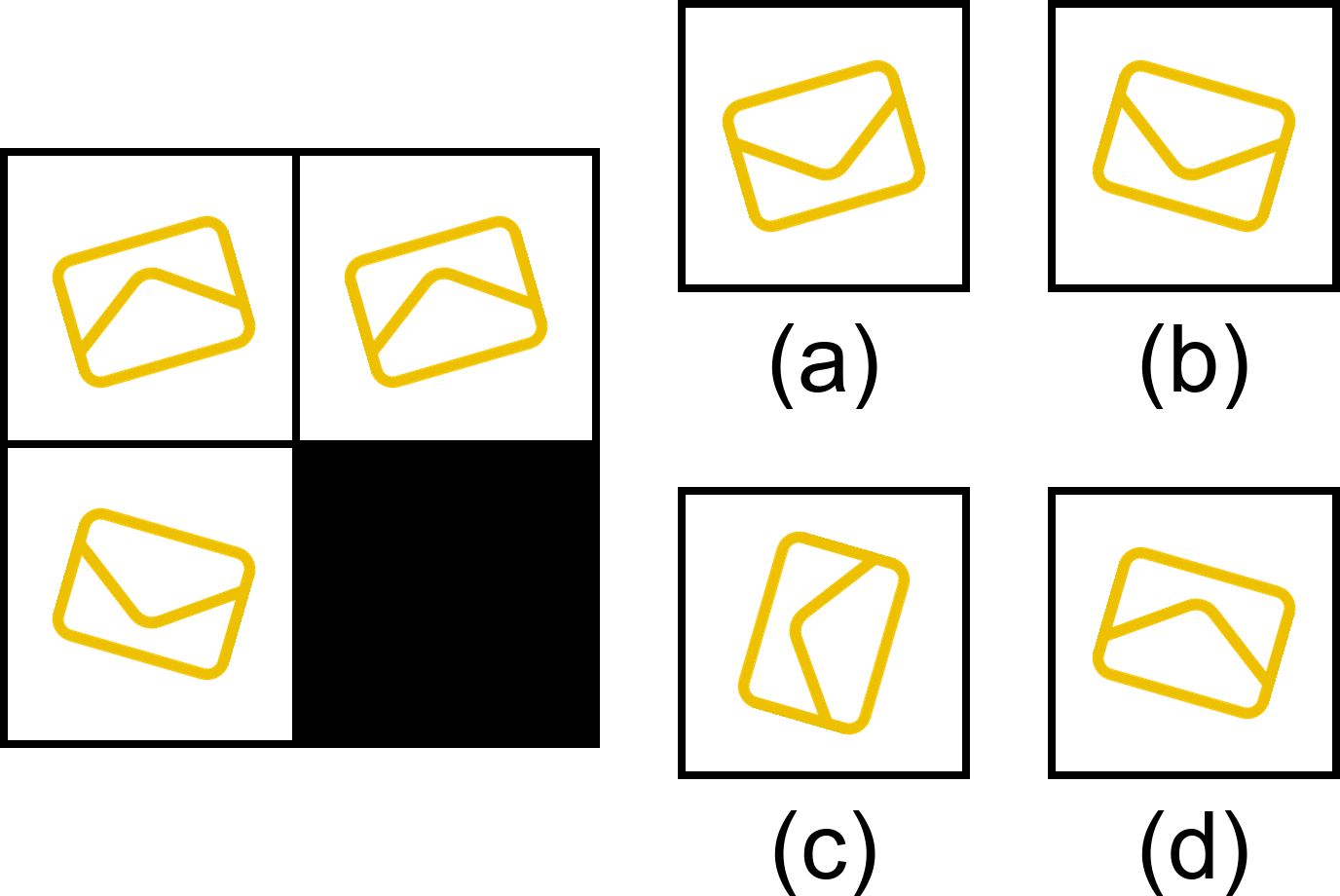} &
            \includegraphics[width=0.3\textwidth]{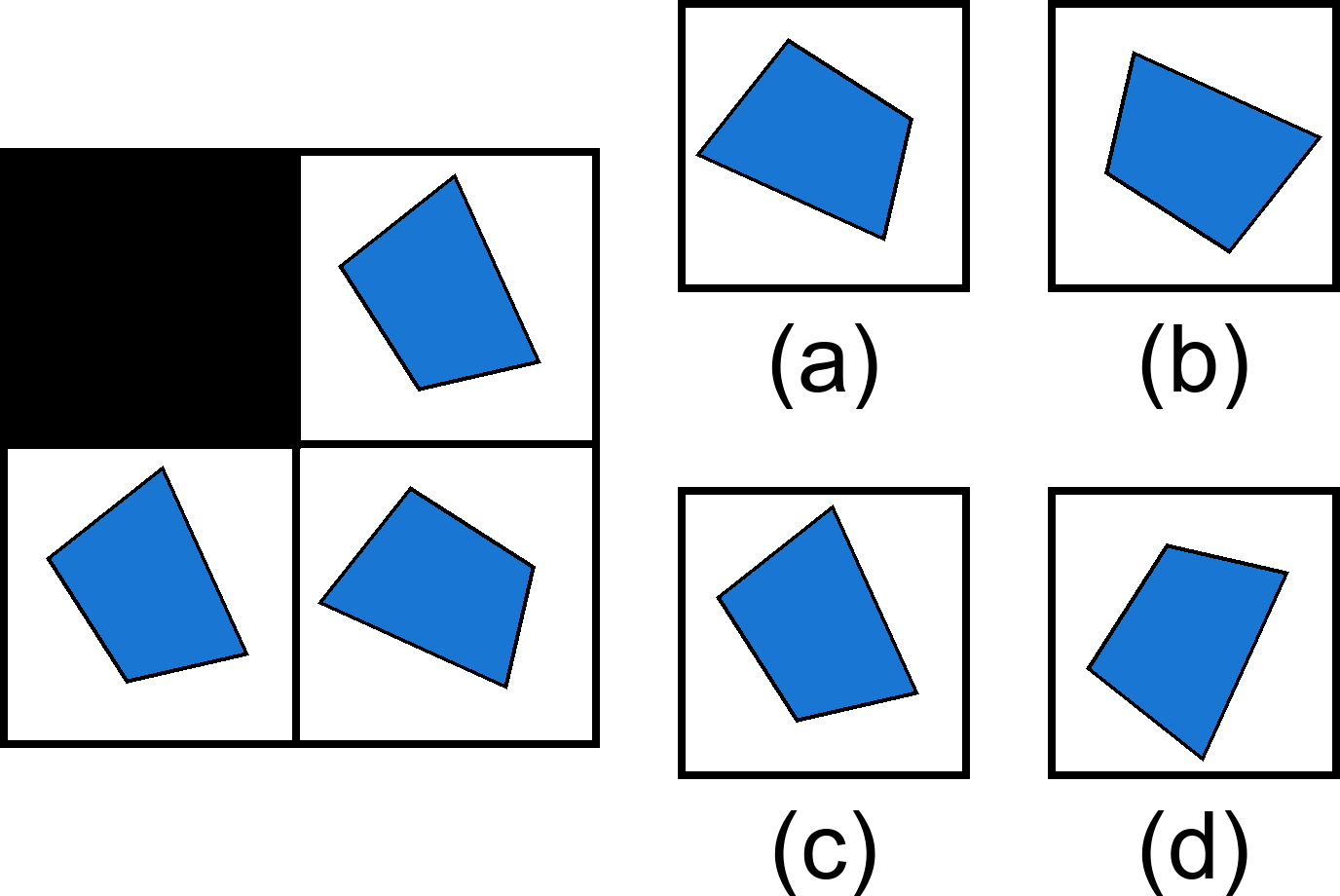} &
            \includegraphics[width=0.3\textwidth]{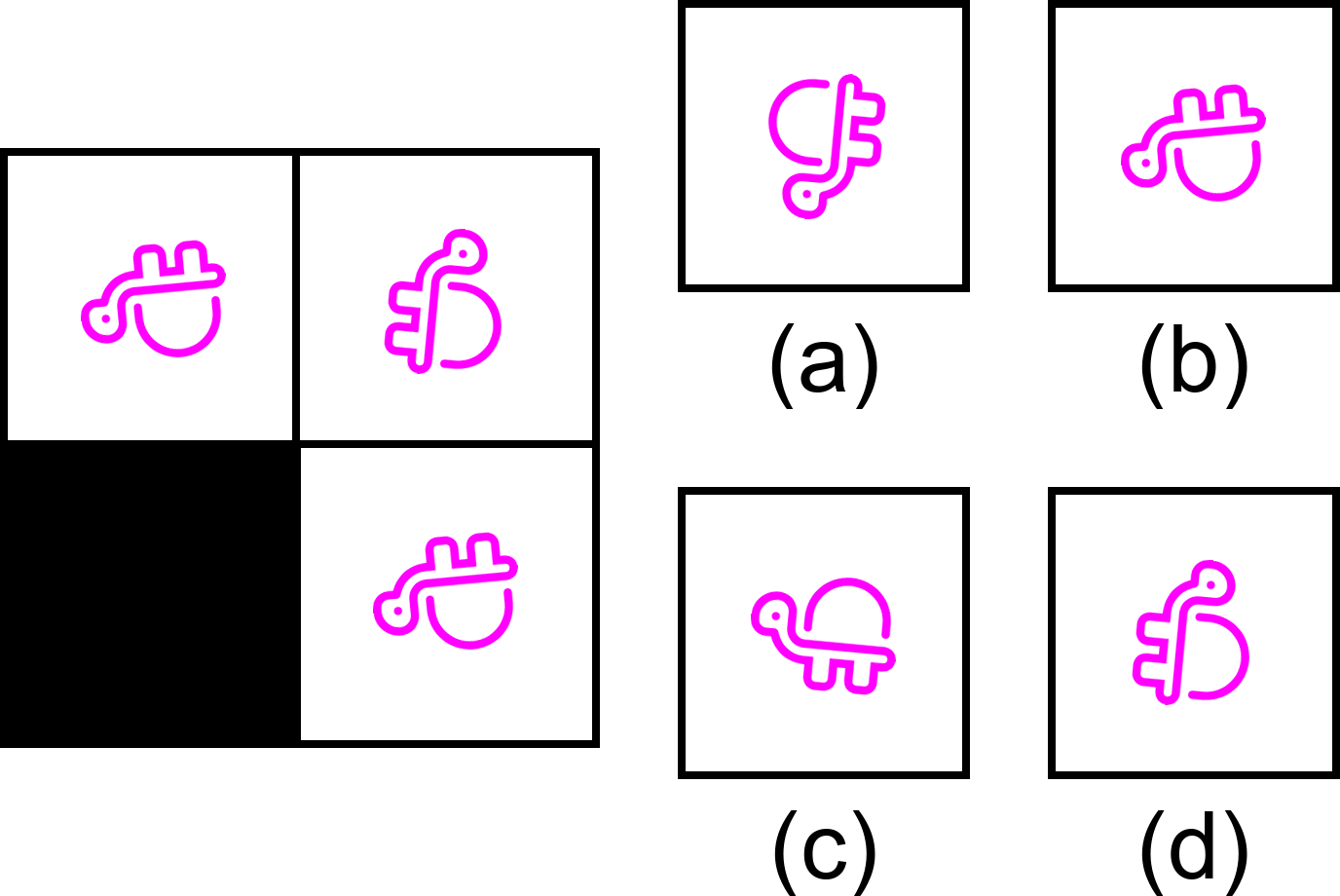} \\
        \end{tabular}
        \subcaption{\textbf{Symmetry Fill} Choose the tile that completes the grid according to the specified symmetry rule.}
        \label{fig:row_symm_fill}
    \end{subfigure}

    \vspace{2em}

    \begin{subfigure}{\textwidth}
        \centering
        \begin{tabular}{ccc}
            \includegraphics[width=0.3\textwidth]{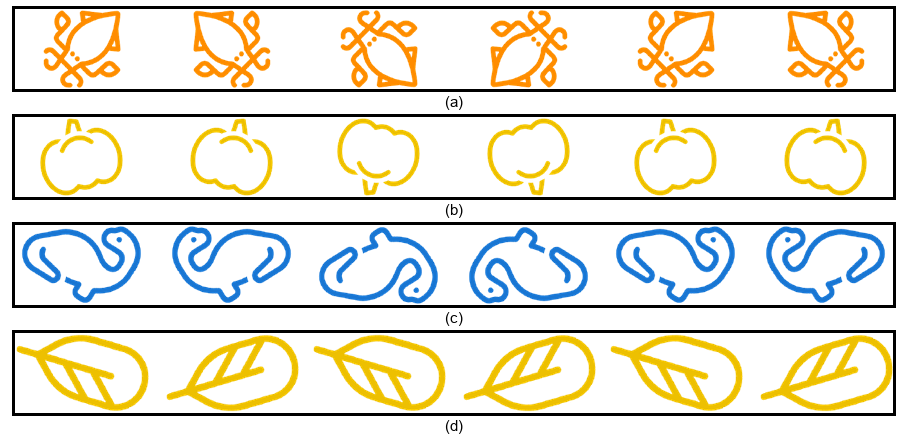} &
            \includegraphics[width=0.3\textwidth]{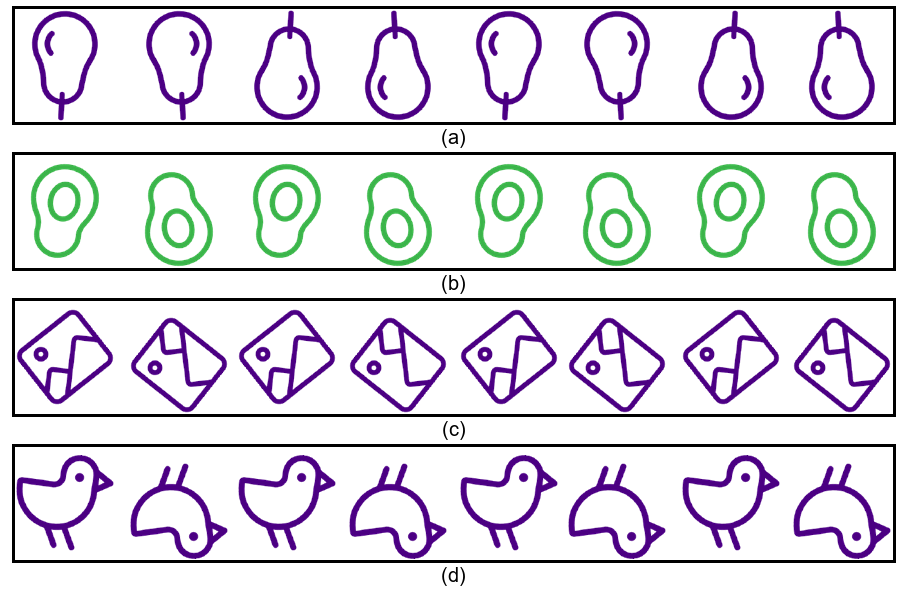} &
            \includegraphics[width=0.3\textwidth]{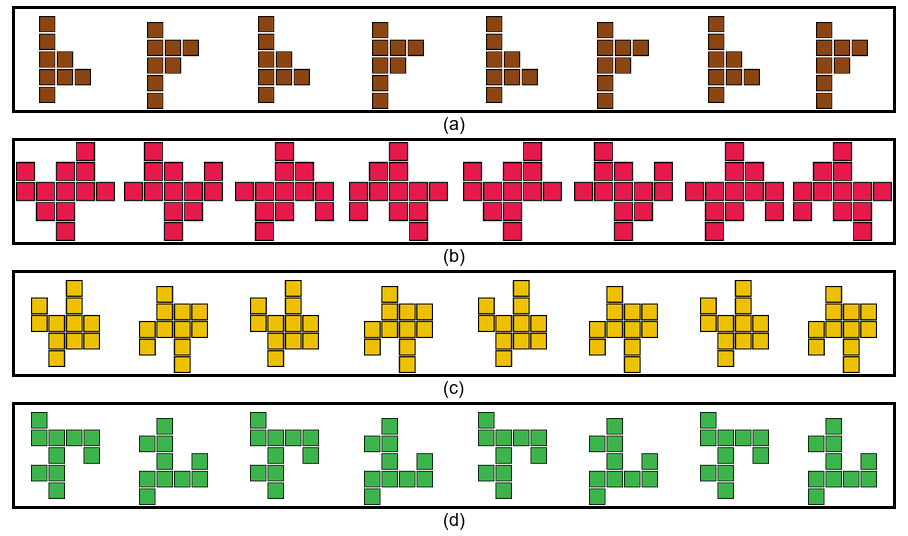} \\
        \end{tabular}
        \subcaption{\textbf{Frieze Groups} Identify which of the four patterns belongs to a different frieze symmetry group.}
        \label{fig:row_frieze}
    \end{subfigure}

    \vspace{2em}

    \begin{subfigure}{\textwidth}
        \centering
        \begin{tabular}{ccc}
            \includegraphics[width=0.3\textwidth]{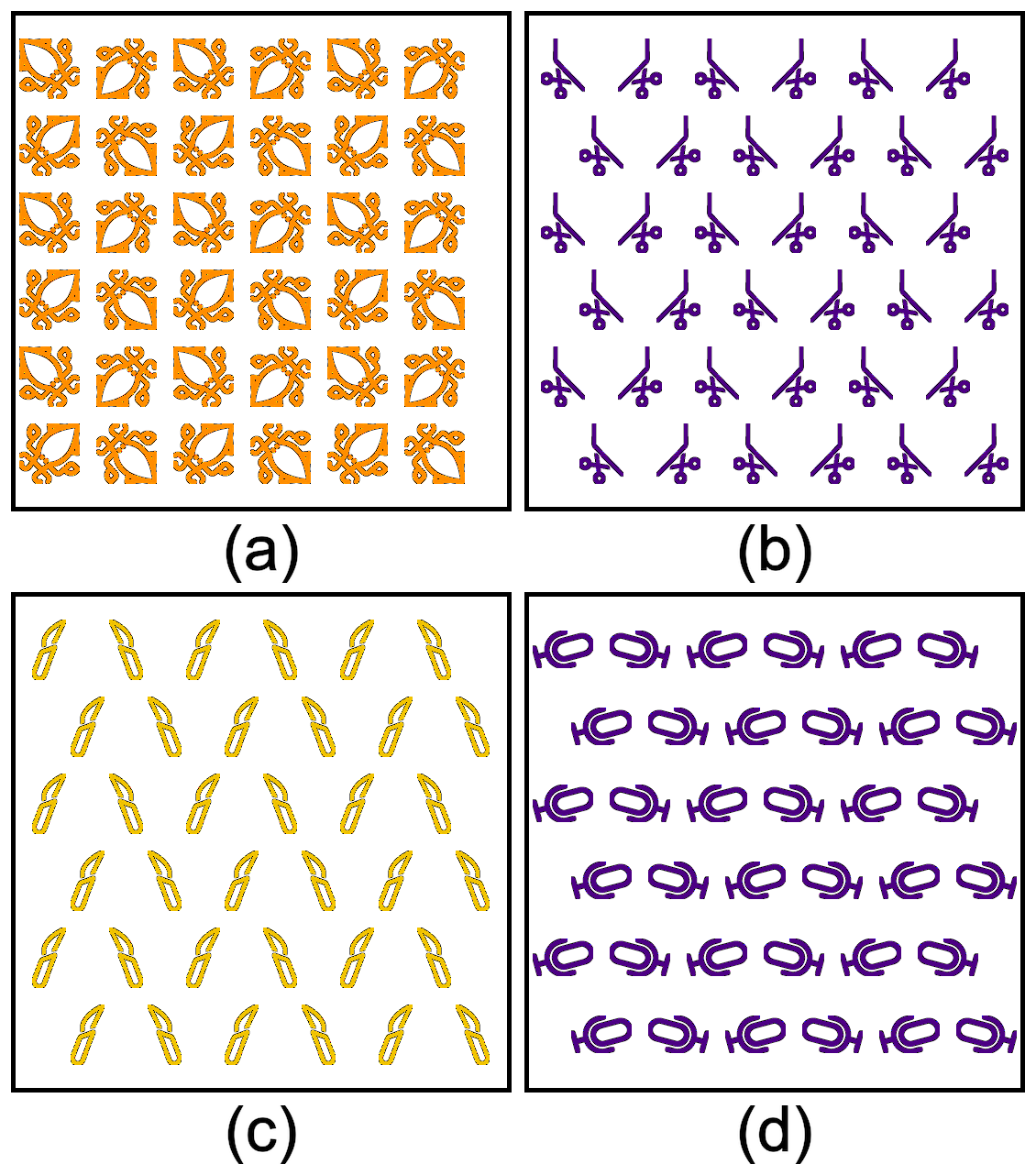} &
            \includegraphics[width=0.3\textwidth]{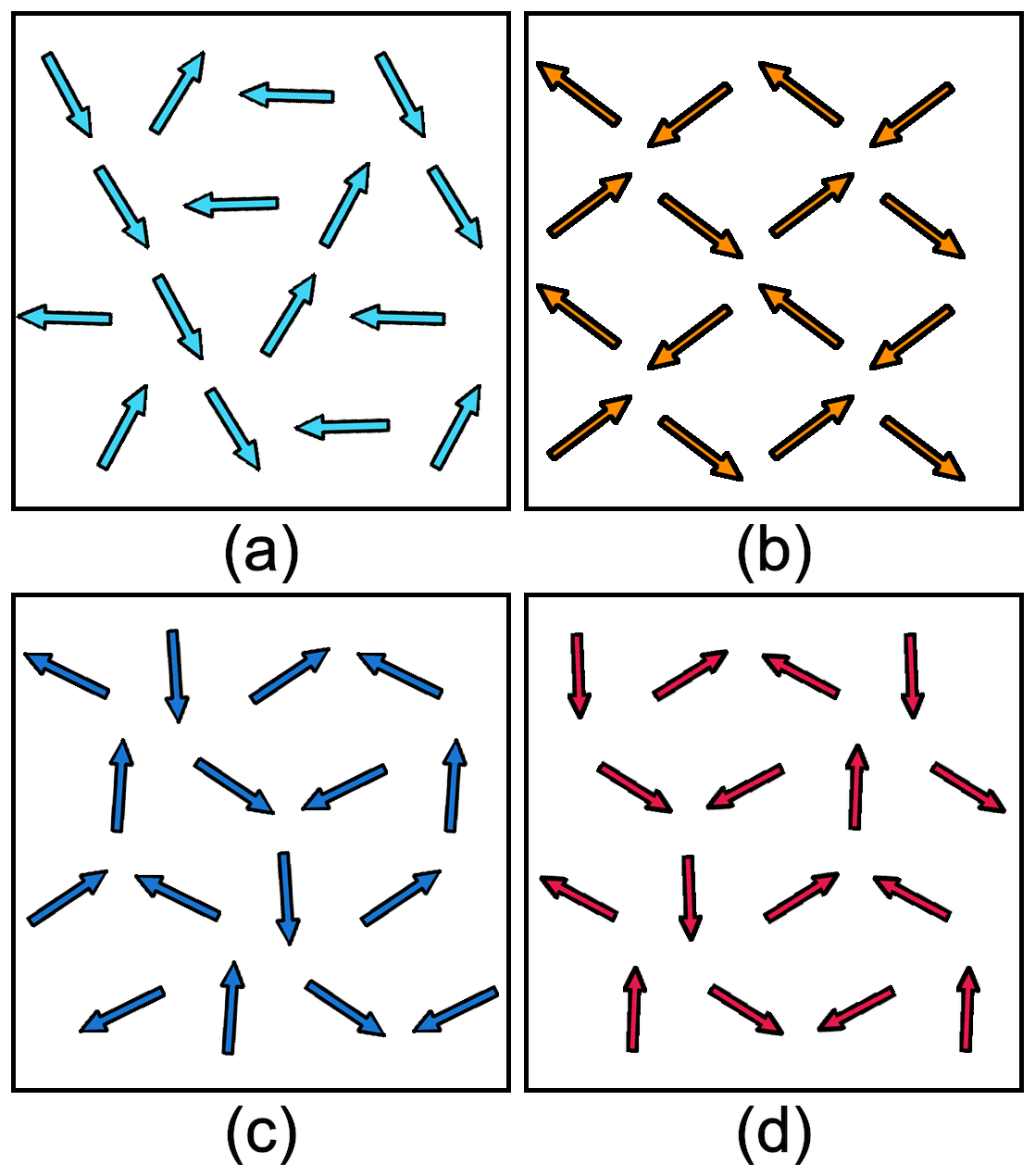} &
            \includegraphics[width=0.3\textwidth]{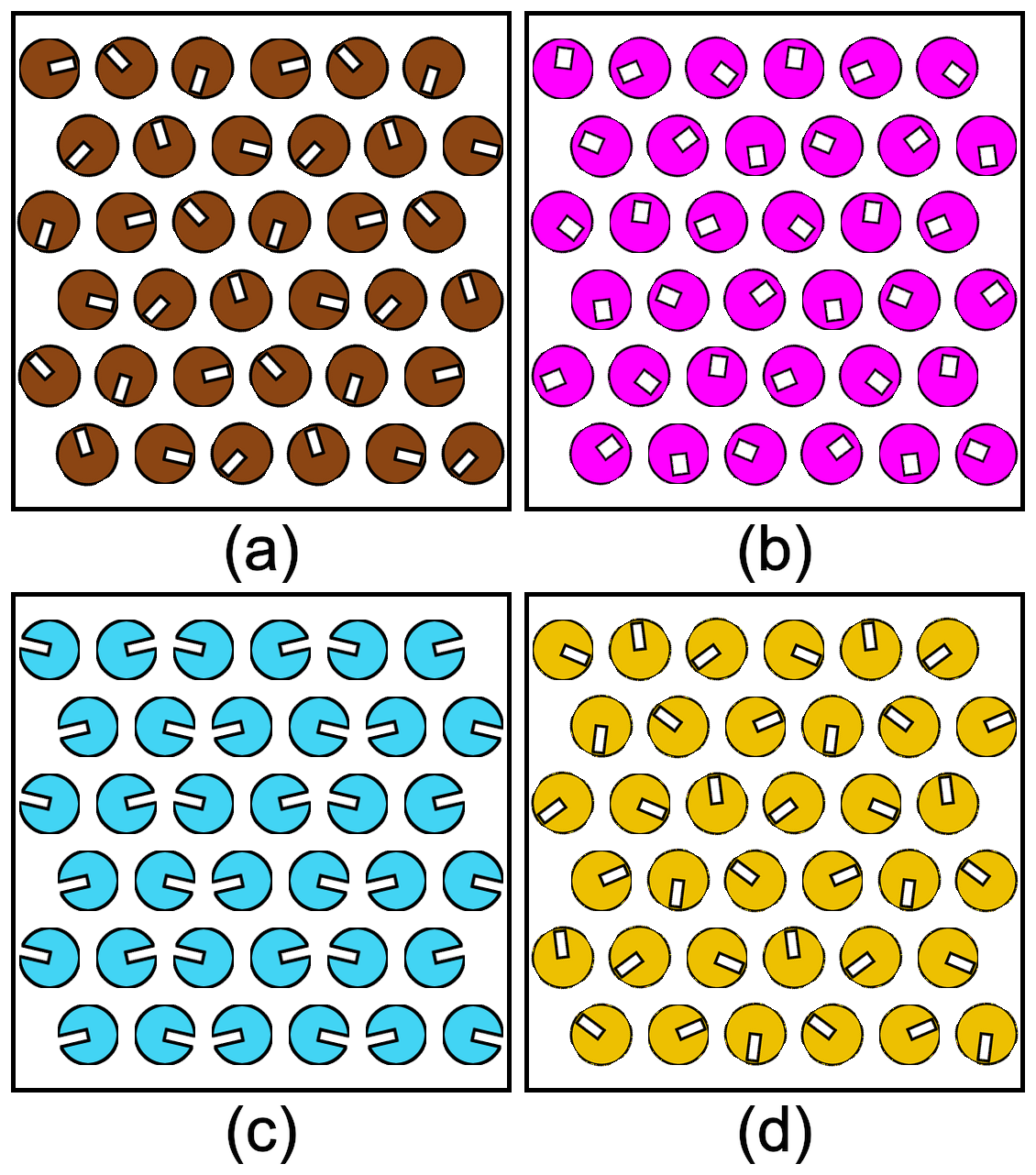} \\
        \end{tabular}
        \subcaption{\textbf{Wallpaper Groups} Identify which of the four patterns belongs to a different wallpaper symmetry group.}
        \label{fig:row_wallpaper}
    \end{subfigure}

    \caption{Examples of Symmetry tasks.}
    \label{fig:grid_symmetry}
\end{figure*}

Figure~\ref{fig:grid_symmetry} shows examples of this type of task.

\subsubsection{Mirror Identification}
\noindent\textbf{Problem.} Textual classification of mirror symmetry (including ``none'') for a composite scene.

\noindent\textbf{Construction.} Place motif instances inside class-specific fundamental regions to synthesize scenes. Verify the final bitmap's category via color-aware symmetry tests; pair with six textual options and shuffle.

\noindent\textbf{Variants.} Six labels - vertical, horizontal, main diagonal, anti-diagonal, vertical+horizontal, none - with target count and canvas scale adapted to the class.

\noindent  \textbf{Complexity.} Not measured.

\noindent\textbf{Distractors.} The five incorrect textual descriptions serve as distractors; all six labels are offered.

\subsubsection{Symmetry Fill}
\noindent\textbf{Problem.} Grid completion under a specified mirror constraint. A 2$\times$2 grid is shown with one missing tile; select the tile that restores the target symmetry.

\noindent\textbf{Construction.} Render a base tile, apply the rule (vertical, horizontal, both, main-diagonal, anti-diagonal) to fill the grid, remove one tile, and construct options by applying distinct transforms while enforcing pairwise distinctness.

\noindent\textbf{Variants.} Five rule keys as above; the missing position and motif vary.

\noindent  \textbf{Complexity.} Not measured.

\noindent\textbf{Distractors.} Transform pool filtered to retain only visually distinct candidates; select three and shuffle with the correct transform.

\subsubsection{Frieze Groups}
\noindent\textbf{Problem.} Odd-one-out identification among four horizontal strips, each generated from a frieze symmetry; three share the same neighbor rule, one differs.

\noindent\textbf{Construction.} Sample a motif family; choose a majority frieze group for three strips and a distinct group for the odd strip. Render with consistent spacing and label (a-d).

\noindent\textbf{Variants.} Six Conway frieze groups (step, sidle, jump, spinning hop, spinning sidle, spinning jump). The strip length and option order vary per instance.

\noindent  \textbf{Complexity.} Not measured.

\noindent\textbf{Distractors.} The distractors are simply additional strips from the majority frieze class; the odd class is unique by construction.

\subsubsection{Wallpaper Groups}
\noindent\textbf{Problem.} Odd-one-out among four 2D wallpaper patches; three are sampled from one wallpaper group and one from another.

\noindent\textbf{Construction.} Sample a motif family and wallpaper groups; generate patches under each group, crop to equal square tiles, and compose a labeled 2$\times$2 grid.

\noindent\textbf{Variants.} Seventeen IUC wallpaper groups; the majority/odd selection and the option order are randomized.

\noindent  \textbf{Complexity.} Not measured.

\noindent\textbf{Distractors.} The three majority-group patches form the distractor set by construction.

\subsection{Sequence \& Transformation Reasoning}

\begin{figure*}[htp]
    \centering
    \begin{subfigure}{\textwidth}
        \centering
        \begin{tabular}{ccc}
            \includegraphics[width=0.3\textwidth]{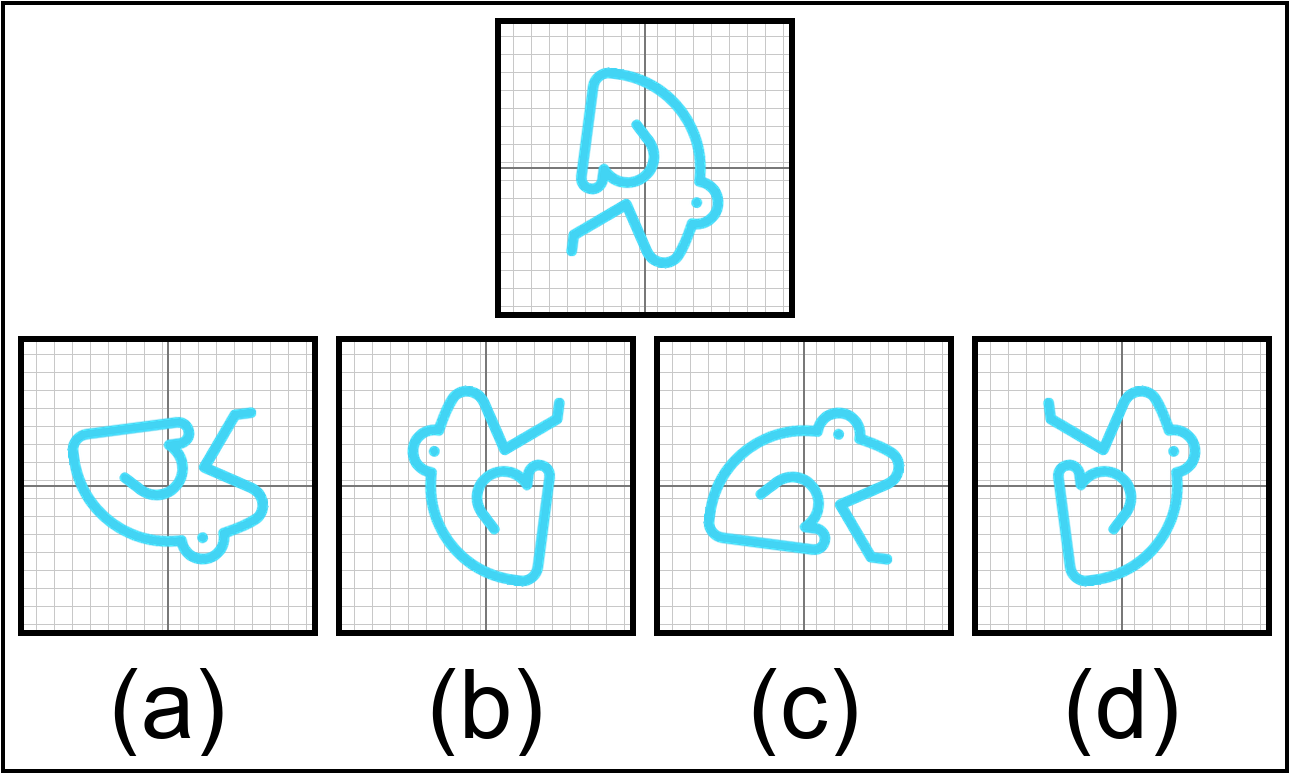} &
            \includegraphics[width=0.3\textwidth]{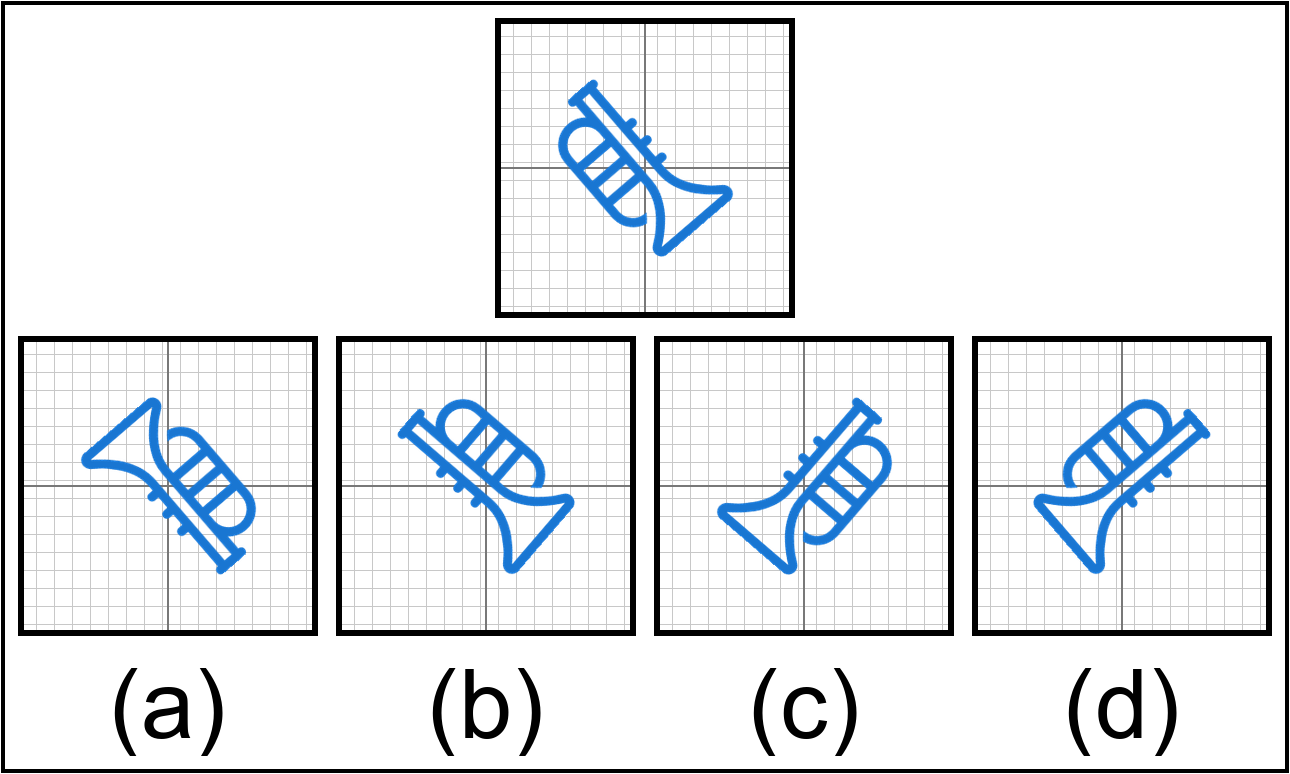} &
            \includegraphics[width=0.3\textwidth]{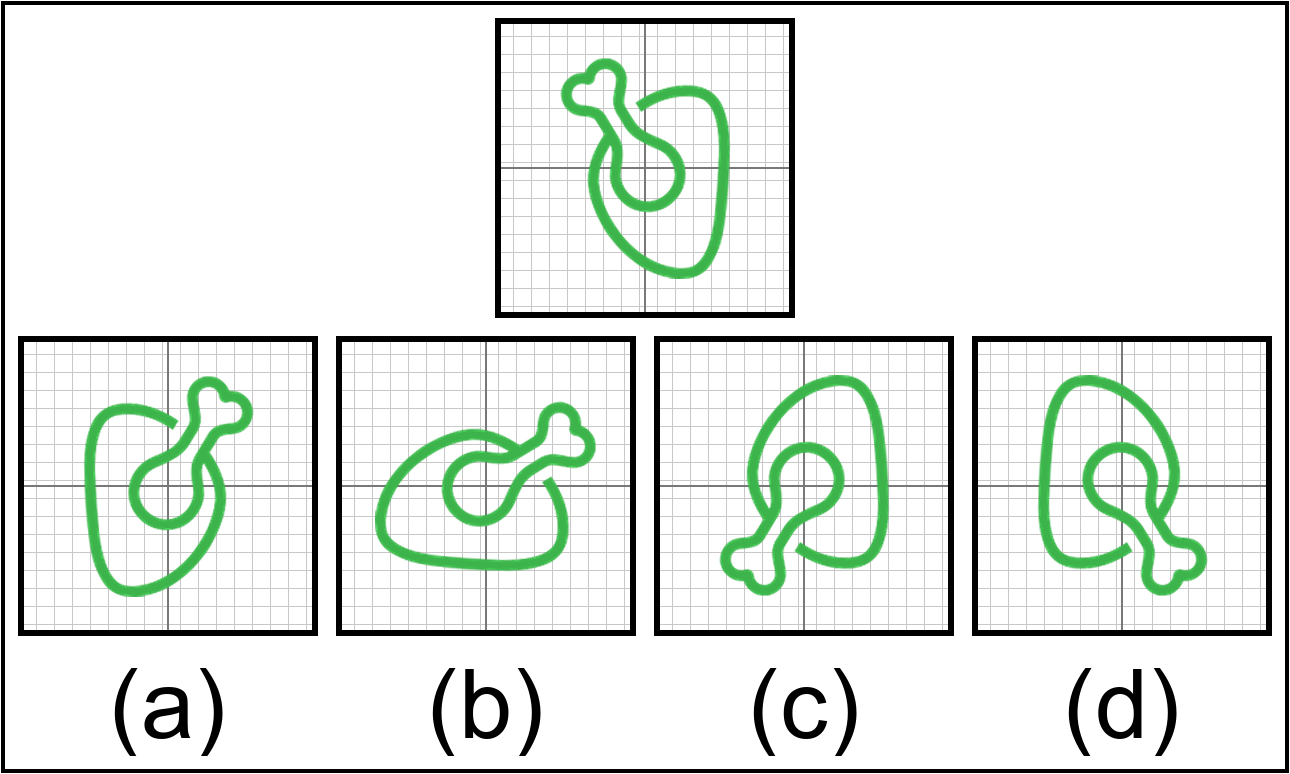} \\
        \end{tabular}

        \subcaption{\textbf{Transform Result Identify} Choose the image that shows the correct result of applying the given transformation.}
        \label{fig:row_trans_result}
    \end{subfigure}

    \vspace{1em}

    \begin{subfigure}{\textwidth}
        \centering
        \begin{tabular}{ccc}
            \includegraphics[width=0.3\textwidth]{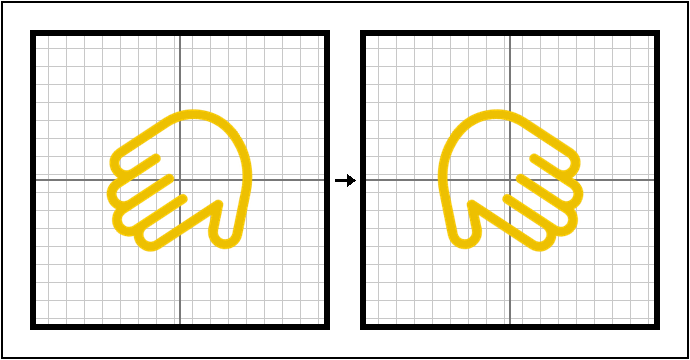} &
            \includegraphics[width=0.3\textwidth]{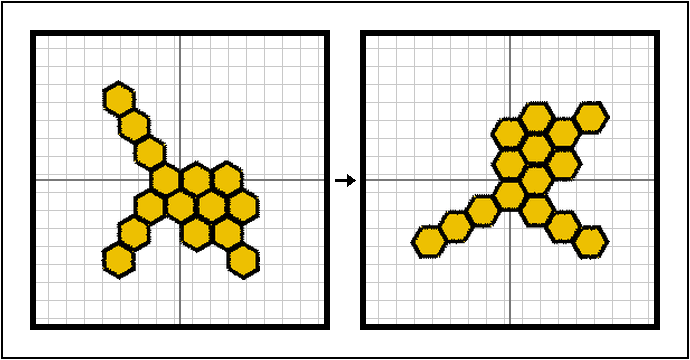} &
            \includegraphics[width=0.3\textwidth]{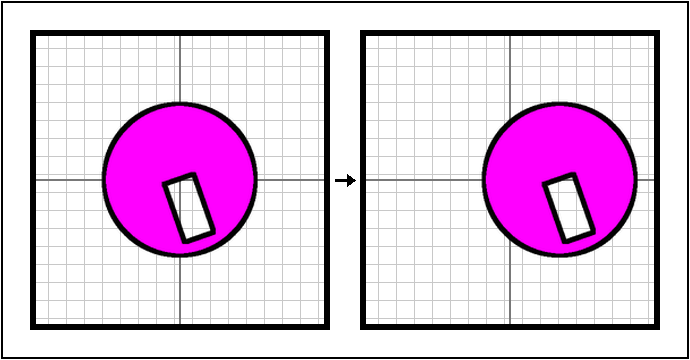} \\
        \end{tabular}
        \subcaption{\textbf{Transform Pair Infer} Identify the transformation that maps the left image to the right image.}
        \label{fig:row_trans_pair}
    \end{subfigure}

    \vspace{1em}

    \begin{subfigure}{\textwidth}
        \centering
        \begin{tabular}{ccc}
            \includegraphics[width=0.3\textwidth]{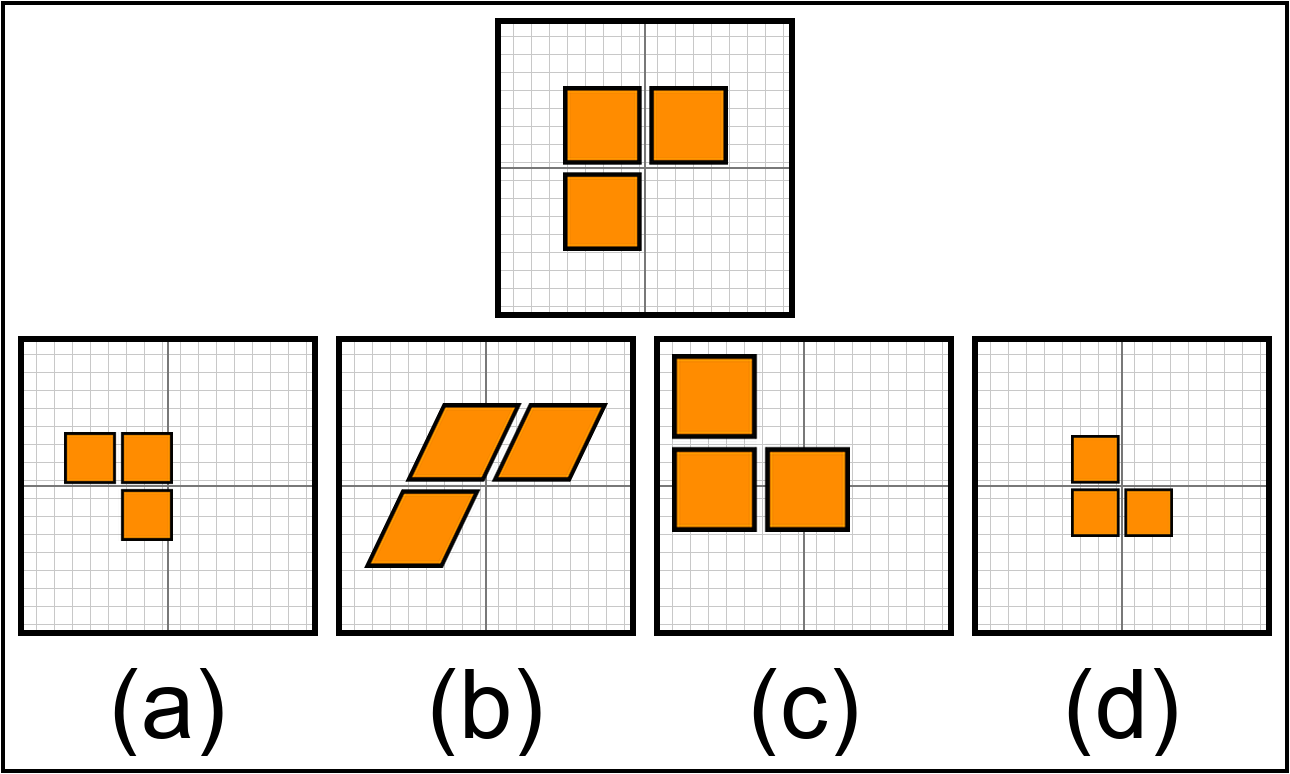} &
            \includegraphics[width=0.3\textwidth]{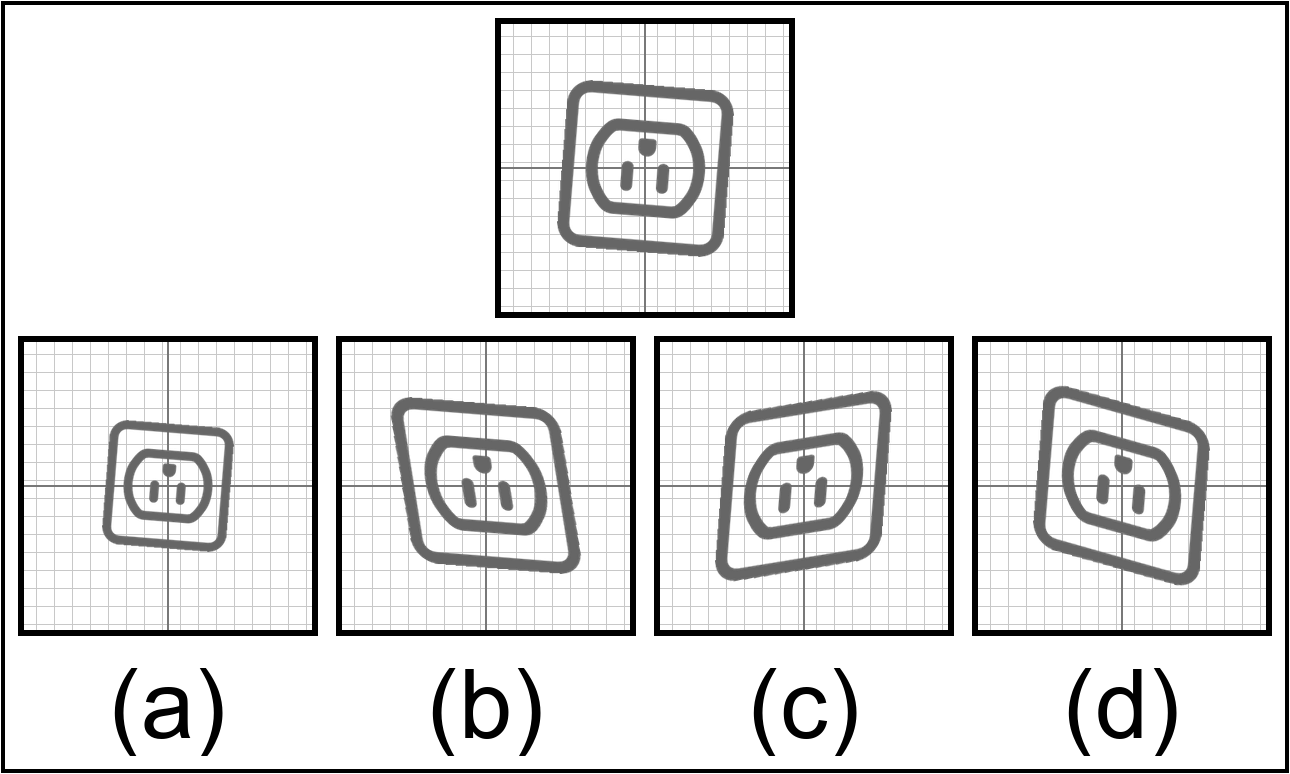} &
            \includegraphics[width=0.3\textwidth]{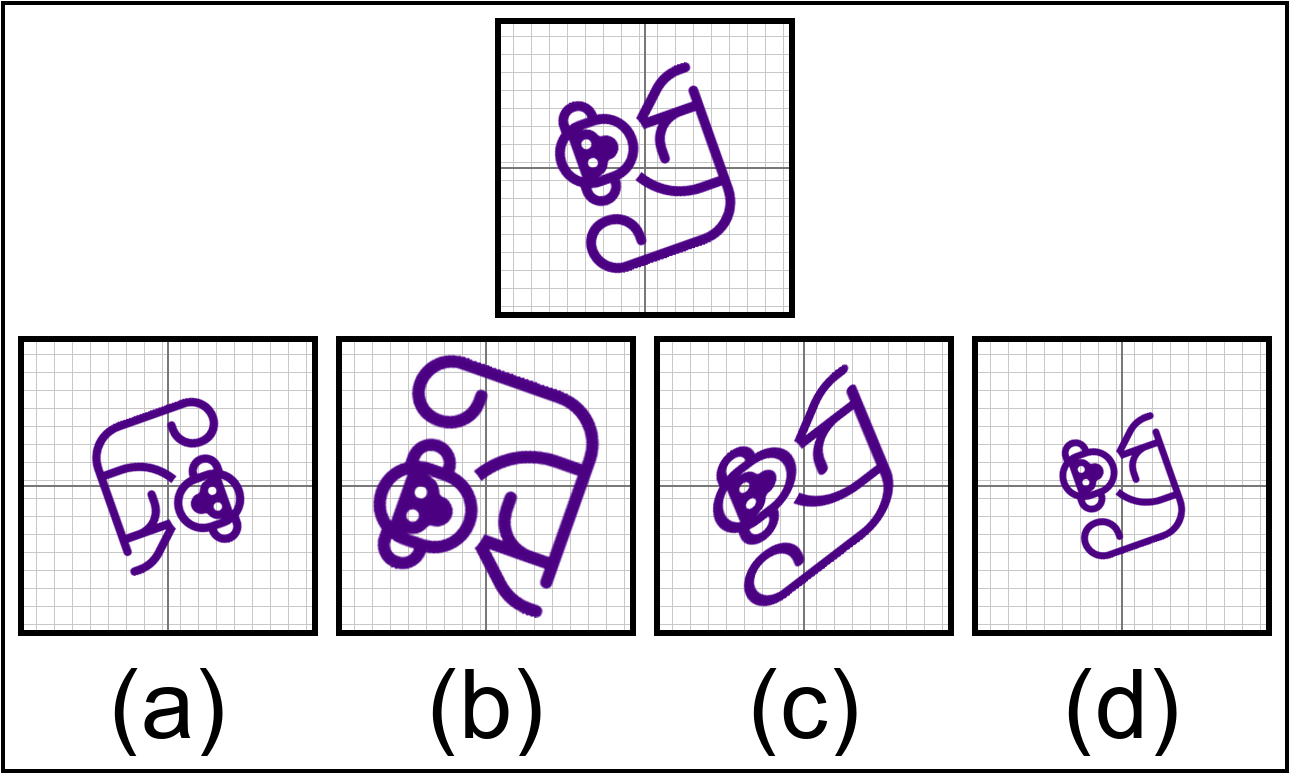} \\
        \end{tabular}
        \subcaption{\textbf{Transform Similarity Identify} Find the shape that is a similar version (rotated, scaled) of the target.}
        \label{fig:row_trans_sim}
    \end{subfigure}

    \vspace{1em}

    \begin{subfigure}{\textwidth}
        \centering
        \begin{tabular}{ccc}
            \includegraphics[width=0.3\textwidth]{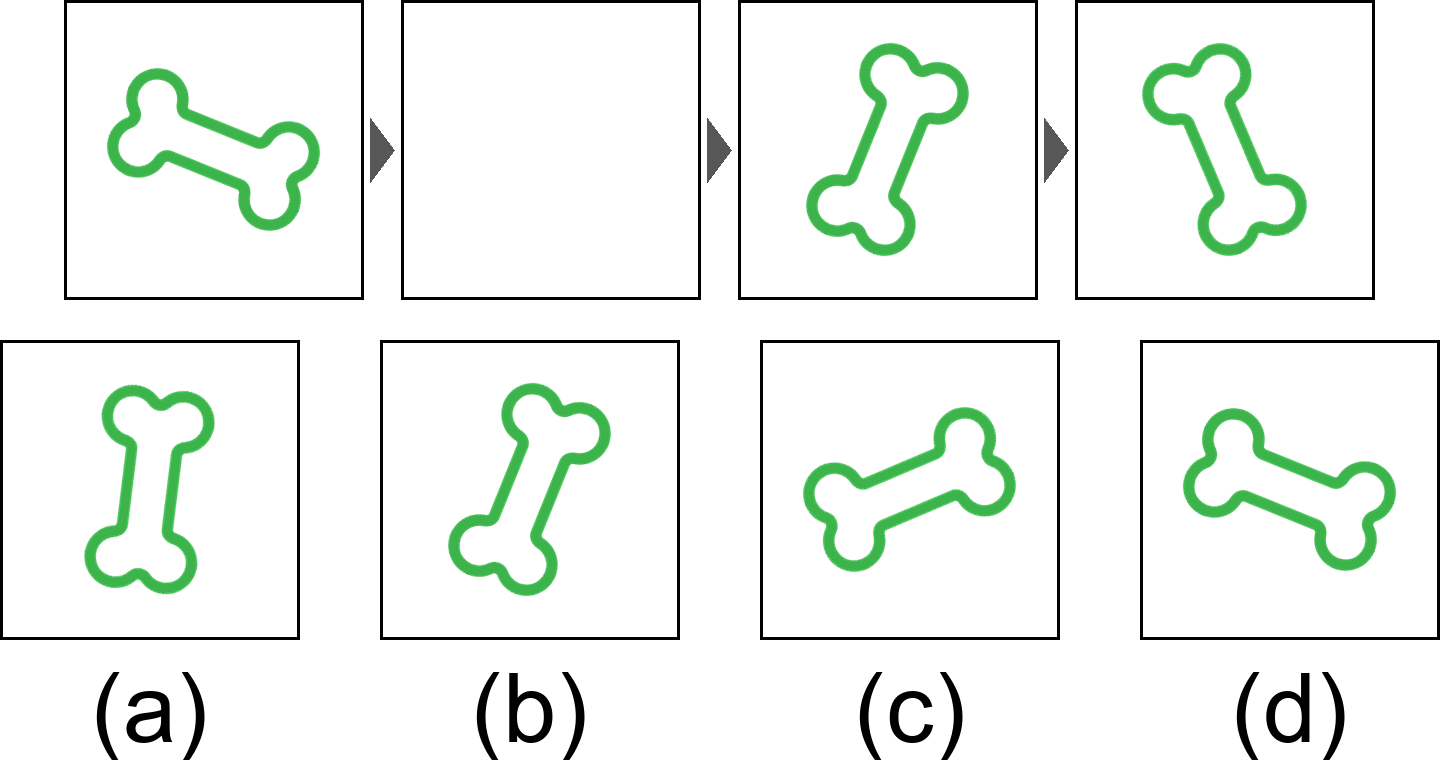} &
            \includegraphics[width=0.3\textwidth]{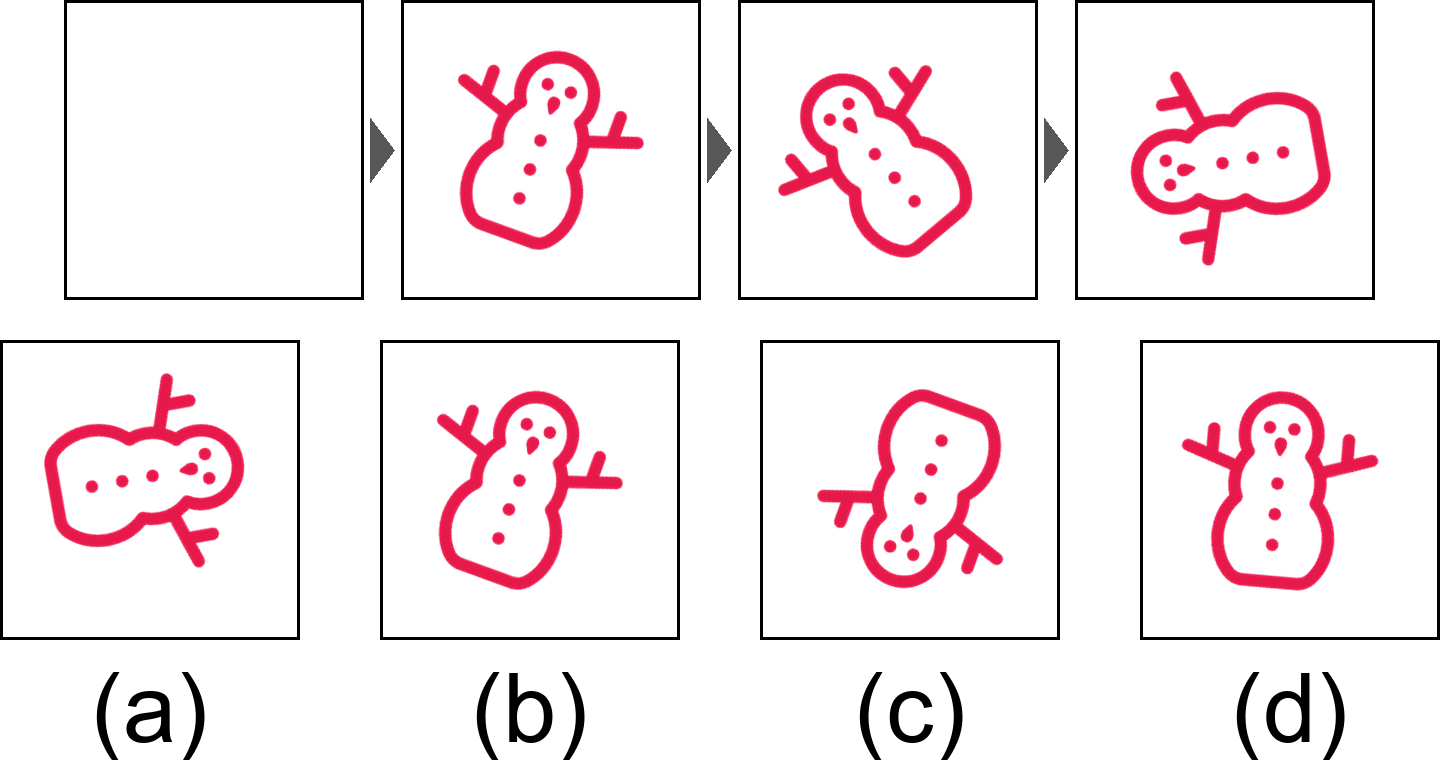} &
            \includegraphics[width=0.3\textwidth]{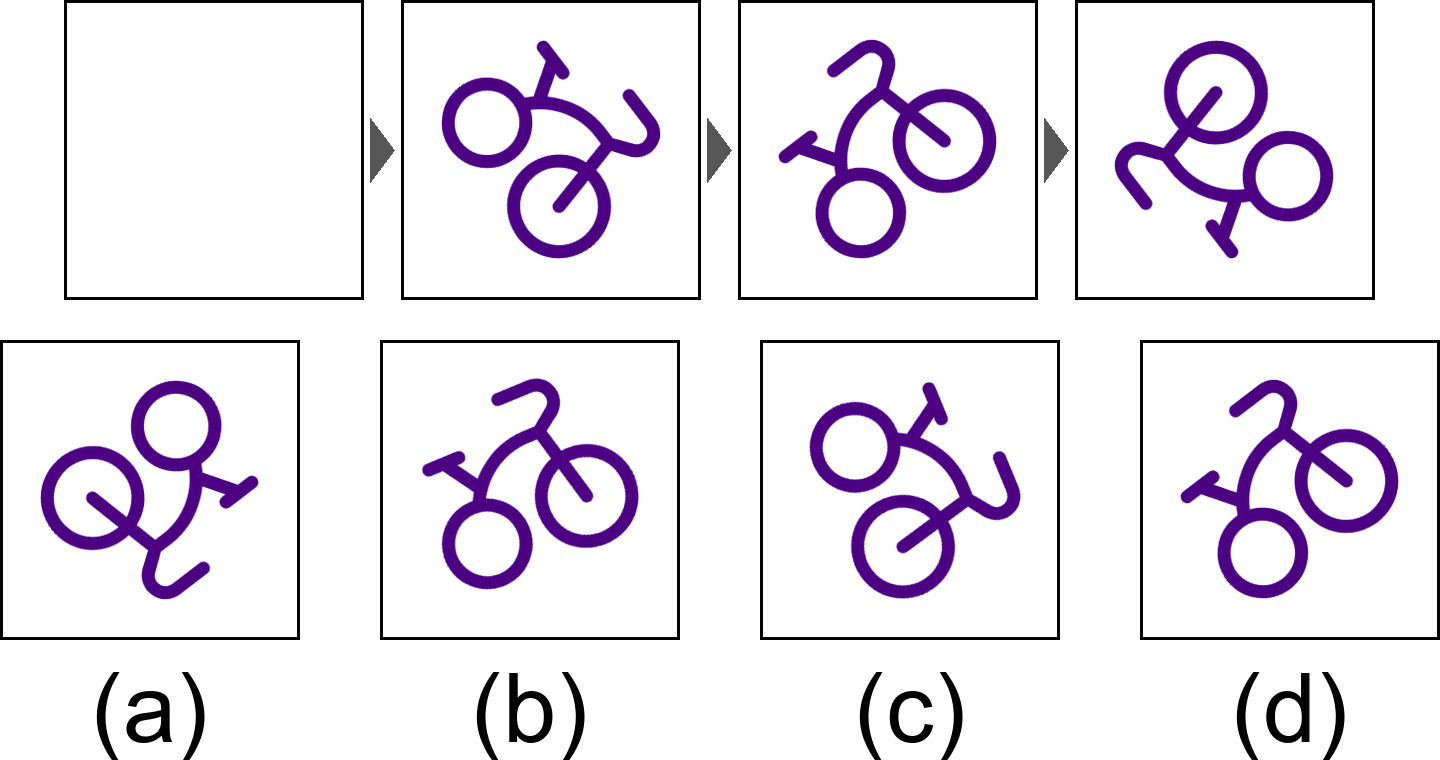} \\
        \end{tabular}
        \subcaption{\textbf{Sequence Rotation} Complete the sequence by finding the missing panel, which follows a constant rotation.}
        \label{fig:row_seq_rot}
    \end{subfigure}

    \begin{subfigure}{\textwidth}
        \centering
        \begin{tabular}{ccc}
            \includegraphics[width=0.3\textwidth]{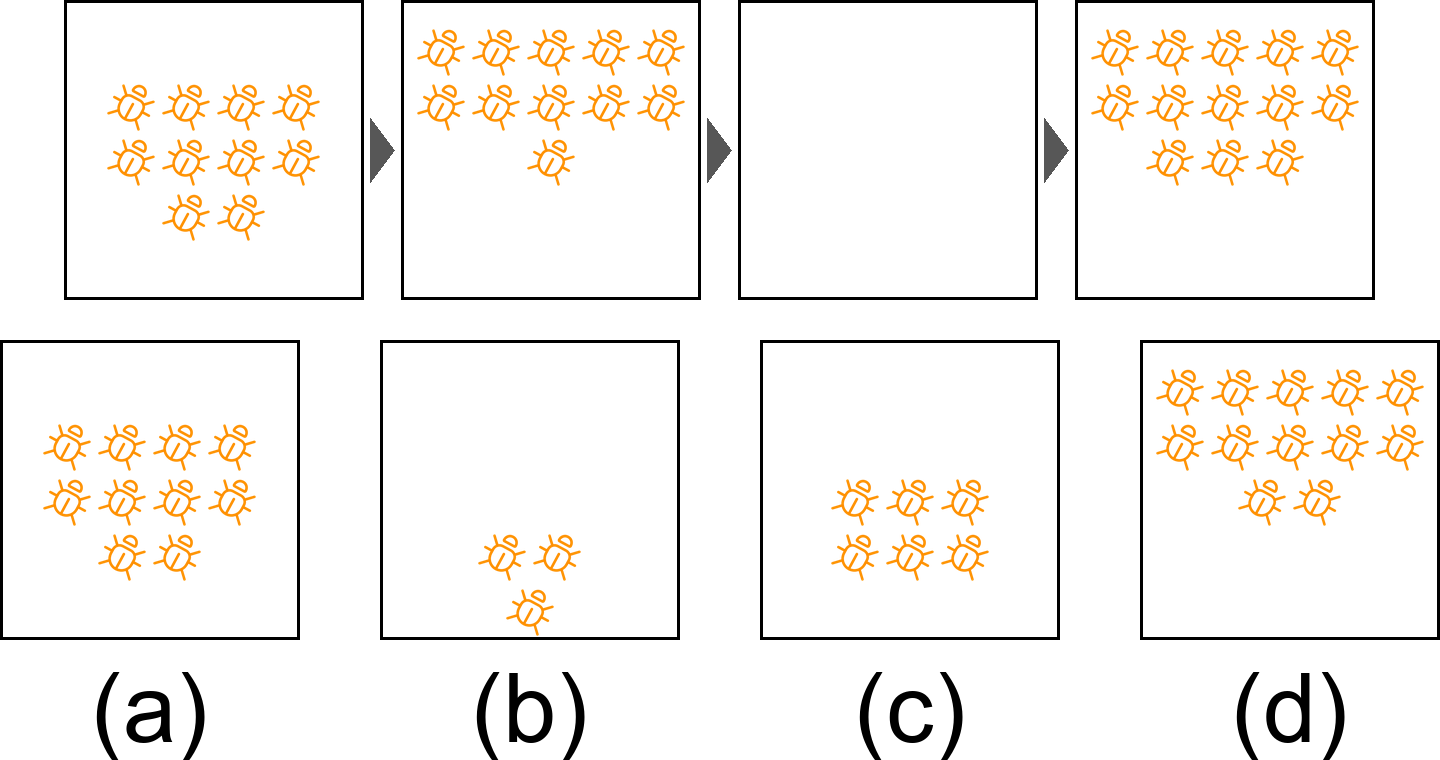} &
            \includegraphics[width=0.3\textwidth]{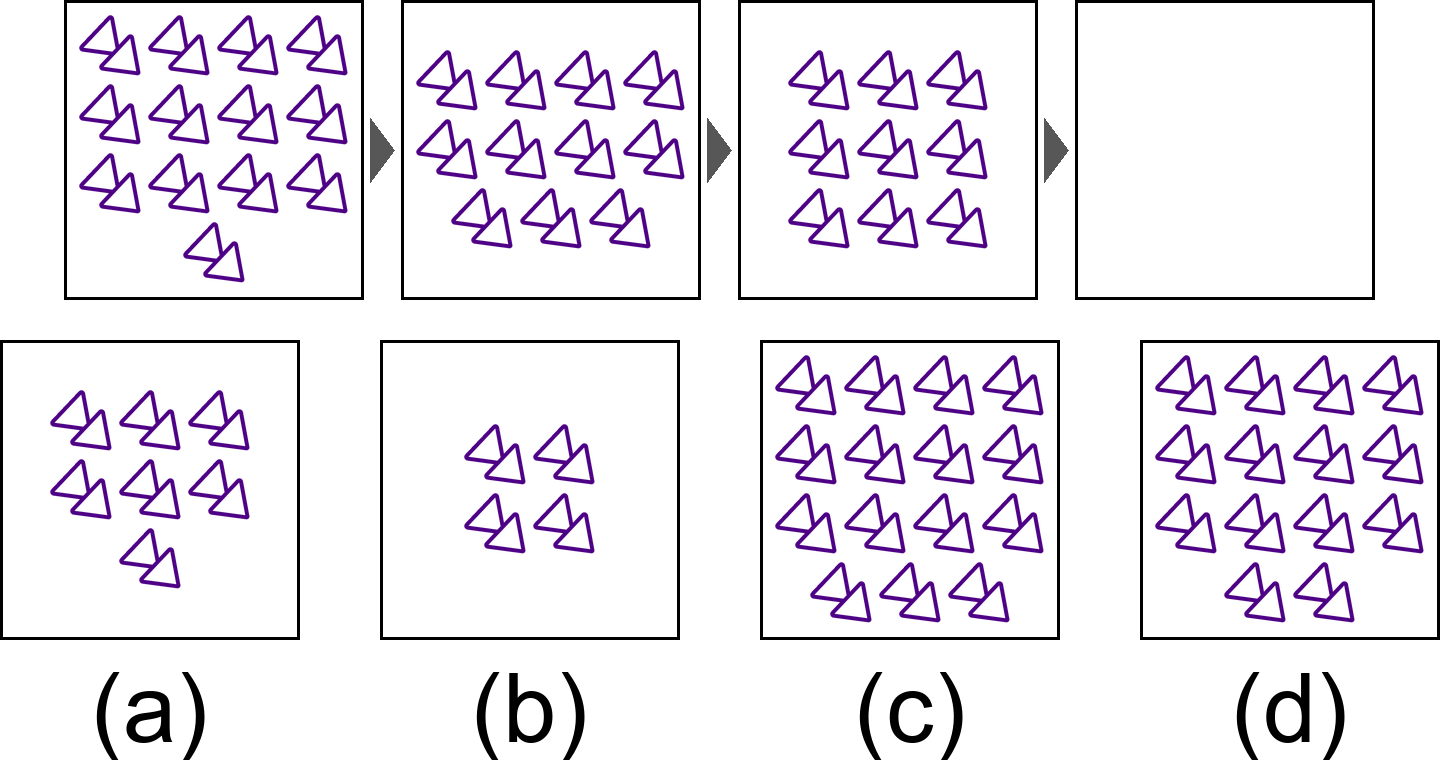} &
            \includegraphics[width=0.3\textwidth]{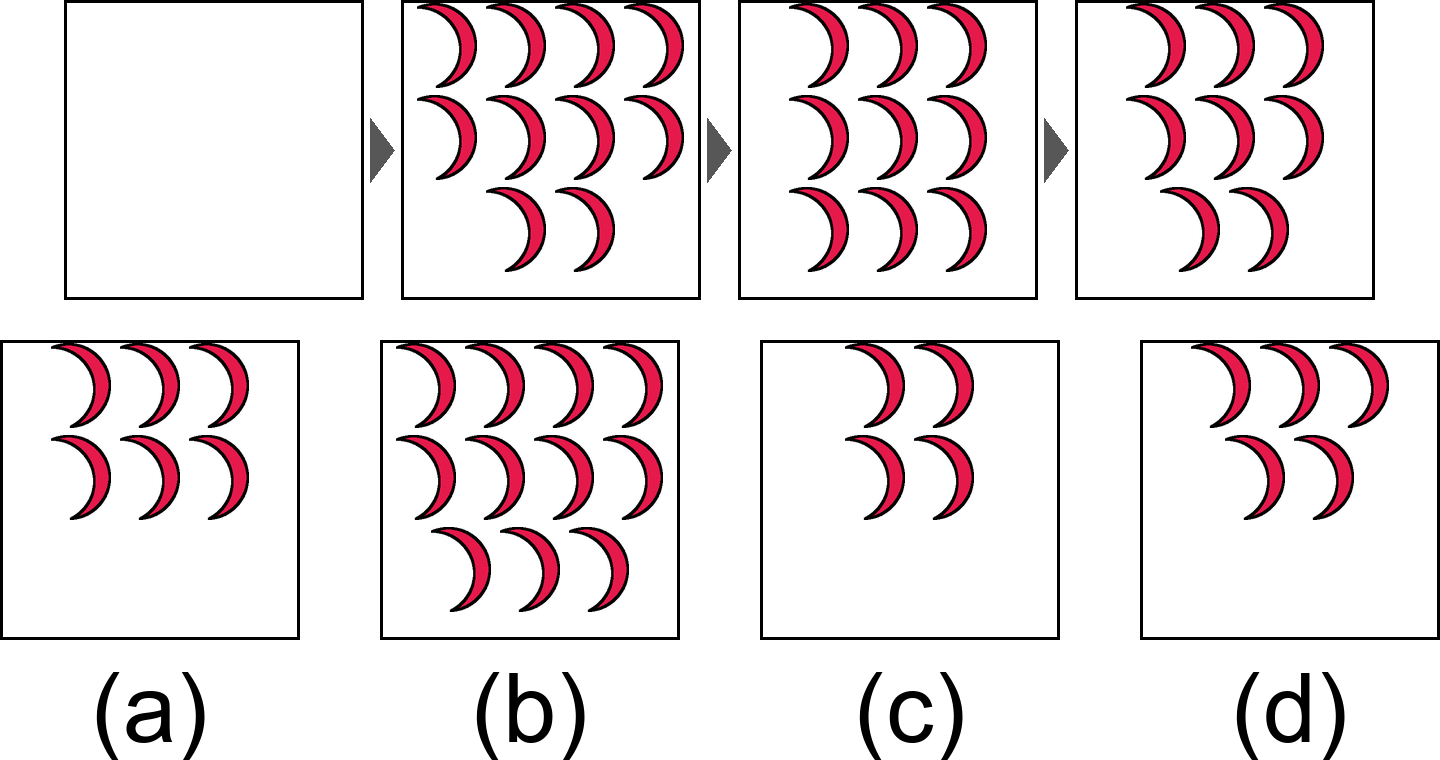} \\
        \end{tabular}
        \subcaption{\textbf{Sequence Arithmetic} Complete the sequence, which follows an arithmetic progression of shapes.}
        \label{fig:row_seq_arith}
    \end{subfigure}

    \begin{subfigure}{\textwidth}
        \centering
        \begin{tabular}{ccc}
            \includegraphics[width=0.3\textwidth]{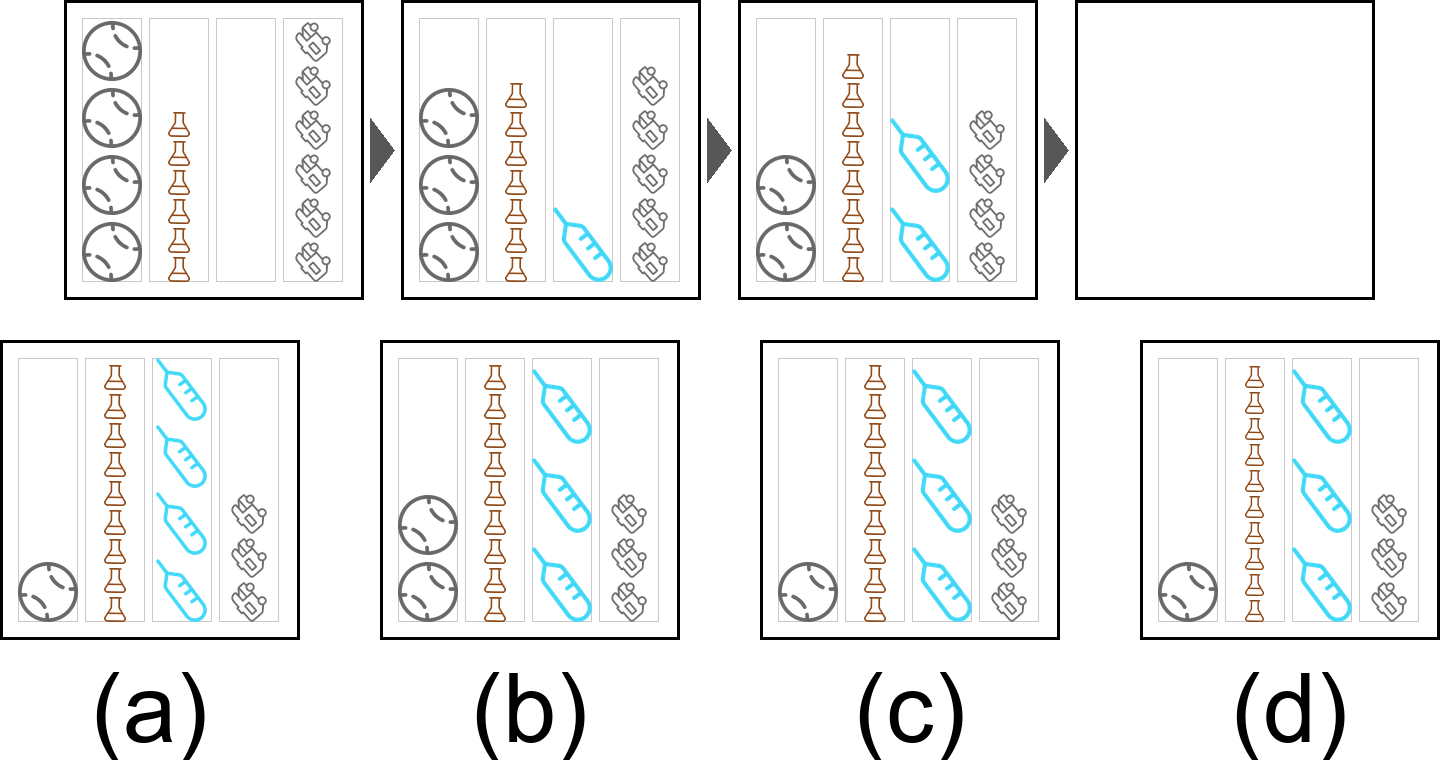} &
            \includegraphics[width=0.3\textwidth]{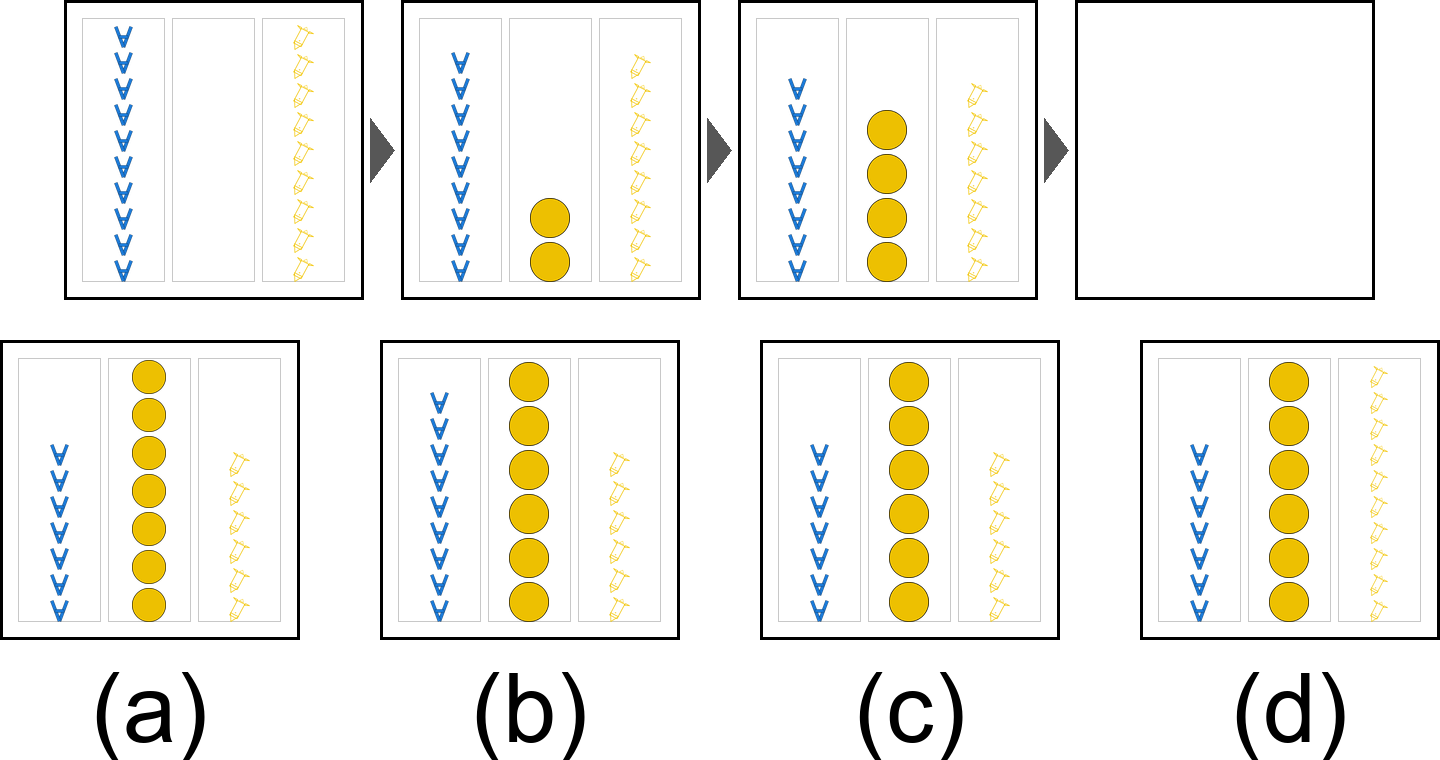} &
            \includegraphics[width=0.3\textwidth]{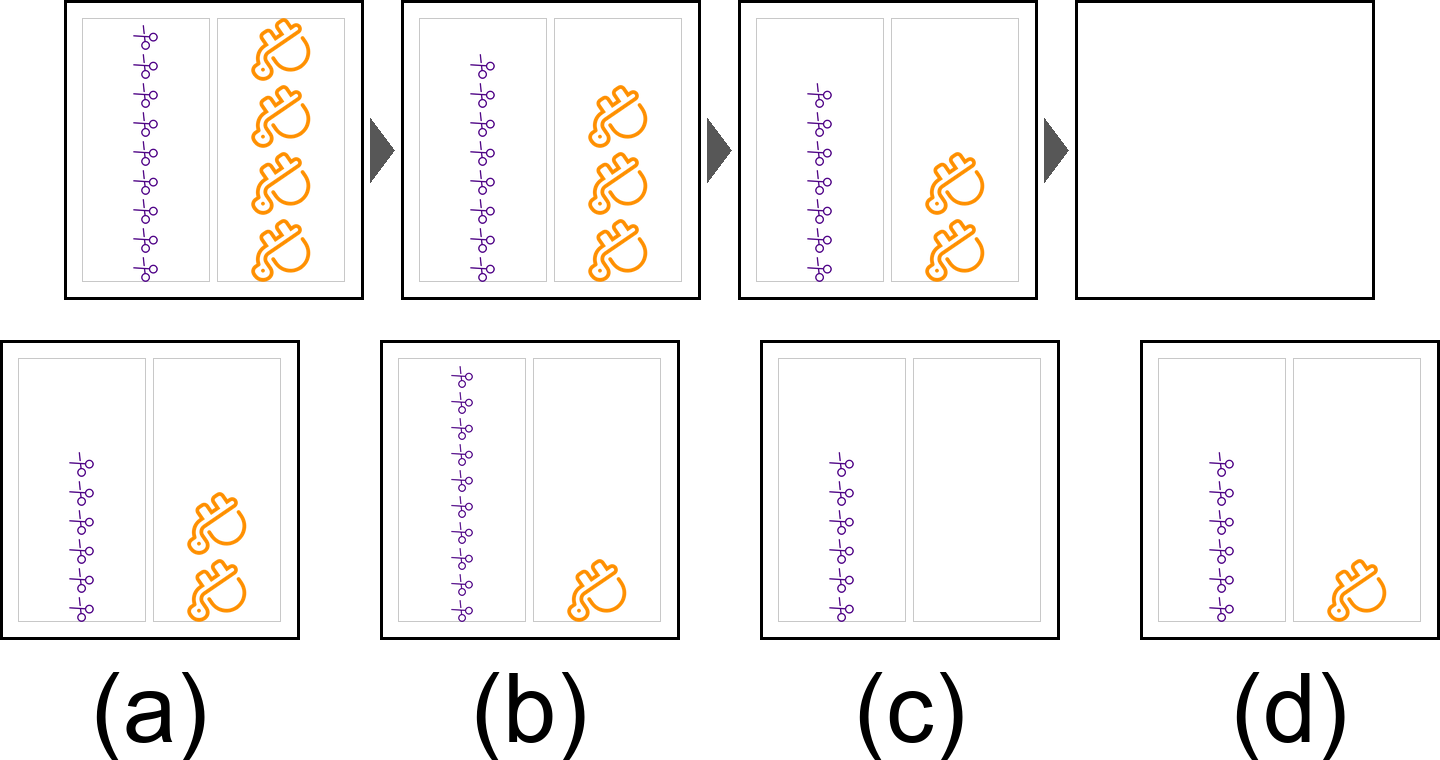} \\
        \end{tabular}
        \subcaption{\textbf{Sequence Multi-Column Arithmetic} Predict the final panel where each column follows its own progression.}
        \label{fig:row_seq_multi}
    \end{subfigure}

    \caption{Examples of Transformation and Sequence tasks.}
    \label{fig:grid_sequence}
\end{figure*}

Figure~\ref{fig:grid_sequence} shows examples of this type of task.

\subsubsection{Transform Result Identify}
\noindent\textbf{Problem.} Visual selection of the result of applying a sampled transform to the original tile.

\noindent\textbf{Construction.} Render a motif patch, center it on graph paper, sample a transformation, and construct one correct and three incorrect image options with consistent placement and borders. Compose a top/bottom layout with labels.

\noindent\textbf{Variants.} Eight transformation families; translations use randomized vectors.

\noindent  \textbf{Complexity.} Not measured.

\noindent\textbf{Distractors.} Render alternative transforms (including alternative translation vectors) and retain only candidates that are pairwise distinct.

\subsubsection{Transform Pair Infer}
\noindent\textbf{Problem.} Identify the single transformation that maps a source tile to a target tile; ``none of the above'' may be correct by omission.

\noindent\textbf{Construction.} Render a motif on graph paper, choose a true transform from mirrors/rotations/translation, synthesize the target, and verify uniqueness against the full rule set. Compose a side-by-side display with an arrow and six labeled textual options.

\noindent\textbf{Variants.} Up to eight answer classes: seven concrete transforms (vertical mirror, horizontal mirror, main diagonal mirror, anti-diagonal mirror, $90^\circ$ rotation, $180^\circ$ rotation, $270^\circ$ rotation, translation) plus none (correct with probability 1/6 when the true transform is withheld).

\noindent  \textbf{Complexity.} Not measured.

\noindent\textbf{Distractors.} When the true transform is present, sample other transforms as distractors with uniqueness filtering; when omitted, append none and select the remainder accordingly (with none fixed to the final slot for clarity).

\subsubsection{Transform Similarity Identify}
\noindent\textbf{Problem.} Similarity-based selection under Euclidean similarity (uniform scale + D$_4$ rigid/mirror motions). Either select the single similar option or the single dissimilar one.

\noindent\textbf{Construction.} Render an asymmetrical motif and produce options using the allowed D$_4$ transformations with optional uniform scaling and translation. For ``dissimilar'', apply enabled breaker warps (e.g., anisotropic scale, shear, perspective) and reject near-similar outcomes via a canonical checker.

\noindent\textbf{Variants.} Two core variants with four options.

\noindent  \textbf{Complexity.} Not measured.

\noindent\textbf{Distractors.} For ``similar'', distractors are other outcomes that remain distinct; for ``dissimilar'', distractors are similar options.

\subsubsection{Sequence Rotation}
\noindent\textbf{Problem.} Rotation-only progression over a single bitmap with a constant angular step; one panel is masked.

\noindent\textbf{Construction.} Render a base motif, compute a global scale fitting all sampled rotations, and generate tiles using a step from $\{30^{\circ}, 45^{\circ}, 60^{\circ}, 90^{\circ}\}$ in either direction. Mask one panel and present four options.

\noindent\textbf{Variants.} Eight rotation regimes (four step sizes $\times$ two directions); mask index is uniform.

\noindent  \textbf{Complexity.} Converts the rotation step into a reversed ordinal (i.e., a smaller rotation difference is more complex).

\noindent\textbf{Distractors.} Alternative rotation angles filtered by separation thresholds; weakly separated candidates are rejected.

\subsubsection{Sequence Arithmetic}
\noindent\textbf{Problem.} Next-step prediction in a count-based progression with one masked panel.

\noindent\textbf{Construction.} Sample a motif by weights. Draw a sequence with the count changing by a set increment/decrement; mask one panel and provide four choices.

\noindent\textbf{Variants.} Four candidate step choices (0, 1, 2, 3), either increasing or decreasing.

\noindent  \textbf{Complexity.} Maximum number of shapes across the sequence.

\noindent\textbf{Distractors.} Different incorrect counts are made and checked for enough visual difference from other options.

\subsubsection{Sequence Multi-Column Arithmetic}
\noindent\textbf{Problem.} Multi-column next-step prediction where each column follows its own arithmetic progression.

\noindent\textbf{Construction.} Sample 2-6 columns, motif kinds, and per-column base specs; draw four time steps using a shared within-column scale set by the maximum count. Hide the final panel and provide four candidates for continuation.

\noindent\textbf{Variants.} Continuous parameterization of column count, motifs, and steps.

\noindent  \textbf{Complexity.}  The number of columns is used to measure complexity.

\noindent\textbf{Distractors.} Edit exactly one column per wrong option, escalating $\pm\Delta$ until the local change exceeds a threshold; reject duplicate/low-contrast candidates.

\subsection{Topological \& Graph Reasoning}

\begin{figure*}[t]
    \centering
    \begin{subfigure}{\textwidth}
        \centering
        \begin{tabular}{ccc}
            \includegraphics[width=0.24\textwidth]{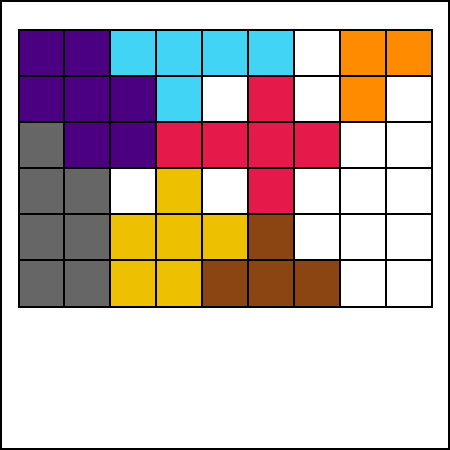} &
            \includegraphics[width=0.24\textwidth]{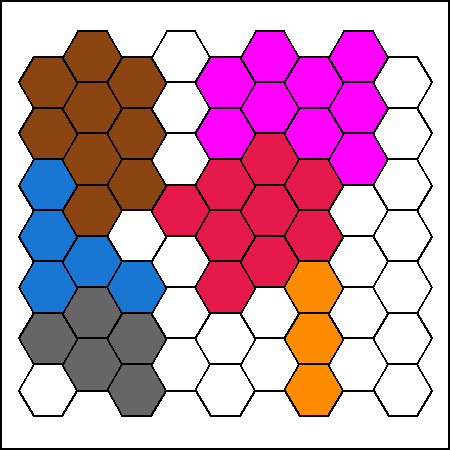} &
            \includegraphics[width=0.24\textwidth]{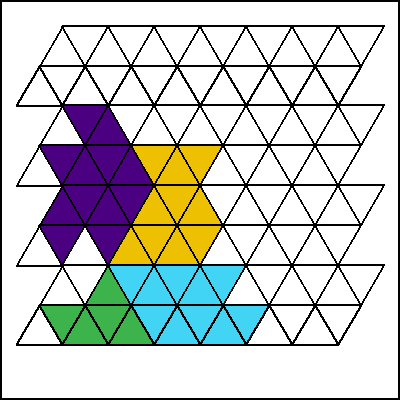} \\
        \end{tabular}
        \subcaption{\textbf{Tiles Geometry} Calculate a geometric property, such as area or perimeter, of a specified region.}
        \label{fig:row_tiles_geo}
    \end{subfigure}

    \begin{subfigure}{\textwidth}
        \centering
        \begin{tabular}{ccc}
            \includegraphics[width=0.24\textwidth]{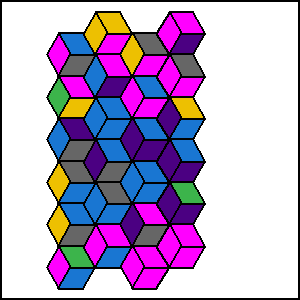} &
            \includegraphics[width=0.24\textwidth]{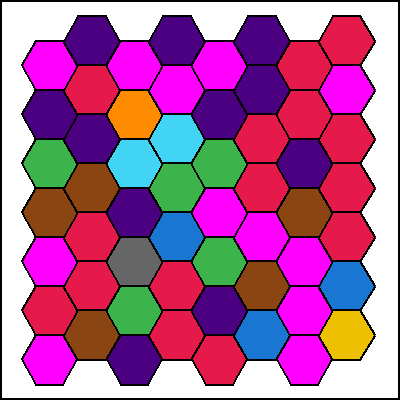} &
            \includegraphics[width=0.24\textwidth]{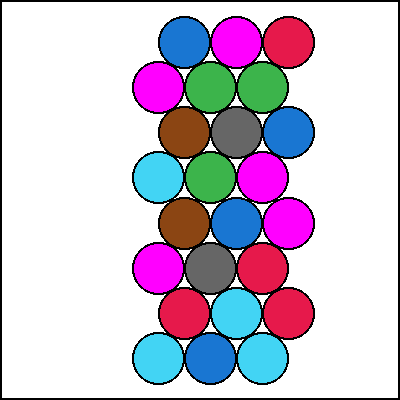} \\
        \end{tabular}
        \subcaption{\textbf{Tiles Connected Component} Find the size of the largest/smallest connected group of colored tiles.}
        \label{fig:row_tiles_conn}
    \end{subfigure}

    \begin{subfigure}{\textwidth}
        \centering
        \begin{tabular}{ccc}
            \includegraphics[width=0.24\textwidth]{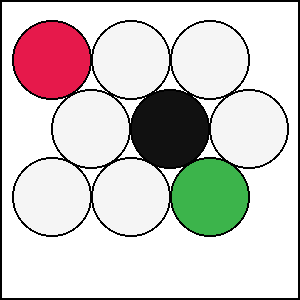} &
            \includegraphics[width=0.24\textwidth]{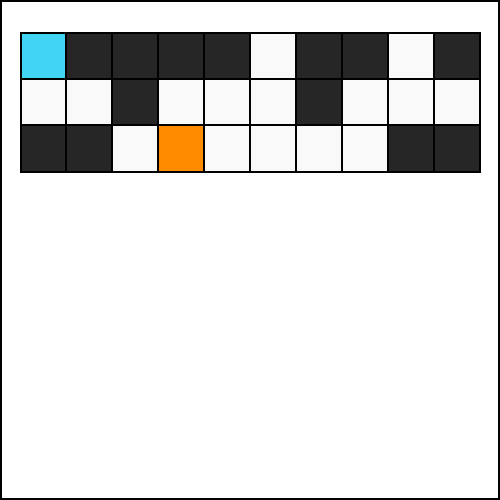} &
            \includegraphics[width=0.24\textwidth]{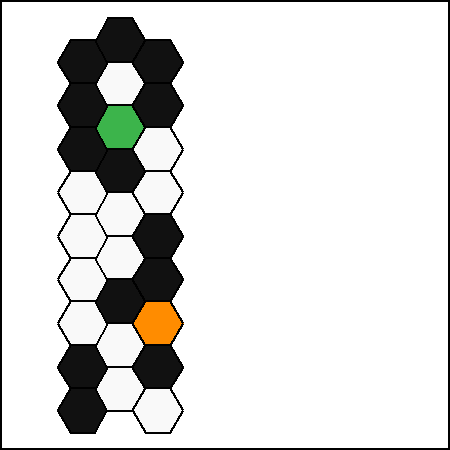} \\
        \end{tabular}
        \subcaption{\textbf{Tiles Shortest Path} Find the length of the shortest path between the two marked tiles if it exists.}
        \label{fig:row_tiles_path}
    \end{subfigure}

    \begin{subfigure}{\textwidth}
        \centering
        \begin{tabular}{ccc}
            \includegraphics[width=0.3\textwidth]{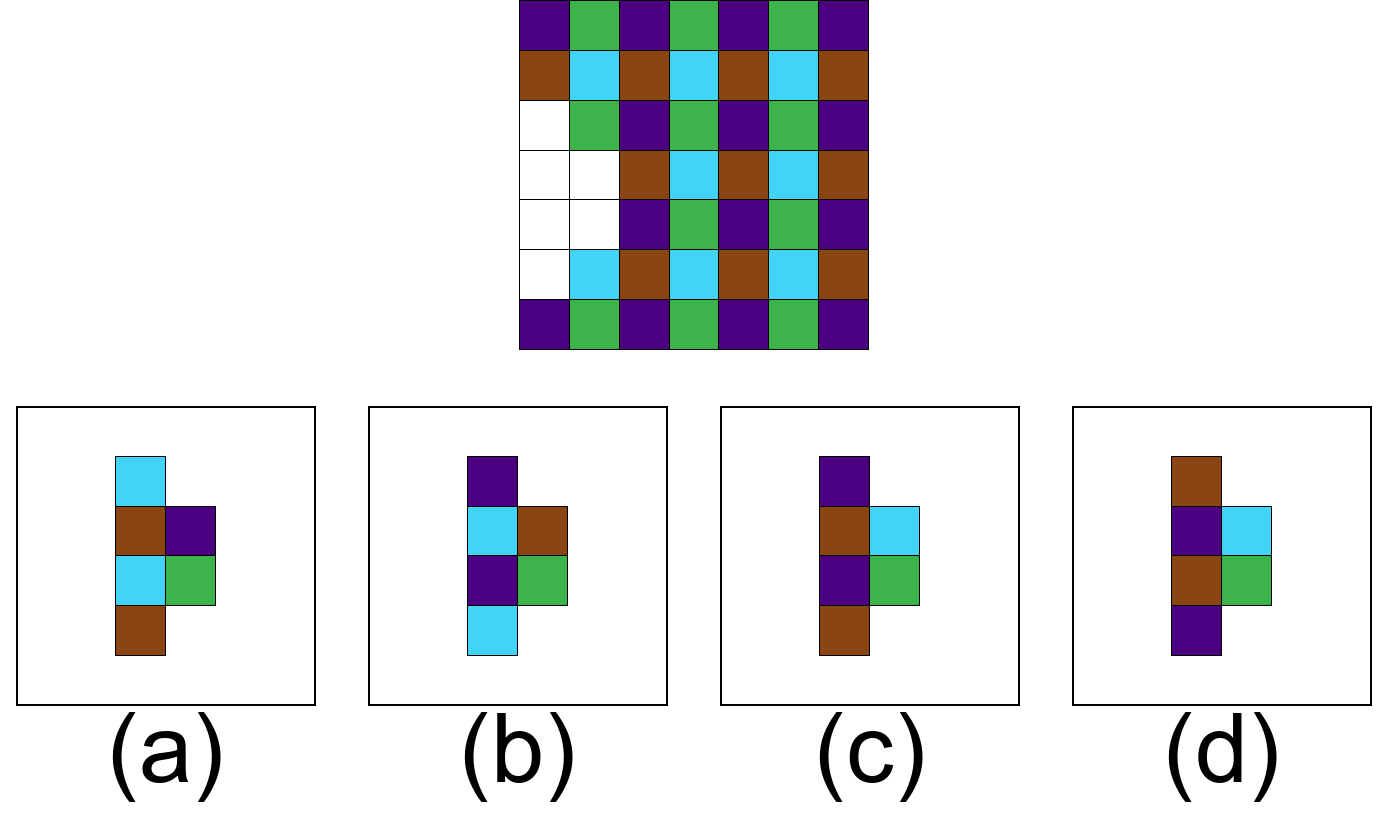} &
            \includegraphics[width=0.3\textwidth]{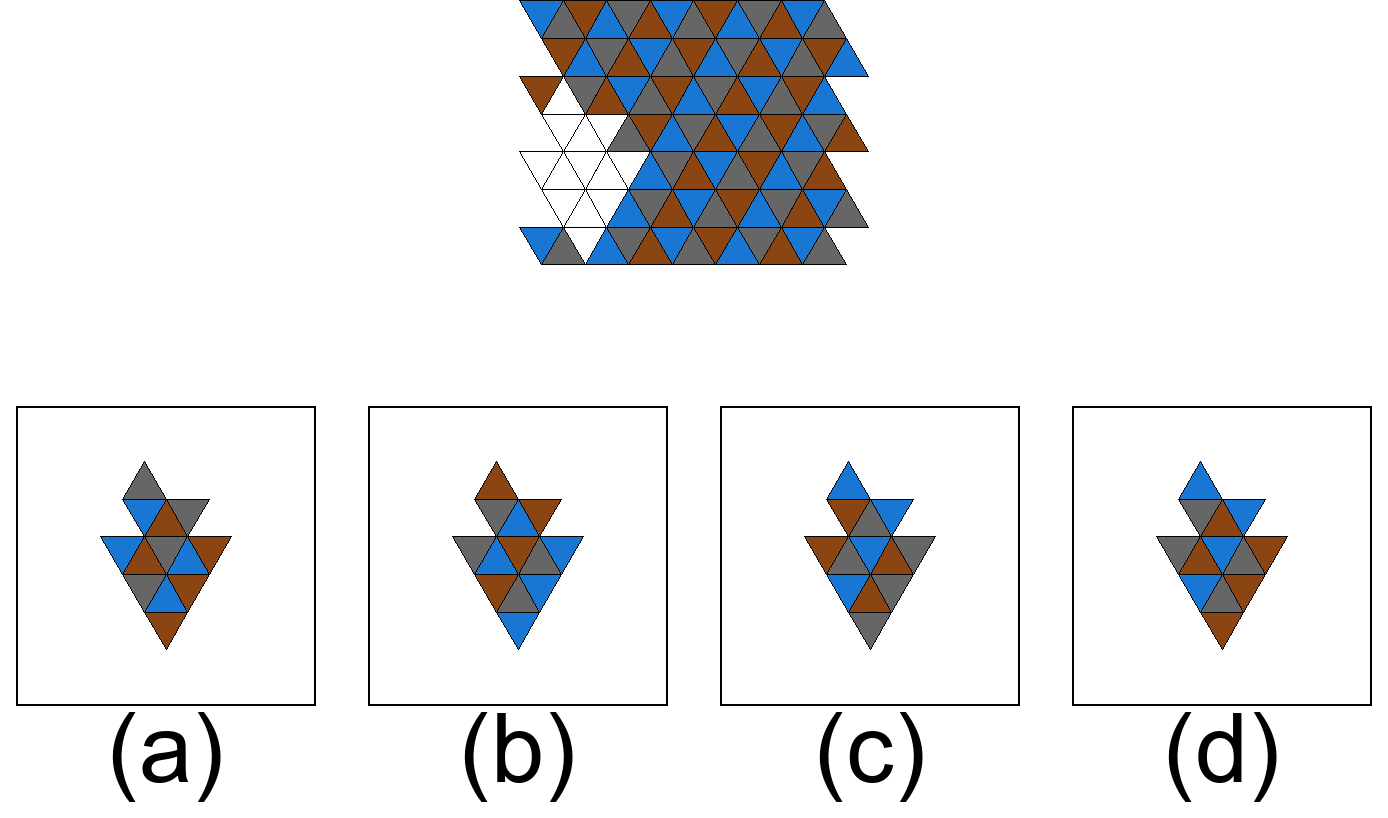} &
            \includegraphics[width=0.3\textwidth]{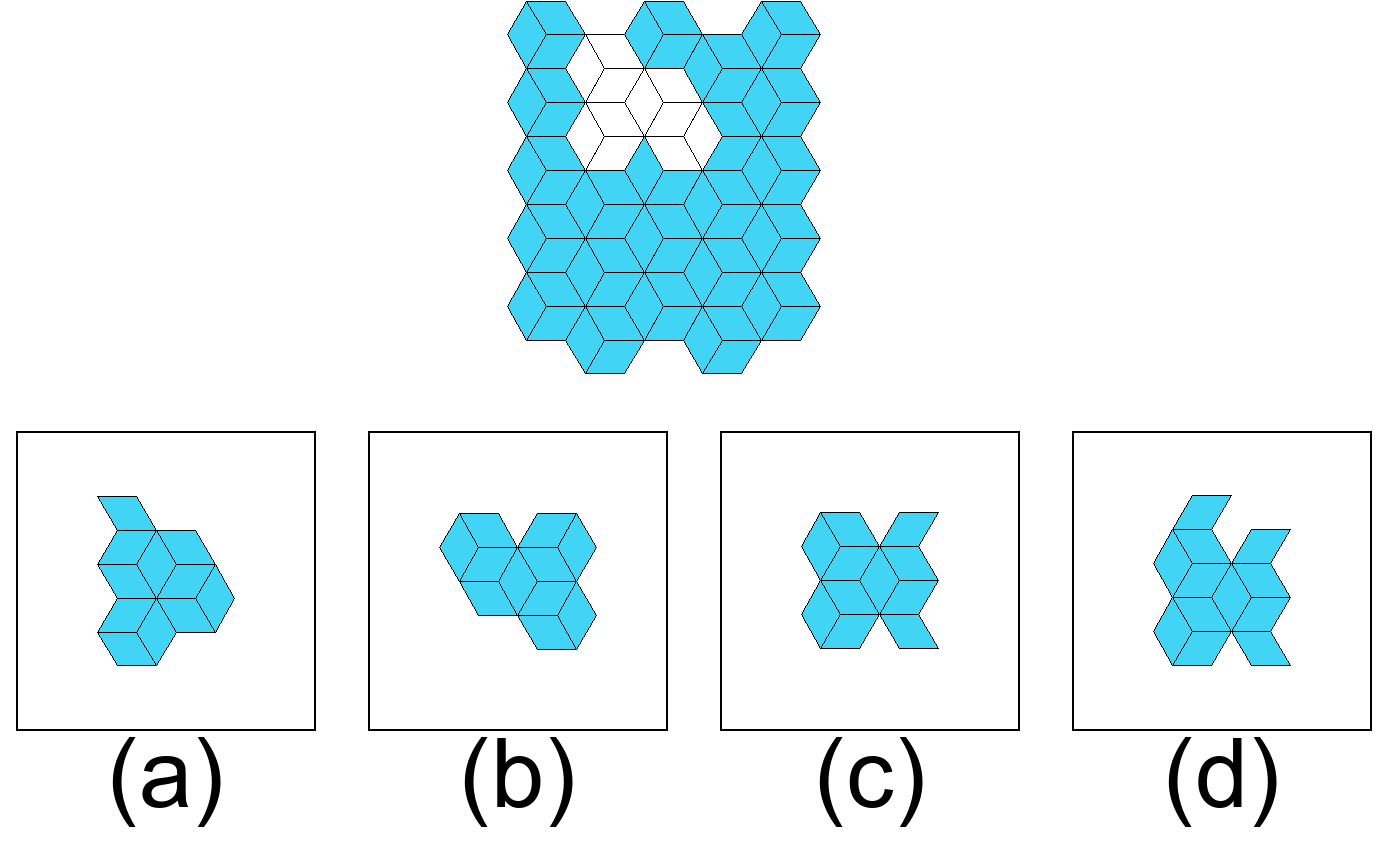} \\
        \end{tabular}
        \subcaption{\textbf{Missing Tiles} Find the correct shape and color of tiles to fill the blank region.}
        \label{fig:row_missing_tiles}
    \end{subfigure}

    \begin{subfigure}{\textwidth}
        \centering
        \begin{tabular}{ccc}
            \includegraphics[width=0.3\textwidth]{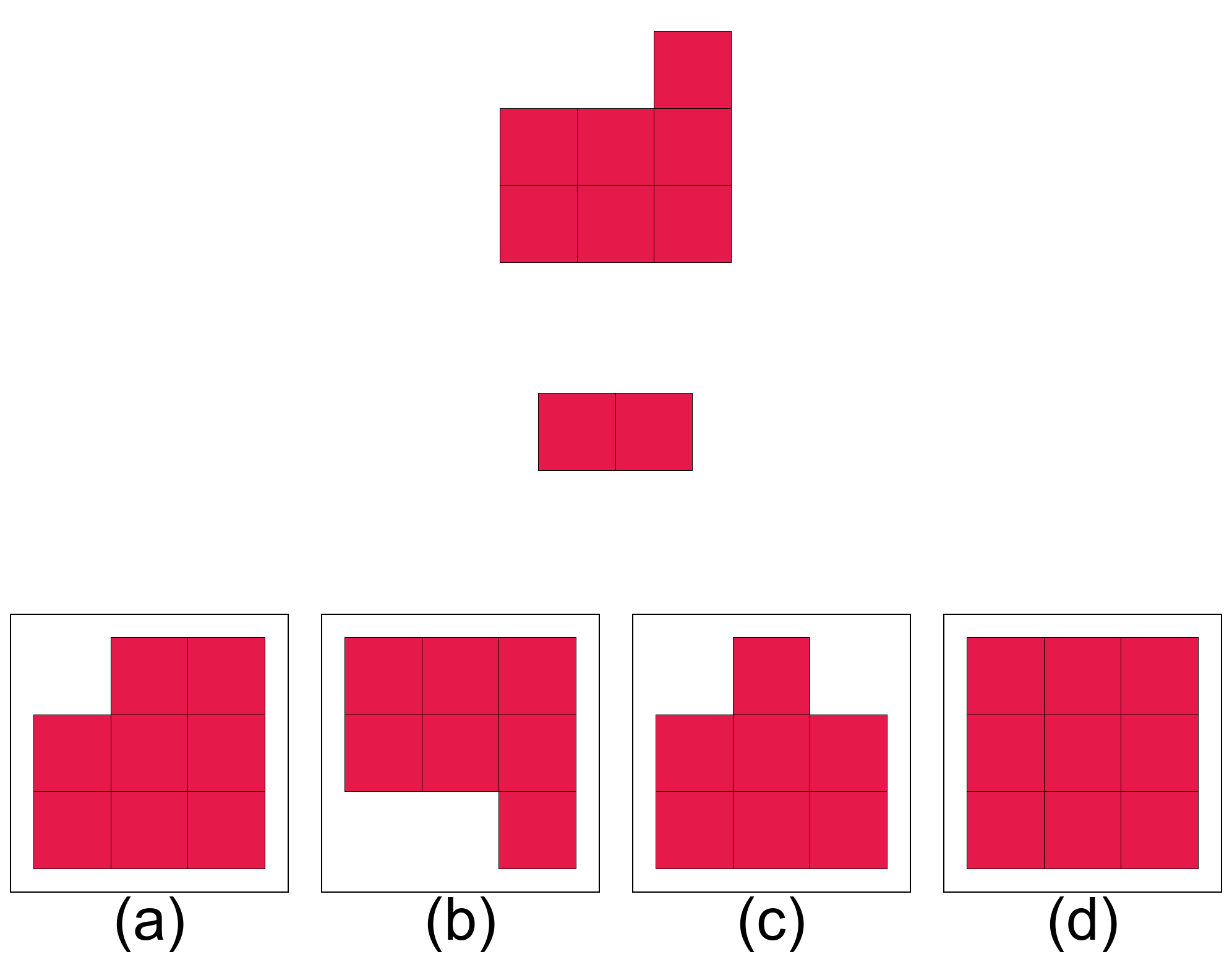} &
            \includegraphics[width=0.3\textwidth]{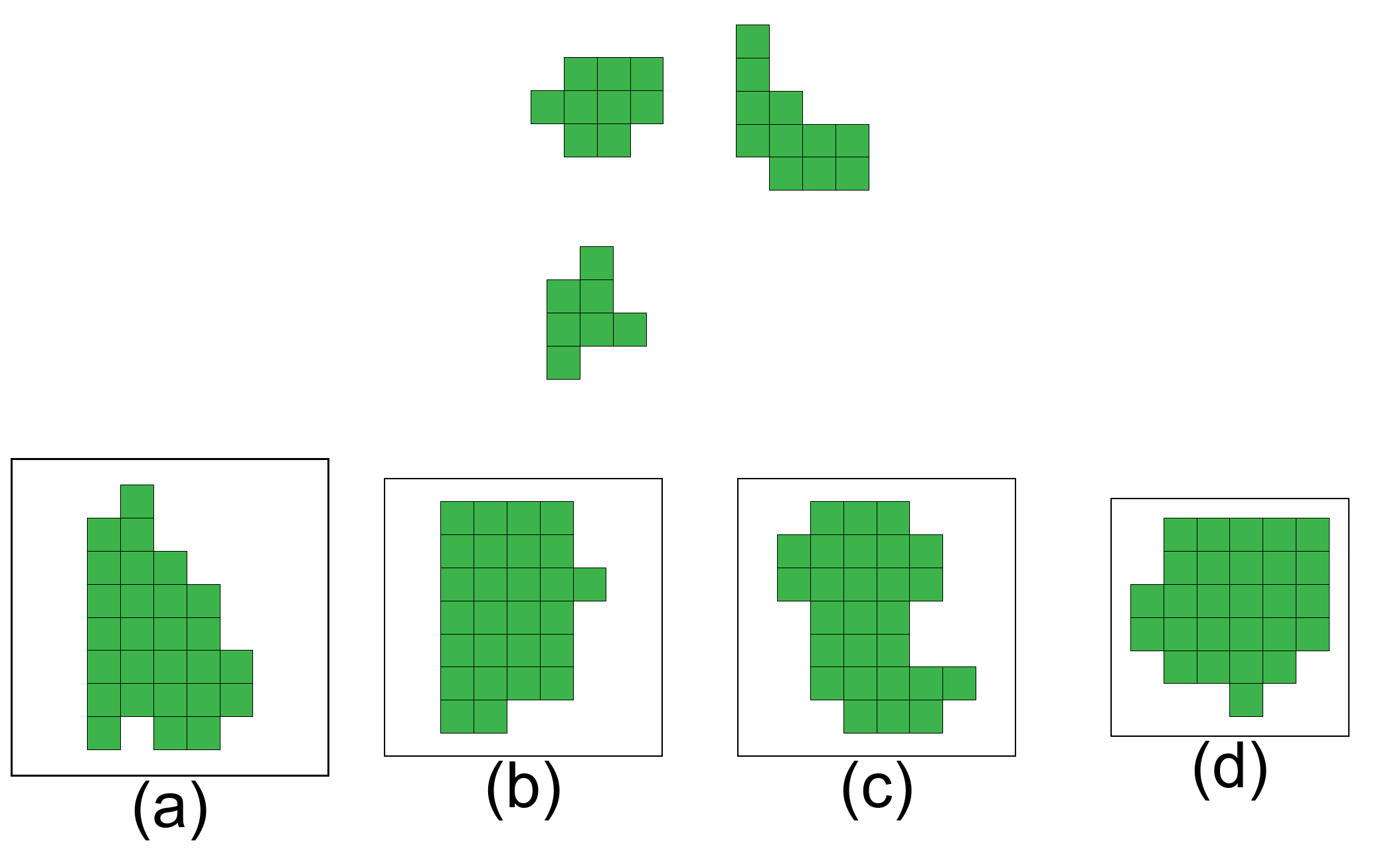} &
            \includegraphics[width=0.3\textwidth]{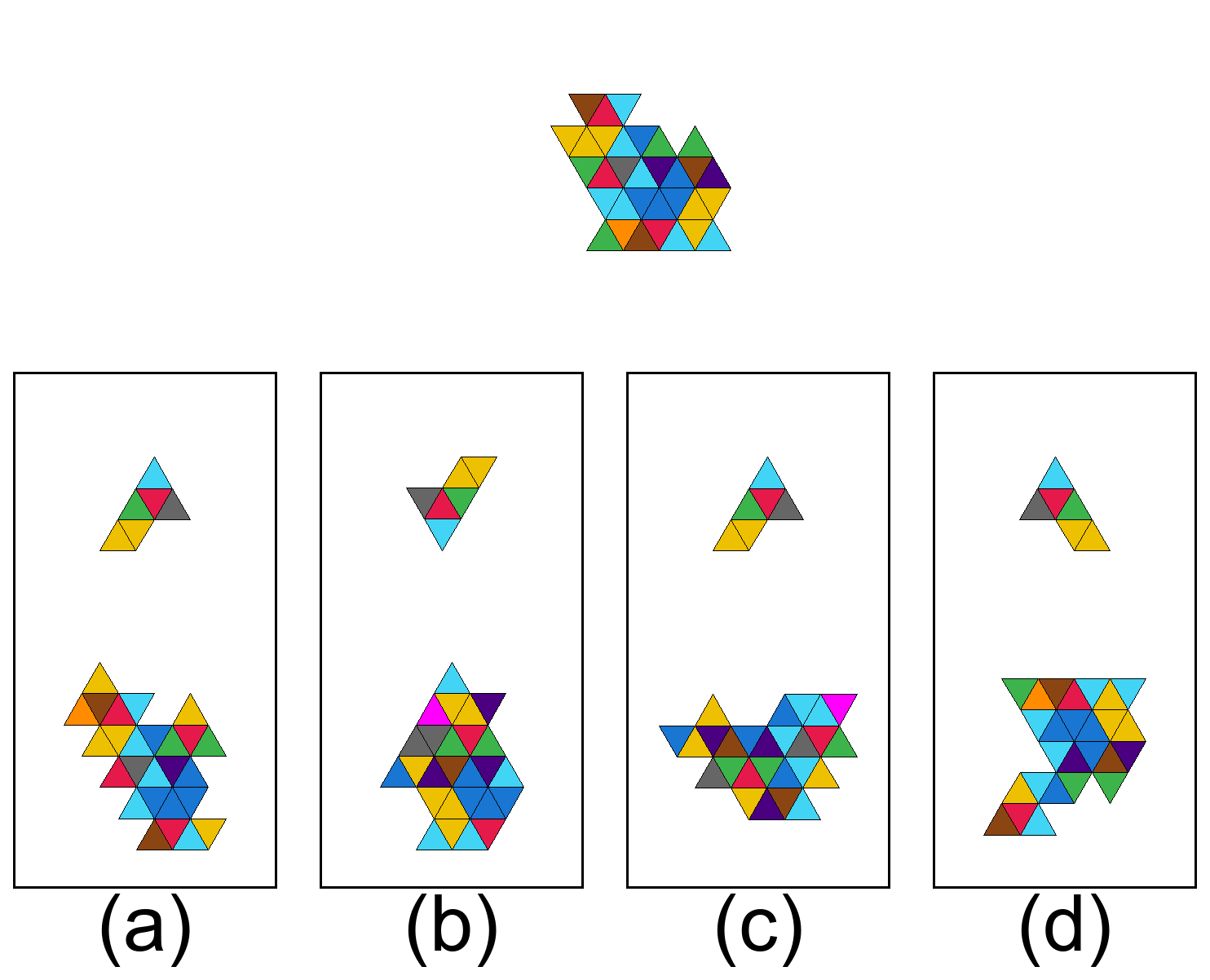} \\
        \end{tabular}
        \subcaption{\textbf{Tiles Composition} Choose the option that composes or decomposes into the top figure.}
        \label{fig:row_tiles_comp}
    \end{subfigure}

    \caption{Examples of Topological and Tiling tasks.}
    \label{fig:grid_topological}
\end{figure*}

Figure~\ref{fig:grid_topological} shows examples of this type of task.

\subsubsection{Tiles Geometry}
\noindent\textbf{Problem.} Geometric measurement over colored regions on a tiling (area, perimeter, holes, area difference, union perimeter).

\noindent\textbf{Construction.} Sample a tiling, paint disjoint regions, compute region graphs, and evaluate the requested measure. Render a white board with a natural-language prompt.

\noindent\textbf{Variants.} Five query types: single region area, single region perimeter, single region hole, two region area difference, union of two region perimeter, with per-instance color selection.

\noindent\textbf{Complexity.} The size of the tiling is the measure of complexity.

\noindent\textbf{Answer type.} Integer.

\subsubsection{Tiles Connected Component}
\noindent\textbf{Problem.} Component analysis on a colored tiling. Query the size of the largest/smallest component or the number of components within a specified color under a given adjacency notion.

\noindent\textbf{Construction.} Sample a tiling and a non-uniform coloring; build the dual graph with edge adjacency (or point-touch for circular tilings). Compute per-color connected components and select a query with a unique answer (enforced for extreme queries).

\noindent\textbf{Variants.} Three different types of measures (largest size, smallest size, count components) on five different tilings.

\noindent\textbf{Complexity.} The number of components measures complexity. 

\noindent\textbf{Answer type.} Integer.

\subsubsection{Tiles Shortest Path}
\noindent\textbf{Problem.} Shortest-path computation on a cell graph with obstacles; return the minimum number of edge-steps or -1 if unreachable.

\noindent\textbf{Construction.} Sample a tiling, build the dual graph, sample an obstacle field from beta-regime priors (sparse, dense, balanced, patchy), choose start/end tiles, and use BFS to verify distance or enforce unreachable cases.

\noindent\textbf{Variants.} Reachable (0.9 probability) vs. unreachable (0.1 probability) variety, five different tilings.

\noindent\textbf{Complexity.} The size of the tiling.

\noindent\textbf{Answer type.} Integer (distance) or -1 if unreachable.

\subsubsection{Missing Tiles}
\noindent\textbf{Problem.} Completion of a partially blanked tiling via color restoration or shape fitting (orientation changes allowed).

\noindent\textbf{Construction.} Sample a tiling and remove a connected region of bounded size. In the color variant, recover the exact color assignment for the missing cells. In the shape variant, recover the exact shape up to the tiling's dihedral symmetries.

\noindent\textbf{Variants.} Two balanced variants (color vs. shape) across four tilings (square, triangular, hexagonal, rhombille).

\noindent\textbf{Complexity.} The size of the tiling is used to measure complexity.

\noindent\textbf{Distractors.} Color variant performs pairwise color swaps or Dirichlet-weighted palette shuffles; shape variant samples alternative connected subsets of equal size that are non-congruent under allowed symmetries.

\subsubsection{Tiles Composition}
\noindent\textbf{Problem.} Piece equivalence and assembly. Either decompose a connected region into a multiset of connected pieces (bags), or compose a bag into a single connected target.

\noindent\textbf{Construction.} Sample a tiling and connected region; split into 2-4 connected pieces via randomized BFS growth. In ``decompose'', show the region on top and candidate bags below; in ``compose'', show a bag on top and candidate target shapes below. Normalize framing across options.

\noindent\textbf{Variants.} Two modes (decompose vs. compose) $\times$ two color modes (uniform, random\_per\_cell). Additional variation from piece counts and tiling families.

\noindent\textbf{Complexity.} The number of connected pieces is used as a measure for complexity.

\noindent\textbf{Distractors.} For decompose, bags reuse piece cardinalities but alter piece shapes. For compose, candidates match area but do not correspond to the true union of pieces.

\begin{figure*}[t]
    \centering
    \includegraphics[width=0.8\textwidth]{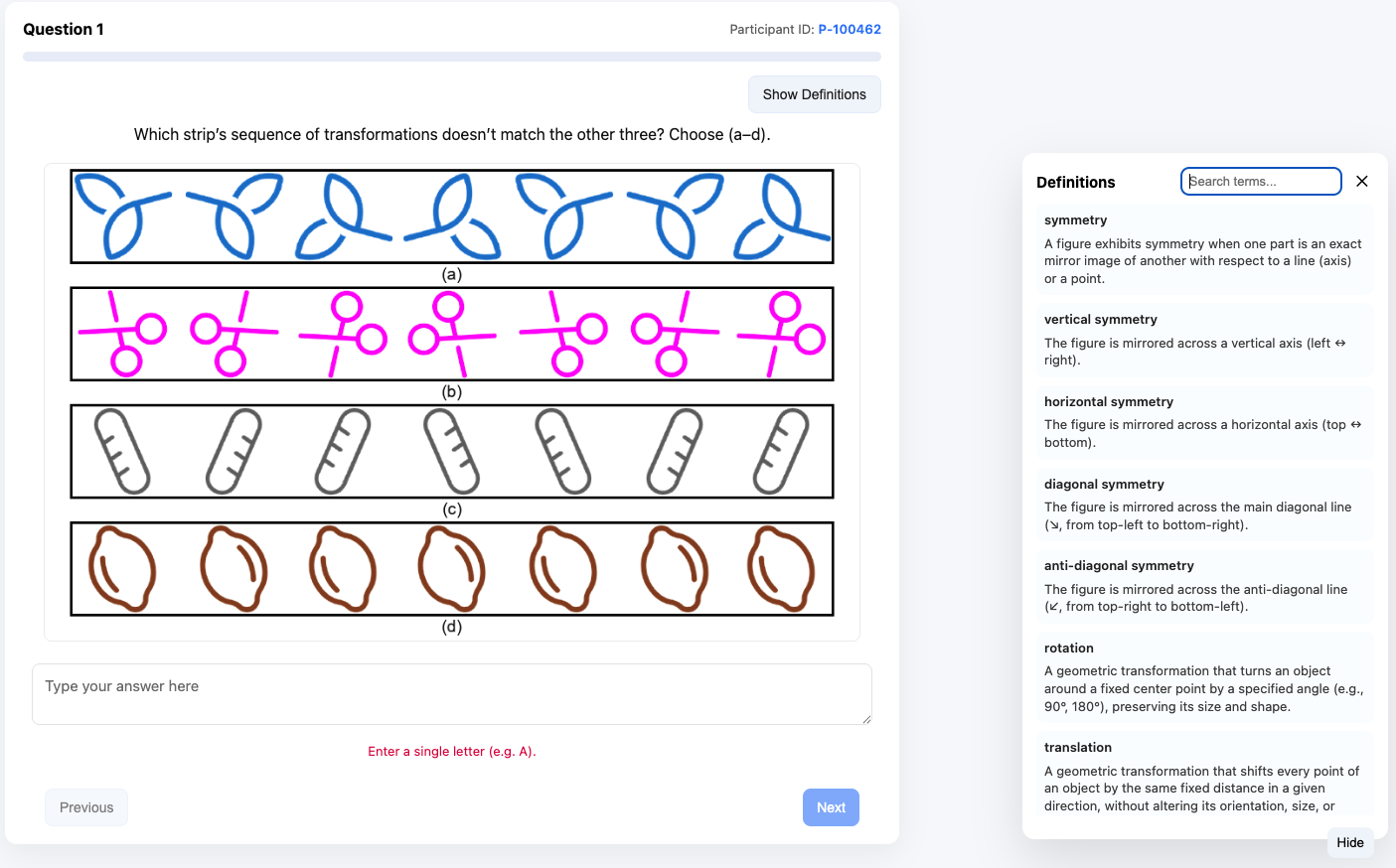}
    \caption{Web application interface used for the human evaluation. Participants were shown a visual prompt (image and/or text) and provided responses in the answer box.}
    \label{fig:human-interface}
\end{figure*}

\begin{figure*}[t]
  \centering
  \begin{subfigure}[b]{0.48\linewidth}
    \centering
    \includegraphics[width=\linewidth]{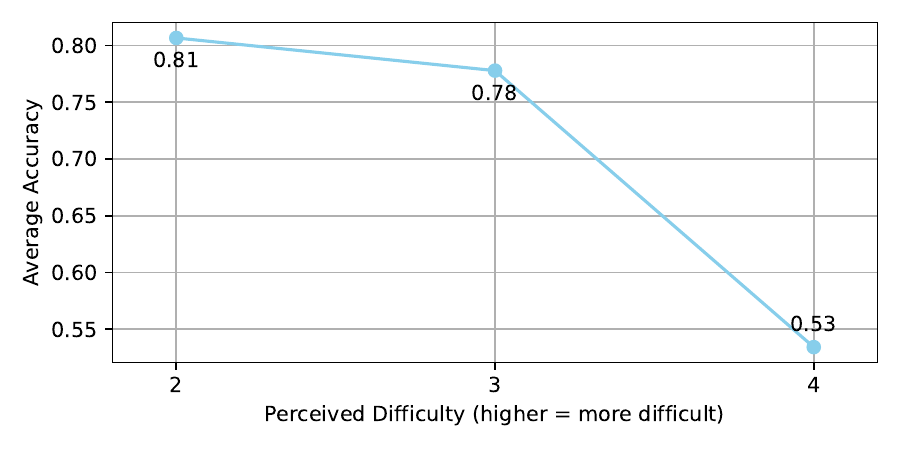}
    \caption{Perceived difficulty vs accuracy.}
    \label{fig:diff-acc}
  \end{subfigure}
  \hfill
  \begin{subfigure}[b]{0.48\linewidth}
    \centering
    \includegraphics[width=\linewidth]{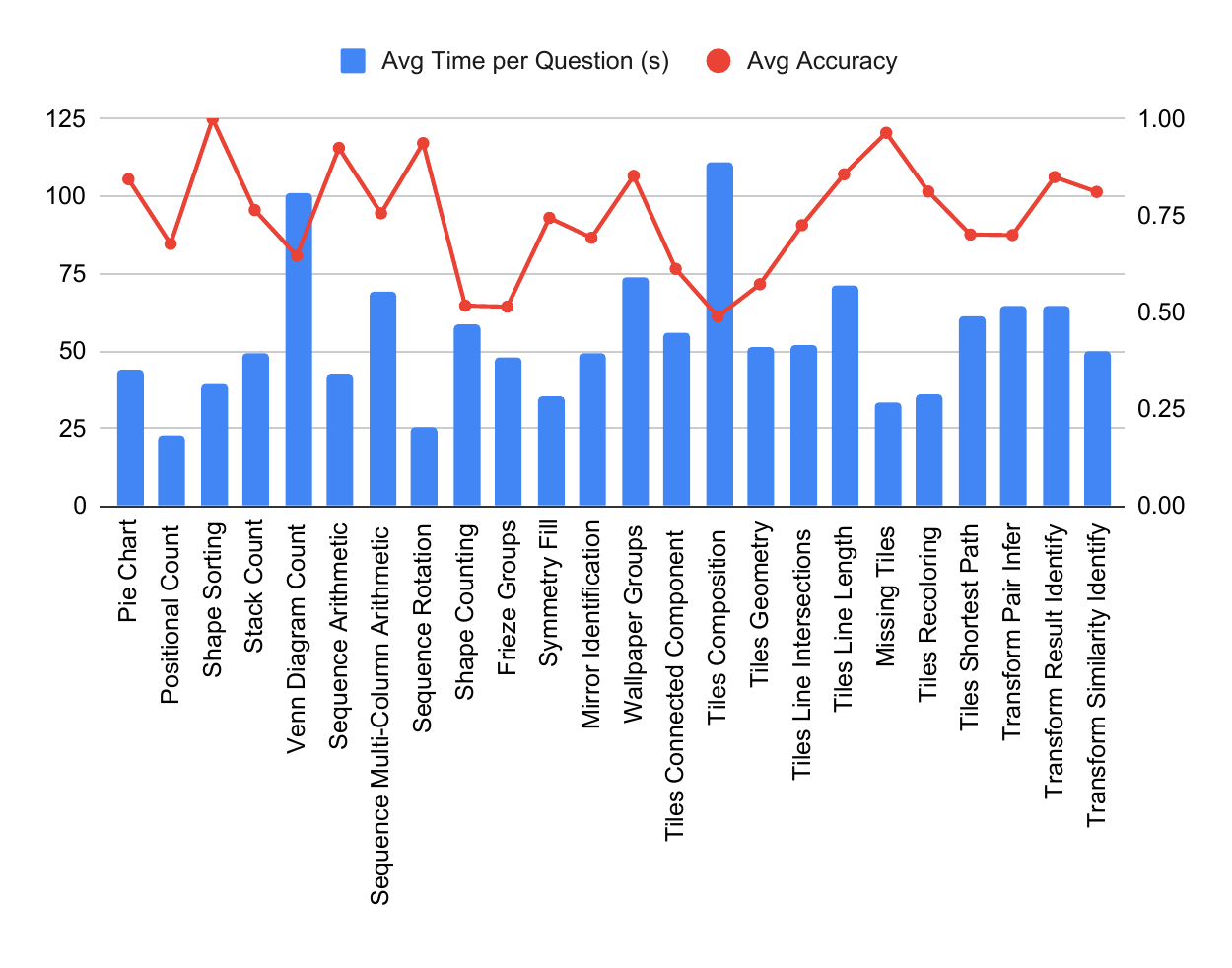}
    \caption{Average time per question (bars) and accuracy (line) per task.}
    \label{fig:task-time-acc}
  \end{subfigure}

  \caption{Human evaluation results. (a) Plot of participant perceived difficulty versus accuracy (b) Task-level time and accuracy.}
  \label{fig:human-results}
\end{figure*}

\section{Human Evaluation}
\label{app:humaneval}

We conducted a human evaluation using a custom-built web application. The participants accessed the app through a browser and were assigned a set of 25 problems (or 10 problems if they explicitly chose the shorter version). Each problem consisted of a visual prompt (image and/or text) and an input field for responses. In total, we collected 32 completed evaluations. 
% This setup allowed us to systematically measure accuracy, timing, and subjective feedback between 
% participants and tasks, allowing comparison of human performance against large language models (LLMs).

The application enforced basic validation (e.g., number formats, single-choice letters, or ordered lists) 
to ensure that responses were well-formed. For each participant, we record the following.
\begin{itemize}
    \item Response text
    \item Correctness (with respect to the ground truth)
    \item Per-question time taken
    \item Overall completion time
    \item Types of tasks assigned
\end{itemize}

To reduce variability in prior knowledge, the interface also provided a dedicated \textit{Definitions} panel containing concise explanations of key terms and concepts 
(e.g., symmetry, rotation, translation). This feature ensured that all participants could engage in tasks from a comparable baseline of conceptual understanding, thereby minimizing confounding effects due to varying background knowledge.

After completing the problem set, participants completed a \textit{post-questionnaire survey} in which they rated 
perceived difficulty, clarity, familiarity, and participation, along with providing optional feedback.

\paragraph{Human Evaluation Setup}

Figure~\ref{fig:human-interface} shows the web interface used to collect human responses to the specific tasks assigned for \emph{\textsc{Sphinx}}.

\paragraph{Human Performance Analysis}

Figure~\ref{fig:human-results} shows human performance in evaluation tasks, highlighting time–accuracy analysis, and the relationship between  difficulty ratings and accuracy.

\section{Analysis of GPT-5 Response}
\label{app:gpt5-analysis}

\subsection{Shape Counting}

\begin{figure*}[t]
\centering

% --- Row 1 --------------------------------------------------------

% --- Panel 1: Tiles Recoloring (example 1, top-left) ---
\begin{subfigure}[t]{0.495\textwidth}
\begin{panelbox}[colback=black!4, equal height group=RecolorTop]
\textbf{(\romannumeral 1)}

{\centering
\includegraphics[width=0.6\linewidth]{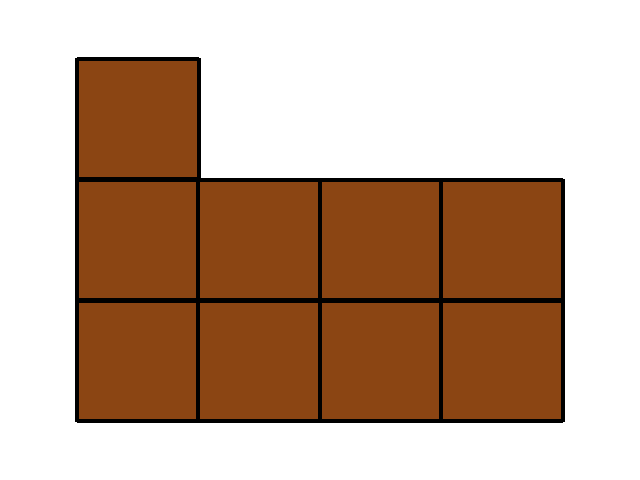}\par}

\textbf{Question:} What is the square count in the figure?

\textbf{Answer:} 12

\textbf{Response from GPT-5 (Correct):}\\
Step-by-step:

- 1×1 squares: Count the unit squares. There are 4 in the bottom row, 4 in the middle row, and 1 on top = 9.

- 2×2 squares: These can only be formed within the 2×4 rectangle at the bottom. There are 3 such squares (columns 1–2, 2–3, and 3–4).

- 3×3 or larger: Not possible because the top row has only one square.

Total squares = 9 + 3 = 12.
\boxed{12}
\end{panelbox}
\end{subfigure}
\hfill
% --- Panel 2: Tiles Recoloring (example 2, top-right) ---
\begin{subfigure}[t]{0.495\textwidth}
\begin{panelbox}[colback=pink!8, equal height group=RecolorTop]
\textbf{(\romannumeral 2)}

{\centering
\includegraphics[width=0.6\linewidth]{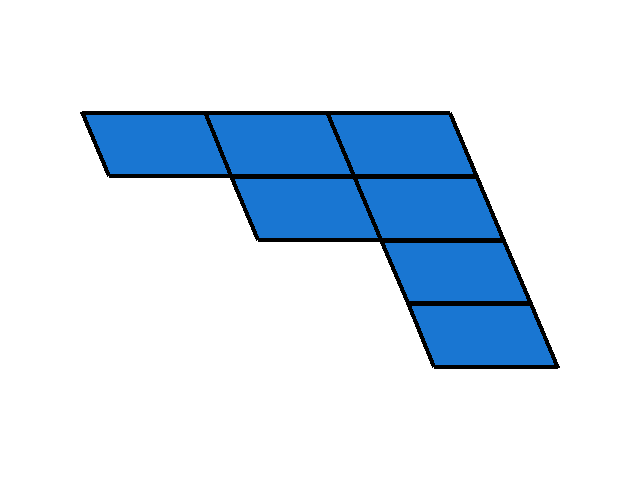}\par}

\textbf{Question:} What is the number of parallelograms?

\textbf{Answer:} 19

\textbf{Response from GPT-5 (Correct):}\\
Step 1: Count 1×1 parallelograms (unit tiles).
- There are 7 unit tiles.

Step 2: Count wider shapes (height 1).
- 2×1: top row has 2, second row has 1 → 3.
- 3×1: only on the top row → 1.

Step 3: Count taller shapes (width 1).
- 1×2: along common columns: three in the rightmost column plus one in the middle column → 4.
- 1×3: rightmost column, starting at rows 0 and 1 → 2.
- 1×4: rightmost column across all rows → 1.

Step 4: Count larger blocks.
- 2×2: only possible using rows 0–1 and columns 1–2 → 1.
- No bigger blocks fit.

Total = 7 + 3 + 1 + 4 + 2 + 1 + 1 = 19.

Final answer: \boxed{19}

\end{panelbox}
\end{subfigure}

% --- Row 2 --------------------------------------------------------

% --- Panel 3: Tiles Recoloring (example 1, bottom-left; duplicate of Panel 1) ---
\begin{subfigure}[t]{0.495\textwidth}
\begin{panelbox}[colback=black!4, equal height group=RecolorBottom]
\textbf{(\romannumeral 3)}

{\centering
\includegraphics[width=0.6\linewidth]{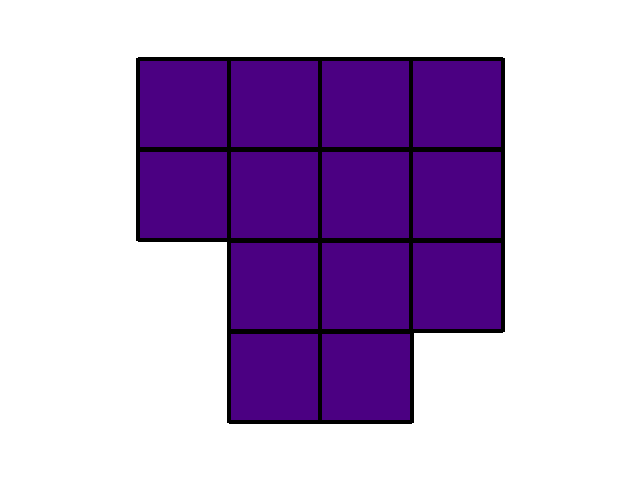}\par}

\textbf{Question:} What is the square count in the figure?

\textbf{Answer:} 20

\textbf{Response from GPT-5 (Incorrect):}\\
Step-by-step count of all axis-aligned squares:

- 1×1 squares:
  Rows have 4, 4, 3, \textcolor{red}{3}, 2 unit squares → 4+4+3+\textcolor{red}{3}+2 = 16.

- 2×2 squares:
  Between rows (1,2): 3
  Between rows (2,3): 2
  \textcolor{red}{Between rows (3,4): 2}
  Between rows (4,5): 1
  Total = 3+2+2+1 = 8.

- 3×3 squares:
  Rows (1–3): 1
  \textcolor{red}{Rows (2–4): 1}
  \textcolor{red}{Rows (3–5)}: 0
  Total = 2.

- 4×4 or larger: none fit entirely in the figure.

Total number of squares = 16 + 8 + 2 = \boxed{26}.

\end{panelbox}
\end{subfigure}
\hfill
% --- Panel 4: Tiles Recoloring (example 2, bottom-right; duplicate of Panel 2) ---
\begin{subfigure}[t]{0.495\textwidth}
\begin{panelbox}[colback=black!4, equal height group=RecolorBottom]
\textbf{(\romannumeral 4)}

{\centering
\includegraphics[width=0.6\linewidth]{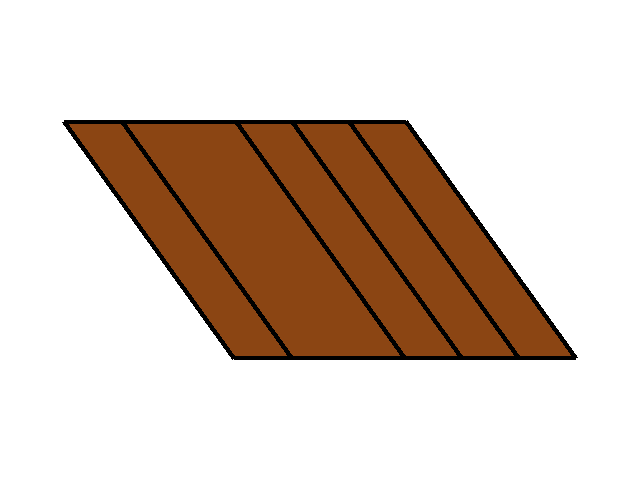}\par}

\textbf{Question:} How many parallelograms are there in the figure?

\textbf{Answer:} 15

\textbf{Response from GPT-5 (Incorrect):}\\
Step-by-step reasoning:

- Inside the parallelogram there are \textcolor{red}{3} lines parallel to one pair of sides. Including the two outer sides, that makes \textcolor{red}{5} parallel lines in that direction.

- In the other direction, there are only the top and bottom sides (2 lines).

- Any parallelogram is formed by choosing two of the 5 parallel lines and using both the top and bottom sides.

Number of parallelograms = C(5,2) × C(2,2) = 10 × 1 = 10.
\boxed{10}

\end{panelbox}
\end{subfigure}

\caption{Example responses from GPT-5 on \emph{Shape Counting} Task, with incorrect reasoning highlighted in red.}
\label{fig:gpt5_shapecounting}
\end{figure*}

Shape counting is the task on which GPT-5 most strongly surpasses human performance (76\% vs 55.2\%). Figure~\ref{fig:gpt5_shapecounting} presents two correct model responses in the top row and two incorrect responses in the bottom row. In the correct cases, GPT-5 produces clear, step-by-step reasoning: it enumerates all relevant sub-shapes, organizes them by size or configuration, and adds the totals in a structured way. The textual reasoning is coherent and reflects an understanding of how larger shapes are made up from smaller ones.

The incorrect examples reveal a different pattern. GPT-5 often reasons correctly in text, but the reasoning is anchored to an inaccurate visual extraction. In example (\romannumeral 3), the model incorrectly states that the figure contains five rows when it actually contains four, and all subsequent steps follow from this mistaken premise. In example (\romannumeral 4), the model identifies five parallel lines from left to right instead of six and then applies a suitable combinatorial formula based on that incorrect count. The final numeric answer would have been correct if the model had extracted the correct number of lines.

These observations also help explain why the human accuracy is lower than that of GPT-5. The participants viewed the images on a monitor and performed all the counting mentally without any external notes. Since shape counting requires keeping track of several intermediate quantities, arithmetic slips or overlooked configurations are common even when the visual perception itself is correct.

\begin{figure*}[t]
\centering

% --- Row 1 --------------------------------------------------------

% --- Panel 1: Tiles Recoloring (example 1, top-left) ---
\begin{subfigure}[t]{0.495\textwidth}
\begin{panelbox}[colback=black!4, equal height group=im2top]
\textbf{(\romannumeral 1)}

{\centering
\includegraphics[width=0.5\linewidth]{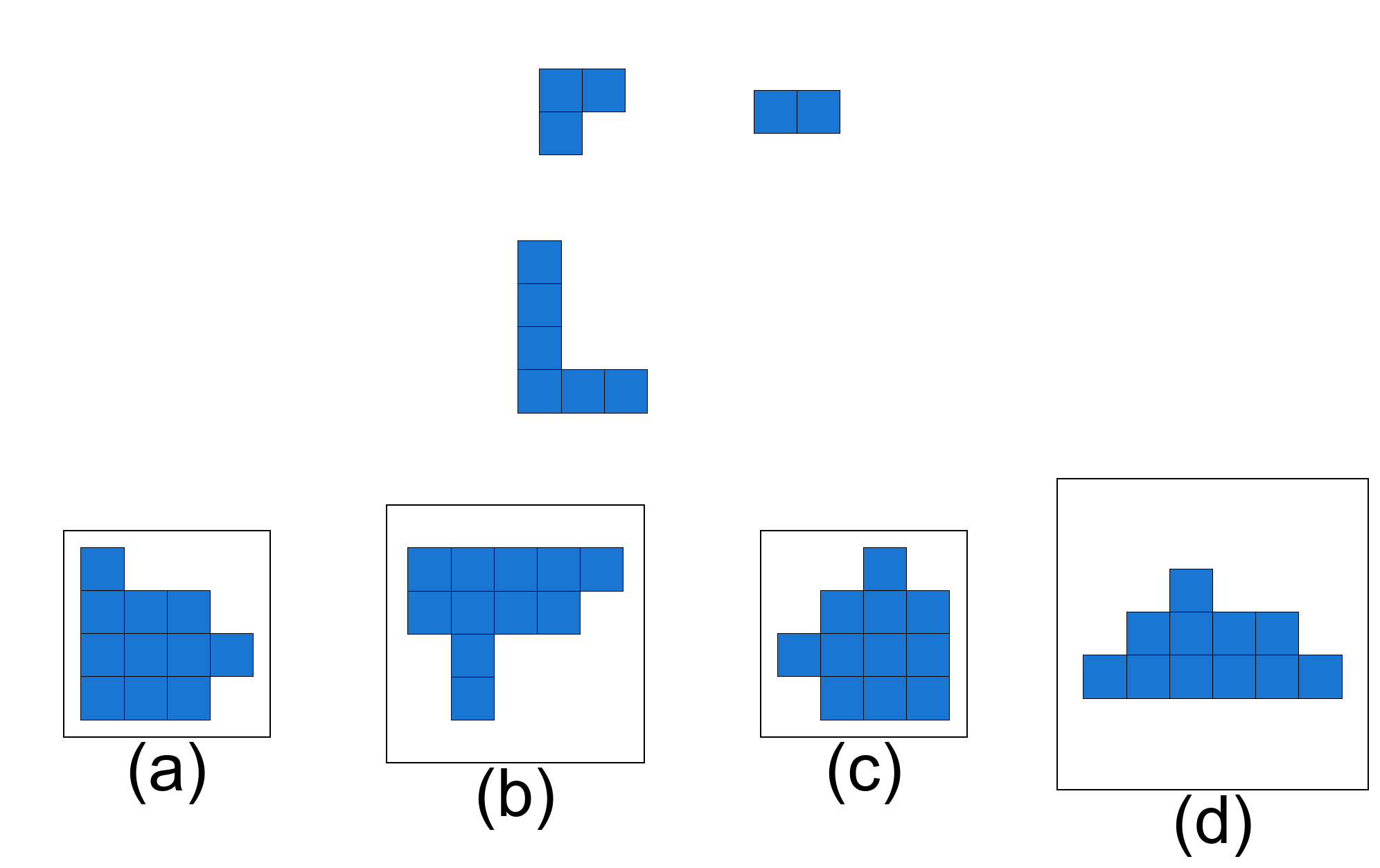}\par}

\textbf{Answer:} (a)

\textbf{Response from GPT-5 (Correct):}\\
Step 1: Determine the areas of the three pieces on top.
- L‑hexomino: 6 unit squares
- L‑triomino: 3 unit squares
- Domino: 2 unit squares
Total area = 6 + 3 + 2 = 11.

Step 2: Count areas of the candidates.
- (a) has 11 squares.
\textcolor{red}{- (b) has 12 squares.}
\textcolor{red}{- (c) has 10 squares.}
\textcolor{red}{- (d) has 10 squares.} Only (a) matches the required area of 11; hence (b), (c) and (d) cannot be tiled by the given pieces.

Step 3: Feasibility check.
Coloring the board in a checkerboard pattern: the hexomino and the domino cover equal numbers of each color, while the triomino covers one more of one color than the other, so the pieces together cover a region with black–white difference 1. Any region with 11 squares (odd area) has such a difference, so (a) is tileable (e.g., place the hexomino along the left and bottom forming an L, the triomino in the upper-left 2×2 corner, and the domino on the right protrusion).

Therefore the only candidate that can be tiled by the top pieces is
\boxed{(a)}.
\end{panelbox}
\end{subfigure}
\hfill
% --- Panel 2: Tiles Recoloring (example 2, top-right) ---
\begin{subfigure}[t]{0.495\textwidth}
\begin{panelbox}[colback=pink!8, equal height group=im2top]
\textbf{(\romannumeral 2)}

{\centering
\includegraphics[width=0.6\linewidth]{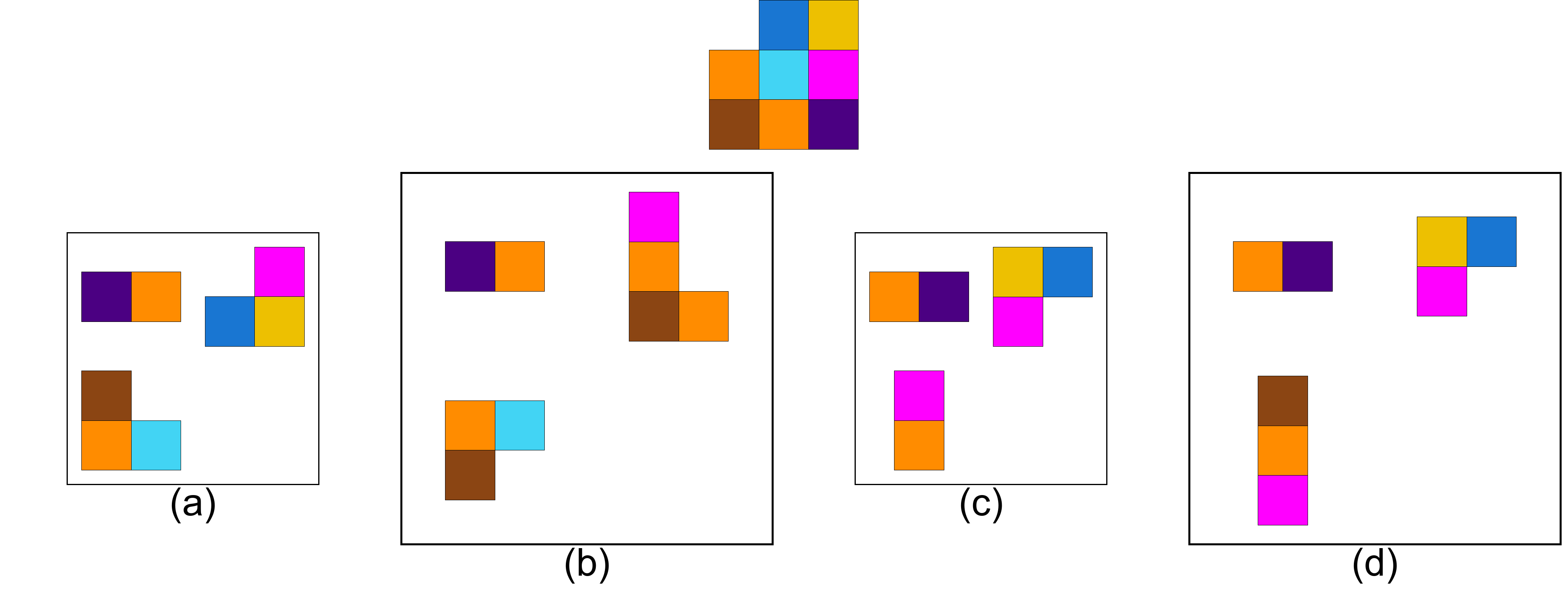}\par}

\textbf{Answer:} (a)

\textbf{Response from GPT-5 (Correct):}\\
Step-by-step reasoning:

- The top figure is a 3×3 grid missing the top-left square (8 unit squares total).

- It can be split into three parts:
  1) An L-shaped triomino formed by the blue–yellow–magenta corner (top-right).
  2) An L-shaped triomino formed by brown–orange–cyan at the lower-left.
  3) A horizontal domino formed by orange–purple at the bottom-right.

- Only bag (a) contains these exact three pieces (two L triominoes and one domino) with the matching color groupings. Other bags have the wrong shapes or number of squares.

Final answer: \boxed{a}

\end{panelbox}
\end{subfigure}

% --- Row 2 --------------------------------------------------------

% --- Panel 3: Tiles Recoloring (example 1, bottom-left; duplicate of Panel 1) ---
\begin{subfigure}[t]{0.495\textwidth}
\begin{panelbox}[colback=black!4, equal height group=im2bot]
\textbf{(\romannumeral 3)}

{\centering
\includegraphics[width=0.5\linewidth]{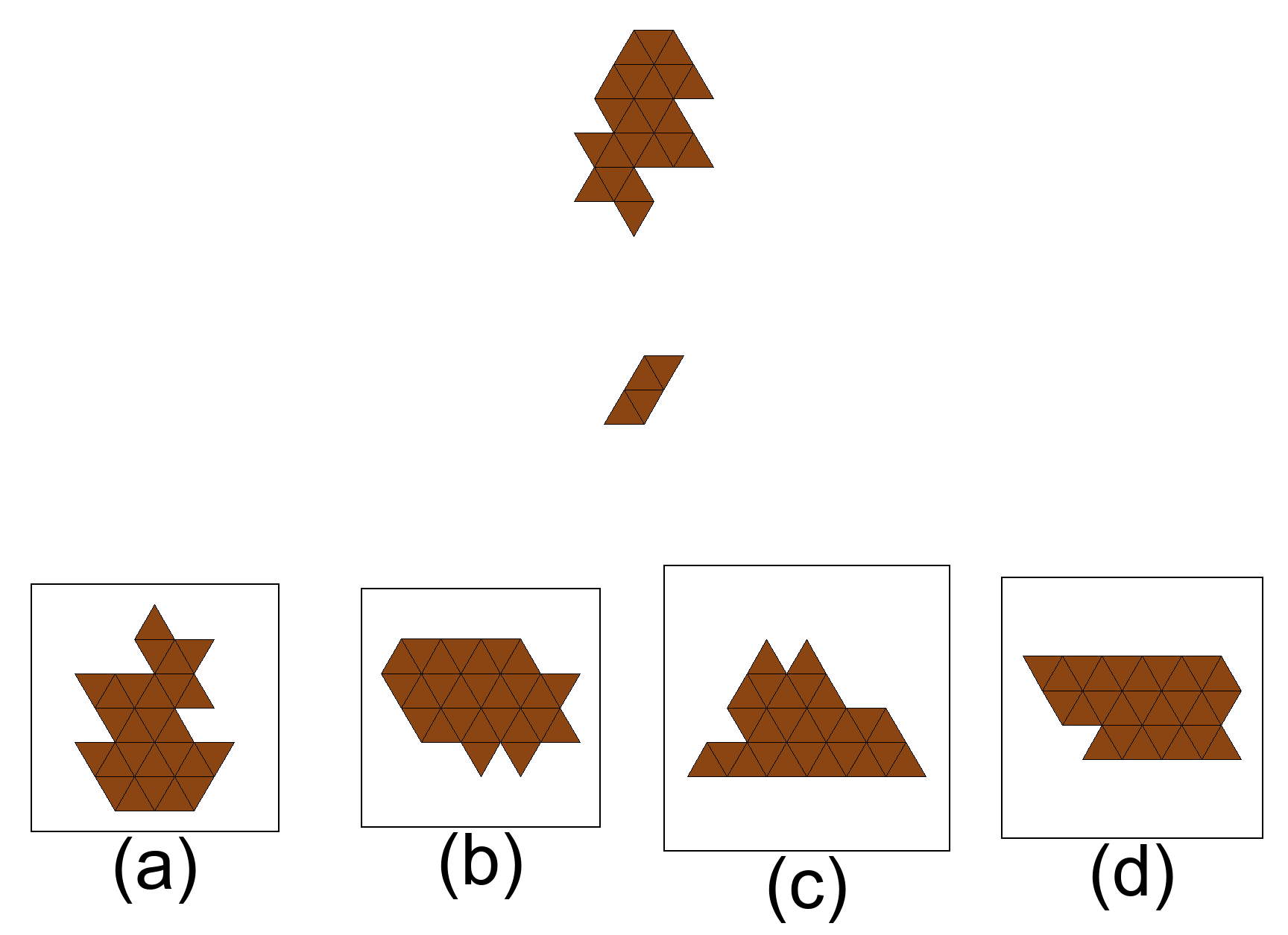}\par}

\textbf{Answer:} (a)

\textbf{Response from GPT-5 (Incorrect):}\\
Reasoning (sketch):

- On a triangular grid, color the unit triangles with three colors in the usual 3‑coloring (so adjacent unit triangles get different colors).  
- For any polyiamond, the triple (number of color‑1, color‑2, color‑3 triangles) is invariant under cutting and reassembling with rotations/reflections.
\textcolor{red}{- Counting the pieces in the “bag” gives one specific triple of counts; comparing these totals with the four candidate tiles, only tile (b) has the same color distribution. 
- Therefore the pieces can be assembled to form (b) and none of the other shapes.}

So the only possible connected tile is \boxed{b}.

\end{panelbox}
\end{subfigure}
\hfill
% --- Panel 4: Tiles Recoloring (example 2, bottom-right; duplicate of Panel 2) ---
\begin{subfigure}[t]{0.495\textwidth}
\begin{panelbox}[colback=black!4, equal height group=im2bot]
\textbf{(\romannumeral 4)}

{\centering
\includegraphics[width=0.6\linewidth]{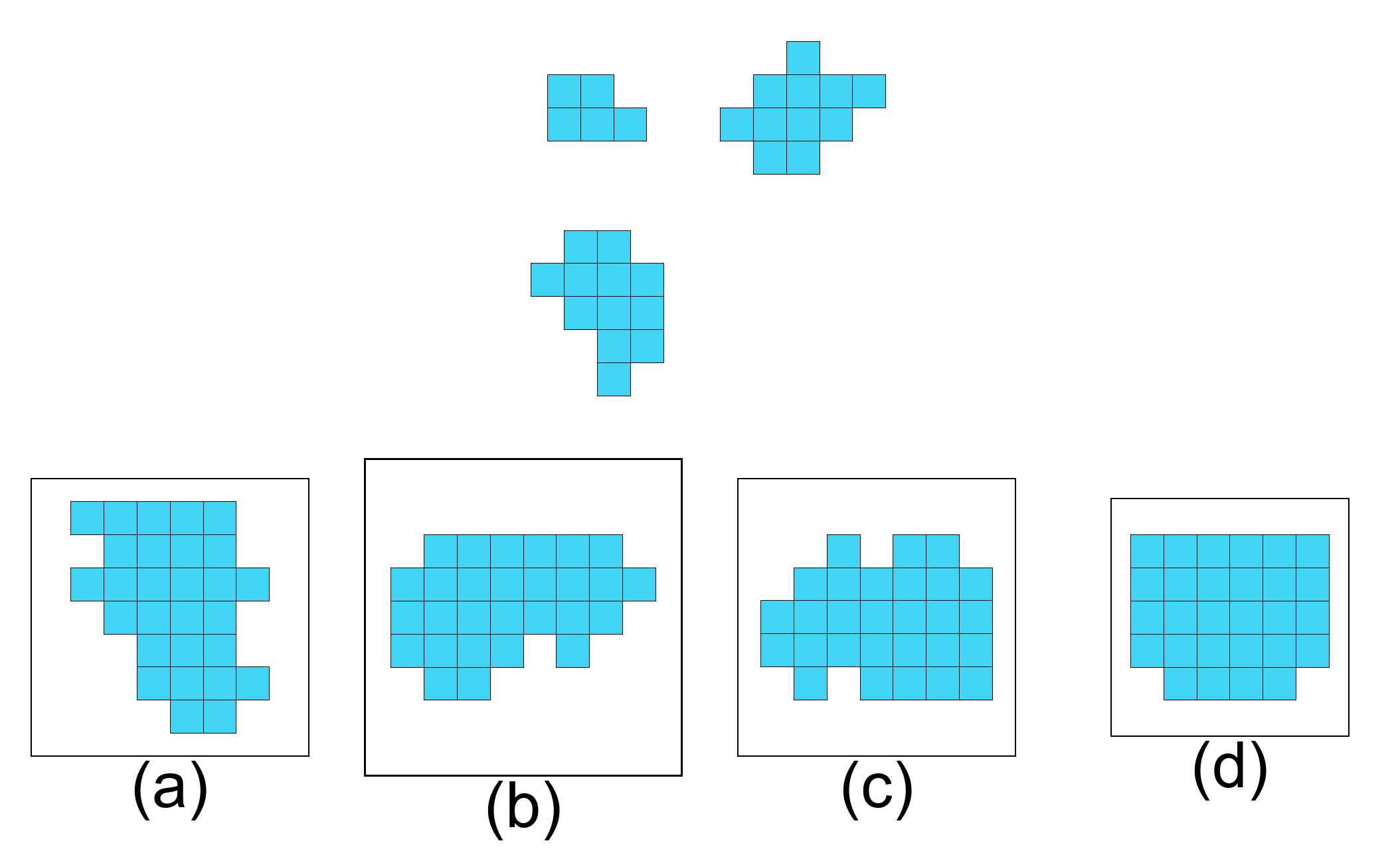}\par}

\textbf{Answer:} (b)

\textbf{Response from GPT-5 (Incorrect):}\\
Step 1: Count the area (number of unit squares) of the pieces in the bag.

- Small piece (2-by-3 with a missing corner): 5 squares.
\textcolor{red}{- Plus-shaped piece with a small extension: 8 squares.}

- Large irregular piece: 12 squares.
\textcolor{red}{Total = 5 + 8 + 12 = 25 squares.}

Step 2: Count the area of each candidate tile.
\textcolor{red}{(a) 25 squares}
\textcolor{red}{(b) 27 squares}
\textcolor{red}{(c) 26 squares}
\textcolor{red}{(d) 29 squares}

\textcolor{red}{Only option (a) has the same total area as the three pieces, and the shapes can be fitted together (rotations/reflections allowed) to fill it without gaps or overlaps.}

Final answer: \boxed{(a)}

\end{panelbox}
\end{subfigure}

\caption{Example responses from GPT-5 on \emph{Tiles Composition} Task, with incorrect reasoning highlighted in red.}
\label{fig:gpt5_tilescomposition}
\end{figure*}

\subsection{Tiles Composition}

Tiles Composition is the task with the second-highest human accuracy, and GPT-5 exceeds human performance by a small margin (about 2\%). The task itself is nontrivial, even for humans. GPT-5 often produced answers without providing explicit reasoning. Of 100 questions, 30 responses from the API did not contain visible reasoning. Figure~\ref{fig:gpt5_tilescomposition} shows representative examples, with two correct responses in the upper row and two incorrect responses in the lower row.

In the correct cases, GPT-5 sometimes reached the right answer despite imperfect or partially incorrect reasoning. For instance, in example (\romannumeral 1), the model attempted to compare the option tiles by counting the total number of small squares. Although its counts were inconsistent across the choices, it nevertheless provided a correct explanation afterward, describing how the smaller pieces can be assembled to form the larger target shape. This suggests that the model is capable of meaningful geometric reasoning even when the initial steps of its analysis are flawed. Example (\romannumeral 2)\ further illustrates this. There, the model gives a coherent explanation of how the large block can be decomposed into the three constituent parts, which aligns with the intended reasoning for the task.

The incorrect cases reveal different failure modes. In example (\romannumeral 3), the model appears to attempt a more sophisticated internal strategy for reasoning about the triangular tiling, although the explicit details do not appear in the response. In the end, it draws the wrong conclusion. In example (\romannumeral 4), the model again relies on counting the number of small squares in the tiles. All options and the three top blocks contain exactly 28 unit squares, yet the model incorrectly states that the top image contains 25 squares and that option (a) also contains 25. This suggests that the model may be producing a post-hoc justification for an incorrect answer rather than performing a reliable visual count.

Overall, these examples show that GPT-5 has some genuine ability to reason about shape composition, but that its success depends heavily on extracting accurate low-level visual features, which remains an inconsistent step in its pipeline.

\begin{figure*}[t]
\centering

% --- Row 1 --------------------------------------------------------

% --- Panel 1: Tiles Recoloring (example 1, top-left) ---
\begin{subfigure}[t]{0.495\textwidth}
\begin{panelbox}[colback=black!4, equal height group=im3top]
\textbf{(\romannumeral 1)}

{\centering
\includegraphics[width=0.5\linewidth]{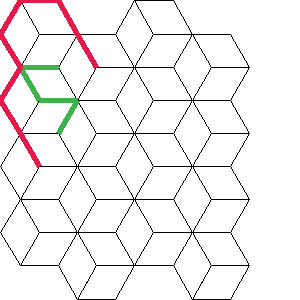}\par}

\textbf{Question:} How many edge segments make up the red (\#e6194b) line? Give your answer as an integer.

\textbf{Answer:} 8

\textbf{Response from GPT-5 (Incorrect):}\\
\textcolor{red}{- Along the left edge, there are 5 stacked rhombi.}

- A vertical zigzag along n rhombi uses 2n edge segments → \textcolor{red}{2×5 = 10. }

\textcolor{red}{- At the top, the red line extends by 2 additional slanted edges.}

Total edge segments = 10 + 2 = 12.

\boxed{12}

\end{panelbox}
\end{subfigure}
\hfill
% --- Panel 2: Tiles Recoloring (example 2, top-right) ---
\begin{subfigure}[t]{0.495\textwidth}
\begin{panelbox}[colback=pink!8, equal height group=im3top]
\textbf{(\romannumeral 2)}

{\centering
\includegraphics[width=0.5\linewidth]{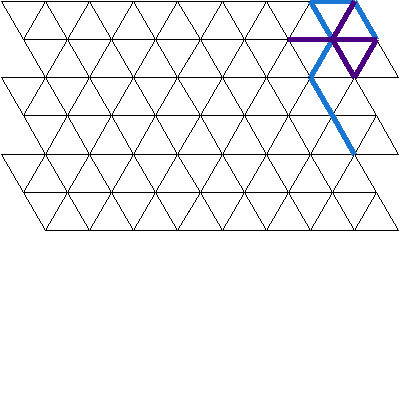}\par}

\textbf{Question:} Measure the blue (\#1976d2) line strictly along tile edges. How many tile-edge units long is it?

\textbf{Answer:} 6

\textbf{Response from GPT-5 (Incorrect):}\\
Step-by-step:

- The blue line is along tile edges, so its length is the number of unit edges it covers.

- Count each straight blue segment:
  - Short top segment: 1 unit edge.
  \textcolor{red}{- Long slanted segment: 7 unit edges (trace each edge along that direction).}
  \textcolor{red}{- Short bottom segment: 1 unit edge.}

\textcolor{red}{- Total length = 1 + 7 + 1 = 9 unit edges.}
\boxed{9}

\end{panelbox}
\end{subfigure}

% --- Row 2 --------------------------------------------------------

% --- Panel 3: Tiles Recoloring (example 1, bottom-left; duplicate of Panel 1) ---
\begin{subfigure}[t]{0.495\textwidth}
\begin{panelbox}[colback=black!4, equal height group=im3bot]
\textbf{(\romannumeral 3)}

{\centering
\includegraphics[width=0.5\linewidth]{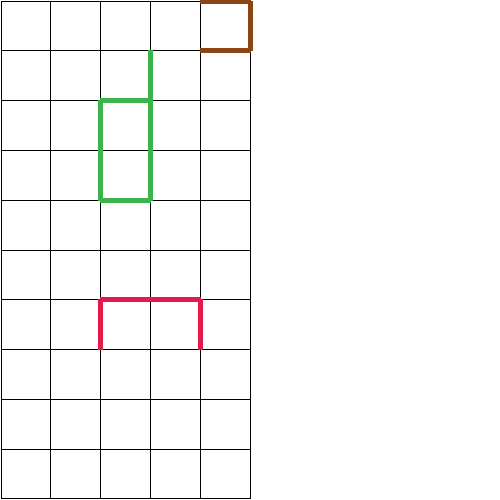}\par}

\textbf{Question:} The brown (\#8b4513) line is a continuous path along grid edges. What is its total length in edge steps?

\textbf{Answer:} 3

\textbf{Response from GPT-5 (Incorrect):}\\

\textcolor{red}{
- Count the grid size: 5 columns by 8 rows.}

\textcolor{red}{- A continuous path along grid edges around the outside has length equal to the perimeter in edge steps: 2(columns + rows) = 2(5 + 8) = 26.
}
\boxed{26}

\end{panelbox}
\end{subfigure}
\hfill
% --- Panel 4: Tiles Recoloring (example 2, bottom-right; duplicate of Panel 2) ---
\begin{subfigure}[t]{0.495\textwidth}
\begin{panelbox}[colback=black!4, equal height group=im3bot]
\textbf{(\romannumeral 4)}

{\centering
\includegraphics[width=0.5\linewidth]{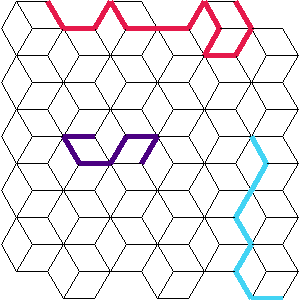}\par}

\textbf{Question:} Colored lines run along the edges of the tiling, from vertex to vertex. What is the total length of the cyan (\#42d4f4) line, measured in edge steps?

\textbf{Answer:} 7

\textbf{Response from GPT-5 (Incorrect):}\\
Reasoning:

- The grid is an isometric (rhombus) tiling with three edge directions.

- The cyan path is a “vertical” zigzag, alternating between the two diagonal edge directions.

- In such a grid, moving down one vertical level requires two edge steps (one down-right + one down-left).

\textcolor{red}{- Counting the vertical levels between the endpoints of the cyan path gives 6 levels.}

\textcolor{red}{Therefore, the cyan line uses 2 × 6 = 12 edge steps.}
\boxed{12}

\end{panelbox}
\end{subfigure}

\caption{Example responses from GPT-5 on \emph{Tiles Line Length} Task, with incorrect reasoning highlighted in red.}
\label{fig:gpt5_tileslinelength}
\end{figure*}

\subsection{Tiles Line Length}

Tiles Line Length is the task where GPT-5 performs the worst relative to humans, reaching only 14\% accuracy compared to the human accuracy of 86.4\%. Figure~\ref{fig:gpt5_tileslinelength} shows four representative examples in which the model produced incorrect responses. These cases illustrate several different types of visual reasoning failures.

In example (\romannumeral 1), the model incorrectly reports five stacked rhombi rather than four and also assigns the wrong colors, stating that all segments are red. In example (\romannumeral 2), the model overcounts the length of the blue slanted segment. Part of this mistake may stem from conflating the blue and purple segments and treating them as a single continuous piece. In example (\romannumeral 3), the model appears to miss the brown segment entirely and instead attempts to compute the length of a continuous path along the outer boundary. This reasoning is also incorrect because the grid contains ten rows rather than eight, which invalidates the model's length calculation. In example (\romannumeral 4), the model seems to infer six distinct vertical levels based on directional changes in the cyan segment, which again leads to an incorrect total.

Taken together, these examples suggest that GPT-5 relies on a collection of heuristics rather than a reliable visual understanding of the line structures. Humans can accurately isolate segments and identify endpoints, but the model does not consistently extract these visual primitives, which prevents it from solving the task correctly.

\subsection{Tiles Recoloring}

\begin{figure*}[t]
\centering

% --- Row 1 --------------------------------------------------------

% --- Panel 1: Tiles Recoloring (example 1, top-left) ---
\begin{subfigure}[t]{0.495\textwidth}
\begin{panelbox}[colback=black!4, equal height group=im4top]
\textbf{(\romannumeral 1)}

{\centering
\includegraphics[width=0.99\linewidth]{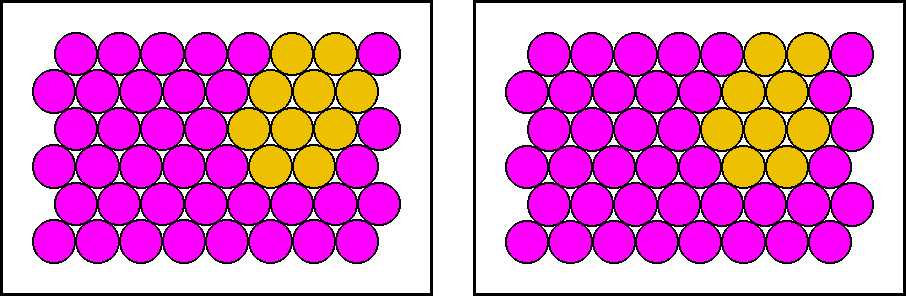}\par}

\medskip

\textbf{Question:} How many cells are different between the left and right images? Answer as an integer.

\medskip
\textbf{Answer:} 1

\medskip
\textbf{Response from GPT-5 (Incorrect):}\\
Step-by-step reasoning:

\textcolor{red}{- The yellow region in each image forms a 3-by-4 rectangle (12 cells).}  

\textcolor{red}{- In the right image this rectangle is shifted one cell down and one cell to the right relative to the left image.}  

- The overlap between a w×h rectangle and the same rectangle shifted by (1,1) has size (w-1)(h-1).

- Cells that differ = 2 × [w·h - (w-1)(h-1)] = 2(w + h - 1).  

\textcolor{red}{- With w = 3 and h = 4: 2(3 + 4 - 1) = 12.}

Final answer: \boxed{12}
\end{panelbox}
\end{subfigure}
\hfill
% --- Panel 2: Tiles Recoloring (example 2, top-right) ---
\begin{subfigure}[t]{0.495\textwidth}
\begin{panelbox}[colback=pink!8, equal height group=im4top]
\textbf{(\romannumeral 2)}

{\centering
\includegraphics[width=0.99\linewidth]{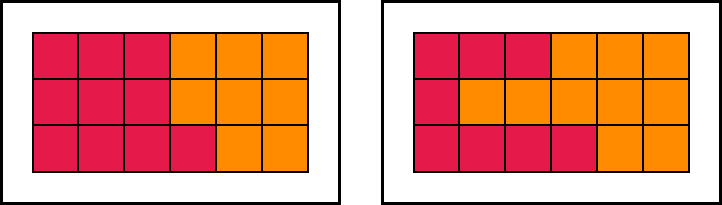}\par}

\medskip

\textbf{Question:} Compare the two tiles. How many cells do not match between left and right? Give an integer.

\medskip
\textbf{Answer:} 2
\medskip
\textbf{Response from GPT-5 (Incorrect):}\\
Step-by-step reasoning:

- Each tile is a 3×6 grid.

- Left tile colors (R=red, O=orange):

  Row1: R R R O O O

  Row2: R R R O O O

  Row3: R R R R O O

- Right tile colors:

  \textcolor{red}{Row1: R R O O O O}

  Row2: R O O O O O

  \textcolor{red}{Row3: R R R O O O}

- Compare cell by cell. The cells that differ are:

  (row1,col3), (row2,col2), (row2,col3), (row3,col4).

\textcolor{red}{- Total mismatches = 4.}

Final answer: \boxed{4}

\end{panelbox}
\end{subfigure}

% --- Row 2 --------------------------------------------------------
% \vspace{5em}
% --- Panel 3: Tiles Recoloring (example 1, bottom-left; duplicate of Panel 1) ---
\begin{subfigure}[t]{0.495\textwidth}
\begin{panelbox}[colback=black!4, equal height group=im4bot]
\textbf{(\romannumeral 3)}

{\centering
\includegraphics[width=0.99\linewidth]{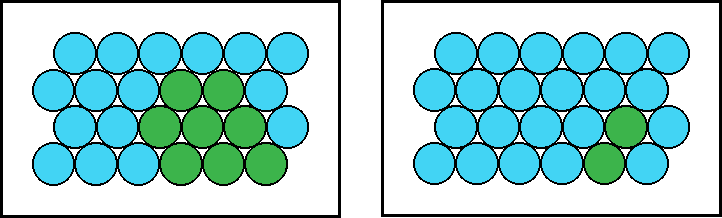}\par}

\medskip

\textbf{Question:} A difference is any cell with unequal colors (including blank vs. filled). How many differences are there?

\medskip

\textbf{Answer:} 6

\medskip

\textbf{Response from GPT-5 (Incorrect):}\\
We compare the two grids cell by cell. Treat each possible circle position as a cell and count a difference whenever the two pictures have different colors at that position.

\textcolor{red}{- There are 26 cells in each picture (6–7–6–7 in the four rows).}

\textcolor{red}{- On the left, 9 of those cells are green; on the right, only 1 cell is green.}

\textcolor{red}{- That one green cell on the right is also green on the left, so it is not a difference.}

\textcolor{red}{- Thus the number of cells with different colors is 9 + 1 - 2 × 1 = 8.}

\boxed{8}

\end{panelbox}
\end{subfigure}
\hfill
% --- Panel 4: Tiles Recoloring (example 2, bottom-right; duplicate of Panel 2) ---
\begin{subfigure}[t]{0.495\textwidth}
\begin{panelbox}[colback=black!4, equal height group=im4bot]
\textbf{(\romannumeral 4)}

{\centering
\includegraphics[width=0.99\linewidth]{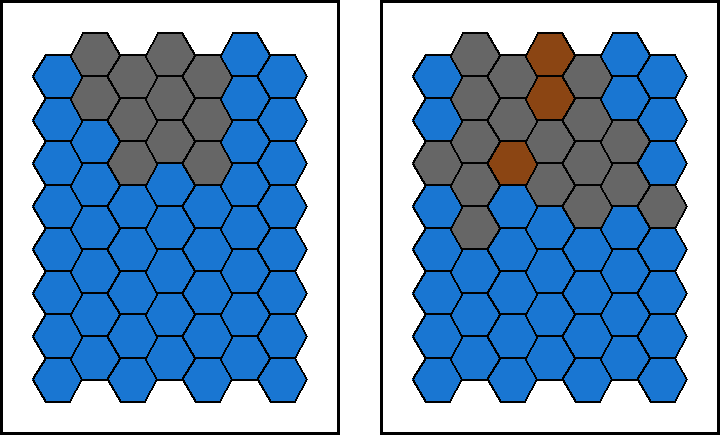}\par}

\medskip
\textbf{Question:} Count cells whose colors do not match between left and right (filled vs. blank also counts). Provide an integer.

\medskip
\textbf{Answer:} 12

\medskip
\textbf{Response from GPT-5 (Incorrect):}\\
\textcolor{red}{Only two cells differ. In the right image, two hexes are brown where the left image has grey; all other cells match.} 

\boxed{2}

\end{panelbox}
\end{subfigure}

\caption{Example responses from GPT-5 on \emph{Tiles Recoloring} Task, with incorrect reasoning highlighted in red.}
\label{fig:gpt5_tilesrecoloring}
\end{figure*}

Tiles Recoloring is a challenging task for all models, including GPT-5, and Figure~\ref{fig:gpt5_tilesrecoloring} shows four representative failure cases. Across these examples, GPT-5 appears to attempt several different strategies to solve the task, yet errors are introduced at various points in these strategies.

In example (\romannumeral 1), the model misjudges the size of the yellow region and concludes that it has been shifted to the right in the second image. In reality, the yellow region is almost perfectly aligned between the two images, with only one cell recolored as magenta. This leads the model to output 12 rather than the correct answer of 1. In example (\romannumeral 2), GPT-5 attempts to convert both tiles into a letter-matrix representation with colors assigned to symbols. This is a valid approach, but the model makes errors when transcribing the right tile into matrix form, resulting in an incorrect comparison.

In example (\romannumeral 3), the model switches to a different algorithm. It identifies that some green cells in the left image have been recolored as cyan in the right image, and proposes counting the green cells in both images and taking their difference. However, it overcounts the green cells in both images by one, which leads to the wrong answer. In example (\romannumeral 4), the model fails to detect additional gray cells that appear only in the right image and therefore underestimates the number of mismatches.

Taken together, these examples indicate that GPT-5's low accuracy on this task is driven primarily by failures in color-specific tile counting. Even when the model proposes valid high-level methods for solving the task, it struggles to carry them out reliably, because the core step of accurately identifying and counting colored cells across two tiles is error prone.

\subsection{Frieze Groups} 
\begin{figure*}[t]
\centering

% --- Row 1 --------------------------------------------------------

% --- Panel 1: Tiles Recoloring (example 1, top-left) ---
\begin{subfigure}[t]{0.495\textwidth}
\begin{panelbox}[colback=black!4, equal height group=im5top]
\textbf{(\romannumeral 1)}

{\centering
\includegraphics[width=0.9\linewidth]{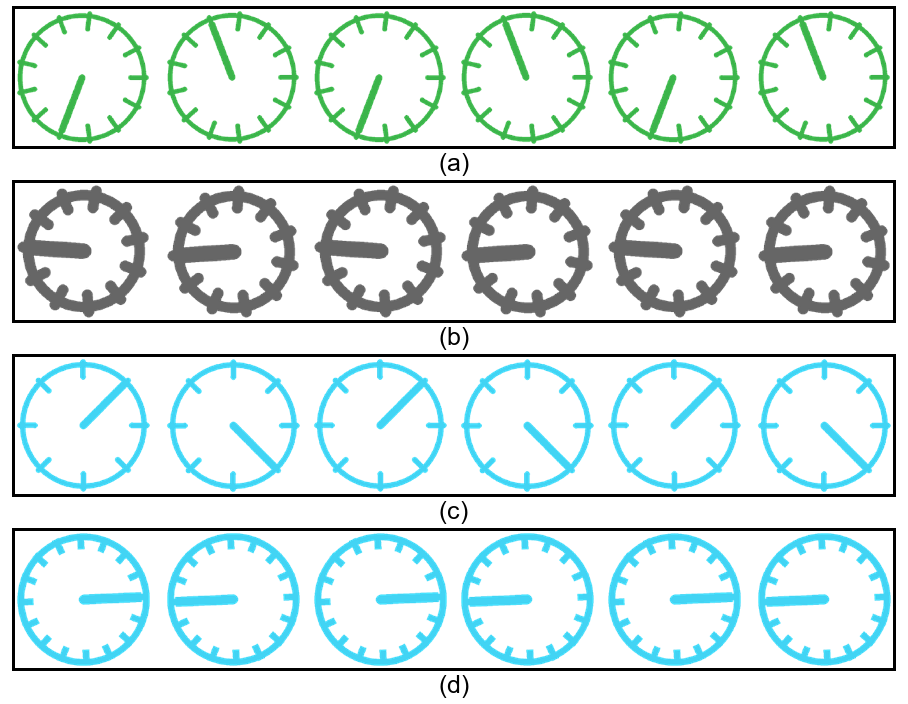}\par}

\textbf{Question:}  Inspect how shapes relate to their neighbors. Which strip follows a different rule than the rest?

\textbf{Answer:} (d)

\textbf{Response from GPT-5 (Incorrect):}\\
\boxed{(b)}

Reason: \textcolor{red}{In strips (a), (c) and (d), adjacent clocks are related by a 180° change of the hand (they alternate between two opposite orientations). In strip (b) all the hands point in the same direction, so it follows a different neighbor relation.}
\end{panelbox}
\end{subfigure}
\hfill
% --- Panel 2: Tiles Recoloring (example 2, top-right) ---
\begin{subfigure}[t]{0.495\textwidth}
\begin{panelbox}[colback=pink!8, equal height group=im5top]
\textbf{(\romannumeral 2)}

{\centering
\includegraphics[width=0.9\linewidth]{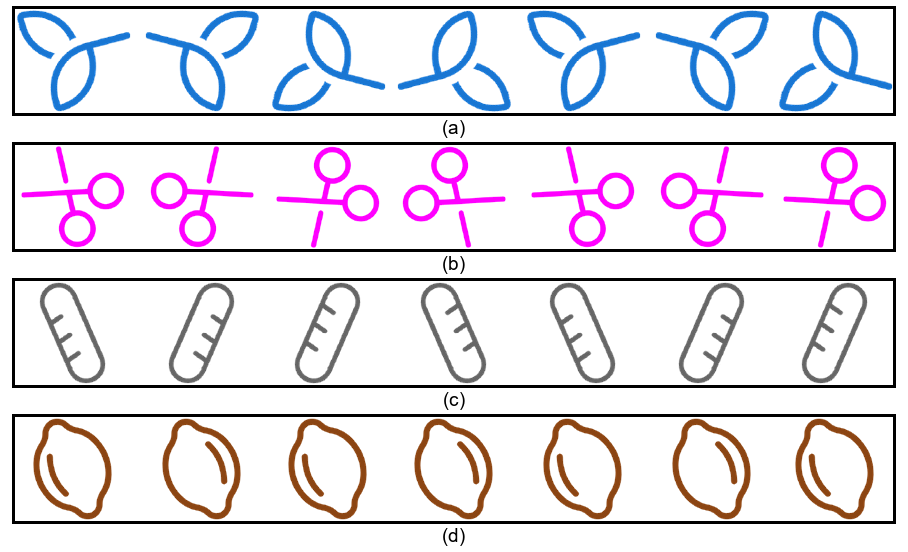}\par}

\textbf{Question:} Which strip’s sequence of transformations doesn’t match the other three? Choose (a–d).

\textbf{Answer:} (d)

\textbf{Response from GPT-5 (Incorrect):}\\
Step-by-step reasoning:

\textcolor{red}{- In strips (a), (b), and (d), each motif alternates between left-right mirror images of the previous one (reflection).}

\textcolor{red}{- In strip (c), the motif repeats with the same orientation each time; there is no alternating reflection.}

\textcolor{red}{Therefore, the sequence of transformations in (c) doesn’t match the other three.}

\boxed{c}

\end{panelbox}
\end{subfigure}

% --- Row 2 --------------------------------------------------------

% --- Panel 3: Tiles Recoloring (example 1, bottom-left; duplicate of Panel 1) ---
\begin{subfigure}[t]{0.495\textwidth}
\begin{panelbox}[colback=black!4, equal height group=im5bot]
\textbf{(\romannumeral 3)}

{\centering
\includegraphics[width=0.9\linewidth]{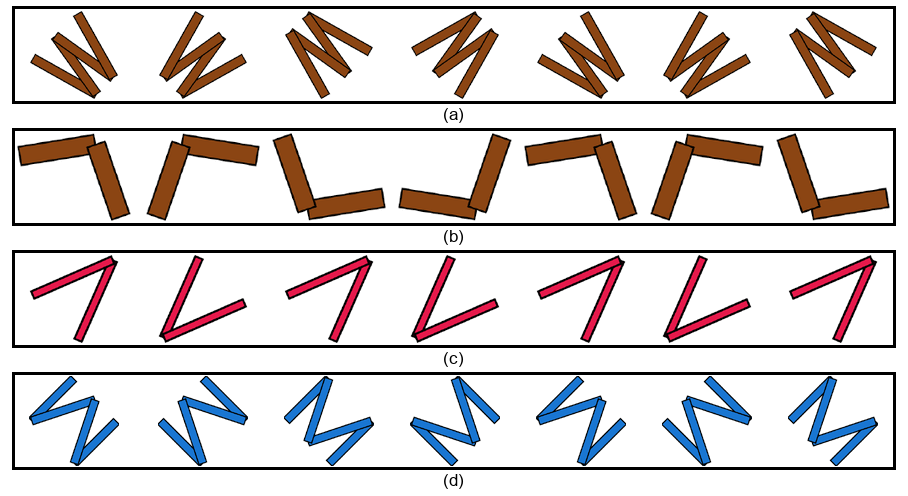}\par}

\textbf{Question:} Only one strip uses a different symmetry between adjacent motifs. Select it (a–d).

\textbf{Answer:} (c)

\textbf{Response from GPT-5 (Incorrect):}\\
Step-by-step reasoning:

- Compare each pair of neighboring motifs and identify the symmetry that maps one to the next.

\textcolor{red}{- In strips (a), (b) and (c), adjacent motifs are mirror images across a vertical line between them (left-right reflection).}

\textcolor{red}{- In strip (d), adjacent motifs alternate as “M” and “W”, i.e., they are mirror images across a horizontal line (top-bottom reflection), not a vertical one.}

Therefore the only strip using a different symmetry is

\boxed{d}

\end{panelbox}
\end{subfigure}
\hfill
% --- Panel 4: Tiles Recoloring (example 2, bottom-right; duplicate of Panel 2) ---
\begin{subfigure}[t]{0.495\textwidth}
\begin{panelbox}[colback=black!4, equal height group=im5bot]
\textbf{(\romannumeral 4)}

{\centering
\includegraphics[width=0.9\linewidth]{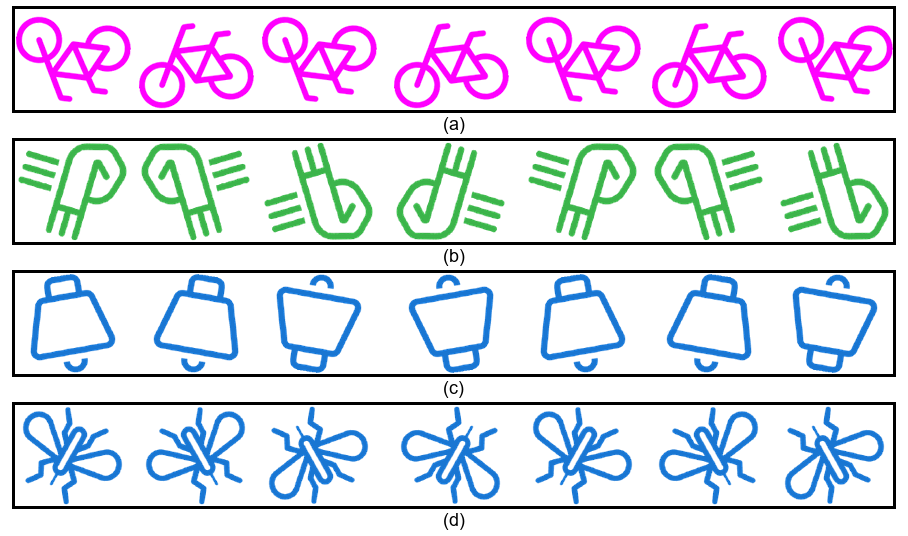}\par}

\textbf{Question:} Only one strip uses a different symmetry between adjacent motifs. Select it (a–d).

\textbf{Answer:} (a)

\textbf{Response from GPT-5 (Incorrect):}\\
Step-by-step reasoning:

- Look at how one motif changes to the next.

\textcolor{red}{- In (a), (b) and (d) the motifs remain upright from one to the next; they just switch left–right. That means adjacent motifs are related by a mirror reflection in a vertical line.}

- In (c) the bell alternates upright and upside‑down: the handle and clapper swap top and bottom. \textcolor{red}{That relation is a 180° rotation (half‑turn), not a mirror.}

Therefore, only strip (c) uses a different symmetry between adjacent motifs.

\boxed{c}

\end{panelbox}
\end{subfigure}

\caption{Example responses from GPT-5 on \emph{Frieze Groups} task, with incorrect reasoning highlighted in red.}
\label{fig:gpt5_frieze}
\end{figure*}

The Frieze Groups task has the lowest human accuracy (48.4\%), and GPT-5 achieves only 26\% accuracy. Since the task is presented in a multiple-choice format with four options, GPT-5's performance is effectively at the level of random guessing. Figure~\ref{fig:gpt5_frieze} presents four representative cases of GPT-5 failure on the Frieze Groups task. The task requires identifying the symmetry pattern that governs the repetition of motifs along a one-dimensional strip, and the examples highlight several recurring challenges for the model. 

In example (\romannumeral 1), GPT-5 reasons that in strips (a), (c), and (d) the clocks differ by a 180° change of the clock hands. In reality, only strip (d) exhibits a 180° rotation, corresponding to group p2. Strips (a), (b), and (c) follow a horizontal reflection pattern consistent with the group p11m. The model, therefore, misclassifies the symmetry in multiple strips. In example (\romannumeral 2), the model makes several errors while analyzing the sequence. Strips (a), (b), and (c) follow a spinning sidle pattern (p2mg), which combines rotation with vertical reflection. Strip (d), in contrast, follows only a 180° rotation. GPT-5 incorrectly concludes that strip (c) repeats in the same orientation, although its visual change in orientation is more pronounced than in strip (d).

In example (\romannumeral 3), strips (a), (b), and (d) follow a combination of horizontal and vertical reflections (p2mm), while strip (c) follows a 180° rotation. The model incorrectly assigns the reflection to the strip (c) and fails to detect the horizontal reflection in the strips (a) and (b) and the vertical reflection present in strip (d). In example~(\romannumeral 4), strips (b), (c) and (d) follow a reflection followed by a rotation (p2mg), while strip (a) uses a horizontal reflection (p11g). GPT-5 incorrectly infers that strips (b) and (d) remain upright and ignores the rotational component. It also incorrectly attributes a vertical reflection to strip (a) and, in strip (c), recognizes the rotation but fails to identify the reflection. These examples suggest that GPT-5 struggles with fine-grained symmetry analysis in one-dimensional sequences.

\begin{figure*}[t]
\centering

% --- Row 1 --------------------------------------------------------

% --- Panel 1: Tiles Recoloring (example 1, top-left) ---
\begin{subfigure}[t]{0.495\textwidth}
\begin{panelbox}[colback=black!4, equal height group=im6top]
\textbf{(\romannumeral 1)}

{\centering
\includegraphics[width=0.45\linewidth]{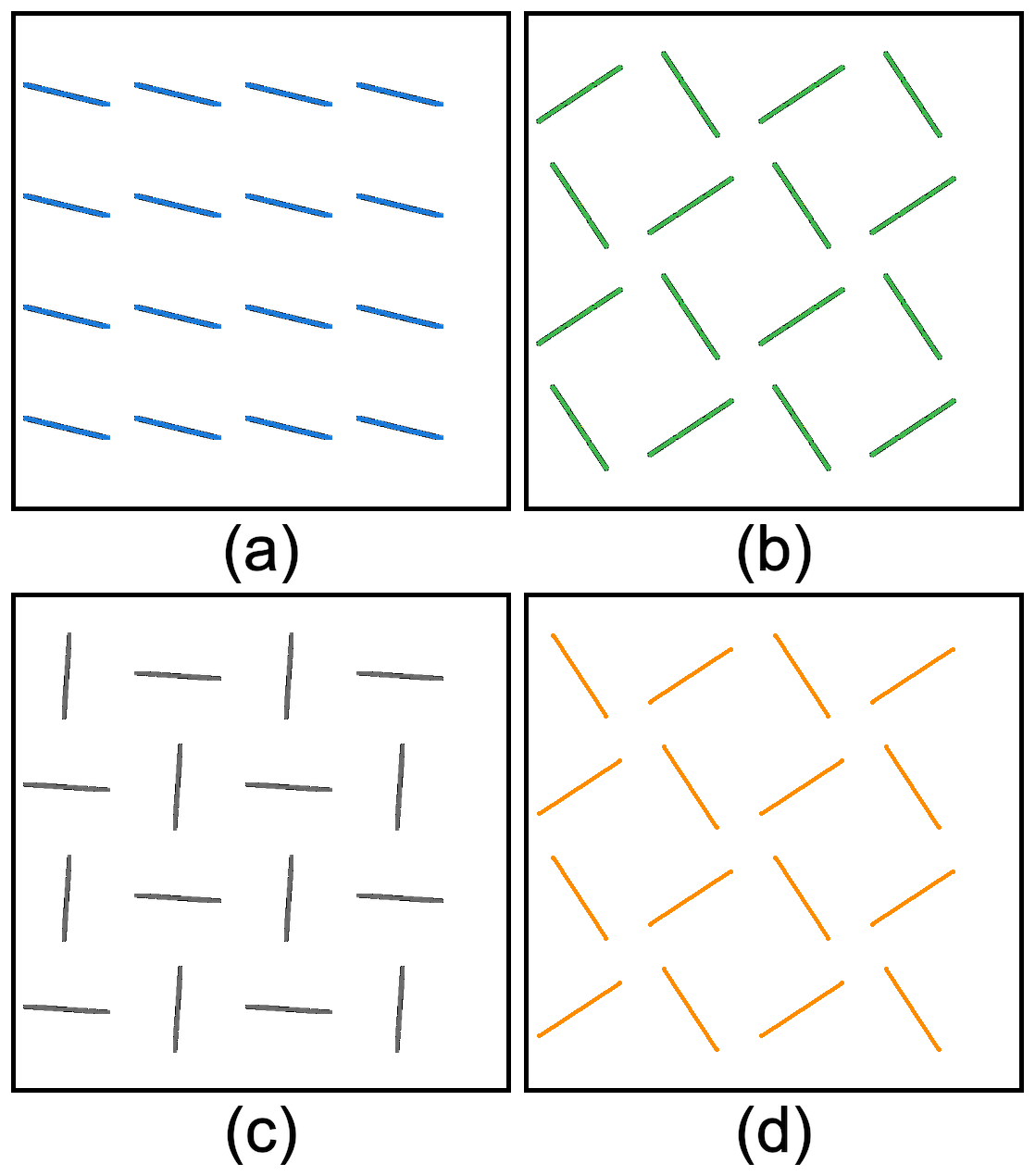}\par}

% \textbf{Question:}  Three panels follow one wallpaper rule, while one uses another. Which panel (a–d) in the 2×2 grid is different?

\textbf{Answer:} (a)

\textbf{Response from GPT-5 (Correct):}\\
Step-by-step:

- Look for rotational symmetries.

- Panels (b), (c) and (d) each have points with 90° rotational (fourfold) symmetry: rotating the pattern by a quarter-turn about the centers of the squares maps the pattern onto itself. They belong to a p4-type wallpaper rule.

- Panel (a) lacks any quarter-turn symmetry; it only has translations and 180° rotations (p2-type).

Therefore, the panel that uses a different wallpaper rule is
\boxed{a}.

\end{panelbox}
\end{subfigure}
\hfill
% --- Panel 2: Tiles Recoloring (example 2, top-right) ---
\begin{subfigure}[t]{0.495\textwidth}
\begin{panelbox}[colback=pink!8, equal height group=im6top]
\textbf{(\romannumeral 2)}

{\centering
\includegraphics[width=0.45\linewidth]{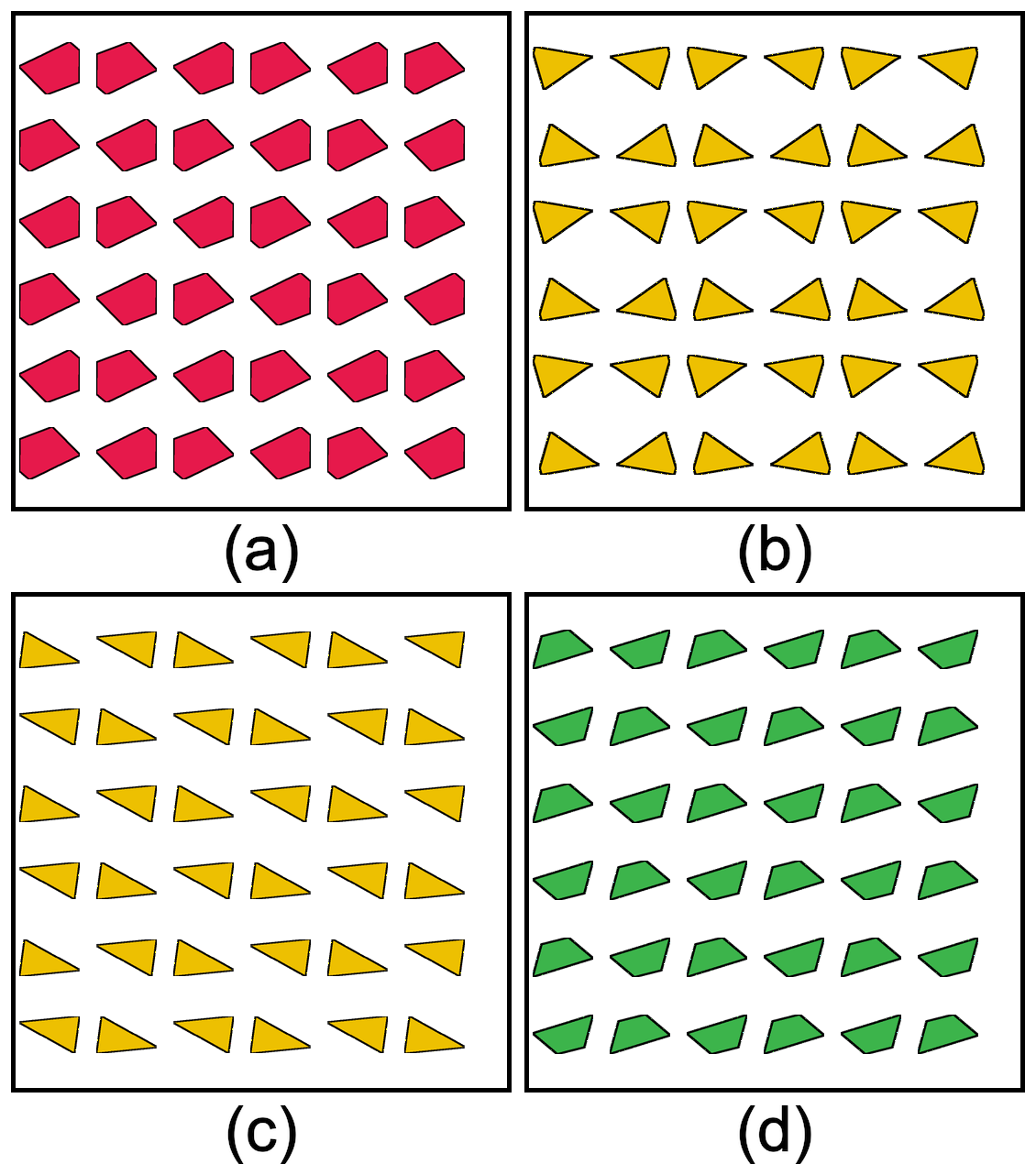}\par}

% \textbf{Question:} Three panels follow one wallpaper rule, while one uses another. Which panel (a–d) in the 2×2 grid is different?

\textbf{Answer:} (b)

\textbf{Response from GPT-5 (Correct):}\\
Step-by-step reasoning:

- Look at how each motif relates to its nearest neighbors.

- In panel (b), every immediate neighbor (left, right, up, and down) is the same motif rotated by 180°. The orientations therefore alternate in a checkerboard pattern; the symmetry is based on half-turns between each pair.

- In panels (a), (c), and (d), \textcolor{red}{motifs within a row keep the same orientation} and the next row is obtained by a glide/mirror of the previous row (not by a 180° rotation between immediate neighbors).

Thus, (b) follows a different wallpaper rule from the other three.\boxed{b}

\end{panelbox}
\end{subfigure}

% --- Row 2 --------------------------------------------------------

% --- Panel 3: Tiles Recoloring (example 1, bottom-left; duplicate of Panel 1) ---
\begin{subfigure}[t]{0.495\textwidth}
\begin{panelbox}[colback=black!4, equal height group=im6bot]
\textbf{(\romannumeral 3)}

{\centering
\includegraphics[width=0.45\linewidth]{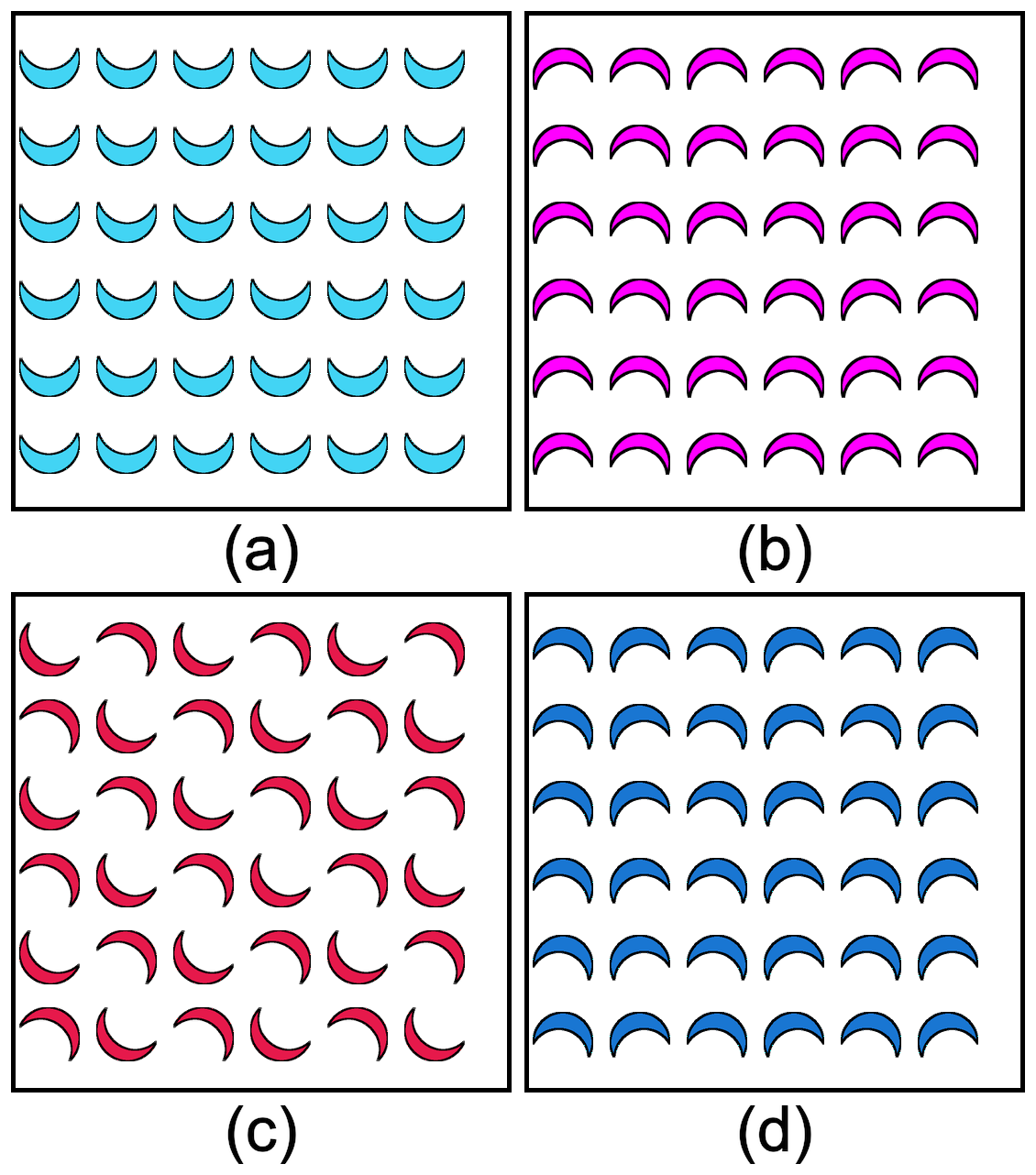}\par}

% \textbf{Question:} In the 2×2 grid of patches, one tile is generated by a different combination of flips/rotations/glides. Which one (a–d) is it?

\textbf{Answer:} (c)

\textbf{Response from GPT-5 (Incorrect):}\\
Step-by-step reasoning:

- Look for symmetries other than translation (flips, rotations, glides) that generate each repetitive patch.

\textcolor{red}{- In (a), every crescent has the same orientation everywhere.} The pattern can be produced by pure translations only; there are no flips/rotations/glides.

\textcolor{red}{- In (b), (c), and (d), neighboring motifs are related by half‑turns or flips (and glides), so these patterns require additional symmetries beyond simple translations.}

Therefore the one generated by a different set of operations is patch (a).

Final answer: \boxed{a}

\end{panelbox}
\end{subfigure}
\hfill
% --- Panel 4: Tiles Recoloring (example 2, bottom-right; duplicate of Panel 2) ---
\begin{subfigure}[t]{0.495\textwidth}
\begin{panelbox}[colback=black!4, equal height group=im6bot]
\textbf{(\romannumeral 4)}

{\centering
\includegraphics[width=0.45\linewidth]{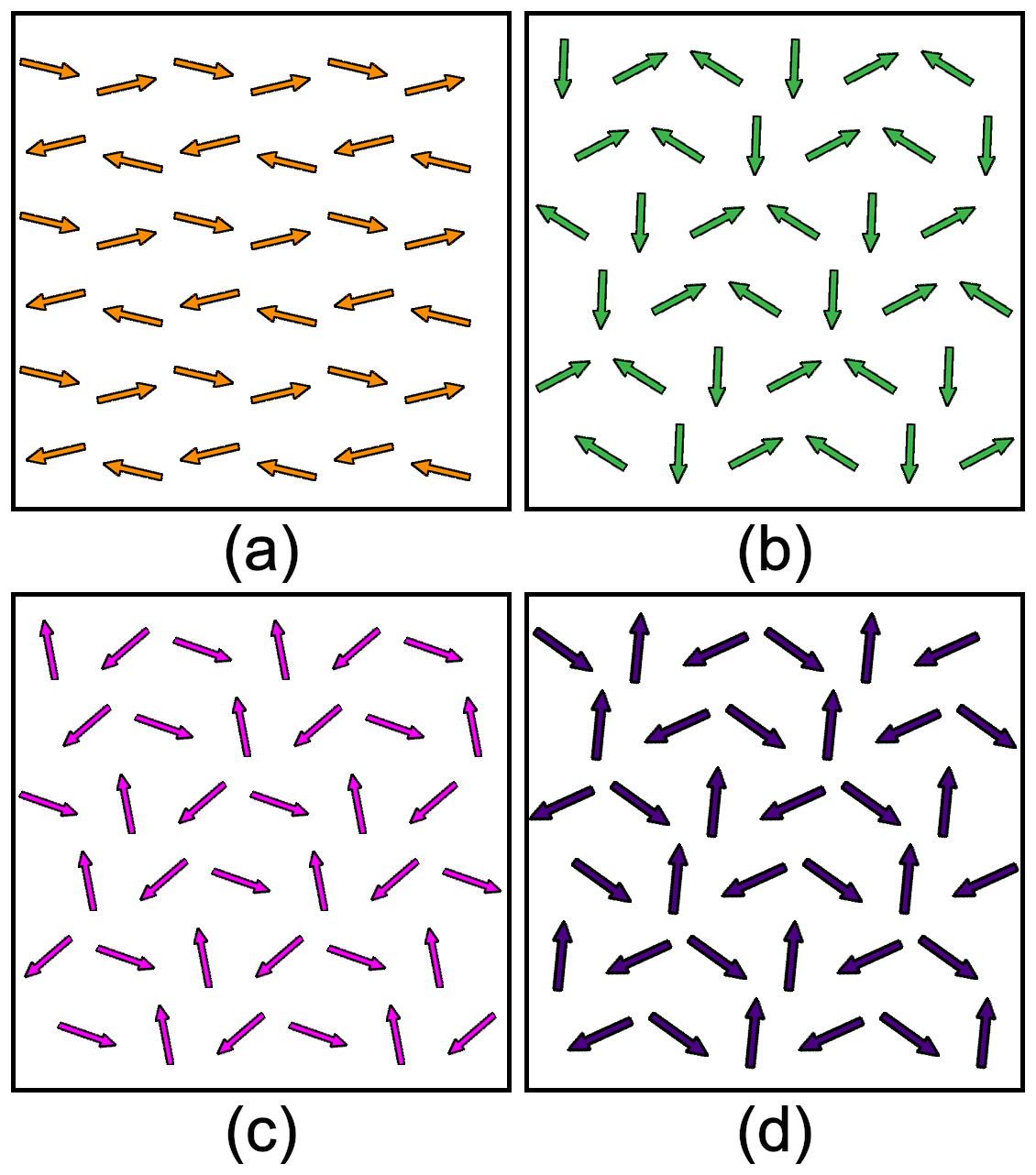}\par}

% \textbf{Question:} Which patch in the 2×2 grid (a–d) does not match the symmetry class of the other three?

\textbf{Answer:} (a)

\textbf{Response from GPT-5 (Incorrect):}\\
\boxed{c}

Reasoning:

\textcolor{red}{- In (a), (b), and (d) each arrow has a partner of opposite direction positioned so that 180° rotations (and corresponding glide reflections) map the pattern to itself.}

\textcolor{red}{- In (c) almost all arrows point in the same NE direction; after a reflection or a 180° rotation they would point SW, which is absent. Thus (c) lacks those symmetries and is the odd one out.}

\end{panelbox}
\end{subfigure}

\caption{Example responses from GPT-5 on \emph{Wallpaper Groups} task, with incorrect reasoning highlighted in red.}
\label{fig:gpt5_wallpaper}
\end{figure*}

\subsection{Wallpaper Groups}
Human accuracy is much higher on the Wallpaper Groups task compared to the Frieze Groups task, reaching 86.4\%, and GPT-5 also performs better at 48\%. Figure~\ref{fig:gpt5_wallpaper} shows four representative GPT-5 responses, with two correct examples in the top row and two incorrect ones in the bottom row. In this task, the two-dimensional layout makes the global symmetry more visually apparent, which likely explains the higher human accuracy. GPT-5 is able to identify the correct symmetry pattern in some cases, but incorrect examples illustrate a recurring limitation. Although humans can easily integrate the global structure of a repeating 2D pattern, GPT-5 often focuses on local neighborhoods and fails to reconcile those local transformations with the global symmetry of the wallpaper. This mismatch leads the model to select the wrong pattern even when the overall structure is clear in the image.

\section{GPT-5 vs GPT-5 Mini}
\label{app:gpt5vsmini}

\begin{figure*}[t]
\centering

%===================== TOP ROW =====================%
\begin{panelbox}[colback=black!4]
\textbf{(\romannumeral 1)}

  %---- Top-left (image) ----%
  {\centering
  \includegraphics[width=0.3\linewidth]{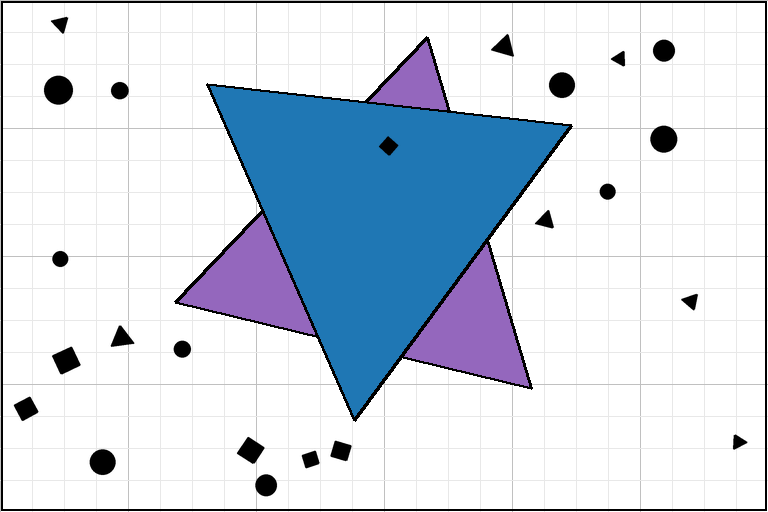}\par}

  %---- Question + Answer ----%
  \textbf{Question:} Equal-area colored sheets are stacked first, and the small black squares are added on top. How many small black squares are strictly inside the purple triangle?

  \textbf{Answer:} 1
  \medskip

  %---- Top row responses ----%
  \noindent
  \begin{minipage}[t]{0.498\linewidth}
    \begin{panelbox}[colback=black!2, equal height group=im11top]

      \textbf{Response from GPT-5 (Correct):}\\
- The black squares are on top of the stacked colored triangles.

- We count only those lying strictly inside the purple triangle’s interior (not on edges).

- Inspecting the figure, only the small black diamond near the center lies inside the purple triangle; all other black squares are outside it.

\boxed{1}

    \end{panelbox}
  \end{minipage}\hfill
  \begin{minipage}[t]{0.498\linewidth}
    \begin{panelbox}[colback=pink!8, equal height group=im11top]

      \textbf{Response from GPT-5 Mini (Incorrect):}\\
Count the little rotated black squares that lie strictly within the purple triangle (not on an edge). \textcolor{red}{Marking them one-by-one on the image gives four such squares.}

Final answer: \boxed{4}

    \end{panelbox}
  \end{minipage}

\end{panelbox}
\vspace{-1em}

%===================== BOTTOM ROW (REPEATED) =====================%
\begin{panelbox}[colback=black!4]
\textbf{(\romannumeral 2)}

  %---- Bottom-left (image repeated) ----%
  {\centering
  \includegraphics[width=0.3\linewidth]{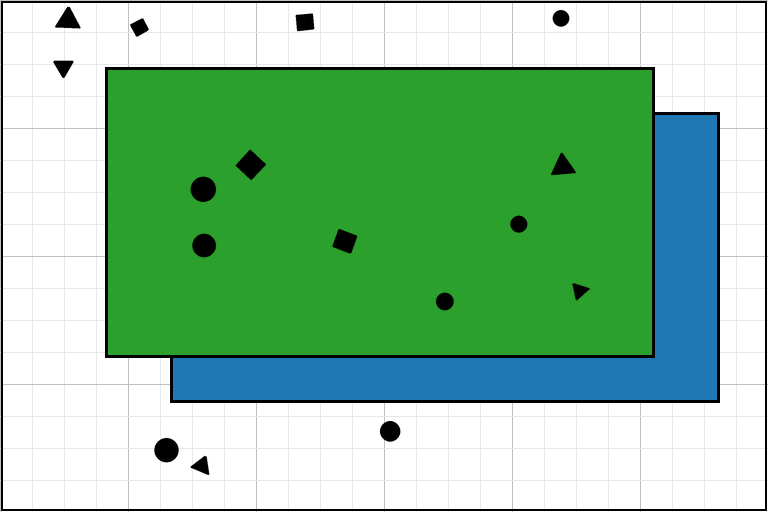}\par}

  %---- Question + Answer repeated ----%
  \textbf{Question:} The equal-area sheets are laid down first and may occlude each other; the black circles are added afterwards on top. How many small black circles are strictly inside the boundary of the blue rectangle?

  \textbf{Answer:} 4
\medskip

  %---- Bottom row responses (same as top) ----%
  \noindent
  \begin{minipage}[t]{0.498\linewidth}
    \begin{panelbox}[colback=black!2, equal height group=im11bot]

      \textbf{Response from GPT-5 (Correct):}\\
Step-by-step reasoning:

- The blue and green rectangles are equal in size; the blue is shifted down and to the right under the green.

- Therefore, the unseen left and top edges of the blue rectangle lie inside the green rectangle by the same amounts as the blue overhang to the right and to the bottom.

- A point (circle) is inside the blue rectangle iff it is to the right of that internal vertical line and below that internal horizontal line.

- Checking the circles, exactly four of the small black circles satisfy both conditions.

Final answer: \boxed{4}

    \end{panelbox}
  \end{minipage}\hfill
  \begin{minipage}[t]{0.498\linewidth}
    \begin{panelbox}[colback=pink!8, equal height group=im11bot]

      \textbf{Response from GPT-5 Mini (Incorrect):}\\
Step 1 — locate the true blue rectangle (including the parts hidden under the green sheet).

- The visible blue strip to the right and bottom shows the blue rectangle’s full horizontal and vertical extents; those edges continue straight under the green sheet.

Step 2 — find which black circle centers lie strictly inside those edges.

\textcolor{red}{- Using the blue rectangle edges inferred from the visible blue border, there are three small black circles whose centers lie strictly inside the blue rectangle (one on the exposed lower/right blue area and two lying on top of the portion of the blue sheet that is hidden under the green).}

Final answer: \boxed{3}

    \end{panelbox}
  \end{minipage}

\end{panelbox}

\caption{GPT-5 vs.\ GPT-5 Mini response on \emph{Stack Count} task, with incorrect reasoning highlighted in red.}
\label{fig:gpt_stack_count}
\end{figure*}

\paragraph{Stack Count.}
GPT-5 outperforms GPT-5 Mini by 36\% on the Stack Count task. In this task, larger shapes of equal area are stacked on top of each other and smaller black shapes are placed on top of the stack. Figure~\ref{fig:gpt_stack_count} shows representative examples of the two models for two questions where GPT-5 gives the correct answer and GPT-5 Mini does not. GPT-5 is able to reason about the equal areas of the stacked shapes and infer the area of the occluded shape by comparing it with the visible one on top. GPT-5 Mini, while able to grasp the need to extrapolate the area of hidden layers, struggles with accurately counting the smaller shapes and therefore arrives at incorrect answers.

\begin{figure*}[t]
\centering

%===================== TOP ROW =====================%
\begin{panelbox}[colback=black!4]
\textbf{(\romannumeral 1)}

  %---- Top-left (image) ----%
  {\centering
  \includegraphics[width=0.46\linewidth]{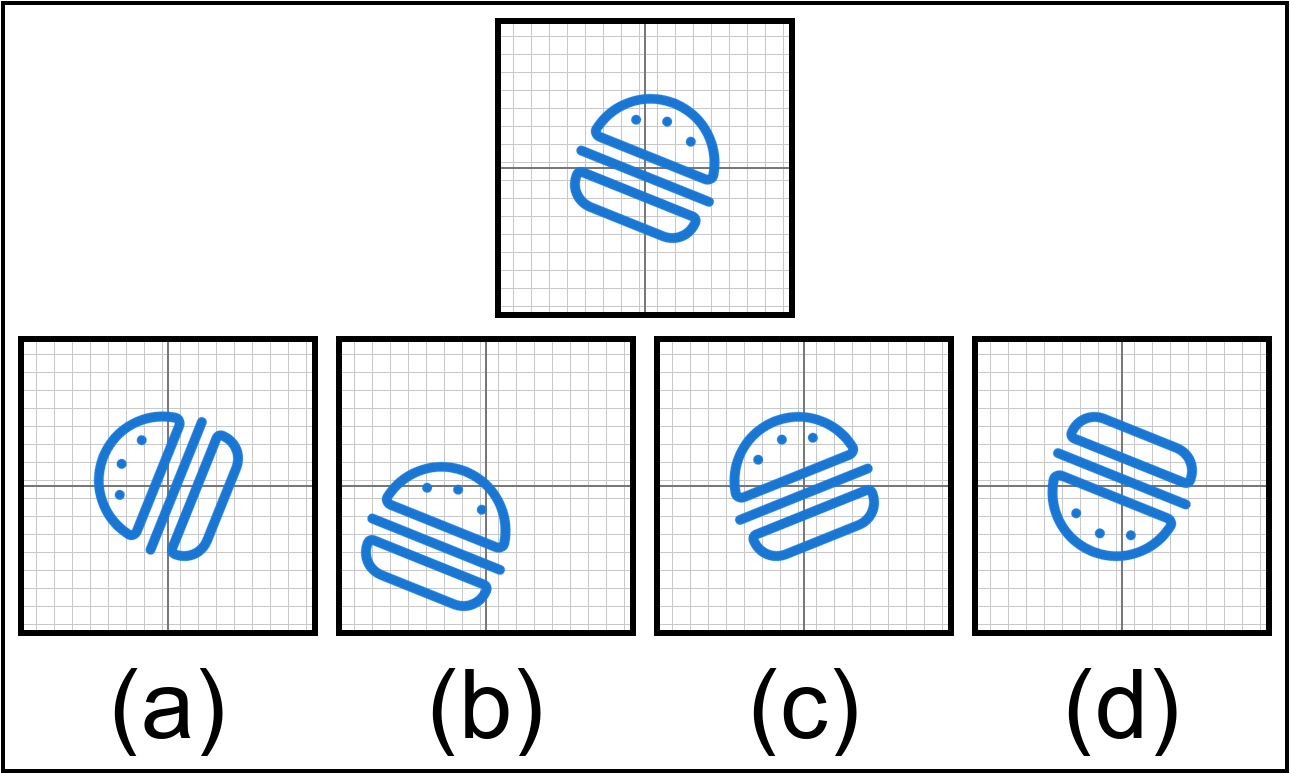}\par}

  %---- Question + Answer ----%
  \textbf{Question:}  After rotate 90° counterclockwise is applied to the top shape, which option (a)–(d) matches the outcome?

  \textbf{Answer:} (a)
  \medskip

  %---- Top row responses ----%
  \noindent
  \begin{minipage}[t]{0.498\linewidth}
    \begin{panelbox}[colback=black!2, equal height group=im12top]

      \textbf{Response from GPT-5 (Correct):}\\
Step-by-step:
- A 90° counterclockwise rotation moves the top semicircular bun to the left side of the figure.

- The two thin layers and the bottom bun, originally below the top bun, become vertical elements to the right of the semicircle.

Only option (a) shows the bun on the left with the three parallel layers to its right.
\boxed{a}

    \end{panelbox}
  \end{minipage}\hfill
  \begin{minipage}[t]{0.498\linewidth}
    \begin{panelbox}[colback=pink!8, equal height group=im12top]

      \textbf{Response from GPT-5 Mini (Incorrect):}\\
Rotate 90° CCW moves the features on the right of the original up and flips the stripes' slope. \textcolor{red}{Option (c) is the only choice where the three dots have moved to the top of the bun and the stripes have the correct rotated orientation.} 

Answer: \boxed{(c)}

    \end{panelbox}
  \end{minipage}

\end{panelbox}
\vspace{-1em}

%===================== BOTTOM ROW (REPEATED) =====================%
\begin{panelbox}[colback=black!4]
\textbf{(\romannumeral 2)}

  %---- Bottom-left (image repeated) ----%
  {\centering
  \includegraphics[width=0.46\linewidth]{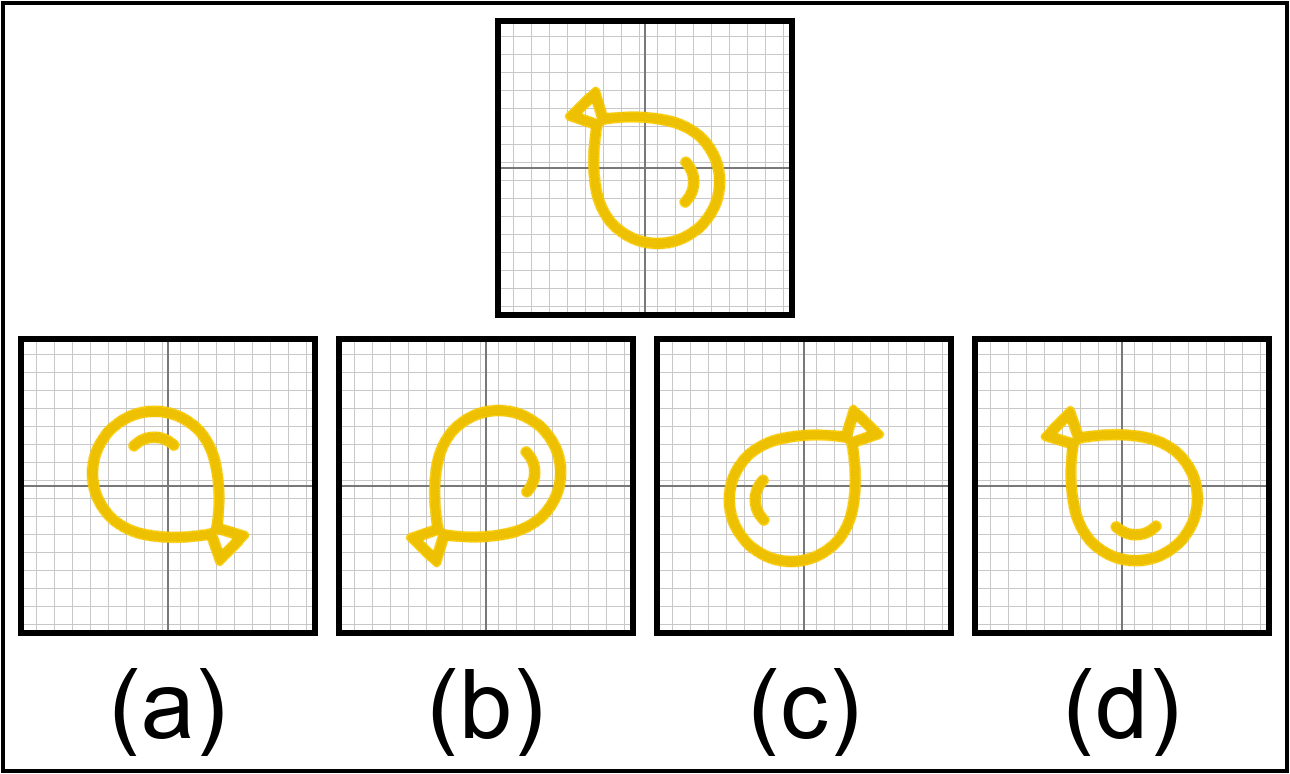}\par}

  %---- Question + Answer repeated ----%
  \textbf{Question:}  Look at the top figure. If you apply reflect across the anti‑diagonal ($↙$), which option (a)–(d) is the correct result?

  \textbf{Answer:} (a)
\medskip

  %---- Bottom row responses (same as top) ----%
  \noindent
  \begin{minipage}[t]{0.498\linewidth}
    \begin{panelbox}[colback=black!2, equal height group=im12bot]

      \textbf{Response from GPT-5 (Correct):}\\
Step-by-step reasoning:

- The anti-diagonal $↙$ is the line y = x (from top-right to bottom-left).

- Reflecting across this line swaps x and y: (x, y) → (y, x).

- So features move: the right-side mark goes to the top, and the top-left arrow moves to the bottom-right.

- Among the options, only (a) shows the mark on top and the arrow at bottom-right.

\boxed{a}

    \end{panelbox}
  \end{minipage}\hfill
  \begin{minipage}[t]{0.498\linewidth}
    \begin{panelbox}[colback=pink!8, equal height group=im12bot]

      \textbf{Response from GPT-5 Mini (Incorrect):}\\
\textcolor{red}{Reflecting across the anti‑diagonal (the line y = -x) sends each point (x,y) to (-y,-x). Applying that to the top figure produces the orientation and placement shown in option (d).}

Final answer: \boxed{(d)}

    \end{panelbox}
  \end{minipage}

\end{panelbox}

\caption{GPT-5 vs.\ GPT-5 Mini response on \emph{Transform Result Identify} task, with incorrect reasoning highlighted in red.}
\label{fig:gpt_transform_result_identify}
\end{figure*}

\paragraph{Transform Result Identify.}
In this task, the model must select the correct output image after a specified transformation is applied to the input motif. GPT-5 outperforms GPT-5 Mini by 17\% on this task. Figure~\ref{fig:gpt_transform_result_identify} shows two representative examples in which GPT-5 answers correctly and GPT-5 Mini answers incorrectly. GPT-5 appears to reason more reliably about the global shape and structure of the motif, allowing it to match the transformed result. In contrast, GPT-5 Mini struggles with these global cues and, in example (\romannumeral 2), also fails to recognize or describe the anti-diagonal symmetry present in the image, leading to an incorrect prediction.

\begin{figure*}[t]
\centering

%===================== TOP ROW =====================%
\begin{panelbox}[colback=black!4]
\textbf{(\romannumeral 1)}

  %---- Top-left (image) ----%
  {\centering
  \includegraphics[width=0.23\linewidth]{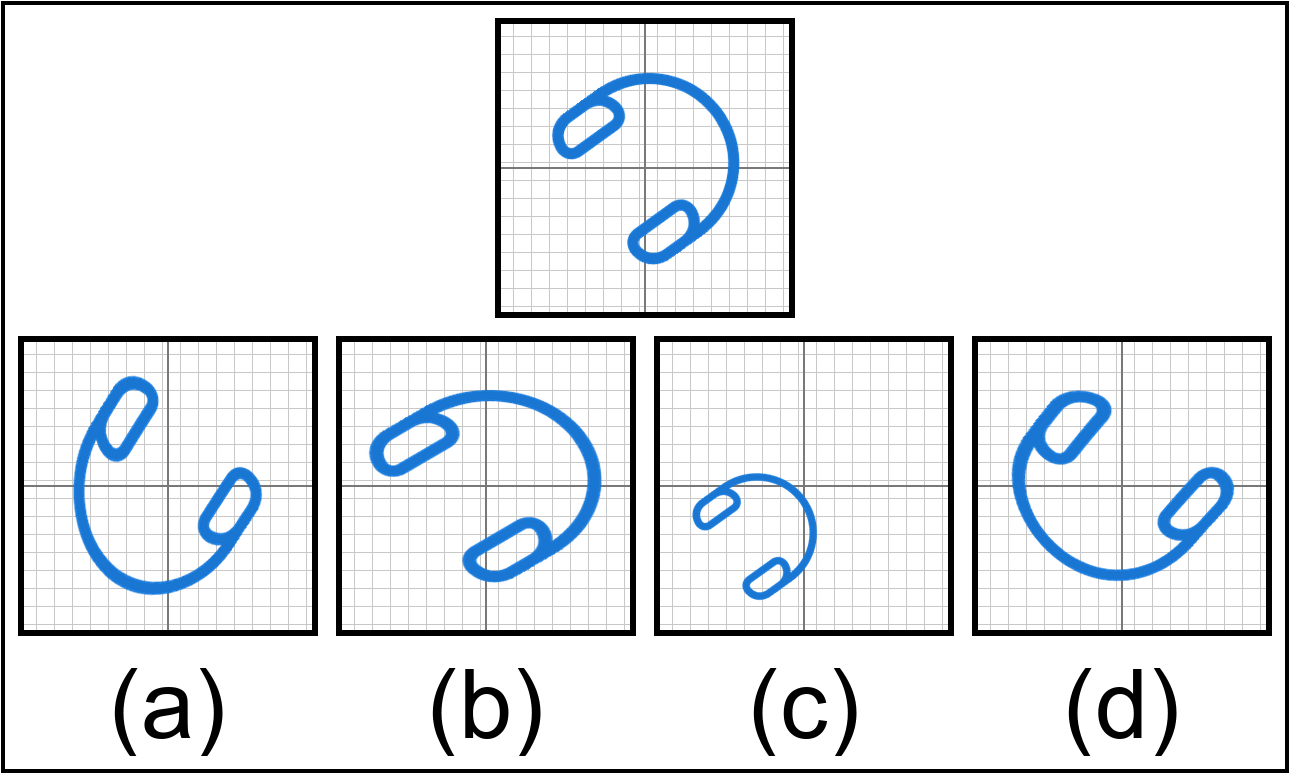}\par}

  %---- Question + Answer ----%
  \textbf{Question:}  Which option (a)–(d) is similar to the top figure under uniform scaling and rigid/mirror motions?

  \textbf{Answer:} (c)
  % \medskip

  %---- Top row responses ----%
  \noindent
  \begin{minipage}[t]{0.498\linewidth}
    \begin{panelbox}[colback=black!2, equal height group=im14top]

      \textbf{Response from GPT-5 (Correct):}\\
Step-by-step reasoning:
- The original figure is a C-shaped arc with two small pill-like loops on the inside, one near each end, with the same relative tilt.
- Similarity allows only uniform scaling, rotation, reflection, and translation; thus the curvature of the main arc and the relative positions/tilts of the two small loops must match.

Check the options:
\textcolor{red}{- (a) and (d): one of the small loops has the opposite tilt relative to the arc, so they cannot be obtained by rigid motion/scale.}
- (b): the spacing and curvature are different (the arc is “stretched” and the loops sit at different relative distances), so not a uniform scale of the original.
- (c): matches the curvature and the placement/tilts of both loops; it is simply a scaled copy.

Final answer: \boxed{(c)}

    \end{panelbox}
  \end{minipage}\hfill
  \begin{minipage}[t]{0.498\linewidth}
    \begin{panelbox}[colback=pink!8, equal height group=im14top]

      \textbf{Response from GPT-5 Mini (Incorrect):}\\
Look at three invariant relations that similarity (with rotation/translation/reflection allowed) must preserve: (1) the two small rounded rectangles lie on the same circular arc, (2) their relative angular positions around the arc (one near the upper-left, the other near the lower-right) and (3) the direction each small rectangle is tilted (both are tilted the same way along the arc). \textcolor{red}{Only option (b) preserves all three at once after a rotation/scale — the other choices place one or both small rounded rectangles in the wrong location or with the wrong tilt relative to the arc.}

Therefore the similar figure is (b). \boxed{(b)}

    \end{panelbox}
  \end{minipage}

\end{panelbox}
\vspace{-1em}

%===================== BOTTOM ROW (REPEATED) =====================%
\begin{panelbox}[colback=black!4]
\textbf{(\romannumeral 2)}

  %---- Bottom-left (image repeated) ----%
  {\centering
  \includegraphics[width=0.23\linewidth]{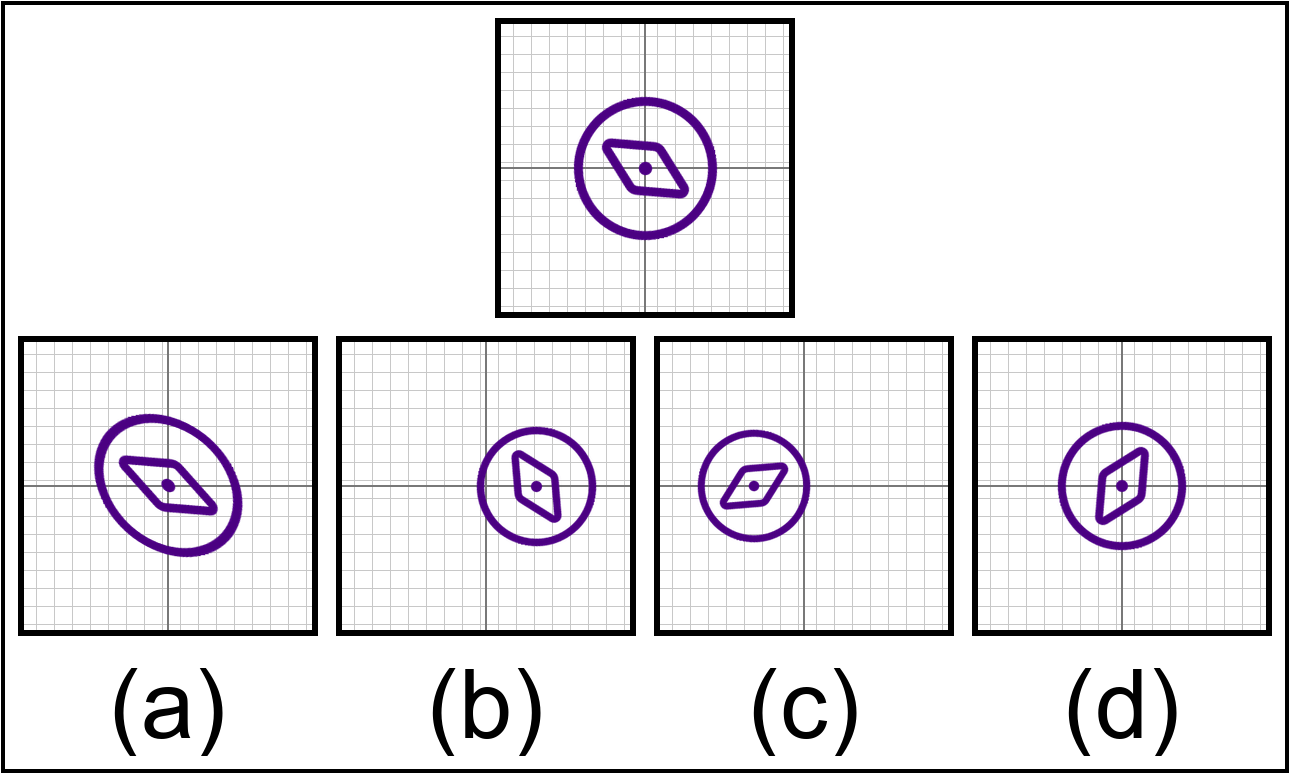}\par}

  %---- Question + Answer repeated ----%
  \textbf{Question:}  Choose the option (a)–(d) that is not obtainable by uniform scaling of the top plus rotation/reflection/translation.

  \textbf{Answer:} (a)
% \medskip

  %---- Bottom row responses (same as top) ----%
  \noindent
  \begin{minipage}[t]{0.498\linewidth}
    \begin{panelbox}[colback=black!2, equal height group=im14bot]

      \textbf{Response from GPT-5 (Correct):}\\
Step-by-step reasoning:

- Allowed operations: uniform scaling + rotation/reflection/translation. These are similarity transformations, which preserve shapes and map circles to circles (no squashing).

- Inspect options: (b), (c), and (d) all keep the outer boundary circular and preserve proportions, so they can come from the top via scaling/rotation/translation.

- Option (a) shows the outer “circle” squashed into an ellipse, with the inner parts distorted similarly. This would require non-uniform scaling/shear, which is not allowed.

Final answer: \boxed{(a)}

    \end{panelbox}
  \end{minipage}\hfill
  \begin{minipage}[t]{0.498\linewidth}
    \begin{panelbox}[colback=pink!8, equal height group=im14bot]

      \textbf{Response from GPT-5 Mini (Incorrect):}\\
Key invariant: uniform scaling + rotation/reflection + translation preserves the relative position of features inside the figure. In particular, the vector from the circle center to the little dot (the diamond's center) scales by the same factor as the circle radius, so the offset as a fraction of the circle radius cannot change.

\textcolor{red}{In the top picture the diamond's center (the small dot) is clearly offset from the circle center. Any allowed transformation must keep a nonzero fractional offset. Option (b) shows the diamond exactly concentric with the circle (offset = 0), which cannot result from the top figure by the allowed operations.}

Therefore the impossible option is \boxed{(b)}.
      
    \end{panelbox}
  \end{minipage}

\end{panelbox}

\caption{GPT-5 vs.\ GPT-5 Mini response on \emph{Transform Similarity Identify} task, with incorrect reasoning highlighted in red.}
\label{fig:gpt_transform_sim_identify}
\end{figure*}

\paragraph{Transform Similarity Identify.}
In this task, the model must identify which image is similar or dissimilar to the reference motif under Euclidean similarity, which includes uniform scaling together with the D4 group of rotations and mirror reflections. Each question contains one correct similar option or one correct dissimilar option. GPT-5 outperforms GPT-5 Mini in this task by 12\%. Figure~\ref{fig:gpt_transform_sim_identify} shows example responses for two questions in which GPT-5 selects the correct answer and GPT-5 Mini does not. As in the Transform Result Identify task, GPT-5 demonstrates a stronger ability to reason about the global structure of the motif and how the transformation affects its overall form. Although GPT-5 occasionally reaches the correct answer with partially flawed reasoning, this is often sufficient in the multiple-choice format. GPT-5 Mini, on the other hand, tends to focus on local features and does not integrate them into a coherent global comparison, which leads to incorrect conclusions in these examples.

\begin{figure*}[t]
\centering

%===================== TOP ROW =====================%
\begin{panelbox}[colback=black!4]
\textbf{(\romannumeral 1)}

  %---- Top-left (image) ----%
  {\centering
  \includegraphics[width=0.49\linewidth]{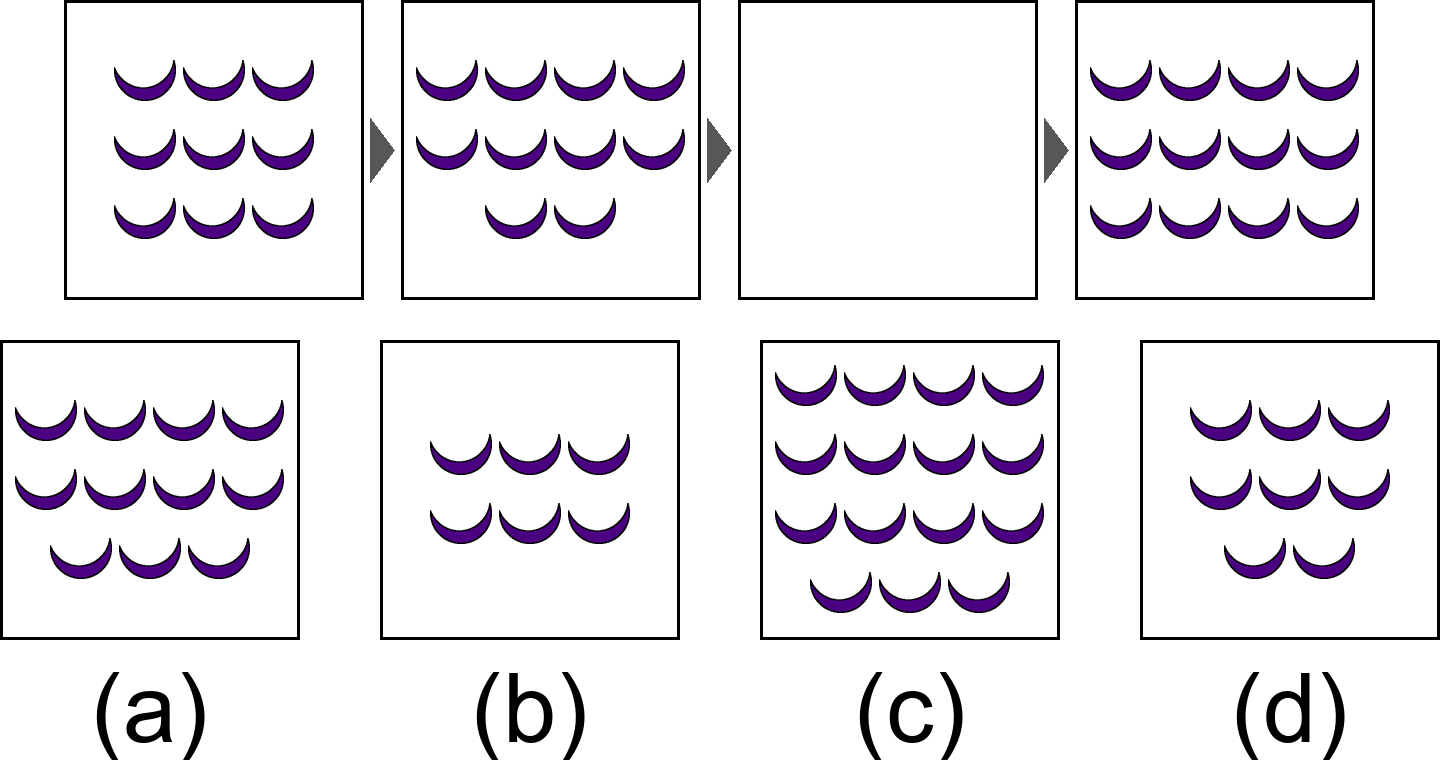}\par}

  %---- Question + Answer ----%
  \textbf{Question:}  Which option (a)–(d) below should replace the blank to complete the pattern shown in the top row?

  \textbf{Answer:} (a)
  \medskip

  %---- Top row responses ----%
  \noindent
  \begin{minipage}[t]{0.498\linewidth}
    \begin{panelbox}[colback=black!2, equal height group=im17top]

      \textbf{Response from GPT-5 (Incorrect):}\\
\textcolor{red}{The top two rows remain full in every panel; only the bottom row changes.}

\textcolor{red}{From panel 1 to 2, two crescents disappear from the bottom-right.  
To return to a full grid in panel 4, one crescent is added back in each step.}

\textcolor{red}{So the missing third panel should have three crescents in the bottom row (still missing the last one).}

\textcolor{red}{Therefore the correct option is \boxed{c}.}

    \end{panelbox}
  \end{minipage}\hfill
  \begin{minipage}[t]{0.498\linewidth}
    \begin{panelbox}[colback=pink!8, equal height group=im17top]

      \textbf{Response from GPT-5 Mini (Correct):}\\
Count the crescents in the top row: left = 9, next = 10, final = 12. The sequence must increase by one each step, so the blank should contain 11 crescents. Option (a) shows 11 crescents.

Answer: \boxed{(a)}

    \end{panelbox}
  \end{minipage}

\end{panelbox}
\vspace{-1em}

%===================== BOTTOM ROW (REPEATED) =====================%
\begin{panelbox}[colback=black!4]
\textbf{(\romannumeral 2)}

  %---- Bottom-left (image repeated) ----%
  {\centering
  \includegraphics[width=0.49\linewidth]{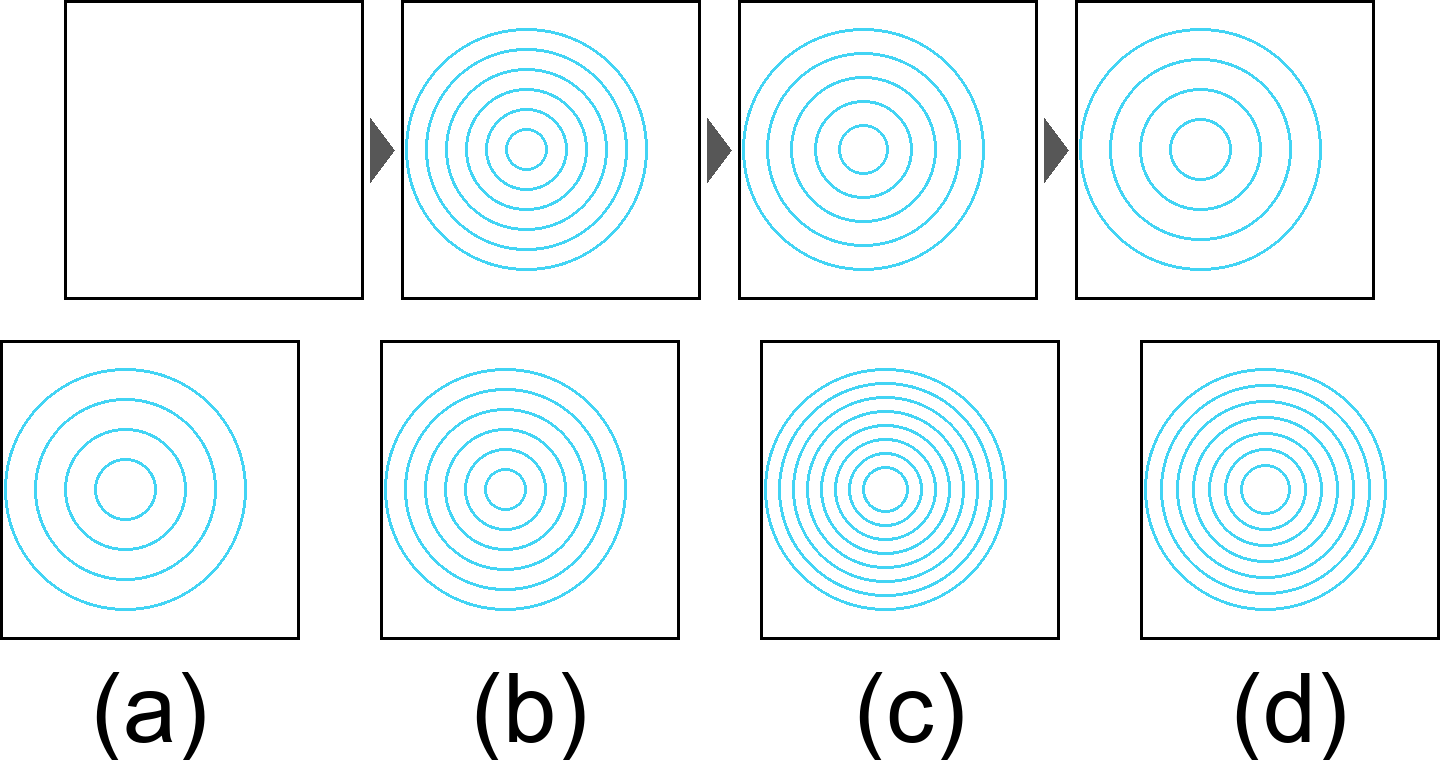}\par}

  %---- Question + Answer repeated ----%
  \textbf{Question:}  Which option (a)–(d) below should replace the blank to complete the pattern shown in the top row?

  \textbf{Answer:} (d)
\medskip

  %---- Bottom row responses (same as top) ----%
  \noindent
  \begin{minipage}[t]{0.498\linewidth}
    \begin{panelbox}[colback=black!2, equal height group=im17bot]

      \textbf{Response from GPT-5 (Incorrect):}\\
Reasoning:

\textcolor{red}{- From the second to the fourth square, the concentric circles expand outward by a fixed amount each step.}

\textcolor{red}{- As they grow, the outermost circle moves out of the frame, so the number of visible circles decreases by one: 5 → 4 → 3.}

\textcolor{red}{- Therefore, the square before the second one should show one more circle than the second: 6 circles, centered and evenly spaced.}

The only option that matches this is option (b).

Final answer: \boxed{b}

    \end{panelbox}
  \end{minipage}\hfill
  \begin{minipage}[t]{0.498\linewidth}
    \begin{panelbox}[colback=pink!8, equal height group=im17bot]

      \textbf{Response from GPT-5 Mini (Correct):}\\
Each step from left to right removes one concentric ring. The first shown image (top row, second position) has 7 circles (center + 6 rings), then 6, then 5. So the missing leftmost image must have one more circle (center + 7 rings = 8 circles). Choice (d) is the only option with that many rings.

Final answer: \boxed{d}

    \end{panelbox}
  \end{minipage}

\end{panelbox}

\caption{GPT-5 vs.\ GPT-5 Mini response on \emph{Sequence Arithmetic} task, with incorrect reasoning highlighted in red.}
\label{fig:gpt_sequence_arithmetic}
\end{figure*}

\paragraph{Sequence Arithmetic.}
GPT-5 Mini outperforms GPT-5 by 18\% on the Sequence Arithmetic task, which requires identifying the missing entry in an arithmetic sequence of motifs. Figure~\ref{fig:gpt_sequence_arithmetic} shows two examples where GPT-5 Mini answers correctly while GPT-5 does not. In both cases, GPT-5 makes multiple mistakes when counting the motifs in the grids and analyzing how their quantities change throughout the sequence. In example (\romannumeral 1), GPT-5 focuses only on the bottom row of each grid and incorrectly counts the number of crescents in the first grid as 1 instead of 3, which leads to the wrong answer. In the second example, the model again miscounts, this time failing to identify the correct number of rings in each grid, and therefore selecting the incorrect option. GPT-5 Mini, on the contrary, appears to perform a more reliable count for this task.

\begin{figure*}[t]
\centering

%===================== TOP ROW =====================%
\begin{panelbox}[colback=black!4]
\textbf{(\romannumeral 1)}

  %---- Top-left (image) ----%
  {\centering
  \includegraphics[width=0.38\linewidth]{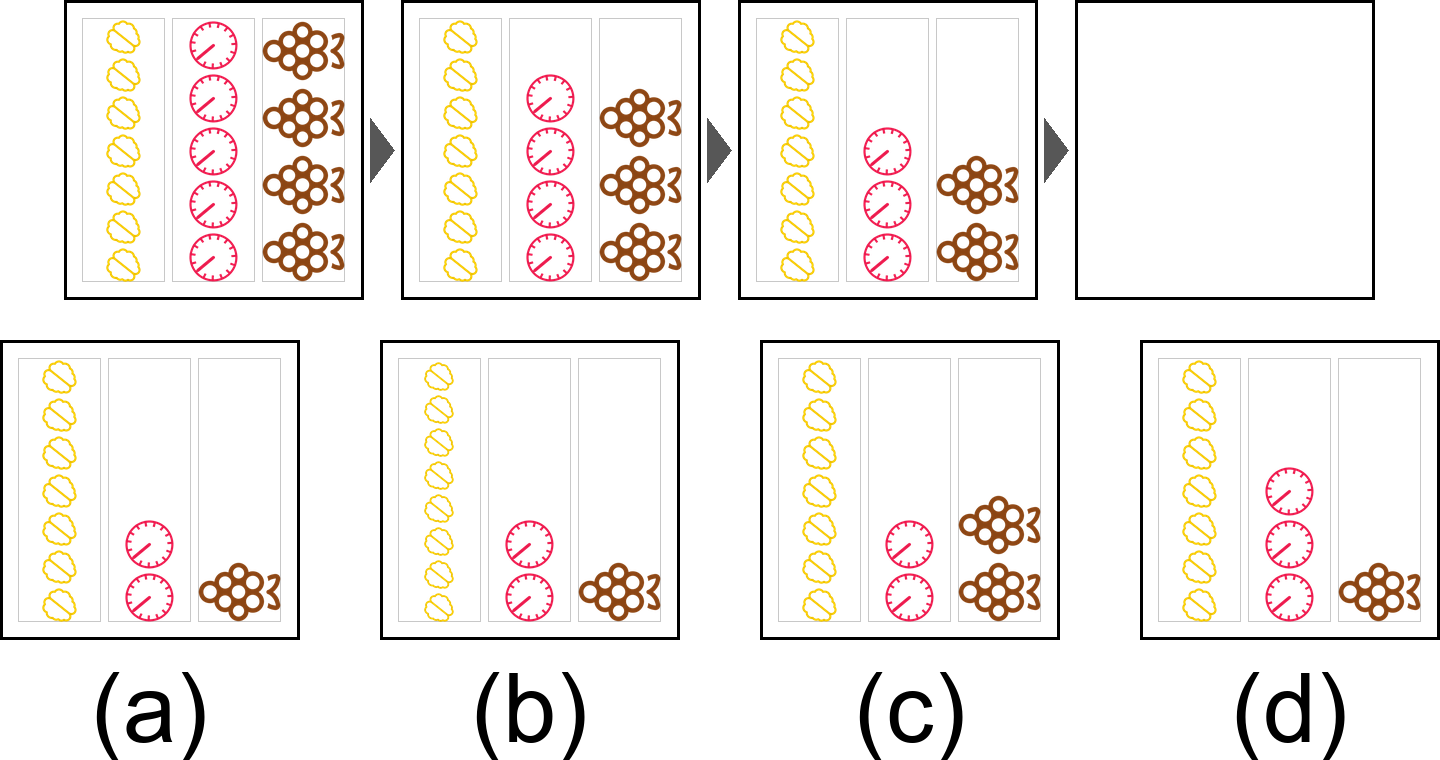}\par}

  %---- Question + Answer ----%
  \textbf{Question:}  Which panel (a)–(d) completes the sequence shown in the top row?

  \textbf{Answer:} (a)

  %---- Top row responses ----%
  \noindent
  \begin{minipage}[t]{0.498\linewidth}
    \begin{panelbox}[colback=black!2, equal height group=im18top]

      \textbf{Response from GPT-5 (Correct):}\\
Step-by-step reasoning:

- In each frame the whole pattern shifts to the right.

- As a column reaches the rightmost narrow area, only the bottom part remains visible: the number of visible objects in that column decreases by one each step.

- From the first to the third frame the pink column goes from \textcolor{red}{4} to 3 items visible, and the brown column from 4 to 2 visible.

- The next step should therefore show 2 pink dials and only 1 brown cluster, both still aligned at the bottom-right.

Only panel (a) matches this continuation.

Final answer: \boxed{a}

    \end{panelbox}
  \end{minipage}\hfill
  \begin{minipage}[t]{0.498\linewidth}
    \begin{panelbox}[colback=pink!8, equal height group=im18top]

      \textbf{Response from GPT-5 Mini (Incorrect):}\\
Answer: \boxed{(b)}

Reason: The yellow column stays the same. Each step the middle (pink clocks) and right (brown clusters) lose one item and shift upward. \textcolor{red}{After three steps the middle has 2 clocks and the right has 1 cluster in the top-shifted positions — matching panel (b).}

    \end{panelbox}
  \end{minipage}

\end{panelbox}
\vspace{-1em}

%===================== BOTTOM ROW (REPEATED) =====================%
\begin{panelbox}[colback=black!4]
\textbf{(\romannumeral 2)}

  %---- Bottom-left (image repeated) ----%
  {\centering
  \includegraphics[width=0.38\linewidth]{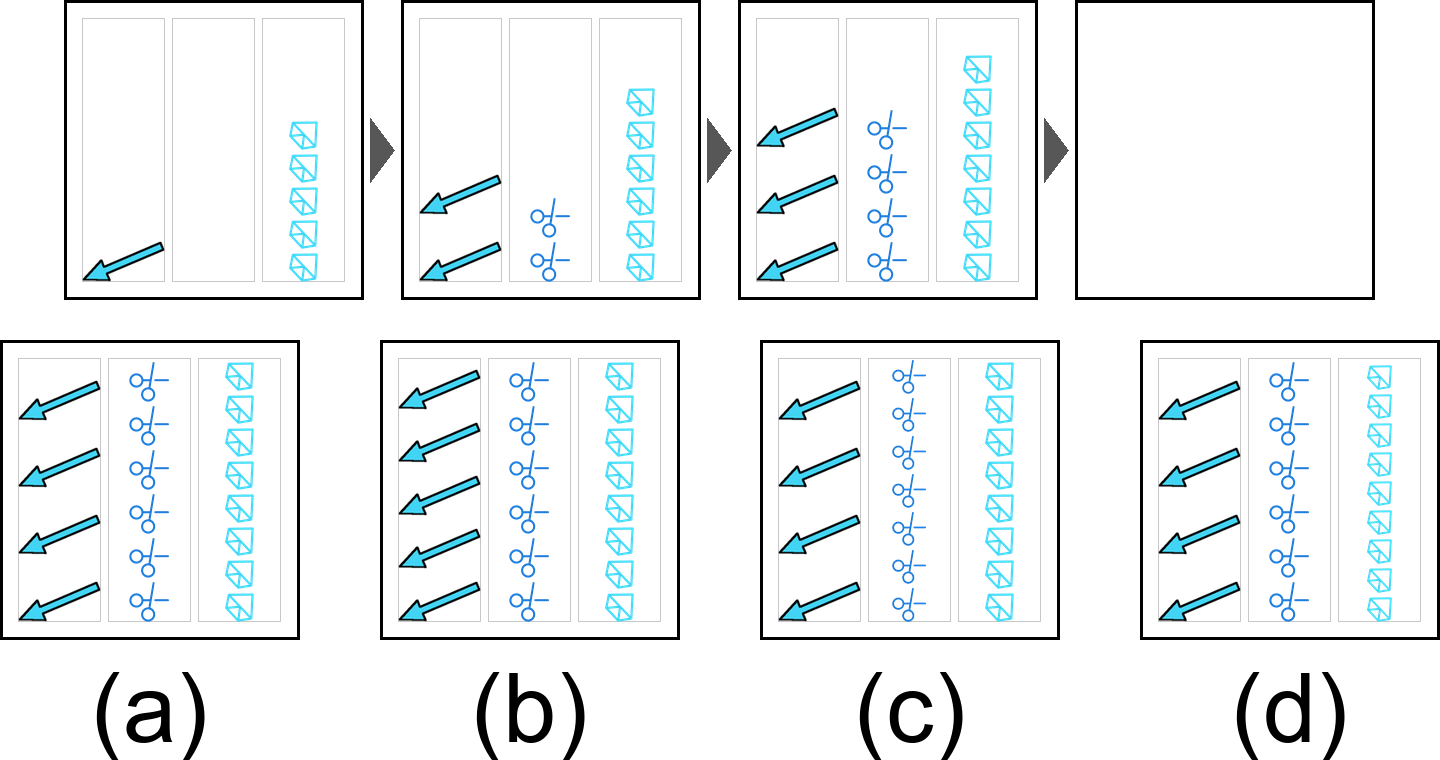}\par}

  %---- Question + Answer repeated ----%
  \textbf{Question:}  Which panel (a)–(d) completes the sequence shown in the top row?

  \textbf{Answer:} (a)

  %---- Bottom row responses (same as top) ----%
  \noindent
  \begin{minipage}[t]{0.498\linewidth}
    \begin{panelbox}[colback=black!2, equal height group=im18bot]

      \textbf{Response from GPT-5 (Incorrect):}\\
Step-by-step:

- In each panel, the three columns contain arrows, circle-stick figures, and cubes respectively.

- The counts increase by one from panel to panel and also from left to right within a panel:

  - Panel 1: 1 arrow, \textcolor{red}{2 circles, 3 cubes.}

  - Panel 2: \textcolor{red}{2 arrows, 3 circles, 4 cubes.}

  - Panel 3: 3 arrows, 4 circles, \textcolor{red}{5 cubes.}

\textcolor{red}{- Therefore, the fourth should have 4 arrows, 5 circles, and 6 cubes.}

\textcolor{red}{Only option (c) matches this.}

\boxed{c}

    \end{panelbox}
  \end{minipage}\hfill
  \begin{minipage}[t]{0.498\linewidth}
    \begin{panelbox}[colback=pink!8, equal height group=im18bot]

      \textbf{Response from GPT-5 Mini (Correct):}\\
Answer: (a)

Reason: Each step adds one more arrow in the left column and \textcolor{red}{one more scissors} in the middle column (scissors = arrows - 1), while \textcolor{red}{the right column of five envelopes stays constant.} After 1, 2, 3 arrows in the first three panels the fourth must show 4 arrows and \textcolor{red}{3 scissors — option (a).}

    \end{panelbox}
  \end{minipage}

\end{panelbox}

\caption{GPT-5 vs.\ GPT-5 Mini response on \emph{Sequence Multi-Column Arithmetic} task, with incorrect reasoning highlighted in red.}
\label{fig:gpt_sequence_mult-arith}
\end{figure*}

\paragraph{Sequence Multi-Column Arithmetic.}
In the Sequence Multi-Column Arithmetic task, each grid contains several columns, and each column follows its own arithmetic progression across the sequence. In contrast to the Sequence Arithmetic task, GPT-5 outperforms GPT-5 Mini by 12\%. Figure~\ref{fig:gpt_sequence_mult-arith} shows two representative cases: in example (\romannumeral 1), GPT-5 is correct while GPT-5 Mini is incorrect, and in example (\romannumeral 2), GPT-5 Mini answers correctly while GPT-5 does not. Interestingly, in both cases, the models display subtle reasoning errors even when selecting the correct answer. Both struggle to track the progression of individual columns and often miscount the motifs within each column. Their explanations reveal that they rely on shallow heuristics or partial observations rather than a full analysis of the multi-column structure.

\paragraph{Summary}
These results, together with the earlier discussion in Section~\ref{app:gpt5-analysis}, reveal several key insights about the GPT-5 model family on the tasks in \textsc{Sphinx}. Both models struggle with counting over abstract symbols, which is evident in motif-based tasks such as \emph{Sequence Arithmetic} and \emph{Sequence Multi-Column Arithmetic}, as well as tile-based tasks such as \emph{Tiles Line Length} and \emph{Tiles Recoloring}. This difficulty becomes even more pronounced when counting interacts with additional structure, as in the \emph{Sequence Multi-Column Arithmetic} task. Although models often produce coherent textual reasoning, errors introduced during the visual counting stage frequently lead to incorrect answers.

For tasks involving symmetry and transformation, GPT-5 is generally better equipped than GPT-5 Mini to reason about the global structure of the image, although both lag behind human performance, as seen in the \emph{Wallpaper Groups} task. The strong performance of GPT-5 on \emph{Shape Counting} also suggests that the model may surpass humans in settings that demand substantial mental arithmetic and relatively less visual extraction.

Overall, these findings show that the success of GPT-5 strongly depends on whether the task requires accurate visual extraction or primarily symbolic reasoning, with its largest failures driven by perceptual errors rather than conceptual ones.

\section{Discussion of RLVR trained models}
\label{app:rlvranalysis}

\begin{figure}[t]
  \centering
  \includegraphics[width=\linewidth]{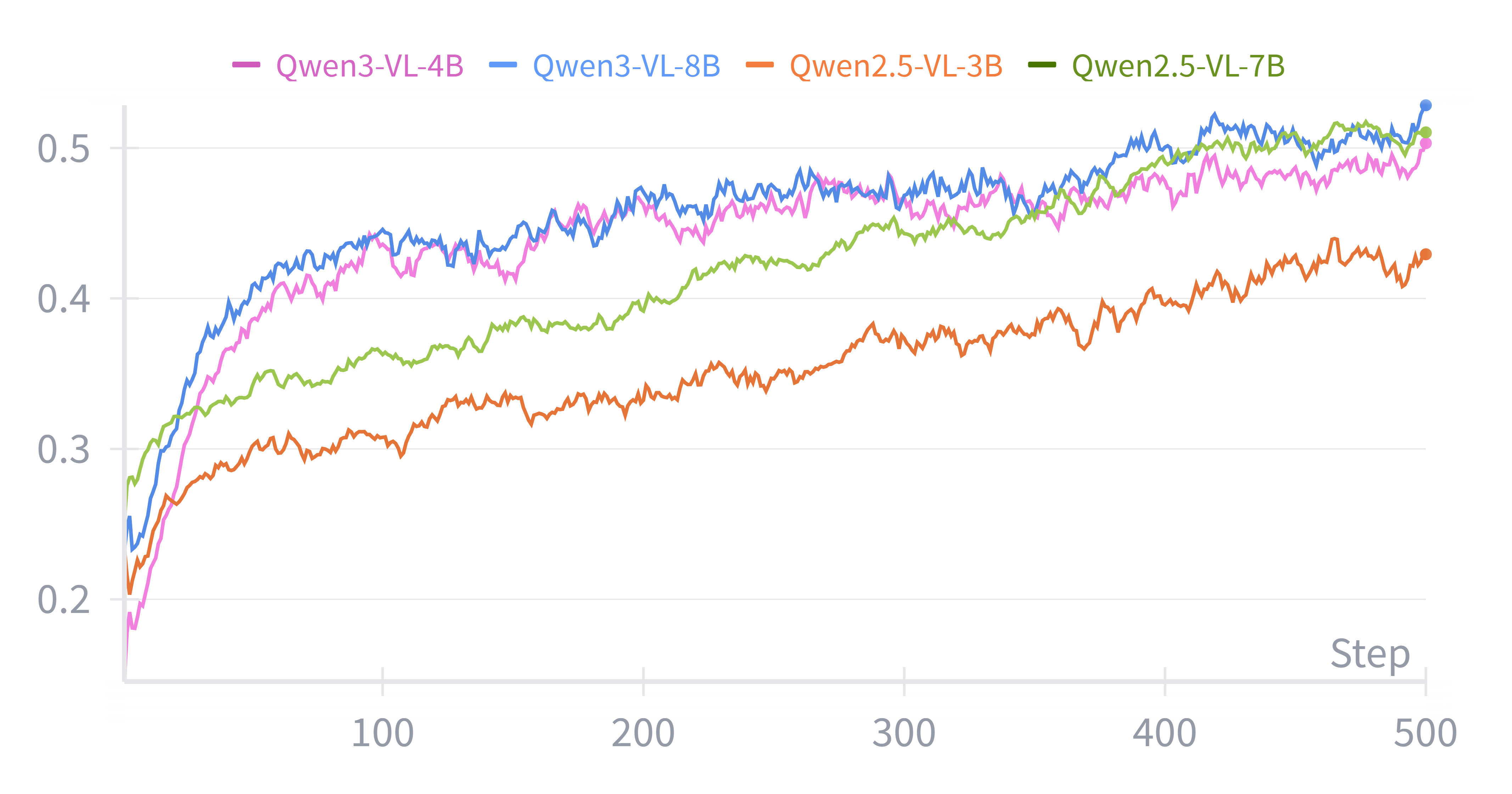}
  \caption{Training reward curves for the four models during RLVR training.}
  \label{fig:train-reward}
\end{figure}

\begin{figure}[t]
  \centering
  \includegraphics[width=\linewidth]{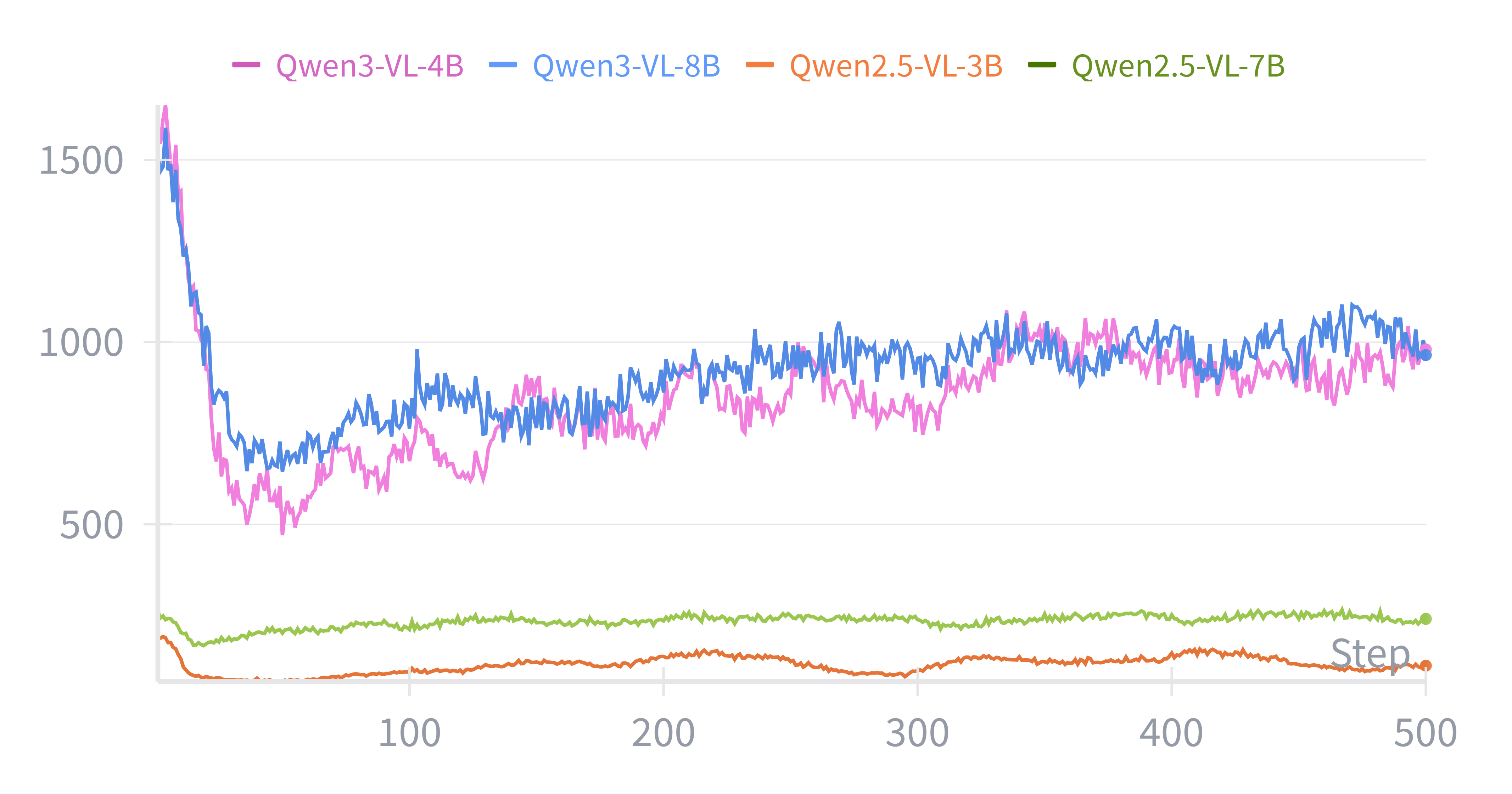}
  \caption{Average response length on the training set during RLVR training.}
  \label{fig:train_response_length}
\end{figure}

\paragraph{Training Dynamics.}
Figure~\ref{fig:train-reward} shows the evolution of the training reward over the course of the RL training for all four models, and Figure~\ref{fig:train_response_length} shows the corresponding average response length measured on the training set. For the Qwen3-VL models, the average response length initially decreases and then gradually increases as training progresses, while for the Qwen2.5-VL models, it remains much closer to that of the base model throughout training. The reward curves also indicate that the models are not fully saturated and could benefit from additional training, which remains a promising direction for future work.

\begin{figure}[t]
  \centering
  
  \begin{subfigure}[t]{0.45\linewidth}
    \centering
    \includegraphics[width=\linewidth]{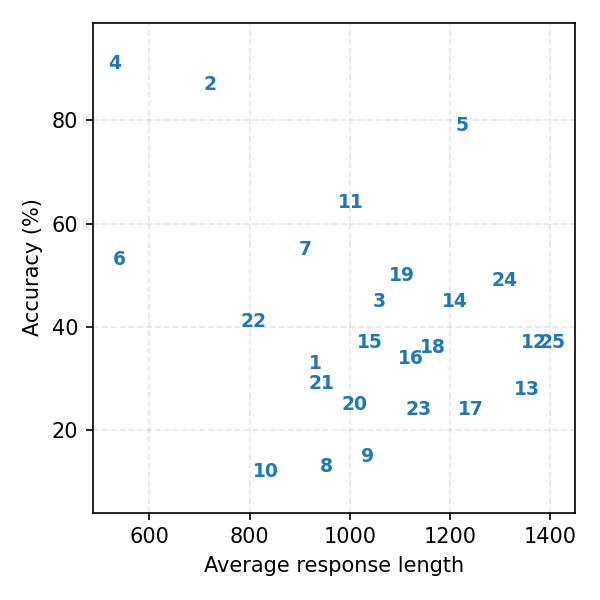}
    \subcaption{Qwen3-VL-8B}
    \label{fig:train_response_length_a}
  \end{subfigure}
  \hfill
  \begin{subfigure}[t]{0.45\linewidth}
    \centering
    \includegraphics[width=\linewidth]{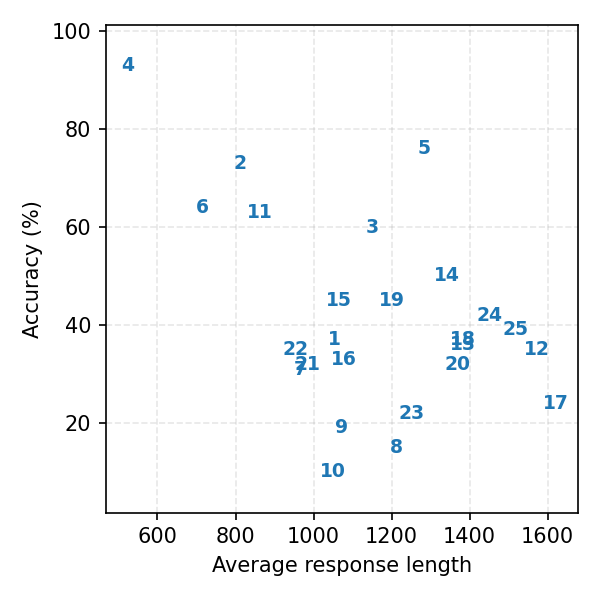}
    \subcaption{Qwen3-VL-4B}
    \label{fig:train_response_length_b}
  \end{subfigure}

  \vspace{0.5em}

  \begin{subfigure}[t]{0.45\linewidth}
    \centering
    \includegraphics[width=\linewidth]{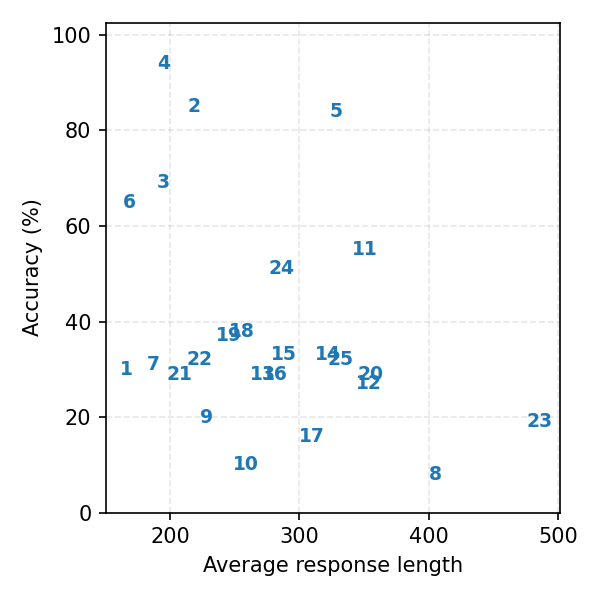}
    \subcaption{Qwen2.5-VL-7B}
    \label{fig:train_response_length_c}
  \end{subfigure}
  \hfill
  \begin{subfigure}[t]{0.45\linewidth}
    \centering
    \includegraphics[width=\linewidth]{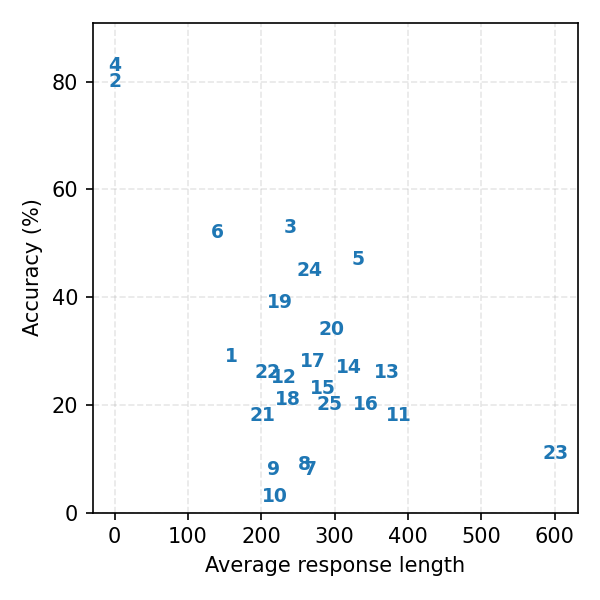}
    \subcaption{Qwen2.5-VL-3B}
    \label{fig:train_response_length_d}
  \end{subfigure}

  \caption{Average response length and accuracy for each task on the \textsc{Sphinx} benchmark for the four RLVR-trained models. Each point corresponds to a task ID and shows the model’s average response length and accuracy for that task.}

   \label{fig:train_response_length_2x2}
\end{figure}

\paragraph{Response Length and Accuracy.}
Figure~\ref{fig:train_response_length_2x2} shows the average response length and accuracy across the \textsc{Sphinx} benchmark tasks for the four RLVR-trained models. The Qwen3-VL models, both base and RLVR-trained, consistently produce longer responses than the Qwen2.5-VL models across all tasks.

Tasks 4 (Pie Chart), 2 (Shape Sorting), and 6 (Venn Diagram) are among the shortest tasks that still achieve high accuracy. Although the Pie Chart task already showed strong performance before training, RLVR training led to clear improvements on the other two tasks. One possible explanation is that these tasks do not rely on Motifs or Tiles, which limits the visual variety of the problem instances and results in test data that are not substantially different from the training data.

In contrast, Task 23 (Tiles Shortest Path) has the longest average description for both Qwen2.5-VL models and yields below 20\% accuracy, with only slightly better performance from the Qwen3-VL models. Tasks 25 (Tiles Composition), 12 (Symmetry Fill), 17 (Transform Similarity Identify), and 13 (Frieze Groups) also show long response lengths for the Qwen3-VL models while remaining below 40\% accuracy. Tiles Composition is not included in the training set, but the other three tasks are, suggesting that these tasks are inherently difficult for the models to learn even with RLVR training.

\begin{figure*}[t]
\centering

%===================== TOP ROW =====================%
\begin{panelbox}[colback=black!4]

  %---- Top-left (image) ----%
  {\centering
  \includegraphics[width=0.48\linewidth]{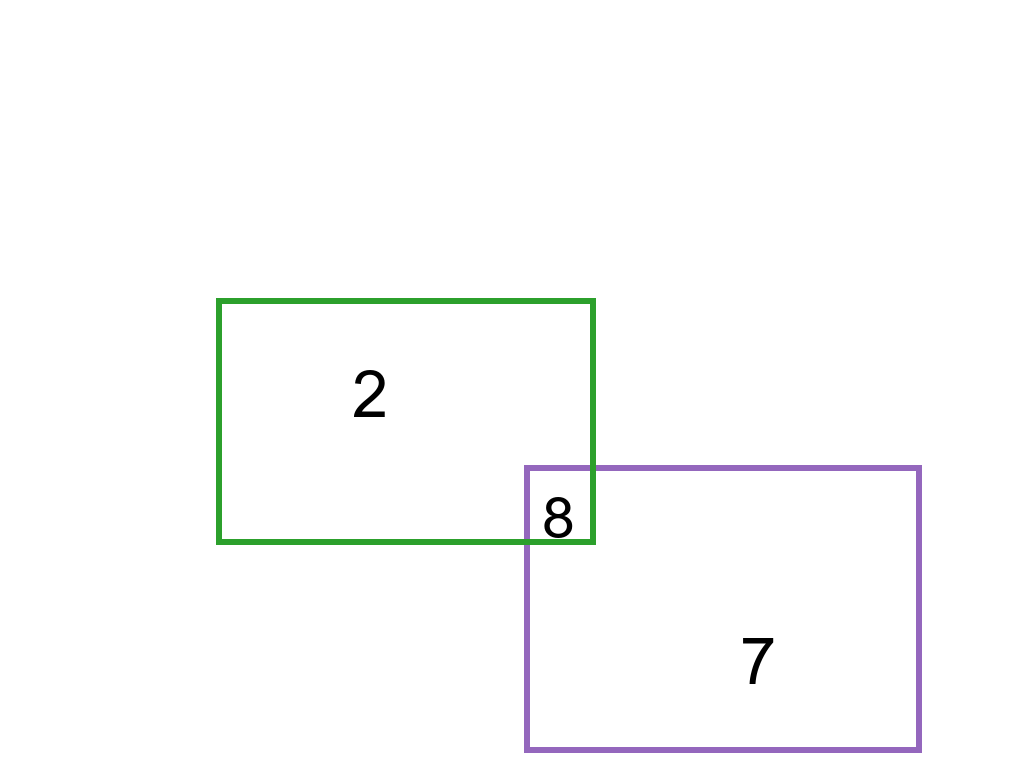}\par}

  %---- Question + Answer ----%
  \textbf{Question:}  What is the total of the numbers present simultaneously in the purple and green rectangle(s)?

  \textbf{Answer:} 8

\medskip

  %---- Top row responses ----%
  \noindent
  \begin{minipage}[t]{0.498\linewidth}
    \begin{panelbox}[colback=black!2, equal height group=im31]

      \textbf{Response from Qwen3-VL-4B (Incorrect):}\\
We are given two rectangles:

- A **green rectangle** containing the number **2**.
- A **purple rectangle** containing the number **8** and **7**.

The question asks: **What is the total of the numbers present simultaneously in the purple and green rectangle(s)?**

\#\#\# Step 1: Understand "simultaneously in the purple and green rectangle(s)"

This means we are looking for numbers that appear **in both** rectangles.

\textcolor{red}{But note: the green rectangle has only the number **2**.  
The purple rectangle has the numbers **8** and **7**.}

\textcolor{red}{There is **no overlap** between the numbers in the green rectangle and the numbers in the purple rectangle.}

\textcolor{red}{So, the set of numbers common to both rectangles is **empty**.}

\#\#\# Step 2: Compute the total

The total of an empty set is **0**.

---

\#\#\# Final Answer:
$\boxed{0}$

    \end{panelbox}
  \end{minipage}\hfill
  \begin{minipage}[t]{0.498\linewidth}
    \begin{panelbox}[colback=pink!8, equal height group=im31]

      \textbf{Response from Qwen3-VL-4B-RL (correct):}\\
Answer: \boxed{(b)}

We are given two overlapping rectangles:

- A **green rectangle** containing the number **2**.
- A **purple rectangle** containing the number **7** and overlapping with the green rectangle at a point where the number **8** is located.

The question asks: **What is the total of the numbers present simultaneously in the purple and green rectangle(s)?**

---
\#\#\# Step-by-step reasoning:

1. The green rectangle has the number **2**.
2. The purple rectangle has the number **7**.
3. The two rectangles overlap, and in the overlapping region, there is the number **8**.

The key phrase is: **“simultaneously in the purple and green rectangle(s)”** — meaning, numbers that are **in both rectangles**.

The number **8** is located **at the intersection** of the two rectangles. Therefore, it is present in **both** the green and purple rectangles.

The numbers **2** and **7** are only in their respective rectangles — **2** is only in the green rectangle, and **7** is only in the purple rectangle. So they are not present in both.

Thus, the only number that is present simultaneously in both rectangles is **8**.

---

\#\#\# Final Answer:

The total of the numbers present simultaneously in both rectangles is just **8**, since it’s the only number shared between them.

\boxed{8}

    \end{panelbox}
  \end{minipage}

\end{panelbox}
\vspace{-1em}

\caption{Qwen3-VL-4B vs.\ Qwen3-VL-4B-RL response on \emph{Venn Diagram} task, with incorrect reasoning highlighted in red.}
\label{fig:qwen3-4b-venn}
\end{figure*}

\begin{figure*}[t]
\centering

%===================== TOP ROW =====================%
\begin{panelbox}[colback=black!4]

  %---- Top-left (image) ----%
  {\centering
  \includegraphics[width=0.4\linewidth]{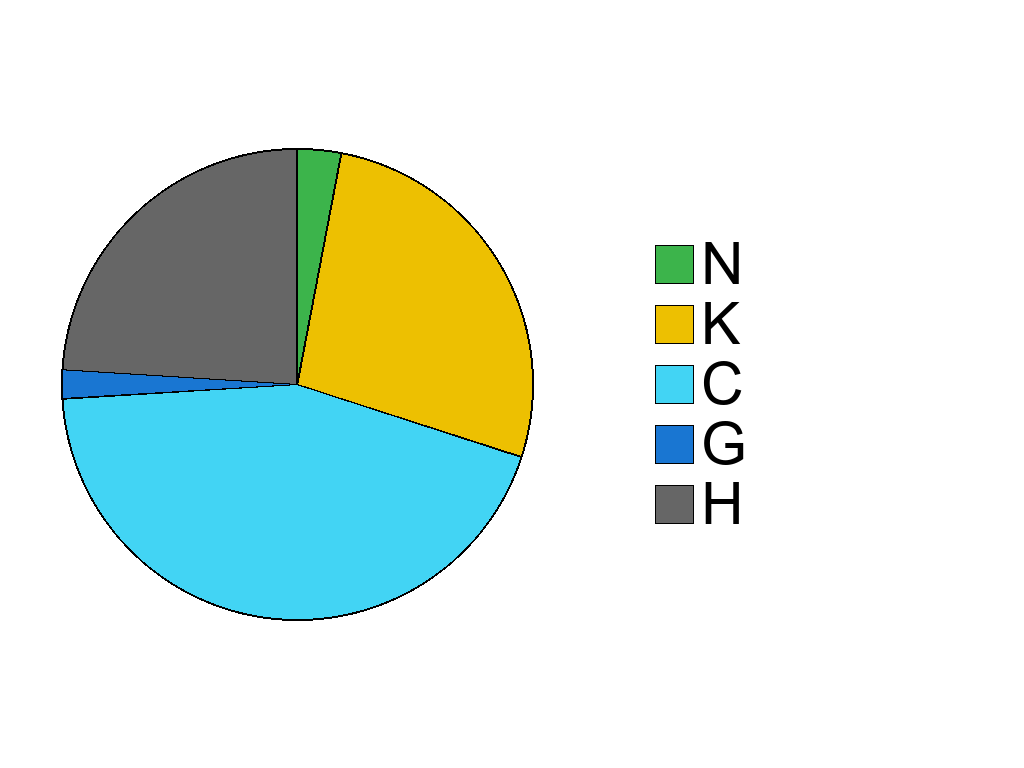}\par}

  %---- Question + Answer ----%
  \textbf{Question:} Select the option (a)–(d) that lists categories in decreasing share (largest → smallest).

(a) C,K,H,N,G
(b) C,N,H,K,G
(c) C,H,K,N,G
(d) C,G,H,N,K

Answer with one letter (a–d).

  \textbf{Answer:} (a)

\medskip

  %---- Top row responses ----%
  \noindent
  \begin{minipage}[t]{0.498\linewidth}
    \begin{panelbox}[colback=black!2, equal height group=im31]

      \textbf{Response from Qwen3-VL-8B (Incorrect):}\\
To solve this, we need to determine the relative sizes of the slices in the pie chart and then order the categories from largest to smallest.

Looking at the pie chart:

- The **light blue** slice (C) is the **largest** portion — it takes up more than half the circle.
\textcolor{red}{- The **dark grey** slice (H) is the second largest.}
\textcolor{red}{- The **yellow** slice (K) is the third largest}
- The **green** slice (N) is a small slice, but larger than the blue slice (G).
- The **blue** slice (G) is the **smallest**.

So, ordering from largest to smallest:

1. C (largest)
2. H
3. K
4. N
5. G (smallest)

This matches option **(c)**: C, H, K, N, G

Let’s verify the other options quickly:

- (a) C,K,H,N,G → K before H? No, H is larger than K.
- (b) C,N,H,K,G → N before H? No, H is larger than N.
- (d) C,G,H,N,K → G before H? No, G is smallest, H is second largest.

Only option (c) has the correct order.

Final Answer:
$\boxed{\text{c}}$

    \end{panelbox}
  \end{minipage}\hfill
  \begin{minipage}[t]{0.498\linewidth}
    \begin{panelbox}[colback=pink!8, equal height group=im31]

      \textbf{Response from Qwen3-VL-8B-RL (correct):}\\
To determine the correct option, we need to analyze the pie chart and rank the categories by their share from largest to smallest.

Looking at the pie chart:

- The **light blue** slice (C) is the **largest** portion.
- The **yellow** slice (K) is the **second largest**.
- The **gray** slice (H) is the **third largest**.
- The **green** slice (N) is the **fourth largest**.
- The **dark blue** slice (G) is the **smallest**.

So, the order from largest to smallest is: **C > K > H > N > G**

Now, let’s check the options:

(a) C,K,H,N,G → This matches our order exactly.

(b) C,N,H,K,G → Incorrect, because N is smaller than H and K.

(c) C,H,K,N,G → Incorrect, because H is smaller than K.

(d) C,G,H,N,K → Incorrect, because G is the smallest, not second.

Therefore, the correct answer is **(a)**.

\boxed{a}

    \end{panelbox}
  \end{minipage}

\end{panelbox}
\vspace{-1em}

\caption{Qwen3-VL-8B vs.\ Qwen3-VL-8B-RL response on \emph{Pie Chart} task, with incorrect reasoning highlighted in red.}
\label{fig:qwen3-7b-pie}
\end{figure*}

\begin{figure*}[t]
\centering

%===================== TOP ROW =====================%
\begin{panelbox}[colback=black!4]

  %---- Top-left (image) ----%
  {\centering
  \includegraphics[width=0.52\linewidth]{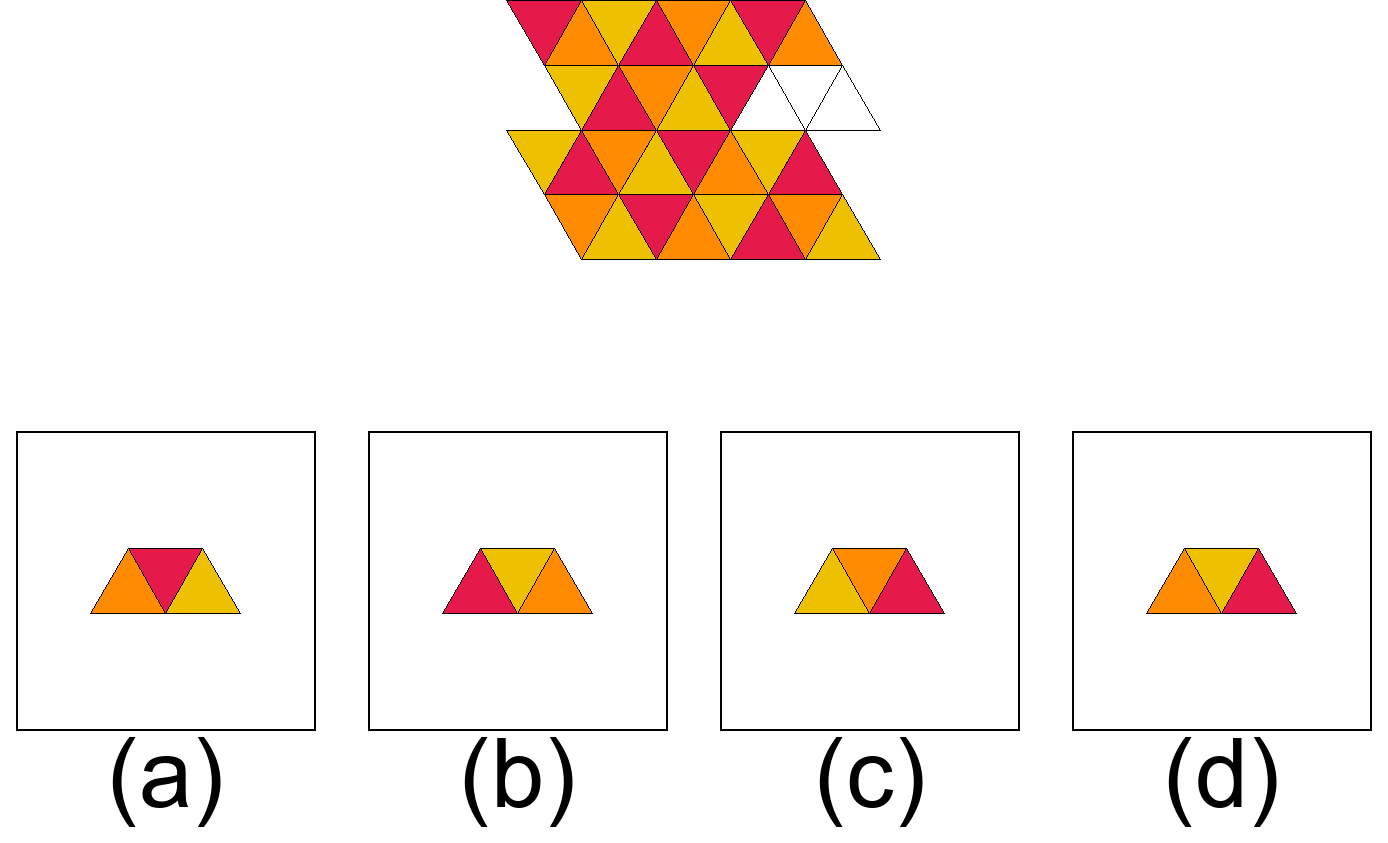}\par}

  %---- Question + Answer ----%
  \textbf{Question:} The shape and position of the blank region are fixed in the top figure. Which option (a)–(d) restores its correct coloring?

  \textbf{Answer:} (d)

  %---- Top row responses ----%
  \noindent
  \begin{minipage}[t]{0.498\linewidth}
    \begin{panelbox}[colback=black!2, equal height group=im31]

      \textbf{Response from Qwen2.5-VL-3B (Incorrect):}\\
To determine which option restores the correct coloring of the blank region in the top figure, we need to analyze the pattern and the given options step by step.

1. **Identify the Pattern:**
   - The top figure shows a pattern where the triangles are colored in a specific sequence.
   - The sequence of colors is: red, yellow, orange, red, yellow, orange, and so on.

 \textcolor{red}{2. **Analyze the Options:**
   - **Option (a):** This option shows a red triangle on the left and a yellow triangle on the right.
   - **Option (b):** This option shows a yellow triangle on the left and an orange triangle on the right.
   - **Option (c):** This option shows an orange triangle on the left and a red triangle on the right.
   - **Option (d):** This option shows a yellow triangle on the left and a red triangle on the right.}

3. **Match the Pattern:**
   - The pattern in the top figure is red, yellow, orange, red, yellow, orange, and so on.
   - We need to find the correct sequence for the blank region.

4. **Determine the Correct Sequence:**
   - The blank region in the top figure is the third triangle from the left in the second row.
   - According to the pattern, the third triangle in the second row should be orange.

 \textcolor{red}{5. **Conclusion:**
   - The correct option that matches the pattern is option (b), which shows an orange triangle on the left and a yellow triangle on the right.}

Therefore, the correct answer is \boxed{b}.

    \end{panelbox}
  \end{minipage}\hfill
  \begin{minipage}[t]{0.498\linewidth}
    \begin{panelbox}[colback=pink!8, equal height group=im31]

      \textbf{Response from Qwen2.5-VL-3B-RL (correct):}\\
To determine the correct coloring for the blank region, we need to analyze the pattern of colors in the given figure and apply it to the blank region.

1. **Identify the pattern in the given figure:**
   - The figure is composed of smaller triangles.
   - The colors of the triangles alternate in a specific pattern.

2. **Analyze the colors in the given figure:**
   - The top-left triangle is red.
   - The triangle to its right is yellow.
   - The triangle below the red triangle is orange.
   - The triangle to the right of the orange triangle is red.
   - This pattern continues throughout the figure.

3. **Determine the color of the blank region:**
   - The blank region is located in the middle of the figure.
   - The pattern suggests that the blank region should be the same color as the triangle directly above it.

4. **Identify the color of the triangle directly above the blank region:**
   - The triangle directly above the blank region is yellow.

5. **Conclusion:**
   - The blank region should be colored yellow to maintain the pattern.

Therefore, the correct answer is \boxed{d}.

    \end{panelbox}
  \end{minipage}

\end{panelbox}
\vspace{-1em}

\caption{Qwen2.5-VL-3B vs.\ Qwen2.5-VL-3B-RL response on \emph{Missing Tiles} task, with incorrect reasoning highlighted in red.}
\label{fig:qwen2-3b-missing}
\end{figure*}

\begin{figure*}[t]
\centering

%===================== TOP ROW =====================%
\begin{panelbox}[colback=black!4]

  %---- Top-left (image) ----%
  {\centering
  \includegraphics[width=0.3\linewidth]{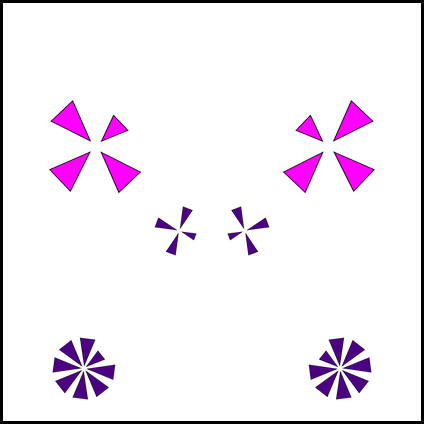}\par}

  %---- Question + Answer ----%
  \textbf{Question:} Exactly one option describes the figure’s mirror symmetry. Which is it? (a) reflection across a vertical line, (b) line symmetry along the anti-diagonal, (c) vertical + horizontal symmetry, (d) main-diagonal ($\searrow$) mirror, (e) horizontal line symmetry, (f) no reflection symmetry

  \textbf{Answer:} (a)

\medskip

  %---- Top row responses ----%
  \noindent
  \begin{minipage}[t]{0.498\linewidth}
    \begin{panelbox}[colback=black!2, equal height group=im34]

      \textbf{Response from Qwen2.5-VL-7B (Incorrect):}\\
To determine the type of symmetry in the given figure, let's analyze it step by step:

1. **Vertical Symmetry**: Check if the figure can be divided into two identical halves by a vertical line.
   \textcolor{red}{- The figure does not have a vertical line of symmetry because the shapes on the left and right sides are not mirror images of each other.}

2. **Horizontal Symmetry**: Check if the figure can be divided into two identical halves by a horizontal line.
   - The figure does not have a horizontal line of symmetry because the shapes above and below the horizontal line are not mirror images of each other.

3. **Anti-Diagonal Symmetry**: Check if the figure can be divided into two identical halves by a diagonal line from the top left to the bottom right.
   - The figure does not have anti-diagonal symmetry because the shapes on the left and right sides are not mirror images of each other along this diagonal.

4. **Main-Diagonal ($\searrow$) Symmetry**: Check if the figure can be divided into two identical halves by a diagonal line from the top right to the bottom left.
   - The figure does not have main-diagonal symmetry because the shapes on the top and bottom are not mirror images of each other along this diagonal.

5. **No Reflection Symmetry**: Check if the figure has any reflection symmetry.
   \textcolor{red}{- The figure does not have any reflection symmetry because none of the lines of symmetry mentioned above apply.}

Given the analysis, the figure does not have any of the described symmetries.

The final answer is \boxed{f}

    \end{panelbox}
  \end{minipage}\hfill
  \begin{minipage}[t]{0.498\linewidth}
    \begin{panelbox}[colback=pink!8, equal height group=im34]

      \textbf{Response from Qwen2.5-VL-7B-RL (correct):}\\
To determine the type of symmetry in the figure, let's analyze it step by step:

1. **Vertical Symmetry (a) and (c):**
   - The figure appears to have vertical symmetry along the vertical axis passing through the center.
   - The shapes on the left and right sides of the vertical axis are mirror images of each other.

2. **Horizontal Symmetry (e):**
   - The figure does not have horizontal symmetry. The shapes on the top and bottom do not mirror each other.

3. **Anti-Diagonal Symmetry (b):**
   - The figure does not have anti-diagonal symmetry. The shapes do not mirror each other along the anti-diagonal line.

4. **Main-Diagonal ($\searrow$) Symmetry (d):**
   - The figure does not have main-diagonal symmetry. The shapes do not mirror each other along the main-diagonal line.

5. **No Reflection Symmetry (f):**
   - The figure does have some form of reflection symmetry, so this option is incorrect.

Given the analysis, the figure has vertical symmetry along the vertical axis passing through the center.

Therefore, the correct answer is (a) reflection across a vertical line.

\boxed{a}

    \end{panelbox}
  \end{minipage}

\end{panelbox}
\vspace{-1em}

\caption{Qwen2.5-VL-7B vs.\ Qwen2.5-VL-7B-RL response on \emph{Mirror Identification} task, incorrect reasoning highlighted in red.}
\label{fig:qwen2-7b-mirror}
\end{figure*}

\paragraph{Example Response.}
We provide several example responses from the base models and the RLVR-trained models in Figure~\ref{fig:qwen3-4b-venn}, Figure~\ref{fig:qwen3-7b-pie}, Figure~\ref{fig:qwen2-3b-missing} and Figure~\ref{fig:qwen2-7b-mirror}.

\section{Evaluation}
\label{app:evalprompt}

\subsection{Prompt}
\begin{tcolorbox}
[colback=black!2,colframe=black!60]
\small
\texttt{Solve the problem by reasoning step by step. Then provide the final answer inside \textbackslash boxed\{\}.}\\[0.4em]
\texttt{Question: \{\}}
\end{tcolorbox}

\begin{table*}[p]
\centering
\newcolumntype{Y}{>{\centering\arraybackslash}X}
\scriptsize
\setlength{\tabcolsep}{2.5pt}
\renewcommand{\arraystretch}{1.03}

\begin{subtable}{\textwidth}
\centering
\begin{tabularx}{\linewidth}{@{\extracolsep{\fill}} p{2.8cm} | Y Y Y Y Y Y Y}
\toprule
\textbf{Task Name} &
\textbf{Hum} &
\textbf{GPT-5} &
\makecell{\textbf{GPT-5}\\\textbf{Mini}} &
\makecell{\textbf{GPT-5}\\\textbf{Nano}} &
\makecell{\textbf{Gemini3}\\\textbf{Flash}} &
\makecell{\textbf{Claude4.5}\\\textbf{Opus}} &
\makecell{\textbf{Claude4.5}\\\textbf{Sonnet}} \\
\midrule
Positional Count & 72.4 & 43 & 34 & 14 & 33 & 33 & 59 \\
Shape Sorting & 100.0 & 78 & 85 & 61 & 29 & 83 & 72 \\
Stack Count & 80.0 & 60 & 24 & 10 & 0 & 45 & 35 \\
Pie Chart & 88.0 & 92 & 91 & 77 & 66 & 77 & 93 \\
Chart Comparison & 96.8 & 97 & 96 & 60 & 22 & 51 & 81 \\
\midrule
Venn Diagram & 65.0 & 52 & 57 & 53 & 51 & 62 & 62 \\
Shape Counting & 55.2 & 76 & 61 & 46 & 61 & 67 & 67 \\
Tiles Line Length & 86.4 & 14 & 18 & 10 & 14 & 16 & 11 \\
Tiles Line Inters. & 75.9 & 26 & 25 & 10 & 1 & 18 & 19 \\
Tiles Recoloring & 69.6 & 15 & 20 & 5 & 26 & 14 & 42 \\
\midrule
Mirror Identification & 68.6 & 65 & 61 & 63 & 40 & 59 & 11 \\
Symmetry Fill & 76.0 & 61 & 54 & 30 & 11 & 41 & 40 \\
Frieze Groups & 48.4 & 26 & 32 & 29 & 21 & 24 & 34 \\
Wallpaper Groups & 86.4 & 48 & 51 & 41 & 32 & 53 & 52 \\
\midrule
Transform Res. Identify & 87.9 & 68 & 51 & 37 & 31 & 43 & 37 \\
Transform Pair Infer & 63.2 & 57 & 51 & 26 & 17 & 30 & 25 \\
Transform Sim. Identify & 82.9 & 38 & 26 & 17 & 11 & 27 & 19 \\
Sequence Rotation & 95.2 & 55 & 50 & 35 & 17 & 29 & 37 \\
Sequence Arithmetic & 93.8 & 50 & 68 & 48 & 24 & 42 & 69 \\
Sequence Multi-Column & 74.3 & 41 & 29 & 28 & 14 & 22 & 37 \\
\midrule
Tiles Geometry & 61.9 & 44 & 41 & 26 & 31 & 33 & 41 \\
Tiles Conn. Component & 57.1 & 44 & 41 & 22 & 29 & 35 & 56 \\
Tiles Shortest Path & 64.9 & 32 & 27 & 17 & 1 & 26 & 33 \\
Missing Tiles & 95.2 & 47 & 36 & 31 & 7 & 30 & 42 \\
Tiles Composition & 50.0 & 48 & 49 & 33 & 1 & 38 & 41 \\
\bottomrule
\end{tabularx}
\caption{Human and proprietary LVLMs.}
\label{tab:visual-reasoning-performance-proprietary}
\end{subtable}

\vspace{0.8em}

\begin{subtable}{\textwidth}
\centering
\begingroup
\fontsize{6.0}{6.8}\selectfont
\setlength{\tabcolsep}{1.8pt}
\renewcommand{\arraystretch}{0.98}
\resizebox{0.97\textwidth}{!}{%
\begin{tabularx}{\linewidth}{@{\extracolsep{\fill}} p{2.3cm} | Y Y Y Y Y Y Y Y Y Y Y Y Y Y}
\toprule
\textbf{Task Name} &
\textbf{Hum} &
\makecell{\textbf{Intern}\\\textbf{VL3}\\\textbf{8B}} &
\makecell{\textbf{Intern}\\\textbf{VL3}\\\textbf{38B}} &
\makecell{\textbf{Llama}\\\textbf{3.2}\\\textbf{11B}} &
\makecell{\textbf{Qwen}\\\textbf{2.5-VL}\\\textbf{3B}} &
\makecell{\textbf{Qwen}\\\textbf{2.5-VL}\\\textbf{7B}} &
\makecell{\textbf{Qwen}\\\textbf{2.5-VL}\\\textbf{32B}} &
\makecell{\textbf{Qwen}\\\textbf{3-VL}\\\textbf{4B}} &
\makecell{\textbf{Qwen}\\\textbf{3-VL}\\\textbf{8B}} &
\makecell{\textbf{Qwen}\\\textbf{3-VL}\\\textbf{30B}} &
\makecell{\textbf{Qwen}\\\textbf{3-VL}\\\textbf{235B}} &
\makecell{\textbf{Molmo}\\\textbf{2}\\\textbf{4B}} &
\makecell{\textbf{Molmo}\\\textbf{2}\\\textbf{8B}} &
\makecell{\textbf{Molmo}\\\textbf{2-O}\\\textbf{7B}} \\
\midrule
Positional Count & 72.4 & 19 & 26 & 14 & 15 & 14 & 36 & 31 & 23 & 26 & 37 & 18 & 30 & 20 \\
Shape Sorting & 100.0 & 51 & 69 & 32 & 56 & 61 & 63 & 41 & 53 & 42 & 91 & 41 & 4 & 0 \\
Stack Count & 80.0 & 21 & 30 & 13 & 6 & 13 & 20 & 25 & 27 & 29 & 45 & 31 & 24 & 39 \\
Pie Chart & 88.0 & 50 & 79 & 30 & 54 & 74 & 91 & 69 & 78 & 83 & 97 & 40 & 2 & 0 \\
Chart Comparison & 96.8 & 26 & 23 & 23 & 24 & 29 & 52 & 46 & 43 & 57 & 64 & 23 & 29 & 20 \\
\midrule
Venn Diagram & 65.0 & 31 & 44 & 15 & 17 & 31 & 45 & 35 & 31 & 50 & 48 & 33 & 30 & 32 \\
Shape Counting & 55.2 & 9 & 9 & 3 & 3 & 17 & 31 & 27 & 32 & 40 & 55 & 6 & 11 & 3 \\
Tiles Line Length & 86.4 & 8 & 10 & 9 & 3 & 4 & 8 & 8 & 8 & 9 & 9 & 7 & 8 & 7 \\
Tiles Line Inters. & 75.9 & 12 & 14 & 14 & 9 & 13 & 15 & 13 & 14 & 12 & 15 & 13 & 17 & 14 \\
Tiles Recoloring & 69.6 & 11 & 17 & 5 & 9 & 8 & 10 & 10 & 15 & 21 & 19 & 8 & 4 & 8 \\
\midrule
Mirror Identification & 68.6 & 31 & 39 & 18 & 20 & 34 & 42 & 54 & 46 & 47 & 54 & 39 & 30 & 33 \\
Symmetry Fill & 76.0 & 22 & 30 & 24 & 24 & 26 & 25 & 27 & 32 & 24 & 34 & 17 & 17 & 26 \\
Frieze Groups & 48.4 & 34 & 30 & 22 & 24 & 27 & 34 & 26 & 20 & 28 & 40 & 21 & 23 & 30 \\
Wallpaper Groups & 86.4 & 33 & 36 & 21 & 23 & 28 & 33 & 38 & 39 & 37 & 45 & 19 & 23 & 25 \\
\midrule
Transform Res. Identify & 87.9 & 26 & 27 & 27 & 21 & 26 & 32 & 37 & 43 & 41 & 37 & 24 & 28 & 30 \\
Transform Pair Infer & 63.2 & 18 & 26 & 22 & 19 & 15 & 18 & 14 & 35 & 25 & 20 & 15 & 18 & 16 \\
Transform Sim. Identify & 82.9 & 23 & 25 & 26 & 23 & 16 & 19 & 29 & 24 & 24 & 24 & 21 & 26 & 16 \\
Sequence Rotation & 95.2 & 30 & 23 & 24 & 18 & 24 & 26 & 32 & 32 & 22 & 45 & 25 & 20 & 27 \\
Sequence Arithmetic & 93.8 & 31 & 31 & 23 & 27 & 33 & 43 & 44 & 38 & 38 & 56 & 24 & 24 & 18 \\
Sequence Multi-Column & 74.3 & 15 & 26 & 20 & 25 & 27 & 22 & 23 & 26 & 23 & 29 & 18 & 21 & 21 \\
\midrule
Tiles Geometry & 61.9 & 22 & 24 & 10 & 10 & 16 & 34 & 19 & 23 & 20 & 36 & 18 & 17 & 17 \\
Tiles Conn. Component & 57.1 & 28 & 29 & 16 & 19 & 16 & 28 & 38 & 29 & 29 & 45 & 28 & 26 & 21 \\
Tiles Shortest Path & 64.9 & 6 & 14 & 10 & 12 & 11 & 9 & 9 & 12 & 12 & 27 & 11 & 7 & 7 \\
Missing Tiles & 95.2 & 21 & 37 & 25 & 23 & 24 & 33 & 18 & 26 & 29 & 25 & 28 & 31 & 23 \\
Tiles Composition & 50.0 & 23 & 26 & 25 & 21 & 31 & 37 & 22 & 29 & 28 & 35 & 27 & 21 & 14 \\
\bottomrule
\end{tabularx}
}
\endgroup
\caption{Human and open-source LVLMs.}
\label{tab:visual-reasoning-performance-open}
\end{subtable}

\caption{Task-level accuracy comparison across the 25 \textsc{Sphinx} tasks. All values are percentages.}
\label{tab:visual-reasoning-performance}
\end{table*}

\subsection{Evaluation Protocol}
We evaluated all model outputs using a two-stage extraction and scoring pipeline. For each question, we first infer the answer type from the ground-truth label (multiple choice, integer, or free-form text) and then apply a set of heuristic rules to extract a canonical answer from the model's raw response. When this heuristic extraction succeeds, the answer is scored directly using \texttt{mathruler} with strict type-consistent normalization. If the heuristic stage cannot produce a valid extraction, we query a small GPT-5-mini model at temperature 0 with a short few-shot prompt to perform answer extraction. The extracted output is then compared to the ground truth.

\section{Task Accuracy}

Tables~\ref{tab:visual-reasoning-performance-proprietary} and~\ref{tab:visual-reasoning-performance-open} present the task-level accuracies of the evaluated models on the 25 tasks in the \textsc{Sphinx} benchmark.

\end{document}